\documentclass[12pt]{settings/ri_thesis}

\usepackage[english]{babel}
\usepackage[autostyle]{csquotes}
\usepackage{microtype}
\usepackage{graphicx}
\usepackage{epigraph}
\usepackage[cmex10]{amsmath}
\usepackage{amssymb,amsfonts,amsthm}
\usepackage{mathrsfs}
\usepackage{appendix}
\usepackage[numbers,sort]{natbib}
\usepackage{setspace}
\usepackage{layout}
\usepackage{tocloft}  
\usepackage[usenames,dvipsnames]{xcolor}

\usepackage[framemethod=TikZ]{mdframed}

\usepackage{tikz}

\usepackage{algorithm}
\usepackage{algpseudocode} 
\usepackage{multirow}

\usepackage{xspace} 
\usepackage[raggedright]{titlesec}  

\usepackage{float}
\newfloat{algorithm}{tbp}{lop}

\colorlet{documentLinkColor}{RoyalBlue}
\colorlet{documentCitationColor}{Turquoise}
\colorlet{documentURLColor}{ProcessBlue}

\usepackage[backref,
        pageanchor=true,
        plainpages=false,
        pdfpagelabels,
        bookmarks,
        bookmarksnumbered,
]{hyperref}
\AtBeginDocument{%
  \hypersetup{
     citecolor = documentCitationColor,
     linkcolor = documentLinkColor,
     urlcolor = documentURLColor,
}}

\usepackage{bm}
\usepackage{nicefrac}

\usepackage{palatino}

\usepackage{mathtools}
\usepackage{booktabs}
\usepackage{array}
\usepackage{colortbl}
\usepackage{ifxetex}

\usepackage{soul}
\usepackage{everysel}
\usepackage{verbatim}

\usepackage[strict]{changepage}

\usepackage[T1]{fontenc}  
\usepackage[b]{esvect}    

\usepackage{xargs}                      

\usepackage{longtable}
\usepackage{enumitem}

\usepackage{subcaption}

\usepackage{ifdraft}
\ifdraft{%
  \usepackage{draftwatermark}
  \SetWatermarkText{DRAFT}
  \SetWatermarkAngle{55}
  \SetWatermarkScale{6.0}
  \SetWatermarkLightness{0.85}
  \SetWatermarkFontSize{12 pt}
}

\usepackage{etoolbox}

\usepackage[nameinlink]{cleveref}

\usepackage[Lenny]{fncychap}

\usepackage{xurl}
\usepackage{fancyhdr}

\colorlet{headergray}{MidnightBlue!50}

\fancypagestyle{plain}{%
  \fancyhf{}
  \fancyfoot[RO,LE]{\thepage}
}

\fancypagestyle{thesis}{%
  \fancyhead{}%
  \fancyhead[RO,LE]{\hyperlink{contents}{\slshape \leftmark}}
  \fancyfoot[RO,LE]{\thepage}%
}

\graphicspath{ {figures/} }

\usepackage[
  reset,
  letterpaper,
  twoside,
  vscale=.75, 
  hscale=.70,  
  nomarginpar,  
  heightrounded,  
]{geometry}

\DeclareMathOperator*{\argmax}{\arg\!\max}

\newcommand{\chapternote}[1]{{%
  \let\thempfn\relax
  \footnotetext[0]{\emph{#1}}
}}

\newlength{\alphabet}
\EverySelectfont{%
  \fontdimen2\font=0.4em
  \fontdimen3\font=0.2em
  \fontdimen4\font=0.1em
  \fontdimen7\font=0.1em
  \hyphenchar\font=`\-
}

\newcolumntype{L}[1]{>{\raggedright\let\newline\\\arraybackslash\hspace{0pt}}m{#1}}
\newcolumntype{C}[1]{>{\centering\let\newline\\\arraybackslash\hspace{0pt}}m{#1}}
\newcolumntype{R}[1]{>{\raggedleft\let\newline\\\arraybackslash\hspace{0pt}}m{#1}}
\newcolumntype{J}[1]{>{\let\newline\\\arraybackslash\hspace{0pt}}m{#1}}
\newcolumntype{j}[1]{>{\let\newline\\\arraybackslash\hspace{0pt}}b{#1}}

\renewcommand{\arraystretch}{1.2}

\definecolor{darkgray}{rgb}{0.2,0.2,0.2}

\hypersetup{%
  pdftoolbar=false,
  pdfmenubar=true,
  pdfstartview={FitH},
  pdftitle={},
  pdfauthor={Hatem Alismail},
  pdfsubject={},
  pdfkeywords={}{},
  colorlinks=true,
  citecolor=blue,
  linkcolor=blue,
  urlcolor=blue
}

\newtheorem{prop}{Proposition}[section]

\newcommand{\T}{^{\mathstrut\scriptscriptstyle{\top}}} 
\newcommand{\mc}[1]{\ensuremath{\mathcal{#1}}}   

\newcommand{\nth}[1]{\ensuremath{#1^\text{th}}}

\makeatletter
\DeclareRobustCommand\onedot{\futurelet\@let@token\@onedot}
\def\@onedot{\ifx\@let@token.\else.\null\fi\xspace}

\apptocmd{\thebibliography}{\csname phantomsection\endcsname\addcontentsline{toc}{chapter}{\bibname}}{}{}

\makeatletter
\newlength\xvec@height%
\newlength\xvec@depth%
\newlength\xvec@width%
\newcommand{\xvec}[2][]{%
  \ifmmode%
    \settoheight{\xvec@height}{$#2$}%
    \settodepth{\xvec@depth}{$#2$}%
    \settowidth{\xvec@width}{$#2$}%
  \else%
    \settoheight{\xvec@height}{#2}%
    \settodepth{\xvec@depth}{#2}%
    \settowidth{\xvec@width}{#2}%
  \fi%
  \def\xvec@arg{#1}%
  \def\xvec@dd{:}%
  \def\xvec@d{.}%
  \raisebox{.2ex}{\raisebox{\xvec@height}{\rlap{%
    \kern.05em
    \begin{tikzpicture}[scale=1]
    \pgfsetroundcap
    \draw (.05em,0)--(\xvec@width-.05em,0);
    \draw (\xvec@width-.05em,0)--(\xvec@width-.15em, .075em);
    \draw (\xvec@width-.05em,0)--(\xvec@width-.15em,-.075em);
    \ifx\xvec@arg\xvec@d%
      \fill(\xvec@width*.45,.5ex) circle (.5pt);%
    \else\ifx\xvec@arg\xvec@dd%
      \fill(\xvec@width*.30,.5ex) circle (.5pt);%
      \fill(\xvec@width*.65,.5ex) circle (.5pt);%
    \fi\fi%
    \end{tikzpicture}%
  }}}%
  #2%
}
\makeatother

\newcommandx{\unsure}[2][1=]{\todo[linecolor=red,backgroundcolor=red!25,bordercolor=red,#1]{#2}}
\newcommandx{\todopicture}[2][1=]{\todo[linecolor=blue,backgroundcolor=blue!25,bordercolor=blue,#1]{#2}}
\newcommandx{\info}[2][1=]{\todo[linecolor=OliveGreen,backgroundcolor=OliveGreen!25,bordercolor=OliveGreen,#1]{#2}}
\newcommandx{\improvement}[2][1=]{\todo[linecolor=Plum,backgroundcolor=Plum!25,bordercolor=Plum,#1]{#2}}
\newcommandx{\thiswillnotshow}[2][1=]{\todo[disable,#1]{#2}}

{ \renewcommand{\arraystretch}{1.5}%
  \begin{tabular}[t]{@{} p{\linewidth} @{}}%
}%
{\end{tabular}}

\newcounter{challenge}[section]
\newenvironment{challenge}[2][]
{ \refstepcounter{challenge}\par\medskip
 \noindent \textbf{Challenge~\thechallenge.~~#2:} \par}
{\medskip}

\newcommand*\matrice[1]{\begin{bmatrix}#1\end{bmatrix}}

\makeatletter
\renewcommand*\env@matrix[1][\arraystretch]{%
  \edef\arraystretch{#1}%
  \hskip -\arraycolsep
  \let\@ifnextchar\new@ifnextchar
  \array{*\c@MaxMatrixCols c}}
\makeatother

\let\stdvec\vec                                         

\newcommand{\FrameSymbol}{\mc{F}}                       
\newcommand{\SetSymbol}{\mc{S}}                       
\renewcommand{\vec}[1]{\xvec[]{#1}}                     
\newcommand{\dvec}[1]{\xvec[.]{#1}}                     
\newcommand{\ddvec}[1]{\xvec[:]{#1}}                    
\newcommand{\frm}[1]{\mc{#1}}                           
\newcommand{\frmsub}[2][]{\mc{#2}_{#1}}                 
\newcommand{\Frm}[1]{{{\FrameSymbol}^{\frm{#1}}}}       
\newcommand{\Frmsub}[2][]{{\FrameSymbol}^{\frmsub[#1]{#2}}} 
\newcommand{\mat}[1]{{\mathbf{#1}}}                     
\newcommand{\numset}[3]{{\mathbb{#1}}^{{#2} \times {#3}}} 
\newcommand{\point}[1]{{#1}}
\newcommand{\axis}[1]{\hat{#1}}
\newcommand{\unit}[1]{\left[ #1 \right]}
\newcommand{\matcomp}[2]{{\left[ #1 \right]}_{#2}}
\newcommand{\vecelem}[2]{{_{#2}{#1}}}
\newcommand{\Project}[2]{{_{#2}{#1}}}
\newcommand{\Plane}[2]{{#1 #2}}

\newcommand{\Set}[1]{\SetSymbol_\mc{#1}}       
\newcommand{\Seti}[2]{\SetSymbol_{\mc{#1}_#2}}       

\definecolor{commentcolor}{gray}{0.5}

\algrenewcommand\algorithmicindent{1.0em}%

\algnewcommand{\LineComment}[1]{\State \textcolor{commentcolor}{\(\triangleright\) #1}}
\algnewcommand{\NewComment}[1]{\textcolor{commentcolor}{\(\triangleright\) #1}}
\algnewcommand{\To}{\textbf{to}}
\algnewcommand{\Break}{\textbf{break}}
\algnewcommand{\Continue}{\textbf{continue}}
\algnewcommand{\IIf}[1]{\State\algorithmicif\ #1\ \algorithmicthen}
\algnewcommand{\EndIIf}{\unskip}
\algnewcommand{\var}[1]{\textit{#1}}
\algnewcommand{\func}[1]{\textsc{#1}}

\newcommand{\FI}{\Frm{I}}
\newcommand{\FB}{\Frm{B}}

\newcommand{\FC}{\Frm{C}}
\newcommand{\FE}{\Frm{E}}
\newcommand{\FR}[1]{\Frmsub[#1]{R}}

\newcommand{\OI}{\point{O}_{\frm{I}}}
\newcommand{\OB}{\point{O}_{\frm{B}}}
\newcommand{\OR}[1]{\point{O}_{\frmsub[#1]{R}}}
\newcommand{\ORB}[1]{_{\frm{B}}\point{O}_{\frmsub[#1]{R}}}

\newcommand{\XI}{\axis{X}_{\frm{I}}}
\newcommand{\YI}{\axis{Y}_{\frm{I}}}
\newcommand{\ZI}{\axis{Z}_{\frm{I}}}
\newcommand{\XB}{\axis{X}_{\frm{B}}}
\newcommand{\YB}{\axis{Y}_{\frm{B}}}
\newcommand{\ZB}{\axis{Z}_{\frm{B}}}
\newcommand{\XR}[1]{\axis{X}_{\frmsub[#1]{R}}}
\newcommand{\YR}[1]{\axis{Y}_{\frmsub[#1]{R}}}
\newcommand{\ZR}[1]{\axis{Z}_{\frmsub[#1]{R}}}
\newcommand{\XS}{\axis{X}_{\frm{S}}}
\newcommand{\YS}{\axis{Y}_{\frm{S}}}
\newcommand{\ZS}{\axis{Z}_{\frm{S}}}
\newcommand{\XC}{\axis{X}_{\frm{C}}}
\newcommand{\YC}{\axis{Y}_{\frm{C}}}
\newcommand{\ZC}{\axis{Z}_{\frm{C}}}
\newcommand{\XE}{\axis{X}_{\frm{E}}}
\newcommand{\YE}{\axis{Y}_{\frm{E}}}
\newcommand{\ZE}{\axis{Z}_{\frm{E}}}

\newcommand{\kIv}{\axis{\vec{k}}_{\frm{I}}}

\newcommand{\kB}{\axis{\mat{k}}_{\frm{B}}}

\newcommand{\iS}{\axis{\mat{i}}_{\frm{S}}}
\newcommand{\jS}{\axis{\mat{j}}_{\frm{S}}}
\newcommand{\kS}{\axis{\mat{k}}_{\frm{S}}}
\newcommand{\iSv}{\axis{\vec{i}}_{\frm{S}}}
\newcommand{\jSv}{\axis{\vec{j}}_{\frm{S}}}
\newcommand{\kSv}{\axis{\vec{k}}_{\frm{S}}}

\newcommand{\kCv}{\axis{\vec{k}}_{\frm{C}}}

\newcommand{\raxis}{\vec{r}}

\newcommand{\RBI}{\mat{R}_{\frm{BI}}}
\newcommand{\RIB}{\mat{R}_{\frm{IB}}}
\newcommand{\RIS}{\mat{R}_{\frm{IS}}}

\newcommand{\RIC}{\mat{R}_{\frm{IC}}}
\newcommand{\RCI}{\mat{R}_{\frm{CI}}}

\newcommand{\RRB}[1]{\mat{R}_{\frmsub[#1]{R}\frm{B}}}
\newcommand{\RBR}[1]{\mat{R}_{\frm{B}\frmsub[#1]{R}}}
\newcommand{\InertB}{\mat{I}_{\frm{B}}}

\newcommand{\phix}[1]{\phi_{x{#1}}}
\newcommand{\phiy}[1]{\phi_{y{#1}}}

\DeclareMathOperator{\cosine}{c}
\DeclareMathOperator{\sine}{s}
\newcommand{\ctheta}{\cosine{\theta}}
\newcommand{\stheta}{\sine{\theta}}
\newcommand{\cphi}{\cosine{\phi}}
\newcommand{\sphi}{\sine{\phi}}
\newcommand{\cpsi}{\cosine{\psi}}
\newcommand{\spsi}{\sine{\psi}}
\newcommand{\cmu}[1]{\cosine{\mu_{#1}}}
\newcommand{\smu}[1]{\sine{\mu_{#1}}}
\newcommand{\cphix}[1]{\cosine{\phix{#1}}}
\newcommand{\sphix}[1]{\sine{\phix{#1}}}
\newcommand{\cphiy}[1]{\cosine{\phiy{#1}}}
\newcommand{\sphiy}[1]{\sine{\phiy{#1}}}

\DeclareMathOperator{\Span}{Span}
\DeclareMathOperator{\Conv}{Conv}

\newcommand{\spt}{_{sp}}
\newcommand{\des}{_{des}}
\newcommand{\ver}{_{ver}}
\newcommand{\hover}{_{hov}}
\newcommand{\hor}{_{hor}}
\newcommand{\lat}{_{lat}}
\newcommand{\nor}{_{nor}}

\newcommand{\Flmax}{F_{lmax}}
\newcommand{\Flmaxnew}{{F'}_{lmax}}

\newcommand{\Fdes}{\vec{F}\des}
\newcommand{\FdesI}{\vec{F}\des^\frm{I}}

\newcommand{\FdesInew}{\vec{F}\des^{\prime\,\frm{I}}}
\newcommand{\FdesCf}{\stdvec{F}\des^{f\,\frm{C}}}
\newcommand{\FdesIf}{\stdvec{F}\des^{f\,\frm{I}}}
\newcommand{\FdesIp}{\stdvec{F}\des^{p\,\frm{I}}}

\newcommand{\Fwind}{\vec{F}_{app}}
\newcommand{\Fgrav}{\vec{F}_{grav}}
\newcommand{\Fext}{\vec{F}_{ext}}

\newcommand{\Fspt}{\vec{F}\spt}
\newcommand{\FsptB}{\vec{F}\spt^\frm{B}}

\newcommand{\Fver}{\vec{F}\ver}
\newcommand{\FverB}{\vec{F}\ver^{\frm{B}}}

\newcommand{\Flat}{\vec{F}\lat}

\newcommand{\Fnor}{\vec{F}\nor}

\newcommand{\Fhor}{\vec{F}\hor}
\newcommand{\FhorB}{\vec{F}\hor^{\frm{B}}}

\newcommand{\FhorBnew}{\vec{F}\hor^{\prime\,\frm{B}}}

\newcommand{\FsptBnew}{\vec{F}\spt^{\prime\,\frm{B}}}
\newcommand{\FsptBnewnew}{\vec{F}\spt^{''\,\frm{B}}}

\newcommand{\PlaneB}{{\Plane{\XB}{\YB}}}
\newcommand{\PlaneI}{{\Plane{\XI}{\YI}}}
\newcommand{\PlaneS}{{\Plane{\XS}{\YS}}}
\newcommand{\PlaneC}{{\Plane{\XC}{\YC}}}

\newcommand{\ProjB}[1]{\Project{#1}{\PlaneB}}
\newcommand{\ProjI}[1]{\Project{#1}{\PlaneI}}

\newcommand{\Fhover}{F\hover}
\newcommand{\FhovB}{\vec{F}\hover^\frm{B}}
\newcommand{\FhovI}{\vec{F}\hover^\frm{I}}

\newcommand{\SetV}{\Set{V}}
\newcommand{\SetUi}{\Seti{U}{i}}
\newcommand{\SetT}{\Set{T}}
\newcommand{\SetTp}{\Set{T}'}
\newcommand{\SetTb}{\Set{T_\textit{b}}}
\newcommand{\SetM}{\Set{M}}
\newcommand{\SetW}{\Set{W}}
\newcommand{\SetWp}{\Set{W}'}
\newcommand{\SetWf}{\Set{W_\textit{6D}}}

\newcommand{\SP}{\mat{S_p}}
\newcommand{\SF}{\mat{S_f}}

\newcommand{\IdentMat}[1]{\mat{I_{#1 \times #1}}}
\newcommand{\ZeroMat}[1]{\mat{0_{#1 \times #1}}}

\begin {document}

\frontmatter

\pagestyle{empty}

\title{\textbf{Physical Interaction and Manipulation of the Environment using Aerial Robots}\\ \vspace{12pt}}
\author{Azarakhsh Keipour}
\date{May 2nd, 2022}
\Year{2022}
\trnumber{CMU-RI-TR-22-17}

\committee{
Sebastian Scherer, \emph{chair} \\
Oliver Kroemer \\
Wennie Tabib \\
Kostas Alexis, \emph{NTNU}
}

\support{}
\disclaimer{}

\permission{All rights reserved.}

\maketitle

\begin{dedication}
To my beloved wife Juwairia and my most wonderful parents Mitra and Iradge.

\end{dedication}

\pagestyle{plain} 

\begin{abstract}
\setlength{\parskip}{1em}
\setlength{\parindent}{0em}

\noindent

The physical interaction of aerial robots with their environment has countless potential applications and is an emerging area with many open challenges. Fully-actuated multirotors have been introduced to tackle some of these challenges. They provide complete control over position and orientation and eliminate the need for attaching a multi-DoF manipulation arm to the robot. However, there are still several open problems before they can be used in real-world applications. 

Researchers have introduced some methods for the physical interaction of fully-actuated multirotors in limited settings. Their experiments primarily use prototype-level software without an efficient path to integrating these methods into real-world applications. This thesis describes a new controller design that provides a cost-effective solution for integrating these robots with the existing software and hardware flight systems for real-world applications. It further expands the controller to physical interaction applications to show its flexibility and effectiveness. 

On the other hand, the existing control approaches for fully-actuated robots assume conservative limits for the thrusts and moments available to the robot. Using conservative assumptions for these already-inefficient robots makes their interactions even less optimal and may even result in many feasible physical interaction applications becoming infeasible. This work proposes a real-time method for estimating the complete set of instantaneously available forces and moments that robots can use to optimize their physical interaction performance.

Finally, many real-world applications where aerial robots can improve the existing manual solutions deal with deformable objects. However, the perception of deformable objects and planning for their manipulation is still challenging. Additionally, no studies have been performed to analyze the requirements of aerial tasks that involve deformable objects. This research explores how aerial physical interaction can be extended to deformable objects. It provides a detection method suitable for manipulating deformable one-dimensional objects and introduces a new perspective on planning the manipulation of these objects. It further studies the viability of working with such objects for aerial manipulators.

\end{abstract}

\begin{acknowledgments}
\setlength{\parskip}{1em}
\setlength{\parindent}{0em}

\noindent
I would like to thank my advisor Sebastian Scherer for his support, advice, and patience during the ups and downs of my graduate journey. His vision and passion for the field have always inspired me to push the boundaries of what robots can do in the real world. His willingness to support his students and his relentless enthusiasm and positive attitude make it really hard to say goodbye to this chapter of my life.

I would also like to thank my committee members for their feedback and suggestions on improving the work along the way up until the end. I would like to thank Kostas Alexis for finding the best ways to improve the work from scientific and storyline perspectives. His invaluable feedback has helped connect the dots of my research together before the proposal and the dissertation. His excellent understanding of details and the value of the work and his eagle-eye perspective have been irreplaceable for me. Oliver Kroemer has always helped with pushing for better and higher-quality work, and his suggestions on how to relate my different parts of work have been highly beneficial. I would like to thank Wennie Tabib for accepting the responsibility to support me at the last minute and for providing excellent feedback on how to make my work better, given the existing time constraints. Thank you to Changliu Liu, who supported me along the way, and with her great control and robotics perspective, I improved my work further. A special thanks to Guilherme A.S. Pereira, who helped start my journey in robotics on my first project and has been a great friend and mentor since. I also received very helpful advice and support during my qualifiers from George Kantor and Maxim Likhachev and am very grateful for all I have learned from them.

During the summer of 2021, in my internship at \textit{X, the moonshot factory} (a.k.a. \textit{Google X}), I had the opportunity to be advised by Maryam Bandari and Stefan Schaal. I would like to thank Stefan for his regular feedback and encouragement on my summer work, his support for publishing the work, and for allowing me to use it in my Ph.D. research. Special thanks go to Maryam, who believed in me and fought hard to make my internship happen, and then gave me space to explore my ideas while providing great feedback and support during and after the summer research.

I would also like to thank Mohammadreza Mousaei for assisting with tests, publications, and lengthy discussions on the concepts, and Junyi (Jenny) Geng for her helpful feedback and helping with the final experiments. Andrew Saba and Greg Armstrong helped in building the UAVs and performing the experiments, Andrew Ashley and John Keller assisted with the flight tests. Thank you to Rogerio Bonatti, Cherie Ho, Jay Patrikar, Weikun Zhen, Kevin Pluckter, Ratnesh Madaan, and Mohammadreza Mousaei for being the most fantastic office mates in the universe. And thank you to all the other former and current AirLab and RI members for their help and support through this work and for making the graduate school a happy time, including but certainly not limited to: Milad Moghassem Hamidi, Sankalp Arora, Rohit Garg, Lucas Nogueira, Brady Moon, Yaoyu Hu, Anish Bhattacharya, Vaibhav Viswanathan, Vishal Dugar, Puru Rastogi, Dong-Ki Kim, Silvio Maeta, Guilherme Pereira, and Geetesh Dubey. No acknowledgment can be complete without thanking Nora Kazour, who was the most supportive friend and lab coordinator who could make magical things happen to facilitate our lives.

Finally, the most special thanks go to my wonderful wife, Juwairia Mulla, and my amazing parents, Iradge and Mitra, for their unconditional support and patience through the whole academic endeavor. 

\end{acknowledgments}

\begin{funding}
\setlength{\parskip}{1em}
\setlength{\parindent}{0em}

This project became possible due to the support and the funding provided by the following institutions:

\begin{itemize}[leftmargin=*]

\setlength\itemsep{1em}
    \item The Robotics Institute Ph.D. program
    \item National Aeronautics and Space Administration (NASA): Grant Number 80NSSC19C010401
    \item X, the moonshot factory (a.k.a., Google X): AI Residency program 
    \item Intrinsic AI, an Alphabet Company
    \item Near Earth Autonomy (NEA)
\end{itemize}

\end{funding}

\cleardoublepage

\setlength{\cftbeforetoctitleskip}{3.0em}
\setlength{\cftaftertoctitleskip}{2.0em}
\setlength{\cftbeforeloftitleskip}{3.0em}  
\setlength{\cftbeforelottitleskip}{3.0em}  
\cftsetpnumwidth{1.75em}  

\hypersetup{linkcolor=black}  
\hypertarget{contents}{}
\microtypesetup{protrusion=false}
\tableofcontents
\clearpage
\listoffigures
\clearpage
\listoftables
\microtypesetup{protrusion=true}

\mainmatter
\pagestyle{thesis}

\hypersetup{linkcolor=documentLinkColor}  


\onehalfspace
\chapter{Introduction} \label{ch:intro}

Our work aims to improve the state-of-the-art in physical interaction and manipulation of the environment using aerial robots and extend it to real-world applications involving rigid and deformable objects. 

This chapter explains the motivation behind choosing the problems we tackled (Section~\ref{sec:intro:motivation}), introduces the current challenges in the field (Section~\ref{sec:intro:challenges}), describes the contributions of our work (Section~\ref{sec:intro:contributions}), and presents the outline of this dissertation (Section~\ref{sec:intro:outline}).
\section{Motivation} \label{sec:intro:motivation}

The last two decades have seen rapid growth in applications for aerial robots, ranging from hobby racing and artistic aerial shows to professional cinematography, security surveillance, and commercial infrastructure inspection. Most of these applications are performed during the flight without any physical contact of the robot with the objects and surfaces around it. 

The unstable nature of the aerial robots makes their controlled physical interaction very challenging. On the other hand, traditional multirotors have a planar arrangement for their rotors, making them underactuated. These multirotors need to tilt in the direction of the generated thrust, affecting the position of their end-effector, making the control of the end-effector position and force even more challenging.

Despite these challenges, the research on applications involving aerial robots' physical interaction with their environment has grown steadily in the last decade. To work around the difficulty of the end-effector control, many researchers attached manipulator arms with multiple degrees of freedom to their aerial robots. This approach increases the payload weight, limits the remaining forces of the robot, and requires a very complex controller to control the end effector reliably. Even with all these challenges, the result is generally a limited reachable workspace for the end-effector, which depends on the arm's properties and attachment location on the robot.

A more novel approach to tackle the challenges of physical interaction is to utilize the recently-introduced fully-actuated multirotors. These robots have a non-planar configuration for their rotors, allowing them to tilt independently from their generated thrust direction. Such an advantage eliminates the need for a heavy and complex manipulation arm attached to the multirotor. Instead, it allows the robot's attitude to be used for controlling the end-effector's position and applied forces. However, there are several drawbacks to this approach: the design of the multirotor is more complex, the robot is less efficient in generating thrusts than the planar designs, the lateral thrust is limited in most common fully-actuated designs, and the integration of the controllers of these new robots into the existing software and hardware flight stacks is not straightforward.

Even with the challenges mentioned above, researchers have found ways to develop and test methods for the physical interaction of aerial robots with their environment in limited settings. The fully-actuated multirotors have been shown to perform tasks such as peg-in-a-hole and contact inspection. However, these experiments have primarily used prototype-level software (such as MATLAB-generated code) without any time- and cost-efficient path to integrating them into an autonomy stack for real-world applications. 

On the other hand, all the work has been performed with conservative assumptions for the possible thrusts that the robot can generate. Since these robots are already inefficient and have limited thrusts, using conservative assumptions for the available thrusts makes the fully-actuated robot interactions even less optimal. It may result in many feasible physical interaction tasks deemed infeasible while resulting in sub-optimal solutions for many other applications. However, this problem is not limited to aerial robots. Estimation of a manipulator's available forces and moments has been used for four decades since the introduction of dynamic manipulability ellipsoids. Ellipsoids can be computed in real-time but are very conservative estimations of the available forces and moments. Estimating the complete set of the available forces and moments can be done using dynamic manipulability polytopes, which currently are computationally expensive and unsuitable for most real-time applications. A real-time solution for estimating the manipulability polytopes can significantly improve the performance of the physical interaction tasks not only for fully-actuated aerial robots but also for the ground manipulators.

The extent of the physical interaction research for aerial robots has been manipulating rigid (or almost-rigid) objects and contact with rigid surfaces, leaving out a major class of objects called deformable objects. Many real-world applications where the multirotors can improve the existing manual solutions include deformable objects. For example, the inspection, maintenance, and manipulation of the cables and wires at the top of the utility pole require working with deformable one-dimensional (a.k.a., deformable linear) objects. 

Several new challenges arise in dealing with these objects, including the perception of their state and planning for their manipulation. Researchers in other robotics communities have been working on these problems for decades and have proposed many methods for perceiving deformable objects. However, many areas, such as detecting the state of the deformable one-dimensional objects (such as wires and cables), lack viable solutions that can be used in fully-automated settings. On the other hand, no studies have been performed to analyze the requirements of the aerial manipulation tasks involving deformable objects.

In this document, we explain our solutions for tackling the challenges mentioned above regarding the physical interaction of aerial robots with their environment. We propose a novel controller design for fully-actuated multirotors that provides a cost-effective solution for integration with the existing software and hardware flight systems for real-world applications. We then expand it to physical interaction applications to show the design's flexibility and effectiveness. 

We propose a set of real-time methods for estimating the instantaneously available forces and moments that an aerial robot can use in its flight and interactions. The methods are proposed to compute the dynamic manipulability polytopes for aerial robots. However, they can be adopted for all kinds of manipulators to replace the conservative estimations made by the dynamic manipulability ellipsoids in real-time applications. We further extend the method to work in physical interactions and illustrate the power of the wrench set estimation methods by improving several aspects of real-world applications. 

Finally, We explore how the physical interaction can be extended to deformable objects. We provide a deformable one-dimensional objects detection method suitable for manipulation tasks and introduce a new perspective on planning for manipulating these objects in specific settings such as utility cable boxes. We further study the viability of aerial manipulation of such objects from the end-effector precision perspective and use the proposed real-time dynamic manipulability polytope computation methods to analyze the feasibility of the wire manipulation task.

\begin{mdframed}[roundcorner=10pt, backgroundcolor=Turquoise!15, innertopmargin=15pt, innerleftmargin=25pt, innerrightmargin=25pt, innerbottommargin=15pt, skipabove=30pt]
    \textit{
    This goal of this work is to improve the state-of-the-art in physical interaction and manipulation using aerial robots by providing faster integration for fully-actuated robots, better estimation of available wrenches and new solutions to allow real-world physical interaction with rigid and deformable objects. 
    }
\end{mdframed}

\section{Challenges and Insights} \label{sec:intro:challenges}

Aerial robots' controlled physical interaction with their environment is still an ongoing research problem. Employing fully-actuated robots has been shown to improve the robustness of the interaction in the presence of real-world uncertainties. However, fully-actuated UAVs create new challenges that need to be addressed before they can effectively be integrated into real-world applications. On the other hand, the autonomous inspection and manipulation of cables and wires using UAVs face additional challenges due to the limited existing research on deformable objects and the lack of solutions for seemingly everyday tasks such as cable detection. 

The major challenges that are currently preventing the physical interaction of fully-actuated UAVs include:

\begin{challenge}{Complete flight stack for fully-actuated UAVs}
The deployment of an aerial robot in an autonomous mission requires a complete system (a.k.a., flight stack) in addition to the flight control subsystem. The flight stack includes software such as a ground control station, planning system, communication protocols, and hardware components such as the safety pilot's remote controller. Due to the nature of fully-actuated UAVs requiring 6-D commands to control as compared to the traditional 4-D commands required for underactuated vehicles, the existing flight stack components cannot be directly used in conjunction with the existing control solutions for fully-actuated UAVs and new solutions should be found to provide a complete flight stack for these vehicles.
\end{challenge}

\begin{challenge}{Wrench limits for most fully-actuated UAV designs}
Most common fully-actuated UAV designs suffer from limited horizontal thrust compared to their overall thrust. On the other hand, controlled physical interaction tasks directly deal with the wrenches generated by the robot and can benefit from the knowledge about these limits and instantaneously available wrenches. However, the existing methods for estimating such limits for UAVs are slow and not suitable for real-time use during flight. Additionally, these methods have almost exclusively been used for UAV architecture design optimization and do not account for the relation between the applied wrenches during flight and the remaining wrenches, making them even less desirable to estimate the available wrenches during the physical interaction.
\end{challenge}

\begin{challenge}{Lack of a complete pipeline for aerial manipulation of deformable objects}
A crucial step in achieving a fully-autonomous mission involving physical interaction with objects such as wires and cables is the perception of these objects. Several segmentation methods exist that can segment these objects, which can be used for visual inspection and obstacle avoidance. On the other hand, many methods have been proposed in the industrial robotics community to track these deformable objects once their initial state is given. However, no method exists to fill the gap between the segmentation and tracking to provide the required initial state for the tracking methods in non-trivial conditions (e.g., a straight line). In order to achieve a fully autonomous mission, this gap needs to be filled.
\end{challenge}

\section{Contributions} \label{sec:intro:contributions}

Applications involving the physical interaction of UAVs with structures involving deformable objects, such as cables and wires, have not been directly addressed in academia or industry yet. There have been efforts on some underlying challenges, especially in recent years. However, several unsolved issues were left out or needed significant improvements to see these types of applications a reality someday.

We tackle the problem of the physical interaction of UAVs by identifying the most critical challenges requiring solutions or improvements to existing solutions. 

We first enable the integration of fully-actuated UAVs into existing flight stacks for real-world applications, cutting the required effort and time for deploying them into the real world. 

Then we address the aerial robots' limited thrust and moment availability to alleviate the issues resulting from the wrench constraints. We propose a method to predict the limits in real-time and use this method to improve different aspects of physical interaction applications, ranging from planning and flight optimization to hardware design and control allocation. 

Finally, the lack of perception methods suitable for physical interaction with wires and cables drove us to propose a detection solution to enable fully-autonomous physical interaction with such objects. Along the way, we also addressed the routing problem for wires and cables from a new angle, improving the performance compared to the existing planning methods, and studied the feasibility of aerial manipulation of these objects from the end-effector's precision and required wrenches perspectives.

A summary of the contributions is as follows:

\begin{enumerate}
    \item A novel controller design is proposed that can extend existing flight controllers to work with the fully-actuated robots. It requires no change to the systems working with the controller and thus provides a short integration timeline with the existing flight stack. A novel set of attitude and thrust strategies and an extension to motion-force control are provided to allow the full actuation of UAVs for indoor and outdoor environments and free flight and physical interaction tasks. See Chapter~\ref{ch:control}.
    
    \item A set of \textit{real-time} wrench set (dynamic manipulability polytopes) estimation methods for calculating the available instantaneous wrenches of robots are proposed. A novel method is proposed to estimate the available wrenches during controlled physical interaction for robotics manipulators. See Chapter~\ref{ch:wrench}.
    
    \item The benefits of the proposed real-time wrench-set estimation methods are showcased by several applications, including: improving the performance and flexibility of the UAV control allocation module; computing the optimal UAV attitude when an external force is exerted during the physical interaction or free flight; improving the planning of physical interaction tasks. See Chapter~\ref{ch:applications}.
    
    \item A novel method for detecting deformable one-dimensional objects (e.g., wires and cables) is proposed to enable autonomous physical interaction and manipulation of such objects. Additionally, a novel approach to spatial representation and routing of deformable one-dimensional objects is proposed that allows significant performance improvement for tasks involving routing on component boards, such as cable boxes being inspected on the utility poles. The first analysis to enable the physical interaction of aerial robots with these deformable objects is performed. See Chapters~\ref{ch:doo-detection} and~\ref{ch:doo-routing}.
\end{enumerate}

\section{Outline} \label{sec:intro:outline}

The outline of this dissertation is as follows:

Chapter~\ref{ch:background} introduces the relevant concepts and the background for the rest of this document. It provides an introduction to fully-actuated multirotors and derives a controller system from the equations of motion of these robots.

Chapter~\ref{ch:control} presents our control design for fully-actuated multirotors to address the current challenges facing the integration of these robots into real-world applications. Then the design is further extended to allow controlled physical interaction with the environment.

Chapter~\ref{ch:wrench} introduces the real-time wrench set estimation methods for fully-actuated multirotors and discusses the different extensions of the method to controlled interaction with the physical world.

Chapter~\ref{ch:applications} illustrates various applications of the real-time wrench set estimation methods to enhance further the use of the fully-actuated multirotors in real-world applications.

Chapter~\ref{ch:doo-detection} presents the novel detection method for deformable one-dimensional objects capable of working with occlusions and an imperfect segmentation, which enables fully-autonomous physical interaction with objects such as cables and wires using ground or aerial manipulators.

Chapter~\ref{ch:doo-routing} proposes a novel approach to spatial representation and routing for deformable one-dimensional objects for manipulation in workspaces such as cable boxes in autonomous tasks such as utility pole manipulation and inspection. It further explores the feasibility of such manipulation tasks from different perspectives for aerial robots. 

Finally, Chapter~\ref{ch:conclusion} provides a summary of the work and outlines the possible further research.

\section{Bibliographical Remarks} \label{sec:intro:bib-remarks}

This thesis only contains the work and research where this author is the primary contributor:

\begin{itemize}[leftmargin=*]
    \item[--] Chapter~\ref{ch:control} is the work done with Sebastian Scherer, Mohammadreza Mousaei, and Junyi Geng, with contributions from John Keller, Andrew Saba, Andrew Ashley, Greg Armstrong, Dongwei (Saeed) Bai, and Near Earth Autonomy. Some of the work has appeared in~\cite{Keipour:2020:arxiv:integration} and~\cite{Keipour:2022:unpub:simulator}.
    
    \item[--] Chapters~\ref{ch:wrench} and~\ref{ch:applications} are based on the work with Sebastian Scherer with contributions from Mohammadreza Mousaei and helpful insights from Oliver Kroemer. The work is to appear in~\cite{Keipour:2022:unpub:wrench}.
    
    \item[--] Chapters~\ref{ch:doo-detection} and~\ref{ch:doo-routing} are based on the work with Maryam Bandari and Stefan Schaal with helpful insights from Kostas Alexis. The work has appeared in~\cite{Keipour:2022:ral:doo}, \cite{Keipour:2022:iros:routing} and~\cite{Keipour:2022:icra-workshop:doo}.
\end{itemize}

\section{Excluded Research} \label{sec:intro:excluded}

The author has excluded a significant portion of his graduate work and publications to keep this thesis focused. The excluded works are listed below:

\begin{enumerate}[leftmargin=*]
    \item Automatic Real-time Anomaly Detection for Autonomous Aerial Vehicles, that appeared in~\cite{Keipour:2019:icra:anomaly}.
    
    \item ALFA: A Dataset for UAV Fault and Anomaly Detection, that appeared in~\cite{Keipour:2021:ijrr:alfa}.
    
    \item Path Planning for Unmanned Fixed-Wing Aircraft in Uncertain Wind Conditions Using Trochoids, that appeared in~\cite{Schopferer:2018:icuas:planning}.
    
    \item Real-Time Ellipse Detection for Robotics Applications, that appeared in~\cite{Keipour:2021:ral:ellipse}.
    
    \item Visual Servoing Approach for Autonomous {UAV} Landing on a Moving Vehicle, that appeared in~\cite{Keipour:2021:arxiv:mbzirc}.
    
    \item Design, Modeling and Control for a Tilt-rotor VTOL UAV in the Presence of Actuator Failure, that appeared in~\cite{Mousaei:2022:iros:vtol}.
    
    \item VTOL Failure Detection and Recovery by Utilizing Redundancy, that appeared in~\cite{Mousaei:2022:icra-workshop:vtol}.

    \item Trajectory Planning for a UAV Wrench Task Considering Vehicle Dynamics and Force Output Capabilities, that appeared in~\cite{Ashley:2021:riss:planning}.
    
    \item Modeling and Control using a Unified MPC Strategy for a Tilt-Rotor VTOL UAV on a Multi-Purpose Platform, that appeared in~\cite{Bai:2022:unpub:mpc}.
\end{enumerate}

\chapter{Background: Fully-Actuated Multirotors and Controllers} \label{ch:background}

This chapter introduces the preliminary definitions and terminology that will set up the infrastructure to discuss the content in the rest of this document. 

Section~\ref{sec:background:control:intro} discusses the related work and provides an introduction to the rest of the chapter. Section~\ref{sec:background:control:modeling} explains the fully-actuated multirotor model and the assumptions used in this work. Sections~\ref{sec:background:control:translational-kinematics}, \ref{sec:background:control:rotational-kinematics} and~\ref{sec:background:control:dynamics} describe the kinematics and dynamics of fully-actuated multirotors. Section~\ref{sec:background:control:control-affine-model} converts the kinematics and dynamics models into a control-affine model. Finally, Section~\ref{sec:background:control:control-allocation} uses the control-affine model to design a controller for fully-actuated multirotors.

\section{Introduction and Related Work} \label{sec:background:control:intro}

\begin{figure}[!htb]
        \centering
        \begin{subfigure}[b]{0.45\textwidth}
            \includegraphics[width=\textwidth]{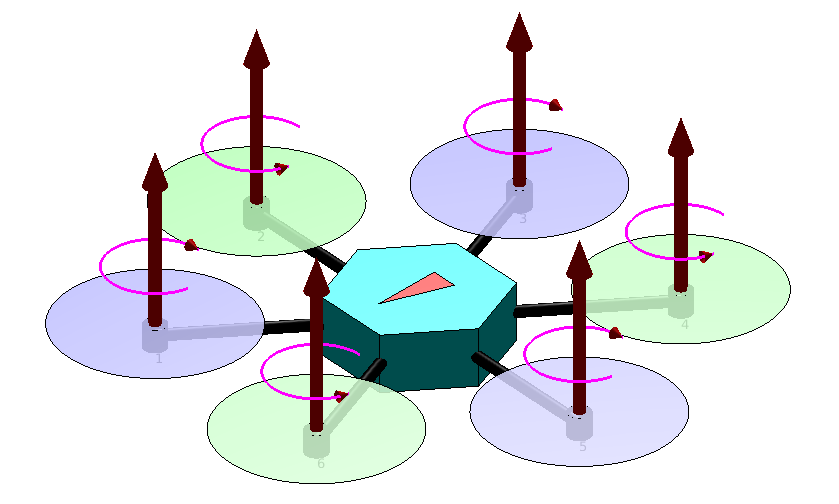}
            \caption{~}
            \label{fig:background:control:coplanar-designs-a}
        \end{subfigure}
        \hfill
        \begin{subfigure}[b]{0.45\textwidth}
            \includegraphics[width=\textwidth]{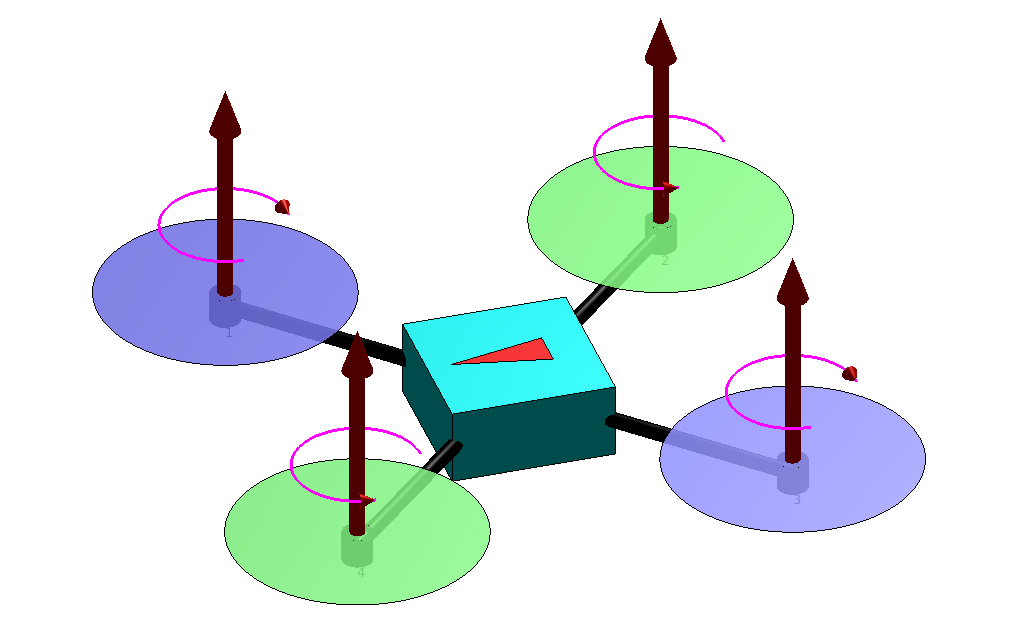}
            \caption{~}
            \label{fig:background:control:coplanar-designs-b}
        \end{subfigure}
            
        \caption[Multirotor designs with fixed co-planar rotors]{In multirotors with fixed co-planar rotors, all the rotors generate thrust in the same direction, and the whole multirotor needs to tilt to generate the desired thrust vector and counteract gravity. (a) A planar hexarotor design. (b) A quadrotor design. }

        \label{fig:background:control:coplanar-designs}
    \end{figure}

In traditional multirotor designs (e.g., quadrotors), the thrusts generated from all rotors are in the same direction, which is the opposite of the gravity vector in the default attitude (see Figure~\ref{fig:background:control:coplanar-designs}). Therefore, the total generated thrust in these robots (called \textit{fixed co-planar} multirotors) is the scalar sum of all the individual rotor thrusts (ignoring some minor aerodynamic effects). This co-planar rotor arrangement makes this design very efficient in compensating for gravity and optimizing energy consumption to generate motion. 

However, fixed co-planar multirotor designs can only generate thrust in a single direction with respect to their body, requiring them to tilt the whole body towards the direction of the desired motion to generate the thrust for the desired acceleration and compensate for gravity. Therefore, these designs cannot fly at a given desired attitude, and their attitude depends on their motion, making them underactuated.

The fixed co-planar multirotors are suitable for many real-world applications, as the precise control of the attitude is not required. Many other applications, such as mapping and cinematography, have found workarounds for the underactuation by adding payloads such as gimbals to allow controlling the camera's attitude separately from the multirotor's attitude~\cite{8968163}. However, good gimbals are expensive and heavy. Moreover, the devices attached to the gimbal have to be controlled to some degree separately from the robot itself~\cite{Keipour:2021:arxiv:mbzirc}. 

As another solution, for physical interactions with the world around the robot, many researchers have been adding manipulator arms with at least two degrees of freedom to achieve the end-effector's complete position and attitude control~\cite{BonyanKhamseh2018, Ruggiero2018}. The same issues with the gimbal also apply here, i.e., the manipulator arm adds the payload weight, increases the robot's cost, and reduces the system's stability. Besides, the control solutions to physical interaction problems are generally very complex and sub-optimal~\cite{Yuksel2014a}.

On the other hand, there are applications where controlling the attitude of the whole robot is required, and there are no good solutions for the problems caused by underactuation. A practical example is a multirotor trying to navigate a disaster site with a very cluttered environment, and some passes narrower than the robot's width.

In recent years, many new designs have been proposed to eliminate the underactuation in multirotors. These designs either control the direction of rotors during the flight (known as \textit{variable-pitch designs}) or set fixed non-zero rotations for the rotors in a way that they are not collinear anymore (known as \textit{fixed-pitch designs})~\cite{Rashad2020, Franchi2019a}. 

The fixed-pitch designs can be made from the existing co-planar multirotors by tilting their rotor arms in different directions, making them easy to construct. To achieve fully independent control over the orientation and position, a robot needs to have at least six rotors, and the arrangement of the rotors should make the allocation matrix full-rank. These designs are classified as \textit{fully-actuated multirotors} as they have full actuation around the hovering point. 

An issue with fixed-pitch designs is the inefficiency due to the different rotor thrusts' directions, causing some portion of the generated forces by different rotors to be used for counteracting each other. Besides, the aerodynamic effects caused by the rotors slightly pointing towards each other further reduce the system's overall efficiency. Finally, these designs can only provide a limited amount of thrust parallel to the ground plane, limiting the maximum horizontal acceleration and the maximum feasible attitude. 

\begin{figure}[!htb]
        \centering
        \begin{subfigure}[b]{0.24\linewidth}
            \includegraphics[width=\textwidth, height=2.5cm]{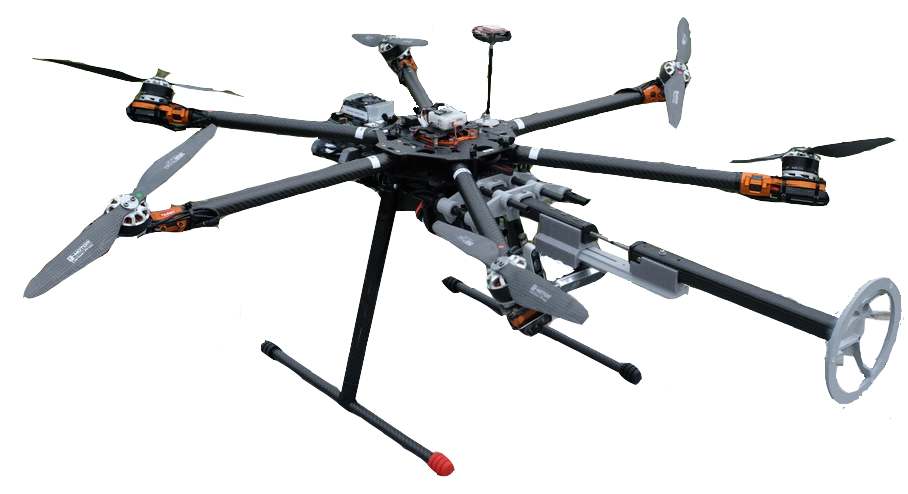}
            \caption{~}
            \label{fig:background:control:fixed-pitch-designs-a}
        \end{subfigure}
        \hfill
        \begin{subfigure}[b]{0.24\linewidth}
            \includegraphics[width=\textwidth, height=2.5cm]{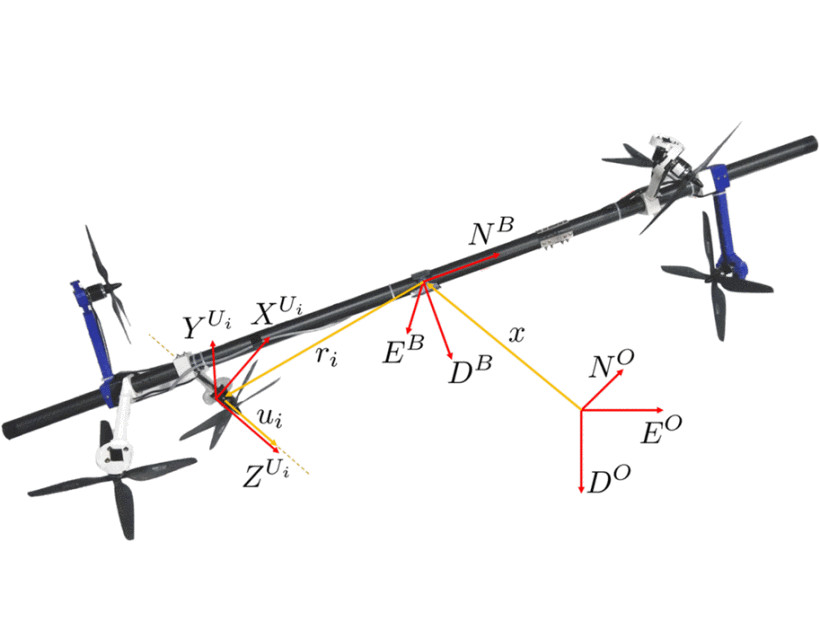}
            \caption{~}
            \label{fig:background:control:fixed-pitch-designs-b}
        \end{subfigure}
        \hfill
        \begin{subfigure}[b]{0.24\linewidth}
            \includegraphics[width=\textwidth, height=2.5cm]{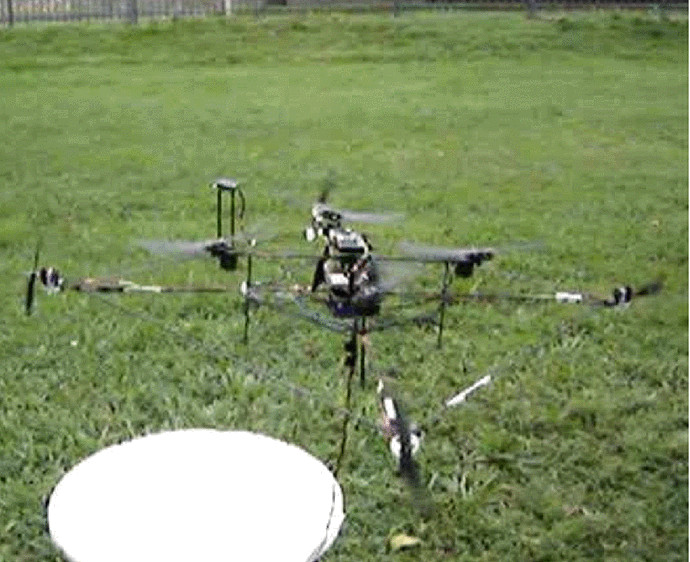}
            \caption{~}
            \label{fig:background:control:fixed-pitch-designs-c}
        \end{subfigure}
        \hfill
        \begin{subfigure}[b]{0.24\linewidth}
            \includegraphics[width=\textwidth, height=2.5cm]{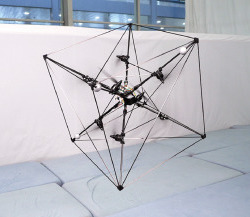}
            \caption{~}
            \label{fig:background:control:fixed-pitch-designs-d}
        \end{subfigure}
            
        \caption[Examples of fixed-pitch fully-actuated multirotor designs]{Some examples of fixed-pitch fully-actuated multirotor designs in the literature. (a)~\cite{Keipour:2020:arxiv:integration} (b)~\cite{Park2016} (c)~\cite{Romero2007} (d)~\cite{Brescianini2018a}.}

        \label{fig:background:control:fixed-pitch-designs}

    \end{figure}

Figure~\ref{fig:background:control:fixed-pitch-designs} shows some fixed-pitch architectures proposed in the literature.

The variable-pitch designs add additional servo motors to control the rotors' direction during the flight. Depending on the design, each servo is responsible for controlling the direction of all rotors together, some of them or just a single rotor. The total number of rotors and servos needs to be at least six for these multirotors to achieve total control over both the orientation and the position. Some of these designs keep the rotors collinear to maximize the efficiency of the generated thrust and simplify the design. 

While most of the variable-pitch designs are fully-actuated, some are \textit{omni-directional}, meaning that they can achieve any desired position and orientation. In addition to the extra servo motors causing a lower flight time, the main issue with the variable-pitch designs is the complexity of both the hardware and the controller for these multirotors, making them expensive and requiring extensive hardware and controller debugging and tuning efforts. 

\begin{figure}[!htb]
        \centering
        \begin{subfigure}[b]{0.24\textwidth}
            \includegraphics[width=\textwidth, height=2.5cm]{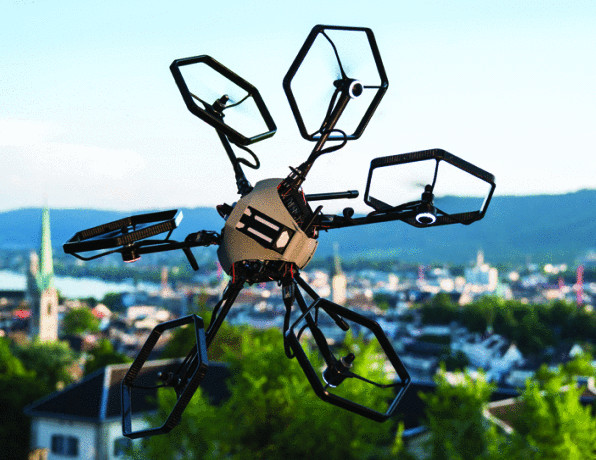}
            \caption{~}
            \label{fig:background:control:variable-pitch-designs-a}
        \end{subfigure}
        \hfill
        \begin{subfigure}[b]{0.24\textwidth}
            \includegraphics[width=\textwidth, height=2.5cm]{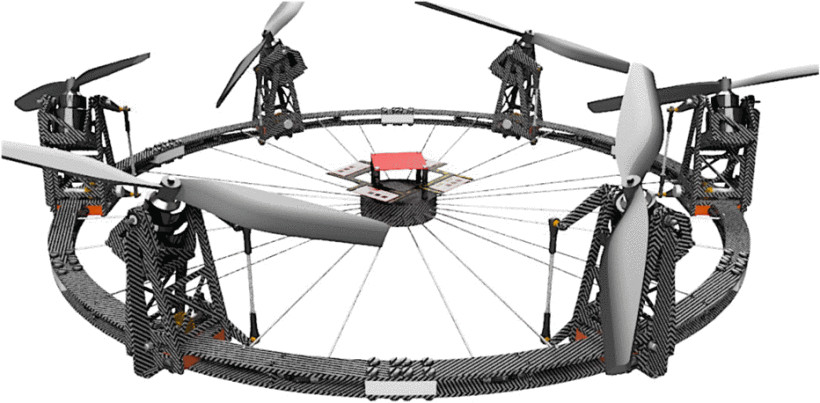}
            \caption{~}
            \label{fig:background:control:variable-pitch-designs-b}
        \end{subfigure}
        \hfill
        \begin{subfigure}[b]{0.24\textwidth}
            \includegraphics[width=\textwidth, height=2.5cm]{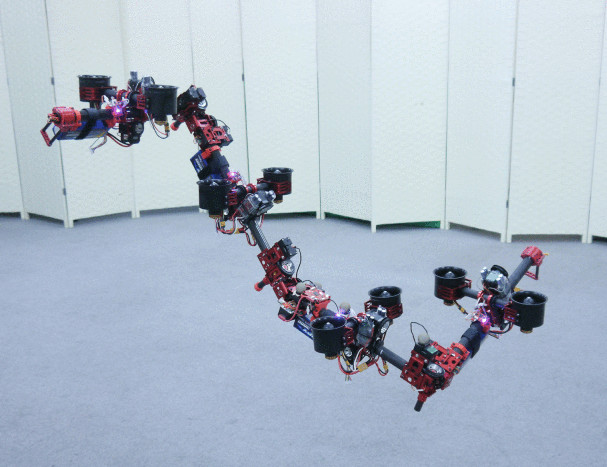}
            \caption{~}
            \label{fig:background:control:variable-pitch-designs-c}
        \end{subfigure}
        \hfill
        \begin{subfigure}[b]{0.24\textwidth}
            \includegraphics[width=\textwidth, height=2.5cm]{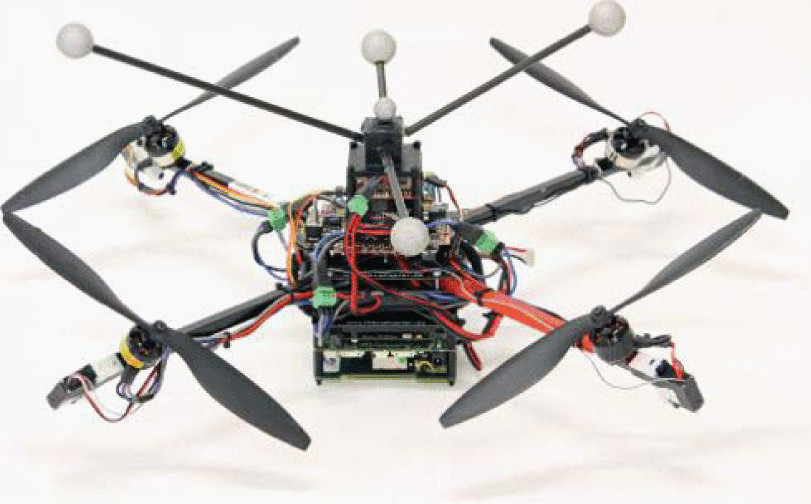}
            \caption{~}
            \label{fig:background:control:variable-pitch-designs-d}
        \end{subfigure}
            
        \caption[Examples of variable-pitch fully-actuated multirotor designs]{Some examples of variable-pitch fully-actuated multirotor designs in the literature. (a)~\cite{Kamel2018a} (b)~\cite{Ryll2016} (c)~\cite{Zhao2018} (d)~\cite{Ryll2015}.}

        \label{fig:background:control:variable-pitch-designs}

    \end{figure}

Figure~\ref{fig:background:control:variable-pitch-designs} shows some sample variable-pitch multirotor designs proposed in the literature.

\section{Modeling and Structure} \label{sec:background:control:modeling}

\begin{figure}[!htb]
        \centering
        \begin{subfigure}[b]{0.49\textwidth}
            \includegraphics[width=\textwidth]{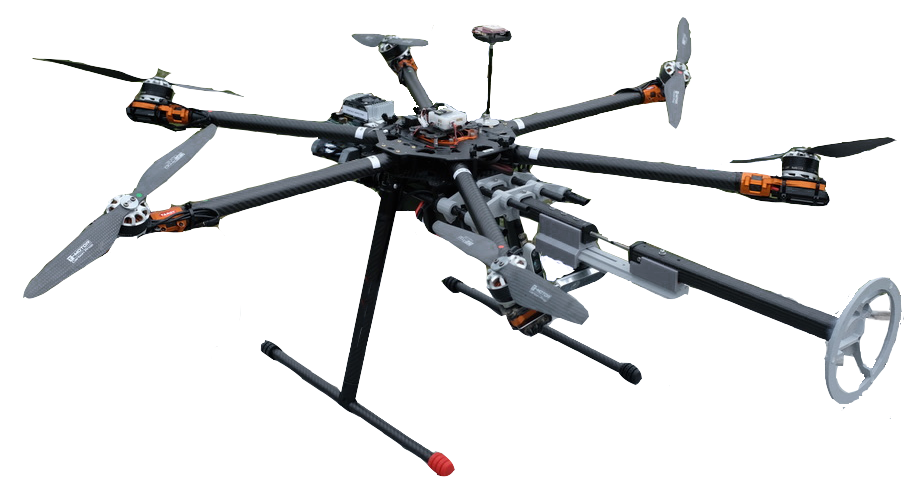}
            \caption{~}
            \label{fig:background:control:our-fixed-pitch-a}
        \end{subfigure}
        \hfill
        \begin{subfigure}[b]{0.45\textwidth}
            \includegraphics[width=\textwidth]{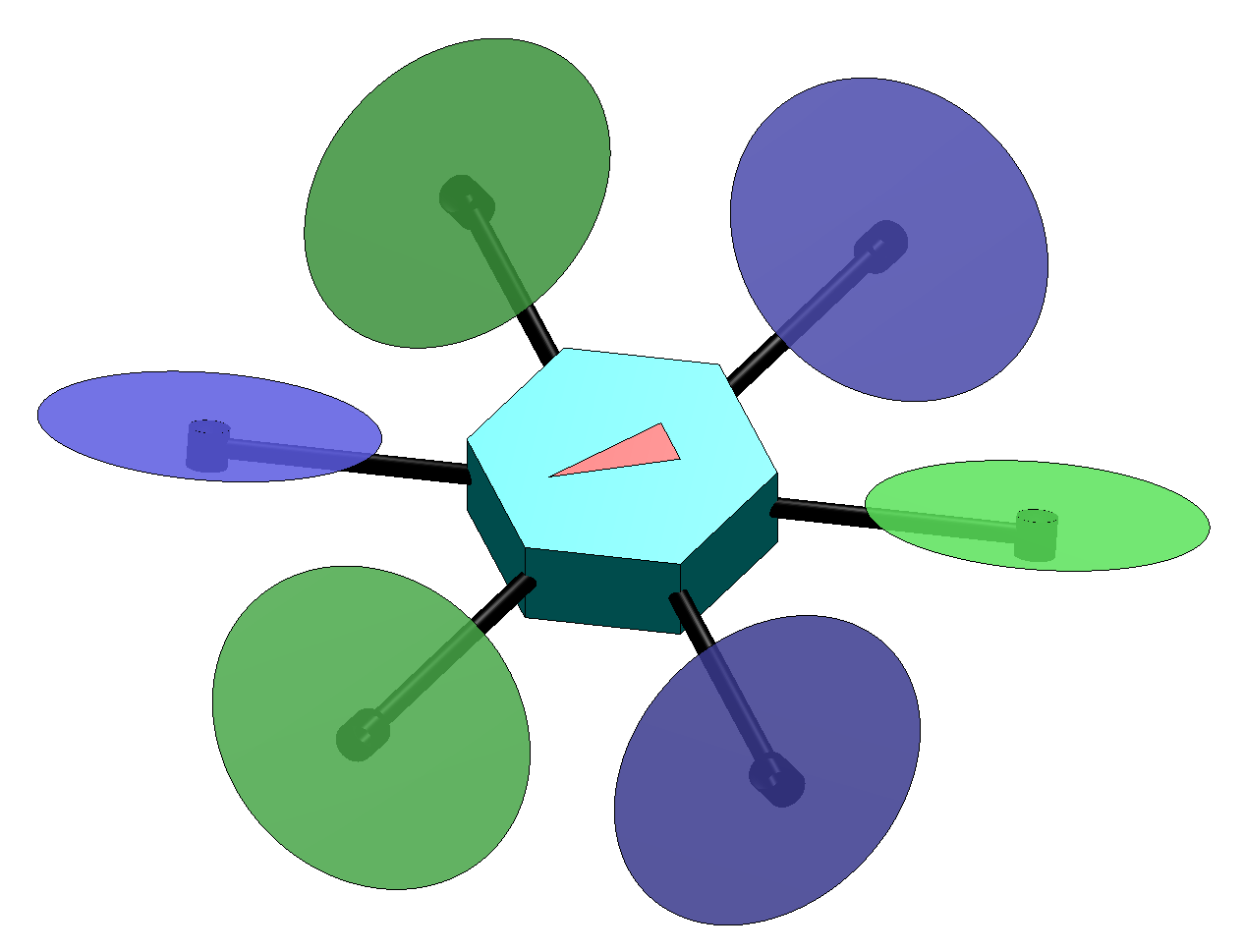}
            \caption{~}
            \label{fig:background:control:our-fixed-pitch-b}
        \end{subfigure}
            
        \caption[A fully-actuated fixed-pitch hexarotor used in this work]{A fully-actuated fixed-pitch hexarotor design used in this work. The shown robot has all its rotors tilted 30 degrees to the side, whereas every other rotor is tilted to the opposite side.}
        
        \label{fig:background:control:our-fixed-pitch}
    \end{figure}

We use the fixed-pitch fully-actuated designs in the current dissertation work, where the rotors can have fixed arbitrary orientations for different tasks. Figure~\ref{fig:background:control:our-fixed-pitch} shows a hexarotor design we have been using for some of our experiments. The fixed-pitch design allows us to develop faster and focus better on the physical interaction challenges. Furthermore, all the methods in this work can also be extended to most of the other fully-actuated and omni-directional robot designs. In contrast, the opposite may not be possible, meaning that the methods developed for more complex omni-directional robots may not be able to get easily adapted to the other types of fully-actuated multirotors. Finally, in many variable-pitch designs, changing the rotor directions is too slow for a suitable reaction to the external disturbances in the applications targeted in this work, and the airflow between the lateral and vertical rotors creates nonlinear dynamics, which is hard to deal with~\citep{Rajappa2015}.

The fixed-pitch model described in this chapter uses a set of assumptions described below. In deriving our fixed-pitch model, we have got our inspiration from~\cite{Mehmood2017} and~\cite{Rajappa2015}. The complete set of the symbols and notations used in this document can be found in Appendix~\ref{appendix:notation}.

\subsection{Assumptions} \label{sec:background:control:modeling-assumptions}

We have made some assumptions to simplify the robot's modeling and control in this work. Many assumptions only affect the low-level controller and the control allocation matrix and do not affect the concepts proposed in other sections. Additionally, some of these assumptions will be lifted later for physical interaction. Therefore, this thesis's work remains valid even with different assumptions for the underlying low-level controller. In the development of the fixed-pitch fully-actuated multirotor, we assume that:

\begin{itemize}
    \item The multirotor is a rigid body.
    \item The geometric center of the multirotor coincides with its center of mass (CoM).
    \item The geometric center of each rotor is the center of spinning and coincides with its actuating motor's center of mass (CoM).
    \item Earth is a perfect homogeneous sphere, so the gravitational vector is directed precisely towards its center. 
    \item The flight time frames and the distances are small enough to assume the Earth as flat and non-rotating. This assumption allows using the North-East-Down (NED) reference frame as an inertial reference frame.
    \item The actuating motors accept rotor angular speeds as inputs (this can be achieved by another low-level controller implemented for the rotors).
    \item The actuating motors have an almost ideal (instantaneous) reaction to the given command with negligible transient time.
    \item The gyroscopic and inertial effects caused by the motors and propellers are small enough to be ignored.
    \item The blade flapping and the rotor-induced drag reactions can be ignored.
    \item The aerodynamic effects resulting from the tilted propellers are small enough to be ignored.
    \item The aerodynamic effects such as the wall, ceiling, and ground effects can be ignored. 
    \item The rotors' arrangement results in a full-ranked allocation matrix (naturally resulting that the robot must have at least six rotors).
\end{itemize}

\section{Translational Kinematics}  \label{sec:background:control:translational-kinematics}

Using the assumptions stated in Section \ref{sec:background:control:modeling-assumptions}, defining three sets of reference frames is enough for modeling the kinematics of a fixed-pitch fully-actuated multirotor: an inertial frame $\FI$, a body-fixed frame $\FB$ and the rotor references $\FR{i}$ ($i=1, 2, \dots ,n_r$), where $n_r$ is the number of rotors in the multirotor. This section defines these reference frames and provides the necessary transformations between them.

\subsection{Inertial Frame}

An inertial frame is a fixed frame defined as $\FI = \{ \OI, \XI, \YI, \ZI \}$, where $\OI$ is an arbitrarily chosen origin, and $\XI$, $\YI$, $\ZI$ are the three right-handed orthogonal axes. For simplicity, using the flat and non-rotating Earth assumptions, we can define the origin as a local point on the ground (e.g., the take-off point). Additionally, it is common to choose the axes in a way that the $\XI$ axis aligns with the North direction, the $\YI$ axis aligns with the east direction, and the $\ZI$ axis is orthogonal to the other axes and points down (parallel to the gravity vector, towards the center of Earth). This reference frame is commonly referred to as the \textit{North-East-Down} (\textit{NED}) frame.

\subsection{Body-Fixed Frame}

The body-fixed frame is a frame fixed to the vehicle and is defined as $\FB = \{ \OB, \XB, \YB, \ZB \}$. $\OB$ is the frame's origin, which coincides with the center of mass (CoM). The $\XB$ axis points to the front of the vehicle, $\YB$ points to the right side, and $\ZB$ is in the direction of the bottom of the UAV, perpendicular to the other two axes. The origin $\OB$ in $\FI$ is called the \textit{position} of the vehicle and can be defined as $\vec{p}^\frm{I} = \matrice{x & y & z}\T \in \numset{R}{3}{1}$.

The rotation from $\FI$ to $\FB$ is performed in three steps: a rotation of $\psi$ (known as \textit{yaw}) around the $\ZI$ axis, then a rotation of $\theta$ (known as \textit{pitch}) around the resulting $\YB$ axis and finally a rotation of $\phi$ (known as \textit{roll}) around the resulting $\XB$ axis. This sequence of rotations is commonly known as $\axis{Z}-\axis{Y}'-\axis{X}''$ or "$3-2-1$" rotation sequence and the resulting roll, pitch and yaw angles are known as \textit{Tait-Bryan} or \textit{Euler-Cardan} angles. We call $\mat{\Phi} = \matrice{\phi & \theta & \psi}\T$ as the \textit{attitude} of the multirotor.

Using $\cosine{(\cdot)}$ as $\cos{(\cdot)}$ and $\sine{(\cdot)}$ as $\sin{(\cdot)}$, the rotation matrix $\RBI \in  SO(3)$ from $\FI$ to $\FB$ can be calculated from the above sequence as:
    
    \begin{equation} \label{eq:background:control:rotation-matrix-bi-zxy}
        \RBI = \begin{bmatrix} 
            1 & 0 & 0 \\
            0 & \cphi & \sphi \\
            0 & -\sphi & \cphi
        \end{bmatrix}
        \begin{bmatrix} 
            \ctheta & 0 & -\stheta \\
            0 & 1 & 0 \\
            \stheta & 0 & \ctheta
        \end{bmatrix}
        \begin{bmatrix} 
            \cpsi & \spsi & 0 \\
            -\spsi & \cpsi & 0 \\
            0 & 0 & 1
        \end{bmatrix}.
    \end{equation}

Simplifying Equation~\ref{eq:background:control:rotation-matrix-bi-zxy}, we will have:

    \begin{equation} \label{eq:background:control:rotation-matrix-bi-full}
        \RIB = (\RBI)\T = \begin{bmatrix} 
            \ctheta\cpsi & \sphi\stheta\cpsi - \cphi\spsi & \cphi\stheta\cpsi + \sphi\spsi\\
            \ctheta\spsi & \sphi\stheta\spsi + \cphi\cpsi & \cphi\stheta\spsi - \sphi\cpsi\\
            -\stheta & \sphi\ctheta & \cphi\ctheta
        \end{bmatrix}.
    \end{equation}

We define the range of the yaw angle $\psi$ to be $-\pi < \psi \leq \pi$. In order to avoid a singularity, both pitch and roll angles are limited to the angles where the UAV is tilted less than $90^\circ$, therefore we have $-\frac{\pi}{2} < \theta < \frac{\pi}{2}$ and $-\frac{\pi}{2} < \phi < \frac{\pi}{2}$. Note that we can avoid this singularity by using other rotation representations (e.g., quaternions) \citep{Diebel2006}. However, there is no practical situation in this project where this singularity can happen, and the Euler angles representation can be devised safely. 

\subsection{Rotor Frames}

The rotor frames are the frames fixed to the rotors and are defined as $\FR{i} = \{ \OR{i}, \XR{i}, \YR{i}, \ZR{i} \}$ ($i=1, 2, \dots ,n_r$). $\OR{i}$ is the center of spinning of the $\nth{i}$ rotor. The $\YR{i}$ axis points towards the vehicle's CoM (from $\OR{i}$ to $\OB$), $\ZR{i}$ is aligned with the axis of rotation of rotor $i$ pointing to the vehicle's bottom direction, and $\XR{i}$ is orthogonal to both and its direction can be obtained from the right-hand rule. The vector from $\OB$ to $\OR{i}$ in $\FB$ can be defined as $\vec{r}_i^\frm{B} = \matrice{r_{i_x} & r_{i_y} & r_{i_z}}\T \in \numset{R}{3}{1}$ ($i = 1, 2, \dots, n_r)$.

\begin{figure}[!htb]
\centering
\includegraphics[width=0.8\linewidth]{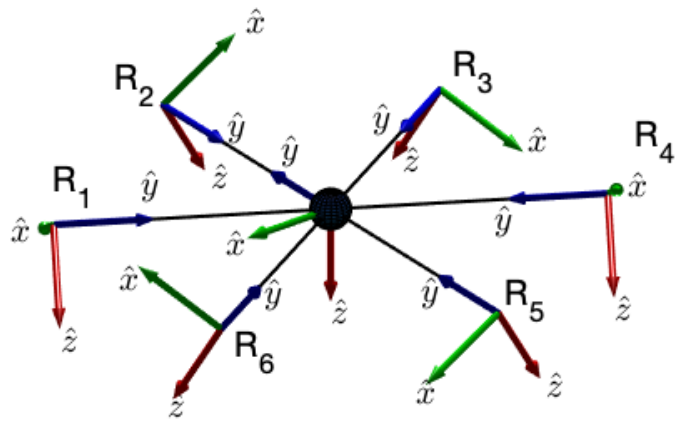}
\caption[The frames defined for the fixed-pitch multirotor]{An illustration of the body-fixed and rotor frames for a fixed-pitch fully-actuated hexarotor.}
\label{fig:background:control:fixed-pitch-frames}
\end{figure}

Figure~\ref{fig:background:control:fixed-pitch-frames} illustrates the frames defined for a fixed-pitch multirotor. 

Assuming that the origin of each rotor $i$ in the body-fixed frame is shown as $\ORB{i}$, the distance $\ell_i$ of each rotor from the vehicle's center of mass can be calculated as $\ell_i = |\ORB{i}| = |\vec{r}_i|$. 
Let us call the vectors from the center of the vehicle to the origin of rotor $i$ as $\vec{r}_i$. Projecting these vectors onto the $\XB\YB$ plane, the angle between each rotor $i$ with the next rotor on the $\XB\YB$ plane in a clockwise direction around $\ZB$ (which is pointing down) can be named $\alpha_i$. For example, the angle between rotors $1$ and $2$ will be called $\alpha_1$. 
In addition, the angle between the projection of $\vec{r}_i$ on $\XB\YB$ plane and $\vec{r}_i$ defines the \textit{dihedral} angle of the rotor $i$'s vector and is named $\phi_{dih_i}$. 

The rotation from $\FB$ to $\FR{i}$ has one additional step compared to the $\RBI$ rotation and is performed using a sequence of four rotations: a $90^\circ$ rotation around the $\ZB$ axis in the positive direction so that an imaginary rotor's frame in front of the vehicle on the $\XB\YB$ plane would have its $Y$ axis pointing towards the vehicle's CoM, next rotation of $\mu_i$ around the $\ZB$ axis to align the $\nth{i}$ rotor's frame to have the projection of its $\axis{Y}$ axis on the $\XB\YB$ plane pointing towards the vehicle's CoM, then a rotation of $\phix{i}$ (called the \textit{inward angle}) around the $\axis{X}$ axis of the new frame, and finally, a rotation of $\phiy{i}$ around the new $\axis{Y}$ axis (called the \textit{sideward angle}). 

Using $\cosine{(\cdot)}$ as $\cos{(\cdot)}$ and $\sine{(\cdot)}$ as $\sin{(\cdot)}$, the rotation matrix $\RRB{i} \in  SO(3)$ from $\FB$ to $\FR{i}$ can be calculated from the above sequence as:

   \begin{equation} \label{eq:background:control:rotation-matrix-rb-zxy}
        \RRB{i} =
        \begin{bmatrix} 
            \cphiy{i} & 0 & -\sphiy{i} \\
            0 & 1 & 0 \\
            \sphiy{i} & 0 & \cphiy{i}
        \end{bmatrix}
        \begin{bmatrix}
            1 & 0 & 0 \\
            0 & \cphix{i} & \sphix{i} \\
            0 & -\sphix{i} & \cphix{i}
        \end{bmatrix}
        \begin{bmatrix} 
            \cmu{i} & \smu{i} & 0 \\
            -\smu{i} & \cmu{i} & 0 \\
            0 & 0 & 1
        \end{bmatrix}
        \begin{bmatrix} 
            0 & 1 & 0 \\
            -1 & 0 & 0 \\
            0 & 0 & 1
        \end{bmatrix}
    \end{equation}

Simplifying Equation~\ref{eq:background:control:rotation-matrix-rb-zxy} and considering that $\RBR{i} = (\RRB{i})\T$, we will have:

    \begin{equation} \label{eq:background:control:rotation-matrix-rb-full}
        \RBR{i} = \begin{bmatrix} 
            - \smu{i} \cphiy{i} - \cmu{i} \sphix{i} \sphiy{i} & - \cmu{i} \cphix{i} & \cmu{i} \sphix{i} \cphiy{i} - \smu{i} \sphiy{i} \\
            \cmu{i} \cphiy{i} - \smu{i} \sphix{i} \sphiy{i} & - \smu{i} \cphix{i} & \cmu{i} \sphiy{i} + \smu{i} \sphix{i} \cphiy{i} \\
            - \cphix{i} \sphiy{i} & \sphix{i} & \cphix{i} \cphiy{i}
        \end{bmatrix}.
    \end{equation}

\section{Rotational Kinematics} \label{sec:background:control:rotational-kinematics}

The angular velocity of the multirotor $\mat{\omega} = \matrice{p & q & r}\T$ is defined as the angular rate in the body-fixed frame. Therefore, $p$, $q$ and $r$ are the angular rates around the $\XB$, $\YB$ and $\ZB$ axes, respectively. As defined in Section~\ref{sec:background:control:translational-kinematics}, the roll angle $\phi$, the pitch angle $\theta$ and the yaw angle $\psi$ are defined in different frames than $p$, $q$ and $r$. The relationship between these variables can be obtained from the rotations between their respective frames \citep{Beard2008}. Starting with $\Dot{\phi}$, $\Dot{\theta}$ and $\Dot{\psi}$ angular rates, and using $\cosine{(\cdot)}$ as $\cos{(\cdot)}$ and $\sine{(\cdot)}$ as $\sin{(\cdot)}$, we have:

    \begin{equation} \label{eq:background:control:euler-to-omega-conversion}
    \begin{split}
        \begin{bmatrix} p \\ q \\ r \end{bmatrix} &= 
        \begin{bmatrix} \Dot{\phi} \\ 0 \\ 0 \end{bmatrix} +
        \begin{bmatrix} 
            1 & 0 & 0 \\
            0 & \cphi & \sphi \\
            0 & -\sphi & \cphi
        \end{bmatrix}
        \begin{bmatrix} 0 \\ \Dot{\theta} \\ 0 \end{bmatrix} +
        \begin{bmatrix} 
            1 & 0 & 0 \\
            0 & \cphi & \sphi \\
            0 & -\sphi & \cphi
        \end{bmatrix}
        \begin{bmatrix} 
            \ctheta & 0 & -\stheta \\
            0 & 1 & 0 \\
            \stheta & 0 & \ctheta
        \end{bmatrix}
        \begin{bmatrix} 0 \\ 0 \\ \Dot{\psi} \end{bmatrix} \\
        &= \begin{bmatrix}
            1 & 0 & -\stheta \\
            0 & \cphi & \sphi \ctheta \\
            0 & -\sphi & \cphi \ctheta
        \end{bmatrix}
        \begin{bmatrix} \Dot{\phi} \\ \Dot{\theta} \\ \Dot{\psi} \end{bmatrix}
    \end{split}
    \end{equation}

From inverting Equation~\ref{eq:background:control:euler-to-omega-conversion} we get:

    \begin{equation} \label{eq:background:control:omega-to-euler-conversion}
        \begin{bmatrix} \Dot{\phi} \\ \Dot{\theta} \\ \Dot{\psi} \end{bmatrix}
        = \begin{bmatrix}
            1 & 0 & -\stheta \\
            0 & \cphi & \sphi \ctheta \\
            0 & -\sphi & \cphi \ctheta
        \end{bmatrix}^{-1}
        \begin{bmatrix} p \\ q \\ r \end{bmatrix}
        = \begin{bmatrix}
            1 & \sin{\phi} \tan{\theta} & \cos{\phi}  \tan{\theta} \\
            0 & \cos{\phi} & -\sin{\phi} \\
            0 & -\sin{\phi} \sec{\theta} & \cos{\phi} \sec{\theta}
        \end{bmatrix}
        \begin{bmatrix} p \\ q \\ r \end{bmatrix}
    \end{equation}

\section{Dynamics} \label{sec:background:control:dynamics}

Assuming that the multirotor is a rigid body, we can use the Newton-Euler formalism to derive the equations of motion for the vehicle dynamics:

    \begin{equation} \label{eq:background:control:neuton-euler-formalism}
    \begin{split}
        &\vec{F}^\frm{I} = m\ddvec{p}^\frm{I} \\
        &\vec{M}^\frm{B} = \InertB \dvec{\omega} + \vec{\omega} \times \InertB \vec{\omega},
    \end{split}
    \end{equation}

\noindent where $\vec{F}^\frm{I} \in \numset{R}{3}{1}$ is the total force vector in $\unit{N}$ applied to robot's center of mass, $\vec{M}^\frm{B} \in \numset{R}{3}{1}$ is the total moment vector in $\unit{N m}$, $\omega \in \numset{R}{3}{1}$ is the body angular velocity vector in $\unit{rad / s}$, $\InertB \in \numset{R}{3}{3}$ is the body-frame inertia tensor in $\unit{N m s^2}$, and $m$ is the total mass of the vehicle measured in $\unit{kg}$.

This section defines the forces and moments acting on the system and derives a model that can be used to control an omni-directional multirotor.

\subsection{Forces}

The two significant forces acting on the vehicle are the \textit{gravity force} and the \textit{thrust force} which is the result of the spinning rotors. Several other less significant forces are applied to the vehicle that can be ignored from the model. For example, the friction between the moving multirotor and air (i.e., \textit{drag force}) is small enough in the low speeds we have in our application and can be considered a disturbance.

Assuming a gravitational acceleration \textit{g} pointing towards the center of Earth, the gravity force of the vehicle will be in the direction of the $\ZI$ axis. Therefore, the total gravity force vector acting on the multirotor can be defined as:

    \begin{equation} \label{eq:background:control:forces-gravity}
        \vec{F}_{grav}^\frm{I} = m g \ZI = 
        \begin{bmatrix} 0 \\ 0 \\ m g \end{bmatrix}
    \end{equation}

Ideally, the thrust $\vec{F}_{{thr}_i}$ generated by the $\nth{i}$ rotor is aligned with the negative direction of the $\ZR{i}$ axis. Ignoring some less significant effects, the magnitude of the generated thrust $F_{{thr}}$ from a spinning rotor can be approximated as $F_{thr} = c_F \Omega^2$, where $c_F > 0$ is a rotor-specific thrust constant in $\unit{N s^2}$ and $\Omega$ is the rotational velocity of the rotor in $\unit{rad / s}$. Therefore, the generated thrust by the $\nth{i}$ rotor in the body-fixed frame can be obtained as:

    \begin{equation} \label{eq:background:control:forces-thrust}
        \vec{F}_{{thr}_i}^\frm{B} = \RBR{i} \left( -F_{{thr}_i} \ZR{i} \right) = 
        \RBR{i} \begin{bmatrix} 0 \\ 0 \\ -c_{Fi} \Omega_i^2 \end{bmatrix}
    \end{equation}

Assuming that the rotors are positioned such that the effect of their airflow on each other's thrust is insignificant, from Equations \ref{eq:background:control:forces-gravity} and \ref{eq:background:control:forces-thrust} we can measure the total force in the inertial frame as:

    \begin{equation} \label{eq:background:control:forces-total}
        \vec{F}^\frm{I} = \vec{F}_{grav}^\frm{I} + \RIB \sum_{i=1}^{n_r} \vec{F}_{{thr}_i}^\frm{B}
        = \begin{bmatrix} 0 \\ 0 \\ m g \end{bmatrix} + 
        \RIB \sum_{i=1}^{n_r} \RBR{i} \begin{bmatrix} 0 \\ 0 \\ -c_{Fi} \Omega_i^2 \end{bmatrix}
    \end{equation}

\subsection{Moments}

The main moments affecting the vehicle are the thrust moment $\vec{M}_{thr}$ acting on the vehicle's CoM (origin) resulting from the rotor thrusts, the reaction moment $\vec{M}_{reac}$ of a spinning rotor acting on the rotor's CoM (origin), and the moment $\vec{M}_{grav}$ acting on the vehicle's CoM resulting from the weights of the individual parts such as legs and rotors. Besides, there are other less significant moments acting on the multirotor, which can be ignored and considered disturbances for the model's simplicity. These moments include the moment resulting from air friction and the drag force acting on the vehicle, the gyroscopic moments of the spinning rotors as rotating masses, and the drag torque of each spinning rotor resulting from the rotor's acceleration.

The moment resulting from a rotor's thrust around the vehicle's body-fixed axes can be calculated as $\vec{M}_{{thr}_{rot}}^\frm{B} = \vec{r} \times \vec{F}_{thr}^\frm{B}$, where $\vec{r}$ is the moment arm in $\unit{m}$ from the vehicle's CoM ($\OB$) to the center of rotor's spinning ($\OR{s}$). Therefore the total moment resulting from the rotor thrusts is:

    \begin{equation} \label{eq:background:control:moments-thrust}
        \vec{M}_{thr}^\frm{B} = \sum_{i=1}^{n_r} \left( \vec{r}_i^\frm{B} \times \vec{F}_{{thr}_i}^\frm{B} \right)
    \end{equation}

Another type of moment is the reaction of the multirotor to a spinning rotor, which is applied to the rotor's center of spinning and has the same magnitude but in the opposite direction of the motor's torque. Similar to the thrust force generated by a spinning rotor $\vec{F}_{thr}$, this reaction moment $\vec{M}_{reac}$ can also be approximated by a quadratic relationship to the rotor speed using $M_{reac} = (-1)^{d} c_\tau \Omega^2$, where $d = 0$ if the rotor is spinning in the positive direction of rotor's $\axis{Z}$ axis (i.e., counter-clockwise around the axis) and $d = 1$ if the rotor is spinning in the negative direction of rotor's $\axis{Z}$ axis. Therefore, the reaction moment of the $\nth{i}$ rotor and the total reaction moment in the body-fixed axes can be calculated as:

    \begin{equation} \label{eq:background:control:moments-reaction-ith}
       \vec{M}_{{reac}_i}^\frm{B} = \RBR{i} \left( M_{{reac}_i} \ZR{i} \right) = 
        \RBR{i} \begin{bmatrix} 0 \\ 0 \\ (-1)^{d_i} c_{\tau i} \Omega_i^2 \end{bmatrix} 
    \end{equation}

    \begin{equation} \label{eq:background:control:moments-reaction}
       \vec{M}_{reac}^\frm{B} = \sum_{i=1}^{n_r} \vec{M}_{{reac}_i}^\frm{B} = 
       \sum_{i=1}^{n_r} \left( \RBR{i} \begin{bmatrix} 0 \\ 0 \\ (-1)^{d_i} c_{\tau i} \Omega_i^2 \end{bmatrix} \right)
    \end{equation}

Finally, the gravity forces of different parts of the multirotor also create a total moment $\vec{M}_{grav}$ around the vehicle's CoM. These moments depend on the structure of the multirotor and can be different for each geometry. For the most common structure of multirotors, where the rotors' legs are extending from the CoM to the rotors, and assuming that $\vec{r}_i$ is the vector connecting the CoM ($\OB$) to the $\nth{i}$ rotor's CoM ($\OR{i}$) and $\vec{r}_{leg_i}$ is the vector connecting the CoM ($\OB$) to the $\nth{i}$ leg's CoM, we have:

    \begin{equation} \label{eq:background:control:moments-gravity}
        \vec{M}_{grav}^\frm{B} = \sum_{i=1}^{n_r} \left[ 
        \left( \vec{r}_i^\frm{B} \times \RBI \begin{bmatrix} 0 \\ 0 \\ m_{rotor_i} g \end{bmatrix} \right) + 
        \left( \vec{r}_{leg_i}^\frm{B} \times \RBI \begin{bmatrix} 0 \\ 0 \\ m_{leg_i} g \end{bmatrix} \right) \right]
    \end{equation}

From the Equations \ref{eq:background:control:moments-thrust}, \ref{eq:background:control:moments-reaction} and \ref{eq:background:control:moments-gravity}, the total moment around the body-fixed axes can be calculated as:

    \begin{equation} \label{eq:background:control:moments-total}
        \vec{M}^\frm{B} = \vec{M}_{thr}^\frm{B} + \vec{M}_{reac}^\frm{B} + \vec{M}_{grav}^\frm{B}
    \end{equation}

\subsection{Equations of Motion}

Let us define the state of the system as $\mat{x} = \matrice{\dvec{p}^\frm{I} & \mat{\Phi} & \omega}\T$. By replacing Equations \ref{eq:background:control:forces-total} and \ref{eq:background:control:moments-total} in Equation~\ref{eq:background:control:neuton-euler-formalism} and by renaming the conversion matrix of Equation~\ref{eq:background:control:omega-to-euler-conversion} to $\mat{\eta}(\mat{\Phi})$, we can get the equations of motion as:

    \begin{equation} \label{eq:background:control:equations-of-motion-linear-acceleration}
    \begin{split}
        &\vec{F}^\frm{I} = \begin{bmatrix} 0 \\ 0 \\ m g \end{bmatrix} + 
        \RIB \sum_{i=1}^{n_r} \RBR{i} \begin{bmatrix} 0 \\ 0 \\ -c_{Fi} \Omega_i^2 \end{bmatrix}
         = m\ddvec{p}^\frm{I} \\
         \Rightarrow \quad
         &\ddvec{p}^\frm{I} = \begin{bmatrix} 0 \\ 0 \\ g \end{bmatrix} + 
         \frac{1}{m} \RIB \sum_{i=1}^{n_r} \RBR{i} \begin{bmatrix} 0 \\ 0 \\ -c_{Fi} \Omega_i^2 \end{bmatrix}
    \end{split}
    \end{equation}

    \begin{equation} \label{eq:background:control:equations-of-motion-angular-acceleration}
        \vec{M}^\frm{B} = \InertB \dvec{\omega} + \vec{\omega} \times \InertB \vec{\omega} \Rightarrow 
        \dvec{\omega} = \InertB^{-1} \left( \vec{M}_{thr}^\frm{B} + \vec{M}_{reac}^\frm{B} + \vec{M}_{grav}^\frm{B} \right) - \InertB^{-1} \left( \vec{\omega} \times \InertB \vec{\omega} \right)
    \end{equation}

    \begin{equation} \label{eq:background:control:equations-of-euler-rate}
        \begin{bmatrix} \Dot{\phi} \\ \Dot{\theta} \\ \Dot{\psi} \end{bmatrix}
        = \underbrace{
        \begin{bmatrix}
            1 & \sin{\phi} \tan{\theta} & \cos{\phi}  \tan{\theta} \\
            0 & \cos{\phi} & -\sin{\phi} \\
            0 & -\sin{\phi} \sec{\theta} & \cos{\phi} \sec{\theta}
        \end{bmatrix}
        }_{\mat{\eta}(\mat{\Phi})}
        \begin{bmatrix} p \\ q \\ r \end{bmatrix}
        \Rightarrow
        \mat{\Dot{\Phi}} = \mat{\eta}(\mat{\Phi}) \cdot \mat{\omega}
    \end{equation}

\section{Control-Affine Model} \label{sec:background:control:control-affine-model}

If the multirotor accepts the rotor rotational velocities $\Omega$ as a control command, we can define the control input $\mat{u}$ as the set of squared rotor speeds $\Omega_i^2$. Then we can rearrange Equations \ref{eq:background:control:forces-total} and \ref{eq:background:control:moments-total} to achieve a control-affine formulation for the system. We have:

    \begin{equation} \label{eq:background:control:control-input-u}
        \mat{u} = \begin{bmatrix} \Omega_1^2 \\ \Omega_2^2 \\ \vdots \\ \Omega_{n_r}^2 \end{bmatrix}
        \in  \numset{R}{n_r}{1}
    \end{equation}
    
From Equations \ref{eq:background:control:rotation-matrix-rb-full} and \ref{eq:background:control:forces-total} we have:

    \begin{equation} 
    \begin{split}
        \vec{F}_{{thr}}^\frm{B} &= \sum_{i=1}^{n_r} \left( \RBR{i} \begin{bmatrix} 0 \\ 0 \\ -c_{Fi} \Omega_i^2 \end{bmatrix} \right)
        = \sum_{i=1}^{n_r} \left( -c_{Fi} \Omega_i^2 
        \begin{bmatrix}
            \left[ \RBR{i} \right]_{13} \\
            \left[ \RBR{i} \right]_{23} \\
            \left[ \RBR{i} \right]_{33}
        \end{bmatrix} \right) \\
        &= \sum_{i=1}^{n_r} \left( -c_{Fi} \Omega_i^2 \begin{bmatrix}
            \cmu{i}\sphiy{i} + \cphiy{i}\smu{i}\sphix{i} \\
            \smu{i}\sphiy{i} - \cmu{i}\cphiy{i}\sphix{i} \\
            \cphix{i}\cphiy{i}
        \end{bmatrix} \right) \\
    \end{split}
    \end{equation}
    
    Expanding the equation further gives:
    
    \begin{equation} \label{eq:background:control:control-affine-thrust-force}
    \begin{split}
        \vec{F}_{{thr}}^\frm{B} 
        &= \underbrace{
        \begin{bmatrix}
            -c_{F1} \left[ \RBR{1} \right]_{13} & -c_{F2} \left[ \RBR{2} \right]_{13} & \cdots & -c_{Fn_r} \left[ \RBR{n_r} \right]_{13} \\
            -c_{F1} \left[ \RBR{1} \right]_{23} & -c_{F2} \left[ \RBR{2} \right]_{23} & \cdots & -c_{Fn_r} \left[ \RBR{n_r} \right]_{23} \\
            -c_{F1} \left[ \RBR{1} \right]_{33} & -c_{F2} \left[ \RBR{2} \right]_{33} & \cdots & -c_{Fn_r} \left[ \RBR{n_r} \right]_{33}
        \end{bmatrix}}_{\mat{L} \in  \numset{R}{3}{n_r}}
        \begin{bmatrix} \Omega_1^2 \\ \Omega_2^2 \\ \vdots \\ \Omega_{n_r}^2 \end{bmatrix} \\
        &= \mat{L} \cdot \mat{u}
    \end{split}
    \end{equation}

    \begin{equation} \label{eq:background:control:control-affine-forces}
        \vec{F}^\frm{I} = \begin{bmatrix} 0 \\ 0 \\ m g \end{bmatrix} + \RIB \vec{F}_{{thr}}^\frm{B} = \begin{bmatrix} 0 \\ 0 \\ m g \end{bmatrix} + \left( \RIB \mat{L} \right) \cdot \mat{u}
    \end{equation}
    
Similar to Equation~\ref{eq:background:control:control-affine-thrust-force}, from Equation~\ref{eq:background:control:moments-reaction} we have:

    \begin{equation} \label{eq:background:control:control-affine-reaction-moment}
    \begin{split}
        \vec{M}_{reac}^\frm{B} 
        &= \sum_{i=1}^{n_r} \left( (-1)^{d_i} c_{\tau i} \Omega_i^2 
        \begin{bmatrix}
            \left[ \RBR{i} \right]_{13} \\
            \left[ \RBR{i} \right]_{23} \\
            \left[ \RBR{i} \right]_{33}
        \end{bmatrix} \right) \\
        &= \sum_{i=1}^{n_r} \left( (-1)^{d_i} c_{\tau i} \Omega_i^2 \begin{bmatrix}
            \cmu{i}\sphiy{i} + \cphiy{i}\smu{i}\sphix{i} \\
            \smu{i}\sphiy{i} - \cmu{i}\cphiy{i}\sphix{i} \\
            \cphix{i}\cphiy{i}
        \end{bmatrix} \right) \\
        &= \underbrace{
        \begin{bmatrix}
            (-1)^{d_1} c_{\tau 1} \left[ \RBR{1} \right]_{13} & \cdots & (-1)^{d n_r} c_{\tau n_r} \left[ \RBR{n_r} \right]_{13} \\
            (-1)^{d_1} c_{\tau 1} \left[ \RBR{1} \right]_{23} & \cdots & (-1)^{d n_r} c_{\tau n_r} \left[ \RBR{n_r} \right]_{23} \\
            (-1)^{d_1} c_{\tau 1} \left[ \RBR{1} \right]_{33} & \cdots & (-1)^{d n_r} c_{\tau n_r} \left[ \RBR{n_r} \right]_{33}
        \end{bmatrix}}_{\mat{G} \in  \numset{R}{3}{n_r}}
        \begin{bmatrix} \Omega_1^2 \\ \Omega_2^2 \\ \vdots \\ \Omega_{n_r}^2 \end{bmatrix}
        = \mat{G} \cdot \mat{u}
    \end{split}
    \end{equation}

Rearranging the Equation~\ref{eq:background:control:moments-thrust}, we have:

    \begin{equation} \label{eq:background:control:control-affine-thrust-moment}
    \begin{split}
        \vec{M}_{thr}^\frm{B} &= \sum_{i=1}^{n_r} \left( \vec{r}_i^\frm{B} \times \vec{F}_{{thr}_i}^\frm{B} \right) 
        = \sum_{i=1}^{n_r} \left( \left(
        \underbrace{
        \vec{r}_i^\frm{B} \times 
        \begin{bmatrix}
            -c_{Fi} \left[ \RBR{i} \right]_{13} \\
            -c_{Fi} \left[ \RBR{i} \right]_{23} \\
            -c_{Fi} \left[ \RBR{i} \right]_{33}
        \end{bmatrix}
        }_{\mat{F}_i \in  \numset{R}{3}{1}} \right) \cdot \Omega_i^2 \right )\\
        &= \underbrace{
        \begin{bmatrix}
            \mat{F}_1 & \cdots & \mat{F}_{n_r}
        \end{bmatrix}}_{\mat{F} \in  \numset{R}{3}{n_r}}
        \begin{bmatrix} \Omega_1^2 \\ \Omega_2^2 \\ \vdots \\ \Omega_{n_r}^2 \end{bmatrix}
        = \mat{F} \cdot \mat{u}
    \end{split}
    \end{equation}

Finally, by replacing Equations \ref{eq:background:control:control-affine-reaction-moment} and \ref{eq:background:control:control-affine-thrust-moment} in Equation~\ref{eq:background:control:moments-total}, we have:

    \begin{equation} \label{eq:background:control:control-affine-moments}
        \vec{M}^\frm{B} = \vec{M}_{grav}^\frm{B} + \underbrace{
        \left( \mat{F} + \mat{G} \right) }_{\mat{M}} \cdot \mat{u}
        = \vec{M}_{grav}^\frm{B} + \mat{M} \cdot \mat{u}
    \end{equation}

\subsection{Control-Affine Equations of Motion}

From Section~\ref{sec:background:control:dynamics} we have the state of the system defined as $\mat{x} = \matrice{\dvec{p}^\frm{I} & \mat{\Phi} & \omega}\T$. From the definition of input command $\mat{u}$ in Equation~\ref{eq:background:control:control-input-u}, and by replacing the control-affine force and moment from Equations~\ref{eq:background:control:control-affine-forces} and \ref{eq:background:control:control-affine-moments} into equations of motion (Equations \ref{eq:background:control:equations-of-motion-linear-acceleration}, \ref{eq:background:control:equations-of-motion-angular-acceleration} and \ref{eq:background:control:equations-of-euler-rate}, we can obtain the systems dynamics $\mat{\Dot{x}}$:

    \begin{equation} \label{eq:background:control:control-affine-equations-of-motion}
        \underbrace{ \begin{bmatrix} \mat{\ddot{p}}^\frm{I} \\ \mat{\dot{\Phi}} \\ \mat{\dot{\omega}} \end{bmatrix}}_{\mat{\dot{x}}}
        = \underbrace{ \begin{bmatrix} 
            \begin{bmatrix} 0 & 0 & g \end{bmatrix}\T \\ 
            \mat{\eta}(\mat{\Phi}) \cdot \mat{\omega} \\ 
            \InertB^{-1} \left( \mat{M}_{grav}^\frm{B} - \left( \mat{\omega} \times \InertB \mat{\omega} \right) \right)  
        \end{bmatrix}}_{\mat{f}(\mat{x})}
        + \underbrace{ \begin{bmatrix} 
            \frac{1}{m} \RIB \mat{L} \\ 
            \mat{0}_{3 \times n_r} \\ 
            \InertB^{-1} \mat{M} 
        \end{bmatrix}}_{\mat{J}(\mat{x})}
        \underbrace{\begin{bmatrix} \Omega_1^2 \\ \Omega_2^2 \\ \vdots \\ \Omega_{n_r}^2 \end{bmatrix}}_{\mat{u}} 
    \end{equation}

\noindent which can be simplified to:

    \begin{equation} \label{eq:background:control:control-affine-equations-of-motion-simplified}
        \mat{\dot{x}} = \mat{f}(\mat{x}) + \mat{J}(\mat{x}) \cdot \mat{u},
    \end{equation}
    
\noindent where $\mat{f}(\mat{x})$ is the \textit{drift vector} due to the gravity and rotational inertia, and $\mat{J}(\mat{x})$ is the \textit{decoupling matrix} mapping the input of the system to the state space.

\section{Controller Design} \label{sec:background:control:control-allocation}

Over the years, many different control methods have been proposed for the conventional underactuated multirotors to control different parameters, such as position, attitude, velocity, accelerations, or precisely track the given trajectories in free flight. Comprehensive reviews of these methods have been published in~\cite{Johansen2013, Sheng2019a, Zulu2014ARO, Kamel2017, Werink2019}.

Many of the methods written initially for underactuated multirotors are already adopted for fully-actuated fixed-pitch and variable-pitch robots. However, due to a broad spectrum of different designs in this category of multirotors, we will only discuss the works that have been proposed or can be easily adapted for the fixed-pitch designs that are the focus of our work.

The most common type of controller in the literature is the \textit{exact feedback linearization and decoupling method} proposed by~\cite{Rajappa2015} which has also been called \textit{nonlinear dynamic inversion} by some authors~\cite{Mehmood2017}. This controller aims to minimize the control effort, and its optimal design parameters depend on the trajectory. The resulting system is linear and suitable for extension to physical interaction applications, which is why we also considered this controller.

Assuming that the goal of our controller is to track the position and the given Euler angles, we can define the output of the system as:

    \begin{equation} \label{eq:background:control:system-output-y}
        \mat{y} = \begin{bmatrix} \mat{p}^\frm{I} \\ \mat{\Phi} \end{bmatrix}
    \end{equation}

By differentiating the output $\mat{y}$ from Equation~\ref{eq:background:control:system-output-y} twice and using the system dynamics equation (Equation~\ref{eq:background:control:control-affine-equations-of-motion}), we can find a relationship between the system input $\mat{u}$ and the output $\mat{y}$ as:

    \begin{equation} \label{eq:background:control:input-to-output-relationship}
        \mat{\ddot{y}} = \begin{bmatrix} \mat{\ddot{p}}^\frm{I} \\ \mat{\ddot{\Phi}} \end{bmatrix}
        = \begin{bmatrix} 
            \begin{bmatrix} 0 & 0 & g \end{bmatrix}\T \\ 
            \mat{\dot{\eta}}(\mat{\Phi}) \mat{\omega} + \mat{\eta}(\mat{\Phi}) \InertB^{-1} \left( \mat{M}_{grav}^\frm{B} - \left( \mat{\omega} \times \InertB \mat{\omega} \right) \right)  
        \end{bmatrix}
        + \begin{bmatrix} 
            \frac{1}{m} \RIB \mat{L} \\ 
            \mat{\eta}(\mat{\Phi})\InertB^{-1} \mat{M} 
        \end{bmatrix}
        \cdot \mat{u}
    \end{equation}
    
From Equation~\ref{eq:background:control:input-to-output-relationship} we can calculate the required input to the system (the controller output) for the desired system output's second derivative (the controller input) as:

    \begin{equation} \label{eq:background:control:output-to-input-relationship}
        \mat{u} =
        \begin{bmatrix} 
            \frac{1}{m} \RIB \mat{L} \\ 
            \mat{\eta}(\mat{\Phi})\InertB^{-1} \mat{M} 
        \end{bmatrix}^{-1}
        \left(
        \mat{\ddot{y}}
        -
        \begin{bmatrix} 
            \begin{bmatrix} 0 & 0 & g \end{bmatrix}\T \\ 
            \mat{\dot{\eta}}(\mat{\Phi}) \mat{\omega} + \mat{\eta}(\mat{\Phi}) \InertB^{-1} \left( \mat{M}_{grav}^\frm{B} - \left( \mat{\omega} \times \InertB \mat{\omega} \right) \right)
        \end{bmatrix}
        \right)
    \end{equation}

Having a controller that can control the position and attitude accelerations, we can use any stabilizing outer controller to control the desired positions and attitudes. The most common is a proportional–integral–derivative (PID) controller used by~\cite{Rajappa2017, Mehmood2017}, and many other researchers. PID has shown to perform well for this task, and its implementation is already available in most traditional controllers, allowing an easier transition to the fully-actuated robots.

The proposed feedback linearization control method assumes that the input $\mat{\ddot{y}}$ can have any value to track a completely 6-DoF independent trajectory. Hence, this controller's limitation is that it does not account for the input saturation, which can result in motion instability when the required inputs for perfect trajectory tracking are not feasible~\cite{bicego:tel-02433940}. \cite{Convens2017} has discussed a solution to this problem by introducing a suitable scheme for systems with linear dynamics and nonlinear state and input constraints. \cite{Franchi2018} specifically takes the saturation into account and addresses the control for fully-actuated vehicles with bounded lateral forces such as the fixed-pitch hexarotor designs we use in this work.

Some other controller methods proposed for the fully-actuated vehicles include the Sliding Mode Control~\cite{Yao2018, Arizaga2019, Nguyen2019}, Nonlinear Model-Predictive Control~\cite{Tadokoro2019}, and Model Reference Adaptive Control~\cite{Sheng2019}.

\chapter{Flexible Control Design for Fully-Actuated Multirotors} \label{ch:control}

The introduction of fully-actuated multirotors has opened the door to new possibilities and more efficient solutions to many real-world applications. However, their integration had been slower than expected, partly due to the need for designing and developing new tools to take full advantage of these robots. 

Despite the recent rise in popularity of fully-actuated robots, many control methods have already been introduced. However, these controllers require full pose 6-D setpoints, forcing the teams working on them to develop full-pose 6-D tools and methods around these controllers to use their robots in real applications. This re-development of the autonomy stack is inefficient, time-consuming, and requires many resources.

We propose a way of bridging the gap between the already available ecosystem for underactuated robots and the new fully-actuated vehicles. This approach can easily extend the existing underactuated flight controllers to support fully-actuated robots or enhance the existing fully-actuated controllers to work with the existing underactuated flight stacks. We introduce attitude strategies that work with the underactuated controllers, tools, planners, and remote control interfaces while taking advantage of the UAV's full actuation. Moreover, thrust strategies are proposed to handle the limited lateral thrust properly, which is a problem with many fully-actuated UAV designs. The strategies are lightweight, simple, and allow rapid integration of the available tools with these new vehicles for the fast development of new real-world applications. 

To show the flexibility of the proposed controller design, we further extend it for physical interaction tasks, making it a hybrid position and force controller. This extension allows precise control over the force and position during contact with the physical world. Such extension provides the ability to physically manipulate the environment during tasks such as inspection and maintenance.

Section~\ref{sec:control:intro} discusses the relevant work and provides an introduction to the rest of the chapter. Section~\ref{sec:control:problem} defines the problem and Section~\ref{sec:control:design} introduces our control design solution in detail. Sections~\ref{sec:control:attitude} and~\ref{sec:control:thrust} explain the attitude and thrust strategies proposed with our controller design. Section~\ref{sec:control:hpfc} extends the control design to a complete force-position controller for controlling physical interactions. Finally, Section~\ref{sec:control:tests} illustrates the controller design's viability and performance with experiments on real and simulated robots for different free-flight and physical interaction tasks.

\section{Introduction and Related Work} \label{sec:control:intro}

The past two decades have seen rapid growth in the number of multirotor applications. Traditional multirotors are designed to have co-planar rotors. Although this design choice is simple and maximizes energy efficiency, these UAVs suffer from underactuation (i.e., their rotational motion is unavoidably coupled with their translational motion).

The first attempts to address the underactuation issue of the multirotors date back to 2007~\cite{Romero2007, Salazar2009}. The authors of these papers proposed a robot with eight rotors: four rotors were used for vertical thrust generation, similar to quadrotors; the other four rotors were used for lateral force generation, helping the robot with lateral displacements. 

The authors of~\cite{Crowther2011} and~\cite{Langkamp2011} presented the first fully-actuated designs with the minimum number of rotors in 2011. They analyzed fully-actuated hexarotor design for fixed-pitch/variable-speed rotors and variable-pitch/fixed-speed rotors. Many other groups followed their lead by replicating the available designs for their research or slightly modifying the hexarotor designs for different optimal criteria~\cite{Kaufman2014, Shimizu2015, Giribet2016, Kotarski2017, Mehmood2017, Lee2018, Lee2018a}.

These new designs allow fully independent control over both linear and angular motions~\cite{Rashad2020, Franchi2019a}. Many other configurations have been proposed to achieve the full actuation while optimizing or simplifying some design or controller aspects. Some examples of these configurations include modifying the quadrotors for full actuation~\cite{Ryll2015}, tetrahedron-shaped hexarotor~\cite{Toratani2012}, omni-directional octocopters~\cite{Brescianini2016,Brescianini2018a}, semi-coaxial hexarotors~\cite{Bershadsky2018}, and hexarotors with additional servos~\cite{Ryll2016,Kamel2018a,Kamel2018}. Some other works have focused on studying the advantages and issues of the new fully-actuated designs for optimal structures~\cite{Das2016,Mehmood2016,Tadokoro2017,Tognon2018,Tadokoro2018a,Tadokoro2018,Schuster2019}, fault tolerance~\cite{Giribet2016,Giribet2018,Giribet2019}, wind tolerance~\cite{Chen2019}, and robustness to aerodynamic effects~\cite{Li2019}.

The fully-actuated robots, in general, have less energy efficiency and generally have more complex hardware designs than underactuated UAVs. However, in many cases, the independent control over all six dimensions greatly simplifies the payload and the approach to a task, ultimately reducing the weight and final costs of the UAV's hardware, software, and development process. Moreover, it makes many new applications possible that were infeasible with underactuated designs. A more comprehensive literature review of fully-actuated multirotors can be found in~\cite{Rashad2020} and~\cite{Franchi2019a}.

\begin{figure}[!htb]
        \centering
        \begin{subfigure}[b]{0.32\textwidth}
            \includegraphics[width=\textwidth, height=3.5cm]{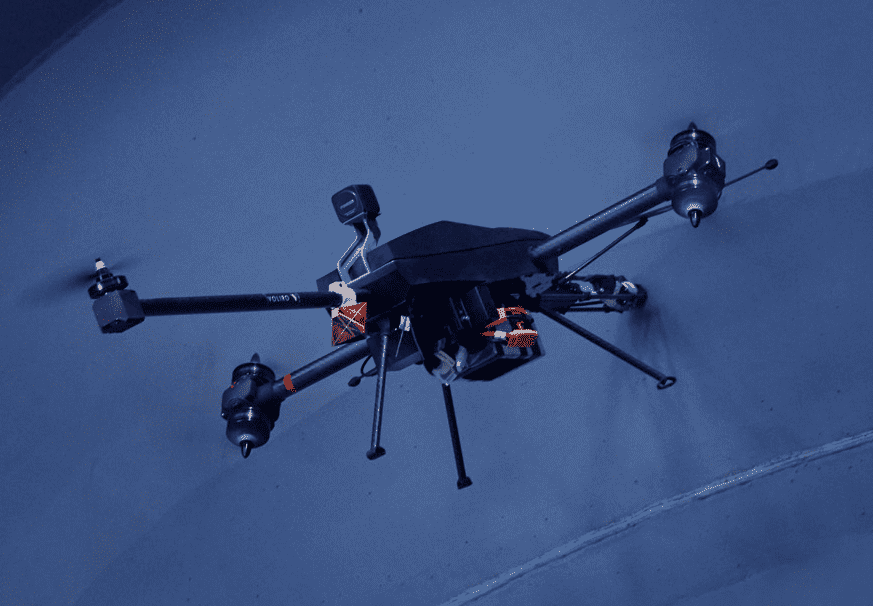}
            \caption{~}
            \label{fig:control:commercial-fully-actuated-a}
        \end{subfigure}
        \hfill
        \begin{subfigure}[b]{0.32\textwidth}
            \includegraphics[width=\textwidth, height=3.5cm]{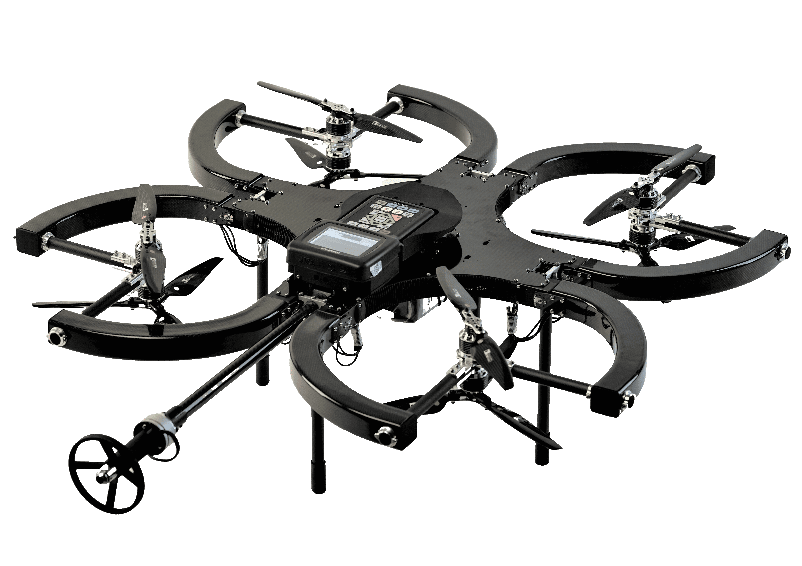}
            \caption{~}
            \label{fig:control:commercial-fully-actuated-b}
        \end{subfigure}
        \hfill
        \begin{subfigure}[b]{0.32\textwidth}
            \includegraphics[width=\textwidth, height=3.5cm]{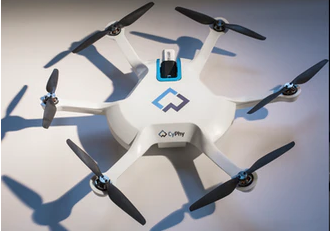}
            \caption{~}
            \label{fig:control:commercial-fully-actuated-c}
        \end{subfigure}

        \caption[Commercial fully-actuated multirotors]{Examples of fully-actuated multirotors being commercialized. (a) Voliro drone for inspection, painting, and assembly. (b) Skygauge drone for inspection. (c) CyPhy drone for hobbyists.}

        \label{fig:control:commercial-fully-actuated}

    \end{figure}

The advantages of fully-actuated multirotors have already encouraged different teams to start developing commercial products. Figure~\ref{fig:control:commercial-fully-actuated} shows some of these designs for different applications. Voliro is an omni-directional multirotor developed by a spin-off from ETH Zurich for applications such as contact inspection and painting~\cite{Voliro2020}; Skygauge is a fully-actuated multirotor specifically designed for non-destructive inspection that is still at the initial development phase~\cite{Skygauge2020}; CyPhy is a failed Kickstarter project by an iRobot cofounder that was targeting the hobbyist community to provide a cheap gimbal-less fixed-tilt hexarotor for stable filming~\cite{CyPhy2018}.

Until recently, aerial robots were only used as contactless remote sensing devices, and the idea of the physical interaction of aerial robots with their environment has only been recently explored. The physical interaction applications are limitless, ranging from contact inspection, maintenance, and assembly to construction, sampling, assistance, delivery, and transportation. The automation of physical interactions not only has the potential to reduce the costs and the danger of many tasks, but it can also make some tasks previously deemed impossible possible, such as exploring disaster sites or operating inside nuclear plants. Hence, governments and large companies have defined large projects to find reliable solutions for these applications. Authors of~\cite{bicego:tel-02433940} list some ongoing and finished projects defined and funded by the European Union.

The physical interaction of aerial robots with the environment can be divided into the following different types:

\begin{enumerate}
    \item \textbf{Grasping and picking:} In this type, the robot's goal is to grasp or pick an external rigid or deformable object using the end-effector. The potential applications for the multirotors can include object retrieval, high-speed courier services, and intelligence gathering.
    The controller for this type of contact generally aims to control the precise pose of the gripper. Depending on the application, it may also have to control the forces and moments exerted by the gripper. 
    
    \item \textbf{Momentary contact:} In this type, the robot's objective is to make brief contact with an external object, usually to hit it at the right time in the right direction. Applications of this contact type can include robots playing ball-based games, such as ping pong and volleyball. The controller generally aims to control the pose and velocity of the end-effector at a precise time.

    \item \textbf{Static contact with rigid surfaces and objects:} This type of contact happens when the robot requires to exert force and torque on a rigid object. The object can be attached to the environment or separated, in which case it can be pushed with a certain amount of force. The most prominent application is non-destructive testing (NDT), which is used for the contact inspection of oil tanks, bridges, and other infrastructure for corrosion, cracks, and other problems. 
    The controllers need to regulate the forces (and/or the moments) applied by the robot to the point of contact based on some contact model. However, precise force control may not be needed depending on the application, and the force can be controlled using passive or active compliance.

    \item \textbf{Moving during contact with rigid surfaces and objects:} In this type, the robot needs to exert forces and torques to the surface while following a trajectory. Many real-world applications include this type of contact, including glass cleaning, wall painting, screw driving, and turning valves and handles.
    The controllers for these tasks need to control the pose and the force (and/or the moment) applied at the surface by the robot while following a pose and force/moment trajectory. 
    
    \item \textbf{Operating on deformable surfaces:} In this type, the surface may deform or change due to the robot's interaction. It requires integrating the prior knowledge about the surface properties into the motion planner and the controller and following the pose and force/moment trajectories to perform a task. Applications include manipulating rigid surfaces, e.g., machining, drilling, hammering, and tasks involving non-rigid (i.e., deformable) objects such as cables, wires, textile, and plants.
    
\end{enumerate}

Many industrial applications require the exertion of force to a surface or an object for repetitive tasks such as machining and assembly. Hence, force control for fixed-base industrial manipulator robots has been studied by researchers for the past several decades. With the progress of aerial robot technology, recently, there have been efforts to adapt the available force control methods or introduce new methods specific to aerial robots. However, fixed-base industrial robots have vital advantages over aerial robots, making it challenging to develop reliable force control methods for aerial robots: 

\begin{itemize}
    \item Fixed-base manipulator robots often operate in extremely controlled settings compared to the operating environment of aerial robots;
    \item The required applied forces for industrial robots are often small compared to their weight and power, while the same cannot be assumed for small aerial robots such as multirotors;
    \item The aerodynamic effects do not affect the industrial robots, while the effects such as the wall effect may become significant during the aerial robots' operation.
\end{itemize}

However, some of the challenges have been addressed before by communities working on other types of robots. The fundamental need in humanoid robots for controlling the interaction forces and moments has driven the research on controllers for physical interactions of manipulators with the environment for many years. Some recently proposed force control methods for multirotors have adapted the methods developed in the humanoid communities to introduce force and motion control working for fully-actuated multirotors (i.e., \textit{aerial manipulation}).

The progress of aerial manipulation has dramatically accelerated in the past few years~\cite{Ruggiero2018, BonyanKhamseh2018, Meng2019, Ding2019a, Suarez2020, Mohiuddin2020}. However, the earliest attempts to apply force to the environment using multirotors go back to 2010. Authors of~\cite{Albers2010} added an extra actuator to a quadrotor to generate forces during physical contact when the robot is hovering.

The work in~\cite{Voyles2012} and~\cite{imav2013:g_jiang_et_al} was one of the first studies of grasping using aerial platforms. The authors later modified their design to apply forces to the environment~\cite{Jiang2013a, Jiang2013}, improved the flight controller~\cite{Jiang2014, Jiang2015}, and enhanced the design~\cite{Jiang2017a} for various applications, such as physical sampling~\cite{Jiang2017} and inspection~\cite{Jiang2015a, Jiang2018}. 

Many different control schemes were initially proposed to optimize the aerial manipulation and physical interaction tasks. These schemes range from the simplest forms such as Proportional–Integral–Derivative (PID) controllers~\cite{Korpela2013,Ruggiero2015}, Linear–Quadratic Regulators (LQR)~\cite{Wopereis2017}, or simply using the standard controllers~\cite{Gioioso2014} to more complex Interconnection and Damping Assignment-Passivity Based Control (IDA-PBC)~\cite{Yuksel2014a,Yuksel2019}, Model Predictive Controller (MPC)~\cite{Lunni2017,Bicego2019}, dynamic decentralized controller~\cite{Tognon2017}, Null Space-based Behavioral (NSB) control~\cite{Baizid2015, Baizid2017}, Model Reference Adaptive Control (MRAC)~\cite{Orsag2013}, and geometric control methods~\cite{Zhou2016,Rashad2017}. The recent trend is to adopt the force control methods of fixed-based manipulators, introducing the ideas of compliant arm force control~\cite{Cataldi2016,Bartelds2016,Yao2019}, parallel force/motion control~\cite{Rashad2019a}, and hybrid force-motion control~\cite{Meng2019b} for aerial manipulators. Authors of~\cite{Yuksel2017},~\cite{tognon:tel-02003048} and~\cite{bicego:tel-02433940} provide comprehensive overviews of the control methods for physical interaction and manipulation.

Some aerodynamic challenges are raised by the fully-actuated designs and the specific requirements of physical interaction with the environment, such as wall and ceiling effects. The different aerodynamic effects on the aerial manipulation task and multirotor flight have been studied in~\cite{Baizid2017} and~\cite{Sanchez-Cuevas2019}.

Traditional underactuated multirotors require a gimbal or a multi-DoF arm to apply forces and moments at the desired direction to the contact point. This requirement increases the complexity and cost of the system, reduces the flight time and the maximum possible force and moment of the robot, and intensifies the uncertainty of the end-effector's estimated pose. Authors of~\cite{Mersha2014} tried to mitigate the effects of the weight added by the manipulator's arms by exploiting the manipulator dynamics. A relatively new idea, which is made possible with the development of fully-actuated robots, is to replace the manipulator arm with a rigidly-attached arm and control the end-effector's pose directly using the multirotor's pose. This paradigm, called \textit{whole-body wrench generation}, eliminates the need for a complex arm controller, lowers the errors introduced by a gimbal or a multi-DoF arm, and reduces the payload of the system, increasing the maximum available forces and moments to apply at the point of contact. As far as we know, the first attempt to design a whole-body manipulator (rigidly-attached manipulator arm) to eliminate the need for the attached manipulation arm altogether was made in 2015~\cite{Nikou2015}.

Many research groups have provided methods to control the end-effector's position and force on multirotors. The following works show the current state of the art in contact control using fully-actuated multirotor robots.

ETH Zürich researchers have proposed a method for Non-Destructive Testing (NDT) inspection. Their vehicle is a complex fully-actuated vehicle with 12 rotors able to take any orientation in 6-D, which is described in~\cite{Bodie2018} and improved in~\cite{Allenspach2020} and~\cite{Bodie2020}. The vehicle has a rigidly mounted manipulator arm, which was initially introduced in~\cite{Bodie2019}. They use variable axis-selective impedance control, which integrates direct force control to control the desired interaction forces actively. They demonstrate force-controlled, peg in a hole, and push-and-slide interactions in different flight orientations. In their previous works, they have been able to demonstrate the use of their robots to control real-world settings such as inspection~\cite{Pfandler2019}.

The researchers at the Laboratory for Analysis and Architecture of Systems (LAAS) have introduced methods for NDT inspection using a simpler fixed-tilt fully-actuated aerial manipulator they have introduced in~\cite{Rajappa2015}. They started with a complex manipulator arm in~\cite{Tognon2019} and simplified the design to a rigidly attached arm in~\cite{Ryll2019} and~\cite{Nava2020}, introducing a whole-body force control strategy instead of controlling the arm. Authors of~\cite{Ryll2019} extend the work from~\cite{Ryll2017} using an outer admittance controller to control the desired admittance behavior while an inner loop controls the position. As a result, there is no direct control over the force and torque at the interaction point. In~\cite{Nava2020}, direct force feedback is used to control the energy, the motion, and the force of the end-effector using a Quadratic Programming optimization method that also respects the bounds and limits of the actuators. They demonstrate their method with force- and position-controlled push and push-and-slide interactions in the presence of external disturbances.

The researchers at the Seoul National University proposed a new omni-directional platform specifically for arbitrary wrench generation~\cite{Park2016, Park2018a}. Previously, their group has also worked on using the quadrotors for tool operations such as driving screws in simulation; however, they never completed their work with real experiments~\cite{Lee2012, Hai-NguyenNguyen2013, Nguyen2015}.

Some other groups have developed fully-actuated aerial manipulators for NDT contact inspection as well, including the researchers at the University of Twente~\cite{Rashad2019, Rashad2019a}, and a team at the University of Seville~\cite{Ollero2018, Trujillo2019}. Besides, some companies have put effort towards commercializing the fully-actuated aerial manipulators for NDT inspection in recent years~\cite{Voliro2020, Skygauge2020}.

The joint force and motion control methods can be divided into three broad categories~\cite{deWit1996}: compliant arm (indirect) force control (e.g., impedance control methods), parallel force/motion control (e.g., passivity-based methods), and hybrid force-motion control (e.g., task optimization-based methods)~\cite{Ortenzi2017}.

In the current work, we use the multirotor's body as the force generator, eliminating the need for having a manipulator arm with multiple degrees of freedom. A fixed-tilt multirotor (see Chapter~\ref{ch:background}) with a rigidly-attached arm is used for this purpose. We focus on the methods requiring the control of force and moments along with the pose of the end-effector rigidly attached to a fixed-tilt fully-actuated multirotor to perform non-destructive and environment-altering tasks such as cable manipulation at the top of a utility pole.

Despite all the progress with the fully-actuated UAVs, integrating these designs into real-life applications and physical interaction tasks has been much slower than the underactuated ones. The added design complexity, the lower energy efficiency, and the additional efforts required to develop new software tools prevent their widespread use in real projects. Even the success of the commercial efforts (e.g.,~\cite{Skygauge2020, CyPhy2018, Voliro2020}) has been minimal so far. 

Over the years, many methods and tools for underactuated multirotors have been developed and are widely available to plan missions and trajectories for different purposes and to control the motion of the robots based on the planned trajectories. The emergence of new architectures has resulted in the introduction of new control methods for the fully-actuated vehicles in different applications, ranging from exact feedback linearization~\cite{Rajappa2015, Mehmood2017} to nonlinear model-predictive control~\cite{Bicego2019}. However, most controllers developed for fully-actuated vehicles require full-pose trajectories and cannot interact with the tools developed for underactuated UAVs. An even more significant challenge is the need for new Remote Control (RC) interfaces that can control the robot in all the six degrees of freedom, making it difficult for the pilots to learn to fly or transition to flying these new robots.

Many research groups utilize tools such as Simulink to design and generate the code for their new fully-actuated controllers. While these tools can also generate the code required for a working flight control system (such as the state estimation), the generated code is too basic for any tasks performed beyond heavily-controlled lab environments. Moreover, the code cannot be directly integrated with the existing autopilots that already have a more powerful flight stack, slowing down the integration of the new robots into real applications even further. 

We have realized that most of the applications do not require a completely free change in orientation and the vehicle's attitude generally needs to follow a set of patterns. We can utilize this observation to define a set of \textit{fully-actuated operation modes} for the attitude (we refer to them as \textit{attitude strategies}). The strategies can be switched during the flight to address the different needs of applications. Identifying such a set of strategies allows using the readily-available flight controllers, software tools, and RC interfaces developed for the underactuated UAVs.

On the other hand, many fully-actuated UAVs, including the most common architectures, such as fixed-pitch hexarotor designs, can generate only limited lateral forces compared to the forces perpendicular to their bodies. Franchi et al.~\cite{Franchi2018} have proposed to call these vehicles \textit{LBF} (vehicles with laterally bounded force), which we will also use in this document. The operation of the LBF robots requires particular attention to handling their lateral thrust limits. We propose a set of methods (\textit{thrust strategies}) that can handle the lateral thrust limits with minimal changes to the available flight controllers, allowing the use of either 6-D (full-pose) or traditional 4-D planners and motion controllers. 

This chapter presents a new way of developing fully-actuated controllers by extending the existing underactuated flight controllers while keeping the entire flight stack intact without any modifications and allowing to take advantage of the full actuation in the UAV applications. Other fully-actuated controller methods can also adopt the approach to provide easier integration into existing flight stacks. It allows the controller to accept the traditional waypoints and simplifies the development of real-world tasks by easy switching between different strategies. Our proposed modifications to the controller have minimal overhead and can be directly integrated into any autopilot, extending only the Position Controller module.

Finally, we illustrate how the proposed controller design can be extended for tasks requiring precise physical interaction with the environment. The controller is extended into a hybrid force and position controller, allowing the control of forces and motions during the contact.

We show our experiments with the new controller design in Gazebo PX4 SITL and our MATLAB simulator as well as on the real robot to illustrate the proposed strategies and methods in free flight and during physical interaction.

\section{Problem Definition} \label{sec:control:problem}

\begin{figure}[!htb]
        \centering
        \includegraphics[width=0.7\linewidth]{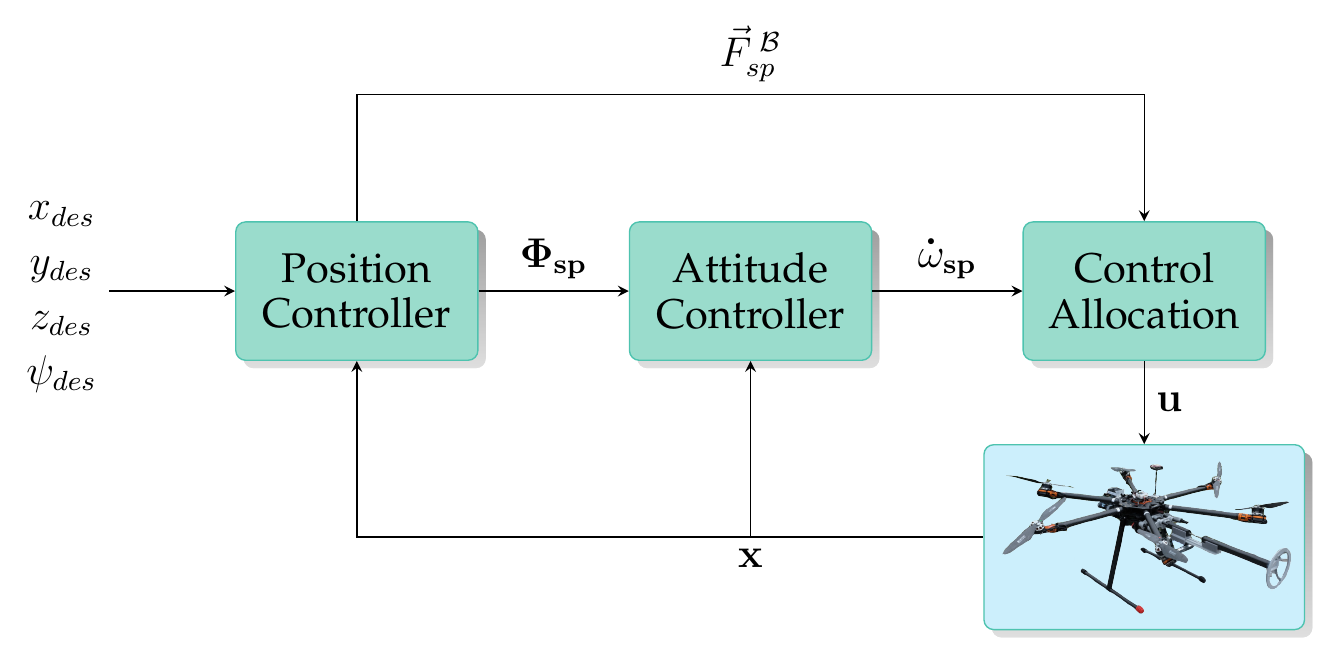}
        
        \caption[High-level illustration of a typical flight controller]{High-level architecture of a typical flight controller. Some architectures consist of more modules, which can be collected together to have the same general architecture with similar inputs and outputs.}
        
        \label{fig:control:typical-controller-architecture}
    \end{figure}

A general flight controller system for multirotors consists of a module for controlling the position (and/or position derivatives), a module for controlling the attitude (and/or angular rates), and a module for control allocation (mixing). Figure~\ref{fig:control:typical-controller-architecture} illustrates the architecture for a typical flight controller for underactuated UAVs. The inputs are normally the desired yaw, the desired position, and/or their derivatives. These inputs are internally utilized to generate the full desired attitude (we call it \textit{attitude setpoint}) for its Attitude Controller module and the desired linear acceleration or forces (we call it \textit{thrust setpoint}) for its Control Allocation module. 

\begin{figure}[!htb]
        \centering
        \includegraphics[width=0.7\linewidth]{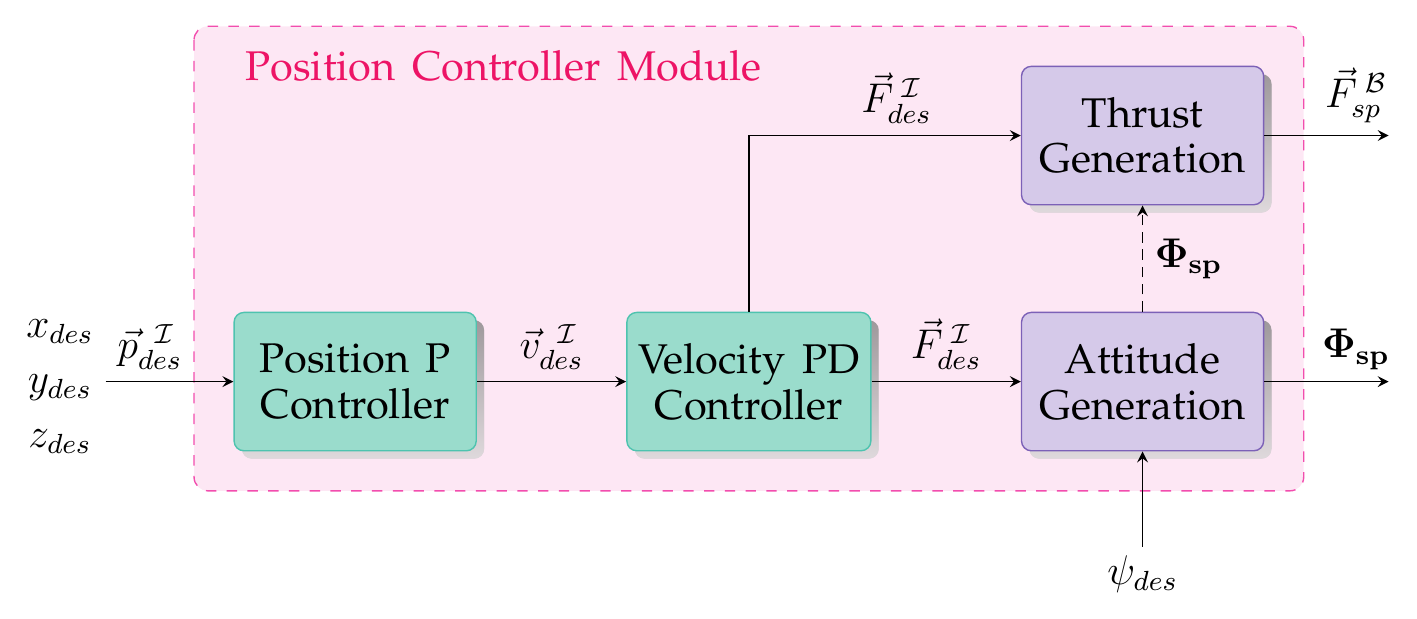}
        
        \caption[An example Position Controller module]{The internal structure of the Position Controller module for the PX4 flight controller as an example of a typical structure utilized in underactuated multirotors. The multirotor's state $\mat{x}$ is used as an input to all the submodules in the figure but is omitted for simplicity.}
        
        \label{fig:control:position-controller-module}
    \end{figure}

The attitude and thrust setpoints generation usually happens within some parent module (e.g., Position Controller). However, without the loss of generality, we assume that they are separate submodules with their inputs and outputs. Figure~\ref{fig:control:position-controller-module} illustrates the internal structure of the Position Controller module of the PX4 flight controller with the thrust and attitude setpoint generation functions depicted as separate submodules.

Some flight controller designs may have slightly different input/output combinations. For example, the forces can be replaced by linear accelerations, which have a linear relationship with forces. Such choices do not change the concepts discussed in this section. Moreover, in some controller designs, the force outputs may be expressed inertial instead of body-fixed frames if the Control Allocation module uses the force in the inertial frame (e.g., see~\cite{Rajappa2015}). In such cases, the Thrust Setpoint Generation function may only apply the thrust limits to the input or not do anything. In this section, we add the Thrust Generation module even when it does nothing in the original controller design to allow the extension of its functions in Section~\ref{sec:control:thrust}.

\section{Our Controller Design} \label{sec:control:design}

Sections~\ref{sec:background:control:intro} and~\ref{sec:control:intro} reviewed the prior work on fully-actuated controllers and described the multirotor kinematics and dynamics model for the fixed-pitch multirotor designs along with a control allocation method based on nonlinear dynamic inversion. Tools such as Simulink and MATLAB allow the design and synthesis of entirely new autopilot systems for the fully-actuated UAVs based on the provided formulations. The resulting autopilots can run directly on a computer connected to the UAV or even compiled for the standard autopilot hardware such as Pixhawk. These tools add the necessary code for functions such as perception, state estimation, and hardware interface, allowing rapid development and testing of new control ideas in controlled lab environments. However, the added code is not as comprehensive as the standard autopilot systems and cannot directly be devised on the UAVs in the real world.

On the other hand, extending the existing autopilots to support a new type of vehicle may seem more challenging and time-consuming. However, the functionalities of the established autopilots are more comprehensive and extensively tested. Therefore, extending the available autopilots to support the new fully-actuated multirotors accelerates the UAVs' integration with real-world applications.

We propose that extending the existing controllers is possible by the following minimal set of changes:

\begin{itemize} 

    \item Modifying or replacing the Control Allocation module to support the new architecture and prioritize angular acceleration over the linear acceleration.
    
    \item Extending the Attitude Setpoint Generation function into an Attitude Setpoint Generator module which allows utilizing the full actuation based on chosen strategies (see Section~\ref{sec:control:attitude}).
    
    \item Extending the Thrust Setpoint Generation function into a Thrust Setpoint Generator module, which manipulates the thrust setpoint to respect the fully-actuated vehicles' thrust limits.
\end{itemize}

\begin{figure}[!htb]
        \centering
        \begin{subfigure}[b]{\textwidth}
            \includegraphics[width=\textwidth]{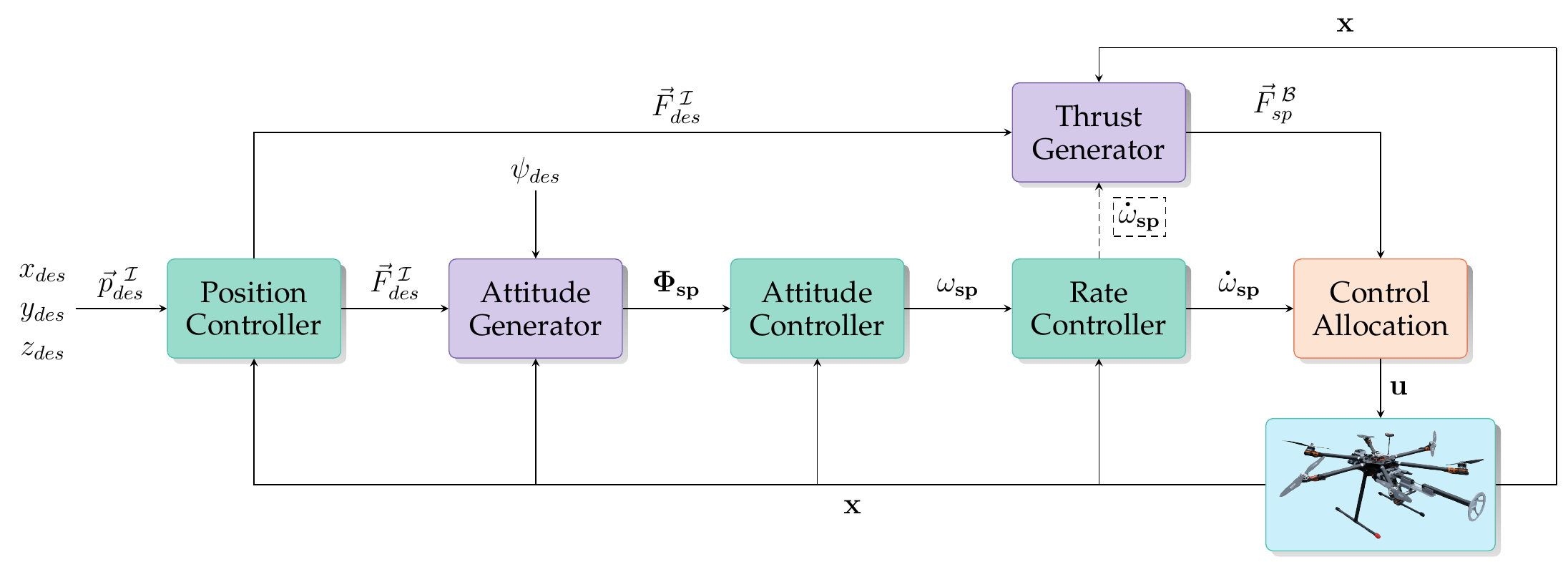}
            \caption{~}
            \label{fig:control:controller-a}
        \end{subfigure}
        \medskip
        \begin{subfigure}[b]{0.75\textwidth}
            \includegraphics[width=\textwidth]{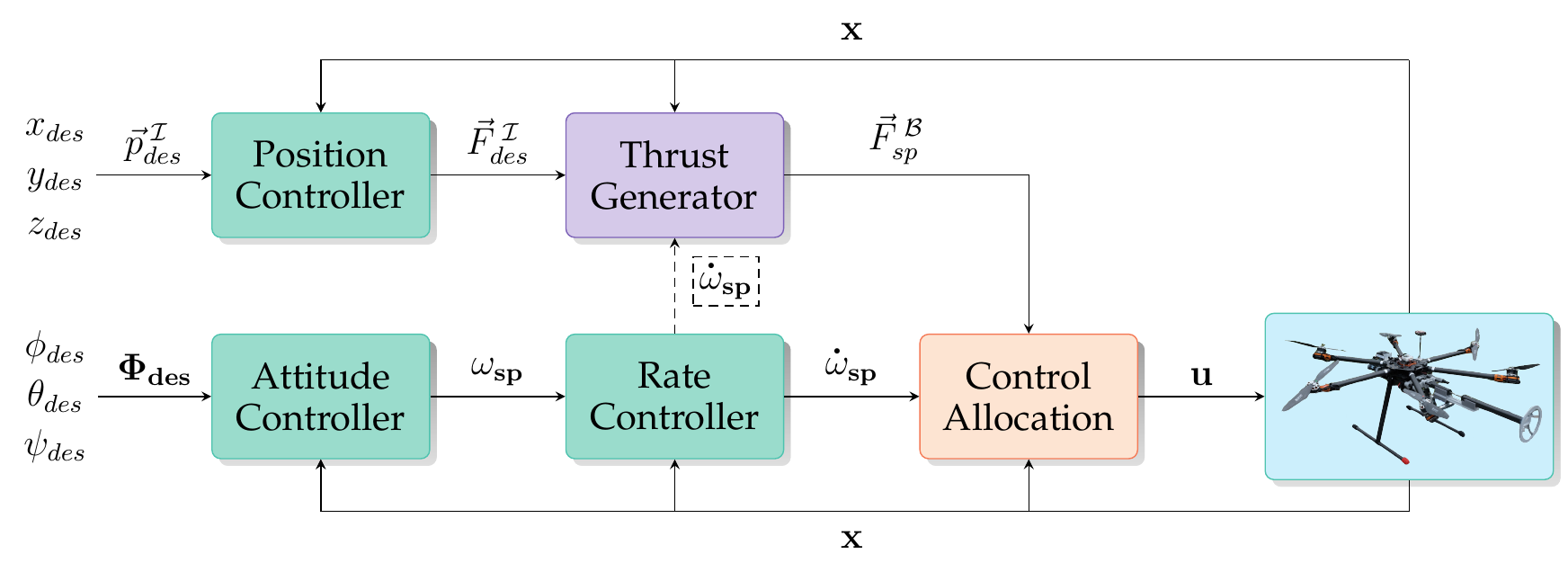}
            \caption{~}
            \label{fig:control:controller-b}
        \end{subfigure}
        
        \caption[Our fully-actuated controller designs]{Our flight controller architecture for fully-actuated multirotors based on the PX4 flight controller design. (a) The controller's input is only the desired position and the desired yaw (and/or their derivatives), requiring an Attitude Generation module to calculate the pitch and roll. The input to the Attitude Controller depends on the Position Controller and the Attitude Setpoint Generator modules. (b) The controller input is the desired pose (position and orientation). The input attitude already includes the full desired attitude and does not require any manipulation, which eliminates the need for the Attitude Setpoint Generator module and makes the Attitude Controller fully independent from the Position Controller. }
        
        \label{fig:control:controller}
    \end{figure}

Figure~\ref{fig:control:controller} illustrates the design of the controller we implemented for fully-actuated multirotors based on the existing PX4 autopilot architecture~\cite{Meier2015}. The controller only modifies the Control Allocation module but devises all other parts of the existing controller on PX4 autopilot for faster implementation, better stability, and integration with other flight controller modules (e.g., perception and state estimation). It also separates the Thrust and Attitude Setpoint Generator modules from the Position Controller and extends them for our purposes.

In addition to the modification of the Control Allocation module to support the new architecture, the overall controller structure depends on the controller's input:

\begin{enumerate}
    \item For inputs with only position and yaw (and/or their derivatives), additional modules for thrust and attitude setpoint generation are used that depend on a desired inertial thrust input from the position controller (Figure~\ref{fig:control:controller-a}). The Attitude Setpoint Generator module produces the full desired attitude for the Attitude Controller to track, making the Attitude Controller an inner loop for the Position Controller. Depending on the application and the priorities defined by the user, a specific attitude generation strategy can be devised. Section~\ref{sec:control:attitude} discusses the attitude and thrust setpoint generation methods we have developed and implemented for the current work. 

    \item For full-pose inputs with both the desired position and complete desired orientation (and/or their derivatives), the Attitude Setpoint Generator module of Figure~\ref{fig:control:controller-a} loses its functionality. Simplifying the structure of Figure~\ref{fig:control:controller-a} shows that now the Position and Attitude Controller modules can work independently to generate the body thrusts and moments required for the UAV to track the desired input (Figure~\ref{fig:control:controller-b}).
\end{enumerate} 

After integrating new fully-actuated UAVs into an existing autopilot (by modifying the Control Allocation module to support the new architecture), we learned that a problem might arise when generating both the thrust setpoint and the angular acceleration setpoint (coming from the Attitude Controller in Figure~\ref{fig:control:typical-controller-architecture}) is not feasible. In this situation, some motor commands calculated by the Control Allocation module will be in the saturation range, which may result in instability for the whole robot if not appropriately handled. 

There are many methods available to handle motor saturation. The default behavior for underactuated multirotors is usually either to bound the motor signal commands or devise a strategy that ensures the $\ZB$-thrust (normal thrust) is prioritized. However, with the fully-actuated UAVs, the strategy needs to be changed to prioritize the moments around the $\XB$ and $\YB$ axes. This change is crucial in keeping the UAV's stability when large commands are given with fixed attitude strategies. Many such strategies have been introduced for underactuated UAVs and can be easily modified for the fully-actuated vehicles (e.g., see~\cite{Smeur2017, Faessler2017, Brescianini2020}). In our implementation, we modified the Airmode functionality of the PX4 firmware and prioritized the moments around the $\XB$ and $\YB$ axes over the thrusts and the moment around the $\ZB$ axis.

\section{Attitude Strategies for Fully-Actuated UAVs} \label{sec:control:attitude}

In traditional coplanar multirotors, the robot can only generate thrust normal to its rotors plane, requiring it to completely tilt towards the total desired thrust direction to align the generated thrust with the desired thrust. However, fully-actuated vehicles are capable of independently controlling their translation and orientation. 

When a full-pose 6-D input is passed to the controller, no additional processing is required on the desired orientation. However, generally, with underactuated controllers, only the yaw is given to the controller, and the other two degrees of orientation are derived from the desired thrusts. We call this calculated desired orientation as \textit{attitude setpoint} which is then sent to the Attitude Controller module. 

The autonomous controller developed in Section~\ref{sec:control:design} has an Attitude Setpoint Generation module for when the given attitude input only specifies the desired yaw (see Figure~\ref{fig:control:controller-a}). This module accepts three inputs: the desired thrust force in the inertial frame ($\FdesI$) coming from the Position Controller module, the desired yaw ($\psi\des$) coming from the motion controller (which may be following a trajectory or converting the user's RC commands to motion commands), and the UAV's current state $\mat{x}$. The module's only output is the complete attitude setpoint ($\mat{\Phi\spt}$), which serves as an input for the Attitude Controller module. 

\begin{figure}[!htb]
\centering
\includegraphics[width=0.6\linewidth]{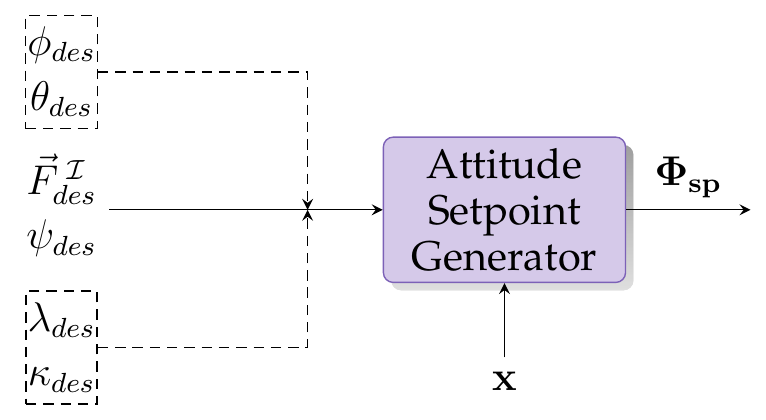}
\caption[Attitude Setpoint Generation module]{An illustration of the Attitude Setpoint Generation module with its inputs and outputs. The optional inputs are enclosed in dashed boxes.}
\label{fig:control:attitude:attitude-setpoint-generation-module}
\end{figure}

Some flight controller designs may include a slightly different set of inputs. For example, the input force vector may be replaced by linear accelerations, a linear conversion that does not change the concepts discussed here. 

For the set of strategies proposed in this section, there are additional strategy-specific inputs (enclosed in dashed boxes in Figure~\ref{fig:control:attitude:attitude-setpoint-generation-module}), which are not present in the underactuated controllers. These inputs will be explained in their relevant strategies later in this section.

Finally, in this problem, it is assumed that the given desired thrusts have already compensated for the gravity force to achieve the desired acceleration, i.e., the thrust equal to the vehicle's weight is added to the upward $z$ component of the desired thrust.

We have implemented several strategies for fully-actuated multirotors:

\begin{enumerate}
\item \textbf{Zero-tilt attitude strategy:} This strategy keeps the robot's tilt at zero at all times, allowing it to stay completely horizontal during the flight.

\item \textbf{Full-tilt attitude strategy:} This is the traditional attitude generation method for multirotors, where the output body-fixed $\axis{Z}$ axis is always in the opposite direction of the desired input thrust.

\item \textbf{Minimum-tilt attitude strategy:} This strategy minimizes the robot's tilt but does not guarantee to keep it at zero, slightly tilting it towards the desired thrust when the desired thrust is significant, allowing the robot to achieve larger accelerations than what is possible in the Zero-tilt strategy.

\item \textbf{Fixed-tilt attitude strategy:} This strategy keeps the robot's tilt at the desired angle and towards the desired direction, independent of the given desired thrust, which can be helpful in some situations, such as flight during a strong wind.

\item \textbf{Fixed-attitude strategy:} This strategy keeps the robot's roll and pitch as desired, independent of the given desired thrust, which can be useful for some situations, such as flight during physical contact with a surface in the wind or at the desired contact angle.
\end{enumerate}

The attitude is usually represented as the set of Euler angles (i.e., roll, pitch, and yaw), rotation matrices, or different types of quaternions, each having its pros and cons. This section uses rotation matrices to represent attitudes and shows the direct ways of calculating the Euler angles when such shortcuts exist. 

For each strategy, it is explained how to derive the attitude setpoint in the rotation matrix form. The rotation matrix is basically the composition of the unit vectors $\iSv$, $\jSv$ and $\kSv$, in the directions of $\XS$, $\YS$ and $\ZS$ axes of the setpoint frame $\Frm{S}$, respectively:

    \begin{equation} \label{eq:control:attitude:rotation-matrix}
        \RIS = \begin{bmatrix} \iS & \jS & \kS \end{bmatrix}
    \end{equation}

This representation can be converted to any other representation used by the Attitude Controller module. For example, the conversion to the Euler angles can be done using Equation~\ref{eq:control:attitude:rotation-matrix-to-euler}:

    \begin{equation} \label{eq:control:attitude:rotation-matrix-to-euler}
        \mat{\Phi\spt} = \begin{bmatrix}[2] 
        \arctan \left( \cfrac{\axis{\mat{j}}_3}{\axis{\mat{k}}_3} \right) \\
        -\arcsin \left( \axis{\mat{i}}_3 \right) \\
        \psi\des 
        \end{bmatrix}
    \end{equation}

The rest of this section describes the proposed attitude strategies and the applications where each can be useful.

\subsection{Zero-Tilt Attitude Strategy} \label{sec:control:attitude:zero-tilt}

Keeping the multirotor's attitude at zero tilt (i.e., keeping it horizontally level) during the flight can be beneficial for many situations, such as making precise contact with the vertical surface or capturing a video using an onboard camera without the need for a gimbal. 

\begin{figure}[!htb]
\centering
\includegraphics[width=0.5\linewidth]{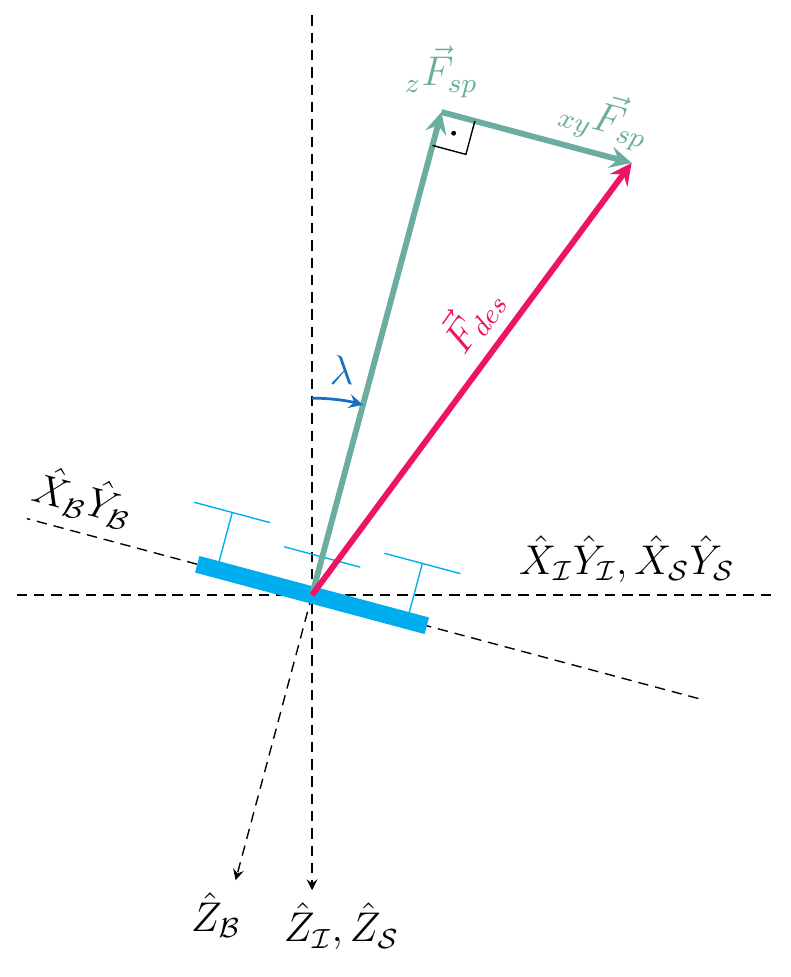}
\caption[Model for zero-tilt attitude strategy]{The model used for zero-tilt attitude calculation for fully-actuated multirotors. The resulting attitude setpoint always has its $\ZS$ axis pointing in the direction of the gravity ($\ZI$).}
\label{fig:control:attitude:zero-tilt-model}
\end{figure}

Calculating the output attitude setpoint in this strategy is straightforward. The output roll and pitch are always zero; therefore, the direction of the $\ZS$ axis for the attitude setpoint would be the same as $\ZI$ (i.e., $\matrice{0 & 0 & 1}\T$), independent of the desired inputs. Additionally, based on our definition of the body-fixed frame, the direction of the $\XS$ axis is the direction of the input desired yaw $\psi\des$. The $\YS$ axis is perpendicular to it, and both lie on the horizontal plane in this strategy. Figure~\ref{fig:control:attitude:zero-tilt-model} shows a model used for the zero-tilt attitude generation strategy.

The rotation matrix for this attitude setpoint can be constructed from the unit vectors in the direction of the setpoint axes as:

    \begin{equation} \label{eq:control:attitude:zero-tilt-axes}
        \RIS = \begin{bmatrix} 
            \cos \psi\des & -\sin \psi\des & 0\\ 
            \sin \psi\des & \cos \psi\des & 0\\ 
            0 & 0 & 1
            \end{bmatrix}
    \end{equation}

While it is possible to use Equation~\ref{eq:control:attitude:rotation-matrix-to-euler} to calculate the Euler angles, in this specific strategy, the Euler angles are directly defined as:

    \begin{equation} \label{eq:control:attitude:zero-tilt-attitude}
        \mat{\Phi\spt} = \begin{bmatrix} 0 \\ 0 \\ \psi\des \end{bmatrix}
    \end{equation}

\subsection{Full-Tilt Attitude Strategy} \label{sec:control:attitude:full-tilt}

With coplanar underactuated multirotor designs, the only way for the robot to achieve the input desired thrust is to tilt so that the direction of the desired thrust is normal to the plane of the rotors. The strategy does not take advantage of the full actuation in the fully-actuated robots, but it is the most helpful strategy to oppose the external forces and disturbances. Additionally, it is readily available in the popular autopilots and can be used for fully-actuated robots right out of the box. 

In most common multirotor architectures, this strategy requires the minimum energy and is the best choice when controlling the robot's attitude is not essential for the task. Section~\ref{sec:applications:wind} describes how this method can further be used to estimate the optimal tilt when an external force is applied to the robot (e.g., in windy conditions). 

Figure~\ref{fig:control:attitude:full-tilt-model} shows a model to demonstrate the full-tilt attitude strategy. In this strategy, the desired input thrust $\FdesI$ should be normal to the robot's output $\PlaneS$ plane and is in the opposite direction of the $\ZS$ axis. Therefore, the unit vector in $\ZS$ direction can be computed as:

\begin{figure}[!htb]
\centering
\includegraphics[width=0.5\linewidth]{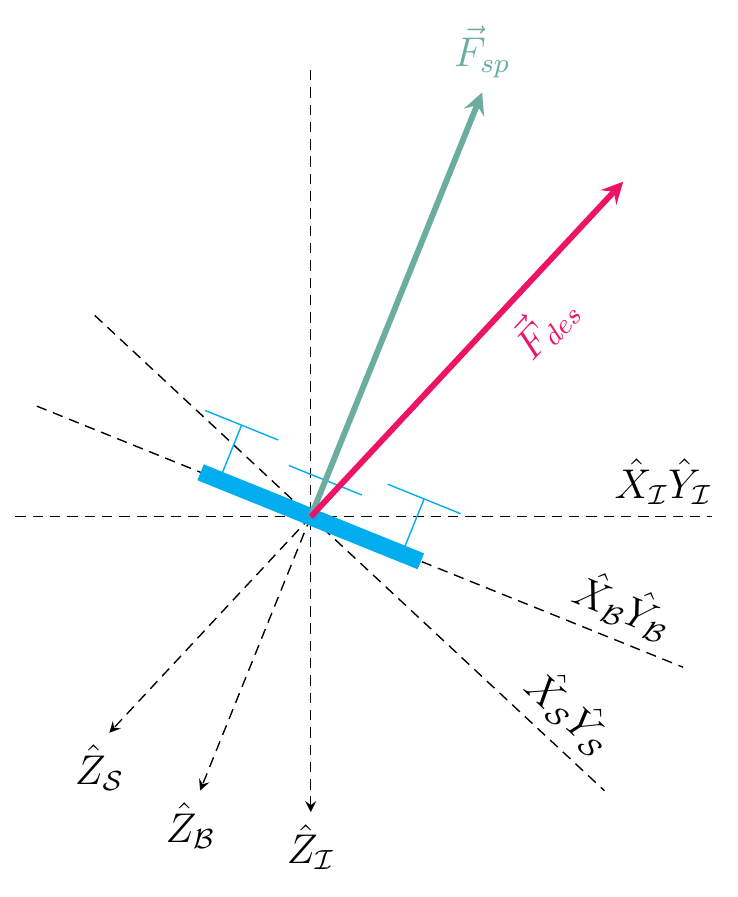}
\caption[Model for full-tilt attitude strategy]{(a) The model used for full-tilt attitude calculation for both fully-actuated and underactuated multirotors. The thrust setpoint is aligned with the current body-fixed $ZB$ axis, while the attitude setpoint is generated based on the desired thrust direction.}
\label{fig:control:attitude:full-tilt-model}
\end{figure}

    \begin{equation} \label{eq:control:attitude:full-tilt-zspt}
        \kSv = - \frac{\FdesI}{\left\| \FdesI \right\|} 
    \end{equation}

If a unit vector in the desired yaw direction on the $\PlaneI$ plane is rotated for $+90^\circ$ around the $\ZI$ axis, it will be on the plane made from axes $\YS$ and $\ZS$. Using this observation, the unit vector in the $\XS$ direction can be calculated as:

    \begin{equation} \label{eq:control:attitude:full-tilt-xspt}
        \iSv = \frac{\matrice{-\sin\psi\des \\ \cos\psi\des \\ 0} \times \kSv}{\left\| \matrice{-\sin\psi\des \\ \cos\psi\des \\ 0} \times \kSv \right\|}
    \end{equation}

From the $\XS$ and $\ZS$ axes calculated in Equations~\ref{eq:control:attitude:full-tilt-zspt} and~\ref{eq:control:attitude:full-tilt-xspt}, the unit vector in the direction of $\axis{Y}\spt$ axis can be computed as:

    \begin{equation} \label{eq:control:attitude:full-tilt-yspt}
        \jSv = \kSv \times \iSv 
    \end{equation}

The rotation matrix and the Euler angles for the attitude setpoint can be computed from Equations~\ref{eq:control:attitude:full-tilt-zspt},~\ref{eq:control:attitude:full-tilt-xspt} and~\ref{eq:control:attitude:full-tilt-yspt} using Equations~\ref{eq:control:attitude:rotation-matrix} and~\ref{eq:control:attitude:rotation-matrix-to-euler}.

\subsection{Minimum-Tilt Attitude Strategy} \label{sec:control:attitude:min-tilt}

The zero-tilt strategy described in Section~\ref{sec:control:attitude:zero-tilt} requires a lateral thrust input that is less than its maximum lateral thrust limit. Section~\ref{sec:control:thrust} describes how to bound the lateral input thrust when it is more than the maximum available lateral thrust $\Flmax$. In some applications, keeping the tilt magnitude as close to zero is desirable, but achieving the input accelerations or rejecting high external forces is more important than maintaining the fixed attitude. An example scenario is filming a highly dynamic target or in high-gust winds. In such cases, a better strategy than full-tilt (Section~\ref{sec:control:attitude:full-tilt}) can be devised that takes advantage of the full actuation to minimize the tilt of the robot by using up the lateral thrust. The strategy is first described in~\cite{Mehmood2017}.

In this strategy, the vehicle keeps its attitude at zero tilt until a larger than maximum lateral thrust is needed, then the vehicle minimally tilts, keeping its lateral thrust at maximum to reduce the tilt as much as possible.

\begin{figure}[!htb]
\centering
\includegraphics[width=0.5\linewidth]{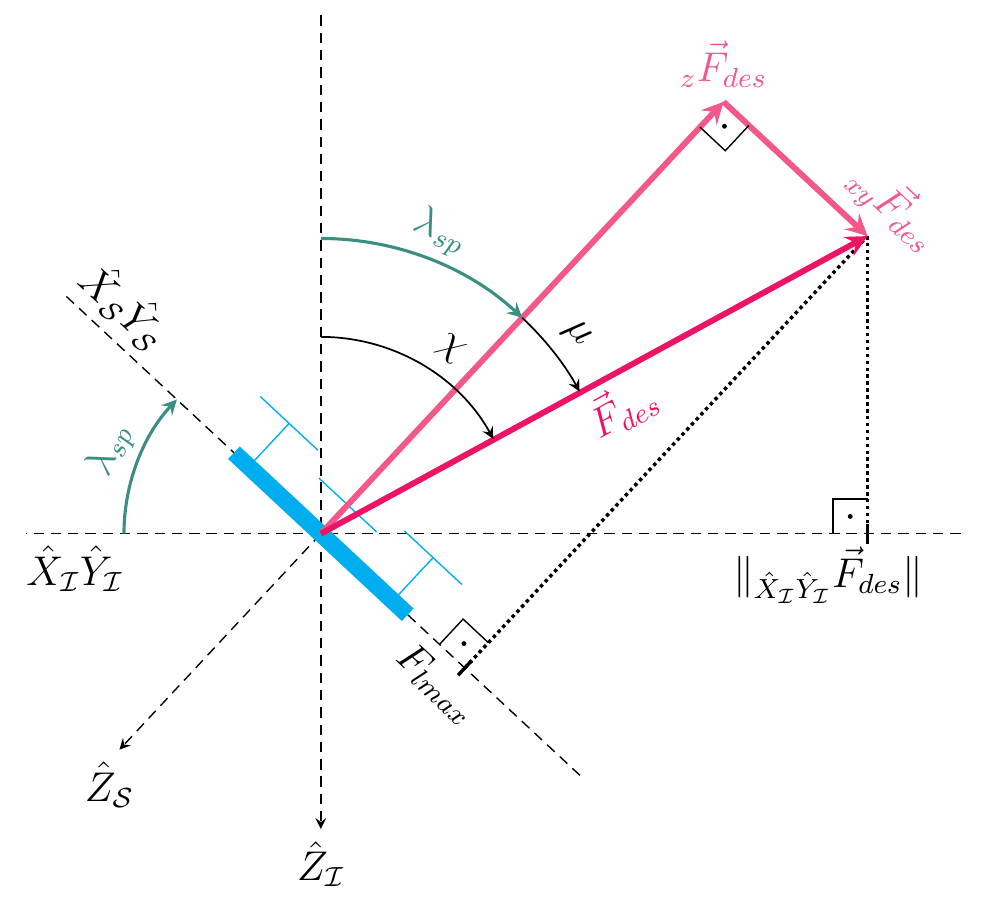}
\caption[Model for minimum-tilt attitude strategy]{An illustration of the model for calculation of the minimum-tilt attitude. At the minimum tilt, the projection of the input desired thrust on robot's body-fixed horizontal plane is at the maximum available lateral thrust. The illustrated vectors $\vecelem{\Fdes}{z}$ and $\vecelem{\Fdes}{xy}$ are respectively the normal and lateral elements of $\Fdes$ in the attitude setpoint frame.}
\label{fig:control:attitude:min-tilt-model}
\end{figure}

Figure~\ref{fig:control:attitude:min-tilt-model} helps demonstrating how the minimum-tilt attitude is calculated. 

Given a desired input thrust vector $\FdesI$, if the desired thrust on the lateral plane $\| \ProjI{\Fdes} \|$ is less than the maximum possible lateral thrust $\Flmax$, then the zero-tilt attitude strategy is used to calculate the attitude setpoint (see Section~\ref{sec:control:attitude:zero-tilt}). However, when the desired thrust on the horizontal plane is larger than the available lateral thrust, all the possible thrust on the lateral plane is used first, then the remaining required thrust determines the required tilt $\lambda\spt$, roll $\phi\spt$ and pitch $\theta\spt$. To calculate the tilt $\lambda\spt$, we have:

    \begin{equation} \label{eq:control:attitude:min-tilt-calculation}
        \lambda\spt = \underbrace{ \arcsin \left(\frac{\left\| \ProjI{\Fdes} \right\|}{\left\| \Fdes \right \|}\right)}_\chi
        - \underbrace{ \arcsin\left( \frac{\Flmax}{\left\| \Fdes \right \|} \right) }_\mu
    \end{equation}

\noindent where $\chi$ and $\mu$ angles are as illustrated in Figure~\ref{fig:control:attitude:min-tilt-model}, and $\ProjI{\Fdes}$ is the projection of the desired thrust on the horizontal plane.

The axis of rotation $\raxis$ for the tilt $\lambda\spt$ is perpendicular to the plane consisting of the desired thrust $\Fdes$ and the inertial $\ZI$ axis. Hence, it can be calculated as:

    \begin{equation} \label{eq:control:attitude:min-tilt-rotation-axis-calculation}
        \raxis = \frac{\Fdes \times \kIv}{\left\| \Fdes \times \kIv \right \|}
    \end{equation}
    
The $\ZS$ axis direction can be computed by rotating the $\kIv$ unit vector around $\raxis$ using the Rodrigues' rotation formula:

    \begin{equation} \label{eq:control:attitude:min-tilt-zdes}
        \kSv = (1 - \cos\lambda\spt) \ (\raxis \cdot \kIv) \ \raxis + \kIv \ \cos\lambda\spt
        + (\raxis \times \kIv) \ \sin\lambda\spt
    \end{equation}

The $\XS$ and $\YS$ axes can be calculated similar to the full-tilt strategy (Section~\ref{sec:control:attitude:full-tilt}) from Equations~\ref{eq:control:attitude:full-tilt-xspt} and~\ref{eq:control:attitude:full-tilt-yspt}. Similarly, the rotation matrix and the Euler angles for the attitude setpoint can be computed using Equations~\ref{eq:control:attitude:rotation-matrix} and~\ref{eq:control:attitude:rotation-matrix-to-euler} from these axes.

\subsection{Fixed-Tilt Attitude Strategy} \label{sec:control:attitude:fixed-tilt}

Some applications require keeping a specific tilt angle for the multirotor. An example scenario is flying in the constant wind where keeping a fixed tilt against the wind is desirable to increase the remaining thrust after opposing the wind, independent of the yaw and the movement direction. 

\begin{figure}[!htb]
\centering
\includegraphics[width=0.5\linewidth]{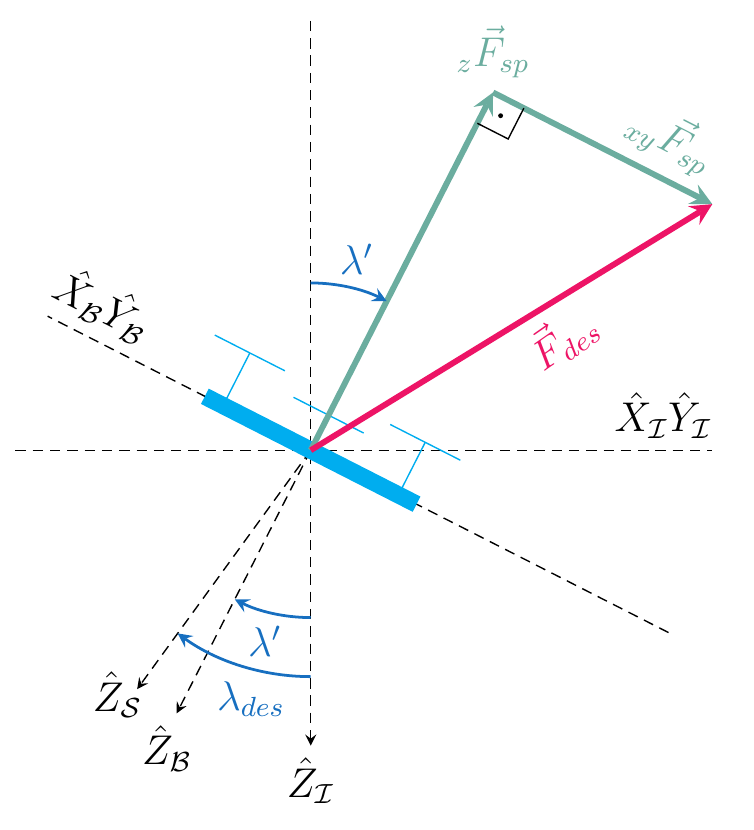}
\caption[Model for fixed-tilt attitude strategy]{An illustration of the model used for the fixed-tilt attitude strategy.}
\label{fig:control:attitude:fixed-tilt-model}
\end{figure}

Figure~\ref{fig:control:attitude:fixed-tilt-model} shows a model demonstrating the axes and angles used in the calculation of the fixed-tilt attitude strategy.

In addition to the desired thrust and yaw inputs, let us assume two new inputs $\lambda\des$ and $\kappa\des$ to the system, representing the angle of the desired tilt and the direction of the tilt, respectively. These two inputs are shown in Figure~\ref{fig:control:attitude:attitude-setpoint-generation-module} in a dashed box and are only used for this attitude strategy. 
We assume that the desired direction $\kappa\des$ is given in the inertial frame (i.e., with respect to the north direction). 

The axis of rotation to tilt the robot is perpendicular to the inertial $\ZI$ axis and the projection of the vector pointing in the direction of tilt on the $\PlaneI$ plane. Hence, the axis of rotation $\raxis$ can be computed as:

    \begin{equation} \label{eq:control:attitude:fixed-tilt-rotation-axis-calculation}
        \raxis = \matrice{\cos\kappa\des \\ \sin\kappa\des \\ 0} \times \kIv
    \end{equation}
    
Having the rotation axis and considering that $\lambda\spt = \lambda\des$, the $\ZS$ axis direction can be calculated using the Rodrigues's rotation formula of  Equation~\ref{eq:control:attitude:min-tilt-zdes}, and the $\XS$ and $\YS$ axes can be calculated from Equations~\ref{eq:control:attitude:full-tilt-xspt} and~\ref{eq:control:attitude:full-tilt-yspt}. Finally, the rotation matrix and the Euler angles for the attitude setpoint can be computed using Equations~\ref{eq:control:attitude:rotation-matrix} and~\ref{eq:control:attitude:rotation-matrix-to-euler} from these axes.

\subsection{Fixed-Attitude Strategy} \label{sec:control:attitude:fixed-attitude}

Suppose the controller's input trajectory from the onboard computer includes full-pose information (both full 3-D position and 3-D attitude). In that case, the attitude generator can be bypassed altogether (see the controller architecture in Figure~\ref{fig:control:controller-b}). However, in practice, some applications require achieving specific attitude angles for the multirotor without the input trajectory explicitly including the roll-pitch angles. An example scenario is during the robot's contact with the wall when a constant orientation can help control the end-effector's pose and wrench. To achieve this goal, the Attitude Generator Module (Figure~\ref{fig:control:attitude:attitude-setpoint-generation-module}) can devise two new inputs for the desired roll $\phi\des$ and the desired pitch $\theta\des$.

In this strategy, the rotation matrix can be directly calculated from the given Euler angles using the "3-2-1"-rotation sequence:

    \begin{equation} \label{eq:control:attitude:rotation-matrix-from-euler-angles}
        \RIS = \begin{bmatrix} 
            \ctheta\cpsi & \sphi\stheta\cpsi - \cphi\spsi & \cphi\stheta\cpsi + \sphi\spsi\\
            \ctheta\spsi & \sphi\stheta\spsi + \cphi\cpsi & \cphi\stheta\spsi - \sphi\cpsi\\
            -\stheta & \sphi\ctheta & \cphi\ctheta
        \end{bmatrix}
    \end{equation}

\noindent where $c$ and $s$ are shorthand for $\cos$ and $\sin$ functions, respectively, and $\phi$, $\theta$, $\psi$ are used for $\phi\des$, $\theta\des$ and $\psi\des$.

\section{Thrust Strategies for Fully-Actuated UAVs} \label{sec:control:thrust}

\begin{figure}[!htb]
        \centering
        \includegraphics[width=0.5\linewidth]{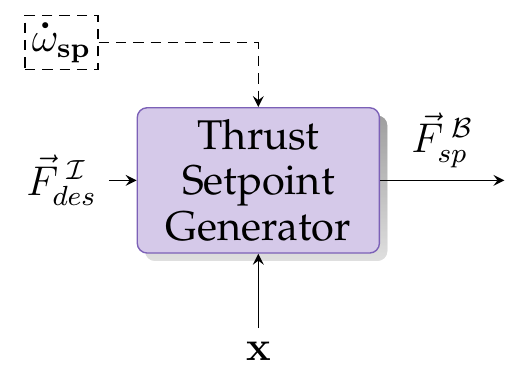}
        
        \caption[Thrust Setpoint Generator module]{An illustration of the Thrust Setpoint Generator module with its inputs and outputs. The optional input is enclosed in a dashed box.}
        
        \label{fig:control:thrust:thrust-setpoint-generation-module}
    \end{figure}

The controller architecture for fully-actuated multirotors, which is described in Section~\ref{sec:control:design}, has a Thrust Setpoint Generator module that takes the desired thrust calculated by the Position Controller and prepares it for the Control Allocation module (see Figure~\ref{fig:control:controller}). Figure~\ref{fig:control:thrust:thrust-setpoint-generation-module} shows this module separately with all of its inputs and outputs. 

The thrust setpoint of the Attitude Setpoint Generator module (Figure~\ref{fig:control:attitude:attitude-setpoint-generation-module}) is expressed in body-fixed frame $\FB$ while here, the input desired thrust is in the inertial frame $\FI$. Assuming unlimited available thrust, the input thrust can be simply rotated from the inertial frame to the current body-fixed frame (i.e., $\FdesI$ can be projected on the current body-fixed axes) to compute the thrust setpoint:

    \begin{equation} \label{eq:control:thrust:thrust-setpoint-calculation}
        \FsptB = 
        \RBI \cdot \FdesI = 
        \begin{bmatrix}[1.5] \FdesI \cdot \XB \\ \FdesI \cdot \YB \\ \FdesI \cdot \ZB \end{bmatrix}
    \end{equation}

Note that in some controller designs, the output thrust setpoint may be described in the inertial frame if the Control Allocation module requires the force to be in the inertial frame (e.g., see~\cite{Rajappa2015}). For such architectures, the thrust setpoint calculated using this section's methods can be simply rotated back from the body-fixed frame to the inertial frame.

In practice, the available thrust is not unlimited, and fully-actuated vehicles have thrust limits that should be considered. Particularly, fully-actuated LBF multirotors (see the definition in Section~\ref{sec:control:intro}) have limited lateral thrust compared to the normal thrust (i.e., in $\ZB$ direction) due to their structure, and if the elements of the input desired thrust $\Fdes$ on the body $\PlaneB$ plane (i.e., $\| \ProjB{\Fdes} \|$) are larger than the maximum possible lateral thrust ($\Flmax$), some motors will saturate, and the whole system may lose its stability. 

One solution proposed so far in~\cite{Franchi2018} can be devised when a full-pose trajectory is available, and it sacrifices the orientation over the position to track the given trajectory. The method requires a full-pose planner and a particular controller architecture and cannot be easily integrated with the available underactuated tools and controllers. The methods proposed in this section are fundamentally different. They can work with all common controller architectures by refining the thrust setpoint to respect the thrust limits and minimize the stability issues. 

Before continuing to describe our methods, we need to define the terminology used in this section:

\textit{Lateral thrust} ($\Flat$): The component of the thrust on the body-fixed $\PlaneB$ plane. It is the vector constructed from the thrust's $x$ and $y$ components in the $\FB$ frame.

\textit{Normal thrust} ($\Fnor$): The component of the thrust on the body-fixed $\ZB$ axis.

\textit{Horizontal thrust} ($\Fhor$): The component of the thrust on the inertial $\PlaneI$ plane. It is the vector constructed from the thrust's $x$ and $y$ components in the $\FI$ frame.

\textit{Vertical thrust} ($\Fver$): The component of the thrust on the inertial $\ZI$ axis.

\textit{Hover thrust} ($\Fhover$): The vertical thrust required to keep the UAV hovering when no wind is acting on it. In other words, the hover thrust of the UAV is the total weight of the UAV and all of its attached components. The hover thrust can be defined differently as the total thrust (or the percentage of the maximum possible thrust) generated by the rotors to keep the UAV hovering; however, in this section, we refer to hover thrust as the required vertical thrust.

\textit{Maximum lateral thrust} ($\Flmax$): The maximum achievable force on the lateral plane ($\PlaneB$) of the UAV in the direction of the desired force at any specific state. Chapter~\ref{ch:wrench} introduces methods to estimate the available thrusts, which can provide an accurate estimation of the lateral thrust limits at each time. For simplicity, here we assume that $\Flmax$ is a constant value independent of both the system state and the desired thrust and is uniform in all directions on the $\PlaneB$ plane.

Given the input desired thrust $\FdesI$ and its rotation, two cases can happen with the thrust setpoint $\FsptB$ from Equation~\ref{eq:control:thrust:thrust-setpoint-calculation}:

\noindent \textit{Case 1.} $\| \Flat \| \leq \Flmax$: In this case, generating the desired input thrust is feasible and the result of Equation~\ref{eq:control:thrust:thrust-setpoint-calculation} can be directly used as the output thrust setpoint.

\noindent \textit{Case 2.} $\| \Flat \| > \Flmax$: In this case, the robot will not be able to achieve the desired thrust. This section provides solutions to address this case based on the application requirements.

We describe two methods with different objectives for handling the latter case when the required lateral thrust is larger than the available thrust:

\begin{enumerate}
\item Only the lateral thrust is bounded to keep the desired vertical thrust.
\item All the thrust is bounded to keep the acceleration directions.
\end{enumerate}

Each method has its applications and can be used depending on the situation.

\subsection{Strategy 1: Keeping the Desired Vertical Thrust}

Just merely cutting the lateral input thrust to the $\Flmax$ value can result in losing a portion of the vertical thrust, leading to altitude tracking error and a crash in extreme cases. Therefore, an essential objective for handling the lateral thrust can be keeping the vertical thrust at the desired input thrust value so the UAV's altitude still follows the input command. 

For this purpose, first, both the vertical and horizontal components of the desired thrust ($\Fver$ and $\Fhor$, respectively) are projected on (rotated to) the body-fixed axes separately to compute the partial thrust setpoint vectors $\FverB$ and $\FhorB$.

We assume that the lateral thrust of $\FverB$ (its components on $\PlaneB$ plane) from the vertical component of the desired input thrust is not greater than $\Flmax$; otherwise, the vertical desired thrust cannot be achieved. The assumption is reasonable if the robot's tilt is limited (the limit can be set based on the dynamics of the multirotor). After consuming the lateral thrust required for the vertical component, the remaining available lateral thrust for $\Fhor$ will be:

    \begin{equation} \label{eq:control:thrust:new-lateral-max-thrust}
        \Flmaxnew = \Flmax - \sqrt{\left(\vecelem{F\ver^\frm{B}}{x}\right)^2 + \left(\vecelem{F\ver^\frm{B}}{y}\right)^2}
    \end{equation}

Next, the horizontal thrust $\FhorB$ is bounded to respect the new lateral thrust limit $\Flmaxnew$: 

    \begin{equation} \label{eq:control:thrust:lateral-thrust-output-bounding-horizontal}
        \vec{F}\hor^{'\frm{B}} =
        \frac{\Flmaxnew}{\sqrt{\left(\vecelem{F\hor^\frm{B}}{x}\right)^2 + \left(\vecelem{F\hor^\frm{B}}{y}\right)^2}} 
        \ \FhorB
    \end{equation}

Finally, the output thrust setpoint is calculated as:

    \begin{equation} \label{eq:control:thrust:new-Fspt-bounded-better}
        \FsptB = \FverB + \FhorBnew
    \end{equation}

A different approach taking advantage of the knowledge that the maximum lateral thrust is already being consumed is explained in~\cite{Mehmood2017}. The method (explained below) directly constructs the thrust setpoint rather than modifying the result of Equation~\ref{eq:control:thrust:thrust-setpoint-calculation}.

Let's assume that the angle between the current $\kB$ unit vector and the desired input thrust projected on $\PlaneB$ plane is $\gamma$. Knowing that the thrust force on the body-fixed horizontal plane $\PlaneB$ is at the maximum, the $x$ and $y$ components of the commanded force $\FsptB$ can be directly calculated as:

    \begin{equation} \label{eq:control:thrust:mehmood-xy}
        \begin{split}
        &\vecelem{F\spt^\frm{B}}{x} = \Flmax \cdot \cos \gamma \\
        &\vecelem{F\spt^\frm{B}}{y} = \Flmax \cdot \sin \gamma
        \end{split}
    \end{equation} 

\begin{figure}[!htb]
        \centering
        \includegraphics[width=0.5\textwidth]{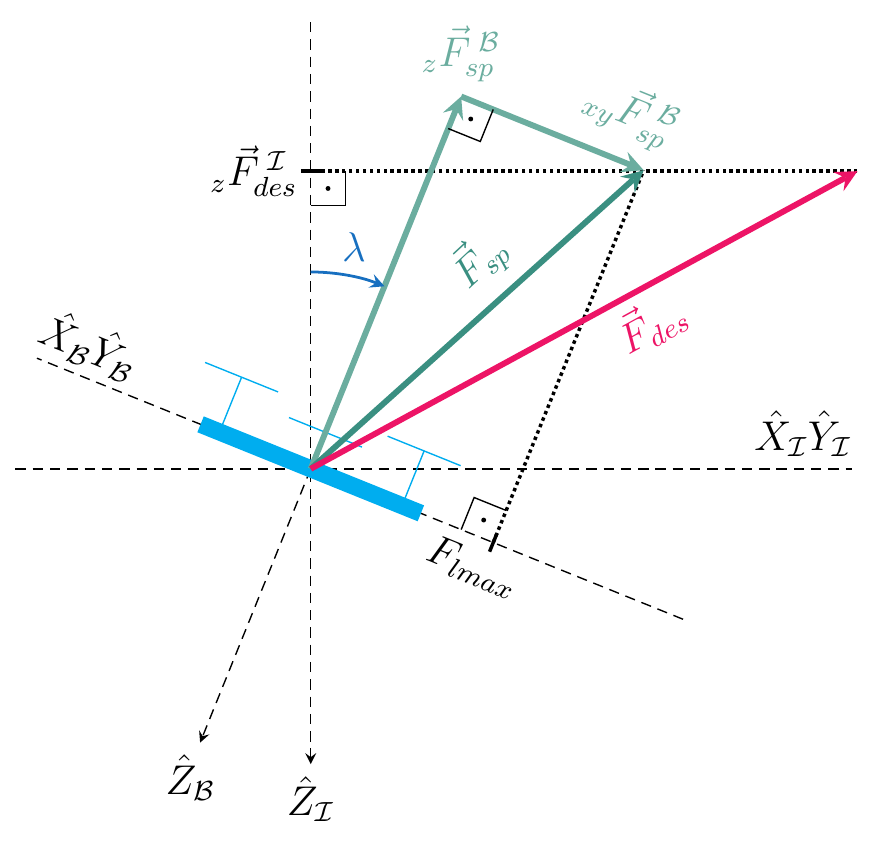}
        
        \caption[Model for handling lateral thrust limit]{An illustration of the model for handling the lateral thrust limit. If generating the desired thrust is not feasible, the desired vertical thrust is prioritized to maintain altitude stability. The illustrated vectors $\vecelem{\Fspt}{z}$ and $\vecelem{\Fspt}{xy}$ are the normal and lateral elements of $\Fspt$ in the \textit{current} body-fixed frame, respectively.}
        
        \label{fig:control:thrust:lateral-thrust-model}
    \end{figure}

As can be observed in Figure~\ref{fig:control:thrust:lateral-thrust-model}, the normal component of the thrust setpoint can be calculated by summing the projections of the vertical component of the desired thrust and the lateral thrust of the setpoint on the $\ZB$ axis. Therefore, assuming the current tilt $\lambda$ for the robot, we have:

    \begin{equation} \label{eq:control:thrust:mehmood-z}
        \vecelem{F\spt^\frm{B}}{z} = \vecelem{F\des^\frm{I}}{z} \sec \lambda - \| \vecelem{F\spt^\frm{B}}{xy} \| \tan \lambda = \vecelem{F\des^\frm{I}}{z} \sec \lambda - \Flmax \tan \lambda
    \end{equation}

Finally, from Equations~\ref{eq:control:thrust:mehmood-xy} and~\ref{eq:control:thrust:mehmood-z} the output thrust command is computed as:

    \begin{equation} \label{eq:control:min-tilt-Fspt}
        \FsptB = \begin{bmatrix} \vecelem{F\spt^\frm{B}}{x} \\ \vecelem{F\spt^\frm{B}}{y} \\ \vecelem{F\spt^\frm{B}}{z} \end{bmatrix}
    \end{equation}

A more straightforward solution exists if the robot's roll and pitch angles are close to zero, which transforms the problem to directly limiting the input's horizontal thrust. This can be achieved by pre-processing the input desired thrust $\FdesI$ to create a new thrust vector $\FdesInew$ with limited \textit{horizontal} thrust before rotating it to the body-fixed frame for thrust setpoint. 

Defining $\Fhor = \matrice{\vecelem{F\des^{\frm{I}}}{x} & \vecelem{F\des^{\frm{I}}}{y} & 0}\T$, the bounded horizontal thrust can be calculated as:

    \begin{equation} \label{eq:control:thrust:horizontal-thrust-bounding}
        \vec{F}\hor^{'\frm{I}} = \frac{\Flmax}{\left\| \Fhor \right\|} \Fhor
    \end{equation}

Replacing the horizontal thrust components with the bounded ones in the original input $\FdesI$, the new desired thrust can be constructed as:

    \begin{equation} \label{eq:control:thrust:horizontal-thrust-bounding-final}
        \FdesInew = \matrice{
        \vecelem{F\hor^{'\frm{I}}}{x}\\
        \vecelem{F\hor^{'\frm{I}}}{y} \\
        \vecelem{\FdesI}{z}}
    \end{equation}

The described approach is computationally efficient; however, it can only be used when the assumption of the robot's tilt being close to zero is valid (e.g., when using the zero-tilt attitude strategy described in Section~\ref{sec:control:attitude:zero-tilt}).

\subsection{Strategy 2: Keeping the Acceleration Directions}

The previously proposed solution guarantees that the output thrust setpoint has the same vertical component as the input desired thrust if it is feasible. While this tactic prevents undesired altitude changes, it may cause a severe reduction in the available lateral thrust for horizontal motion when the vertical thrust command is large, which may prove dangerous in extreme cases or in situations such as flying in the wind. To avoid those issues, the ratio of the horizontal and vertical accelerations can be maintained, or the horizontal acceleration can be given priority over the vertical acceleration.

Knowing the hover thrust $\Fhover$ (it can easily be estimated experimentally), if the $z$ component of the input desired thrust is larger than $\Fhover$, then the hover thrust vector $\FhovI = \matrice{0 & 0 & \Fhover}\T$ is rotated to the current body-fixed attitude to obtain the baseline for zero acceleration:

    \begin{equation} \label{eq:control:thrust:Fspt-thrust-limit-hover-thrust}
        \FhovB = \RBI \cdot \FhovI
    \end{equation} 

Accounting for the consumed lateral thrust by the hover thrust, the remaining lateral thrust $\Flmaxnew$ is computed as:

    \begin{equation} \label{eq:control:thrust:new-lateral-max-thrust-hover-thrust}
        \Flmaxnew = \Flmax - \sqrt{\left(\vecelem{F\hover^\frm{B}}{x}\right)^2 + \left(\vecelem{F\hover^\frm{B}}{y}\right)^2}
    \end{equation}

The rest of the input desired thrust ($\FdesInew = \FdesI - \FhovI$) is used for the output thrust calculation with its lateral bound limited to $\Flmaxnew$. The lateral of the thrust setpoint $\FsptBnew$ rotated from $\FdesInew$ using Equation~\ref{eq:control:thrust:thrust-setpoint-calculation} is computed as:

    \begin{equation} \label{eq:control:thrust:lateral-thrust-calculation-output}
        \left\| \Flat \right\| = 
        \left\| \begin{bmatrix}[1.5]
        \vecelem{F\spt^\frm{B}}{x} \\ \vecelem{F\spt^\frm{B}}{y} \\ 0
        \end{bmatrix}
        \right\|_2 = 
        \sqrt{\left(\vecelem{F\spt^\frm{B}}{x}\right)^2 + \left(\vecelem{F\spt^\frm{B}}{y}\right)^2}
    \end{equation}
    
If the resulting lateral thrust $\| \Flat \|$ is larger than $\Flmaxnew$, we can bound the lateral thrust:

    \begin{equation} \label{eq:control:thrust:lateral-thrust-output-bounding}
        \vec{F}\lat^{'\frm{B}} = \frac{\Flmaxnew}{\|\Flat\|} \Flat = 
        \frac{\Flmaxnew}{\sqrt{\left(\vecelem{F\spt^\frm{B}}{x}\right)^2 + \left(\vecelem{F\spt^\frm{B}}{y}\right)^2}} 
        \begin{bmatrix}[1.5]
        \vecelem{F\spt^\frm{B}}{x} \\ \vecelem{F\spt^\frm{B}}{y} \\ 0
        \end{bmatrix}
    \end{equation}    
    
The partial thrust setpoint with bounded lateral thrust is reconstructed as:

    \begin{equation} \label{eq:control:thrust:final-bounded-thrust}
        \FsptBnewnew = \begin{bmatrix}[1.5]
        \vecelem{F\lat^{'\frm{B}}}{x} \\ \vecelem{F\lat^{'\frm{B}}}{y} \\ \vecelem{F\spt^\frm{B}}{z} \end{bmatrix}
    \end{equation}
    
Finally, the result of Equation~\ref{eq:control:thrust:final-bounded-thrust} is combined with $\FhovB$ from Equation~\ref{eq:control:thrust:Fspt-thrust-limit-hover-thrust} to calculate the feasible thrust setpoint $\FsptB$.

\section{Extending the Controller for Physical Interaction} \label{sec:control:hpfc}

Many potential applications specific to the fully-actuated multirotors, e.g., our application of utility pole maintenance, include physical interaction with the environment. These applications range from non-destructive inspections to environment-altering tasks such as wire manipulation or moving objects. 

This section describes how the proposed controller design of this chapter can be extended to provide simultaneous position and force control of a multirotor with a rigidly attached end-effector during the physical interaction with the environment. We call the multirotor with a rigidly-attached end-effector as \textit{whole-body wrench generator}.

\begin{figure}[!htb]
        \centering
        \begin{subfigure}[b]{0.67\textwidth}
            \includegraphics[width=\textwidth]{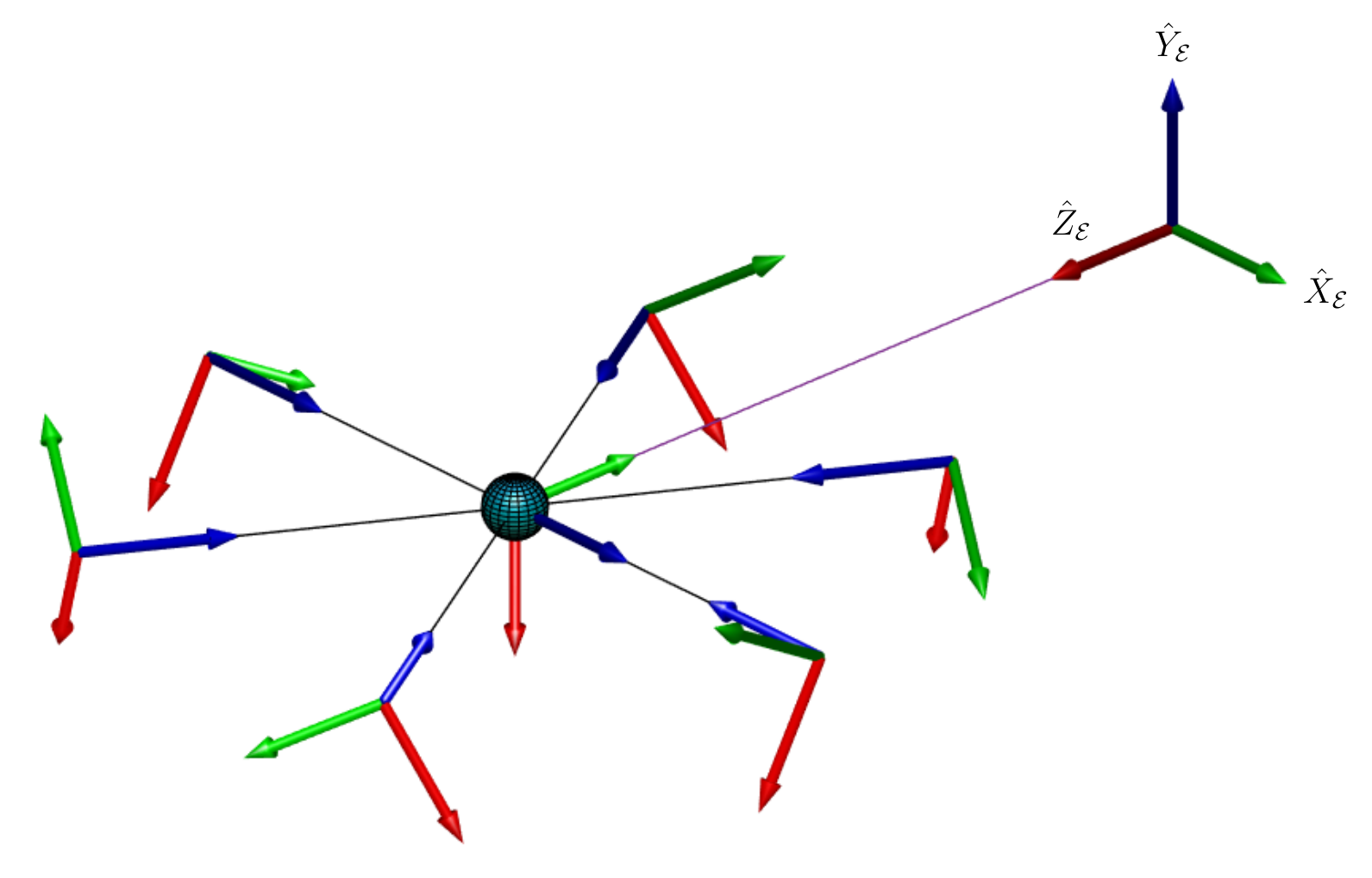}
            \caption{~}
            \label{fig:control:hpfc:frames-a-end-effector}
        \end{subfigure}
        \hfill
        \begin{subfigure}[b]{0.29\textwidth}
            \includegraphics[width=\textwidth]{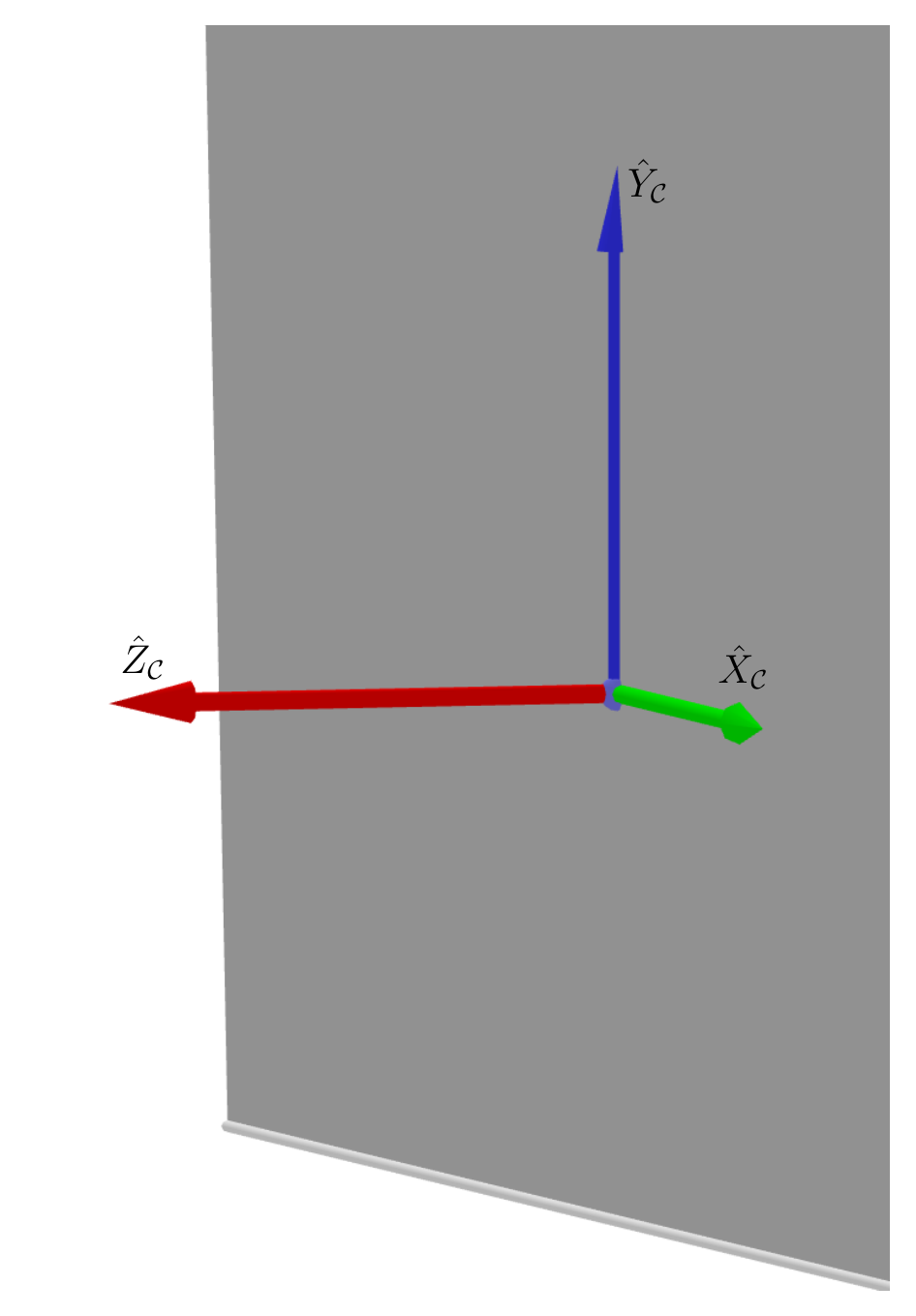}
            \caption{~}
            \label{fig:control:hpfc:frames-b-contact}
        \end{subfigure}
        
        \caption[Coordinate frames for the end effector and the contact point]{Coordinate frames used in a physical interaction: (a) End-effector. (b) The contact point on the surface.}
        
        \label{fig:control:hpfc:frames}
    \end{figure}

The end-effector (rigidly attached to the multirotor) is used for physical contact with the environment. Figure~\ref{fig:control:tests:tilthex-a} shows the end-effector on the hexarotor used in this project. In addition to the frames described in Section~\ref{sec:background:control:translational-kinematics}, there are two more useful frames in this problem:

\begin{itemize}

    \item The end-effector frame $\Frm{E}$ which is attached to the end-effector as illustrated in Figure~\ref{fig:control:hpfc:frames-a-end-effector}. Since the end-effector is rigidly attached to the UAV, its frame is fixed in the body-fixed frame. 
    
    \item The contact frame $\Frm{C}$ which is attached to the contact point as illustrated in Figure~\ref{fig:control:hpfc:frames-b-contact}. This frame moves with the contact point during the interaction, with its $\ZC$ axis always being normal to the surface.
\end{itemize}

Our assumptions in this section are:

\begin{enumerate}
    \item The contact surface is rigid, and the forces applied by the UAV are non-destructive and non-altering.
    
    \item The desired positions and forces are feasible for the UAV.
    
    \item There is direct feedback from the end-effector of the UAV. This feedback can be achieved by devising a force/torque sensor at the end-effector.
    
    \item There are no rotational constraints on the robot's motion at the point of contact.
\end{enumerate}

\subsection{Hybrid Position and Force Controller} \label{sec:control:hpfc:controller}

We implemented a hybrid position-force controller (HPFC) to control both the position and the applied force during contact with a planar surface. The subspace affected by each one at the point of contact is separated using two $3 \times 3$ matrices called \textit{selection matrices} to achieve independent control over the force and the position. Each row represents one of the 3-DoFs of space at the contact point.
    
The position selection matrix $\SP$ defines the directions in the contact frame that are free to move, and the force selection matrix $\SF$ defines the directions in which a force can be applied. The definition of $\SP$ and $\SF$ depends on two types of constraints: the natural constraints, which are due to the environment's geometry, and the artificial constraints, which depend on the task. For example, the end effector cannot move into the wall (along $-\ZC$), creating a natural constraint, and if the motion along $\XC$ is restricted based on the task, that is an artificial constraint.

The selection matrices are square diagonal matrices with only $0$ or $1$ elements, and if there are only natural constraints, the two matrices are complements:
    
    \begin{equation} \label{eq:control:hpfc:selection-matrix-complementary}
        \SP = \IdentMat{3} - \SF
    \end{equation}

\noindent where $\IdentMat{3}$ is the $3 \times 3$ identity matrix.

For example, if only natural constraints are present when facing a planar surface, the selection matrices would be as follows:

    \begin{equation} \label{eq:control:hpfc:selection-matrix-planar-surface}
        \SP = \matrice{1 & 0 & 0\\0 & 1 & 0\\0 & 0 & 0}, \quad  
        \SF = \matrice{0 & 0 & 0\\0 & 0 & 0\\0 & 0 & 1}
    \end{equation}

When artificial constraints are added, some non-zero diagonal elements become zero. Therefore, the Hadamard product of the two selection matrices will always stay zero:

    \begin{equation} \label{eq:control:hpfc:selection-matrices-product}
        \SP \odot \SF = \ZeroMat{3}
    \end{equation}

\noindent where $\ZeroMat{3}$ is the $3 \times 3$ zero matrix.

The selection matrices can be defined based on the task before the execution. As mentioned, they are utilized to separate the subspace for the applied force and the position at the contact frame. The contact frame $\FC$ (i.e., the rotation $\RIC$) can be computed from the normal of the contact surface and is arbitrary as long as it is consistent with the devised selection matrices. 

\begin{figure}[!htb]
        \centering
        \includegraphics[width=\linewidth]{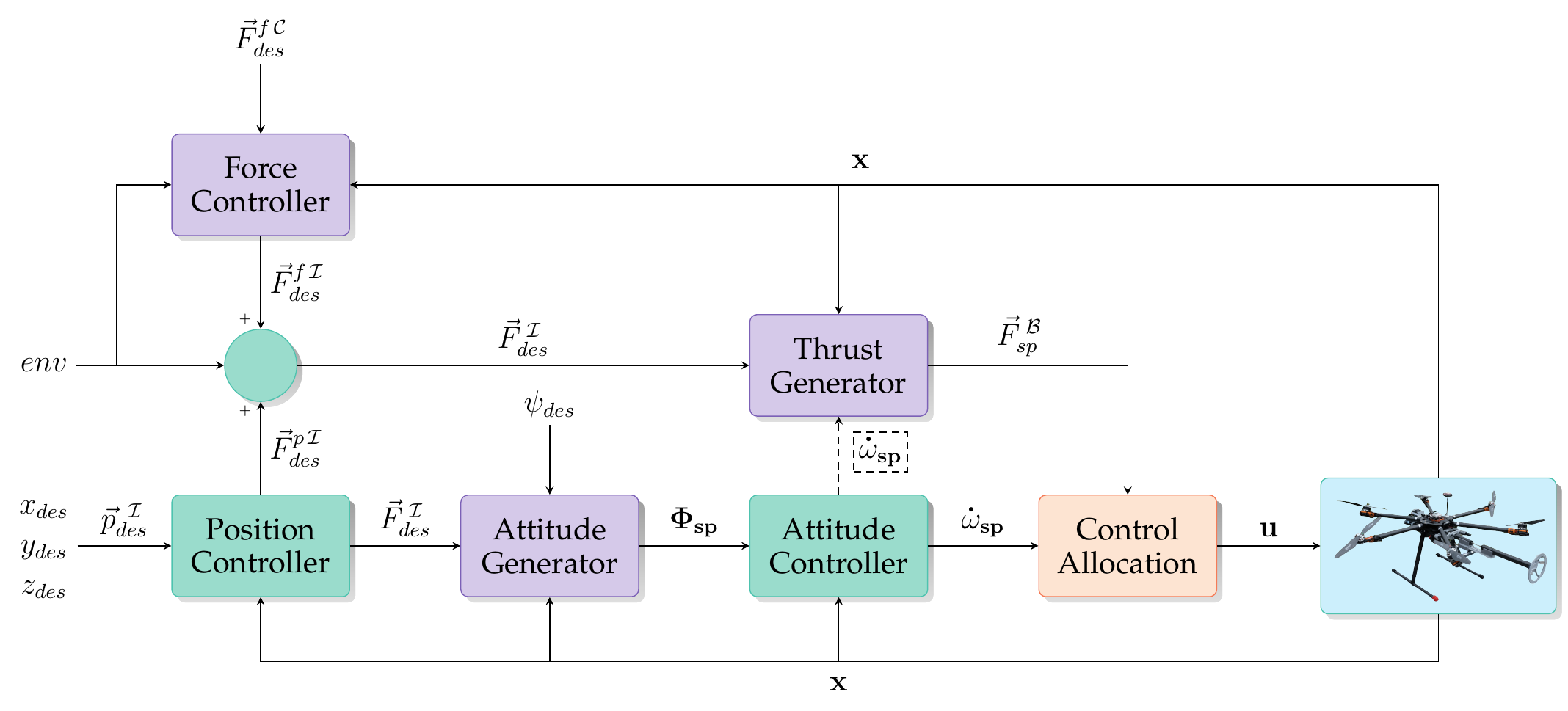}
        
        \caption[Hybrid Position-Force Controller architecture]{The design of our Hybrid Position-Force controller extended based on the controller architecture of Section~\ref{sec:control:design}. The force and position control modules independently calculate the necessary thrusts (accelerations) to achieve the desired inputs, then are combined based on their respective subspaces.}
        
        \label{fig:control:hpfc:hpf-controller}
    \end{figure}

We add a new Force Controller module to control the force during the contact. The module's inputs are the desired force to apply in the contact frame ($\FdesCf$), the environment information that includes the contact frame ($\RIC$), and the state information that includes the force feedback in the end-effector frame measured by the force sensor. Like the Position Controller module, the output is the force in the inertial frame, which is now combined with the Position Controller's output force based on the subspaces before feeding into the Thrust Setpoint Generator module. Figure~\ref{fig:control:hpfc:hpf-controller} illustrates the hybrid position-force controller architecture developed based on our free-flight controller of Figure~\ref{fig:control:controller}.

Note that the force controller computes the desired force in the $\FC$ frame, while the measured force feedback is in the $\FE$ frame. So, independent of the force control method, the force feedback should be transformed into the contact frame. Finally, after the output is calculated in the contact frame, the output is transformed into the inertial frame.

We used a PID controller to follow the reference force in our work. Since both the input and the desired output are of the same type, the PID loop generates changes to the last output and not the output itself.

Finally, the selection matrices are applied to the outputs of force and position controllers ($\FdesIf$ and $\FdesIp$, respectively) in the contact frame and the results are combined together to obtain the input for the Thrust Setpoint Generator module:

\begin{equation} \label{eq:control:hpfc:combine-output-forces}
    \FdesI = \RIC \left( \RCI \ \SP \ \FdesIp \ + \ \RCI \ \SF \ \FdesIf \right)
\end{equation}
    
Note that, since the Hadamard product of the two selection matrices is always zero, they transform $\FdesIp$ and $\FdesIf$ to orthogonal subspaces. Therefore, the outputs of position and force controllers can only affect their respective subspaces and do not affect each other, completely decoupling the force from the position.

\section{Experiments and Results} \label{sec:control:tests}

\subsection{Hardware and Software} \label{sec:control:tests:hardware-software}

We have tested the proposed controller both in simulation and on real robots. 

\begin{figure}[!htb]
        \centering
        \includegraphics[width=0.8\linewidth]{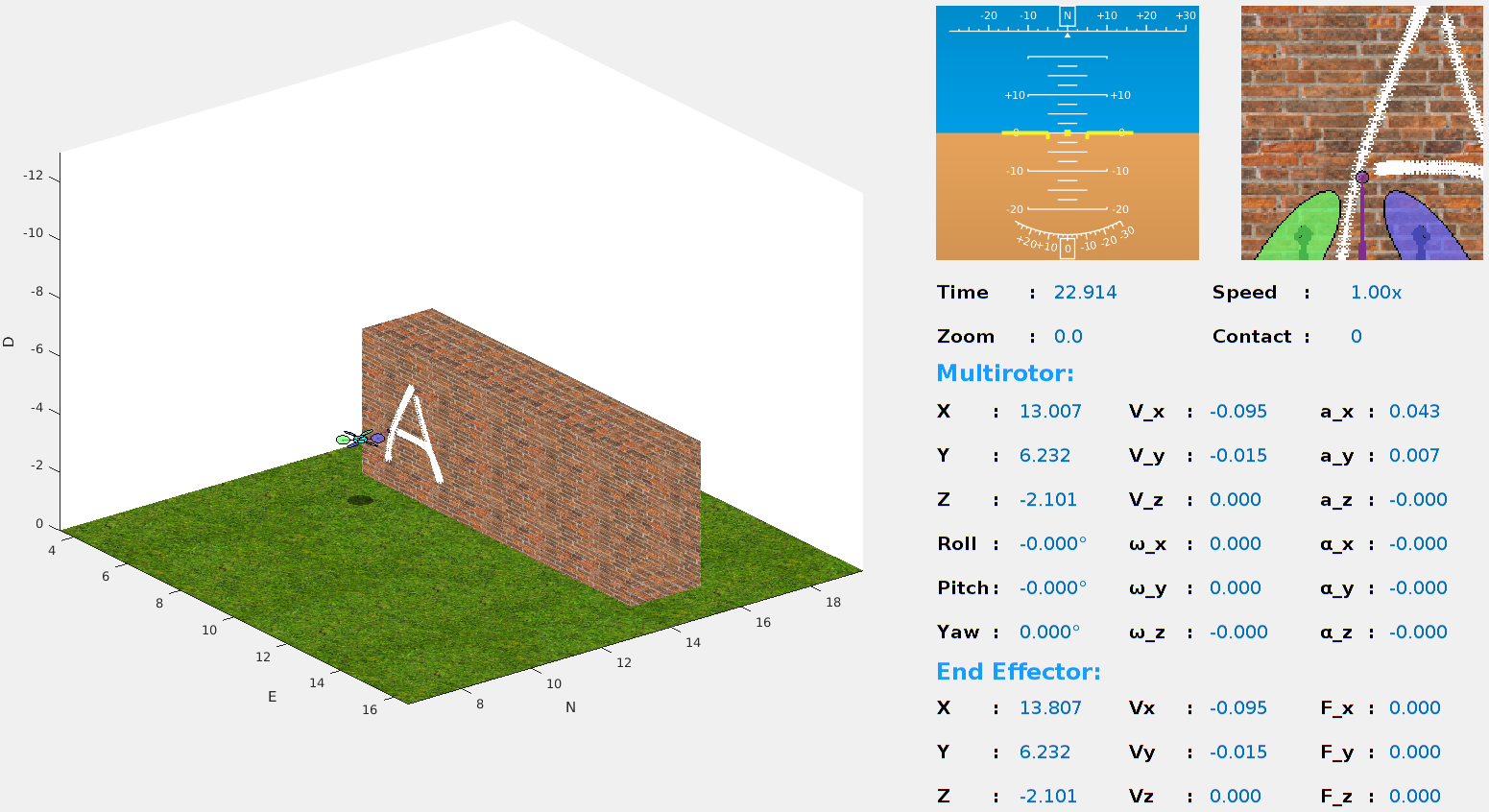}
        
        \caption[Screenshot of the MATLAB simulation with our hexarotor]{A screenshot of the MATLAB simulator for the fully-actuated hexarotor used in our experiments. This environment is used to develop, analyze, and test new ideas, architectures, and control methods.}
        
        \label{fig:control:tests:sim-matlab-sample}
    \end{figure}

\begin{figure}[!htb]
        \centering
        \includegraphics[width=0.8\linewidth]{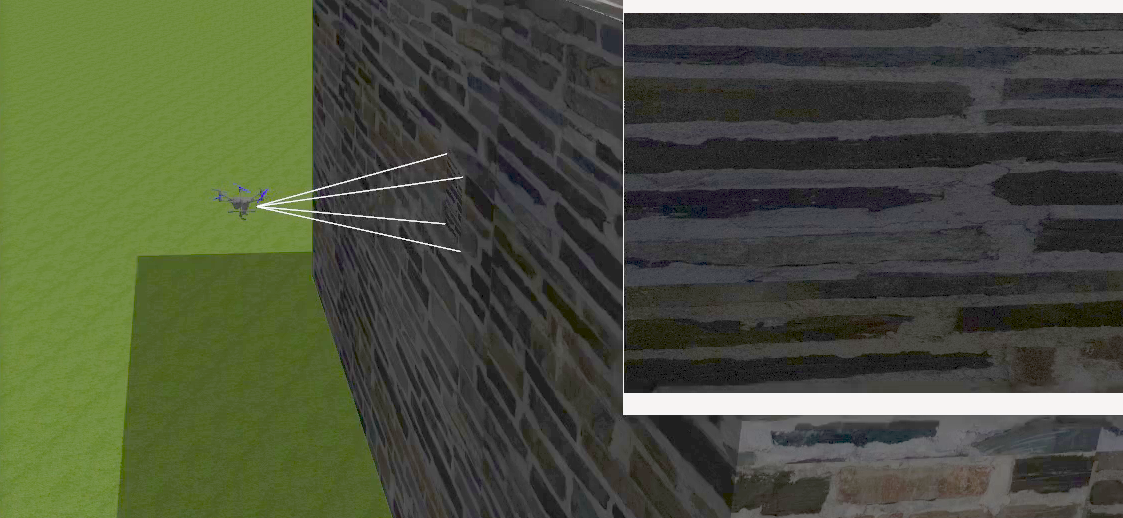}
        
        \caption[Screenshot of the Gazebo PX4 SITL with our hexarotor]{A screenshot of the PX4 SITL Gazebo simulator with the fully-actuated hexarotor used in our experiments. This environment is used for firmware development and testing the autopilot before deploying the code on the real UAV.}
        
        \label{fig:control:tests:sim-gazebo-sample}
    \end{figure}

\begin{figure}[!htb]
        \centering
        \begin{subfigure}[b]{0.6\textwidth}
            \includegraphics[width=\textwidth]{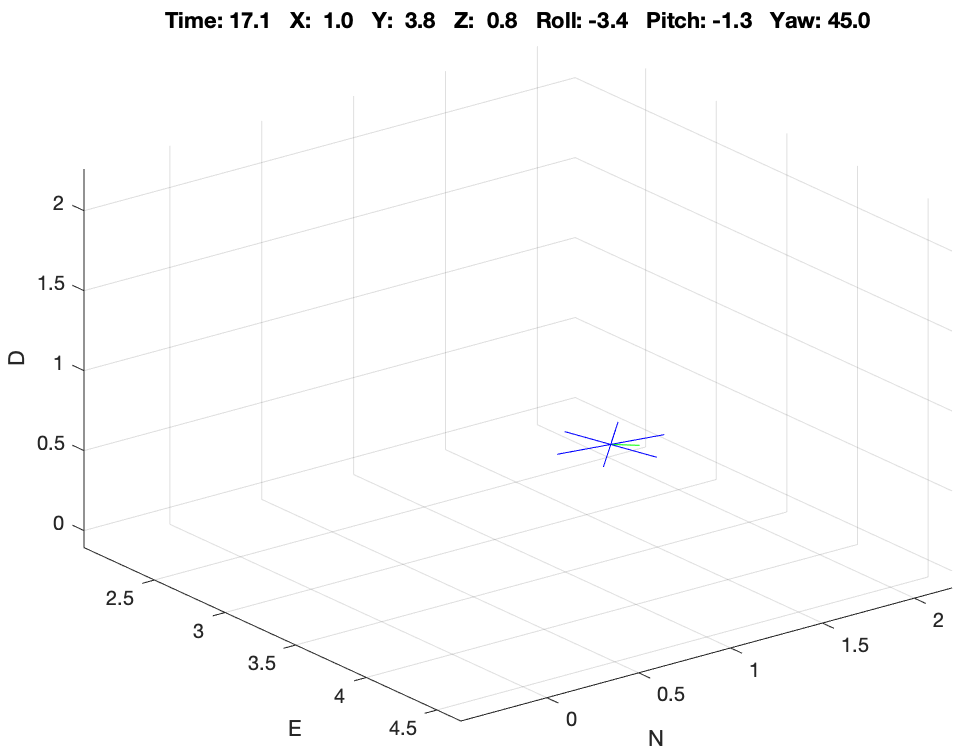}
            \caption{~}
            \label{fig:control:tests:sim-simulink-sample-a}
        \end{subfigure}
        
        \medskip

        \begin{subfigure}[b]{\linewidth}
            \includegraphics[width=\textwidth]{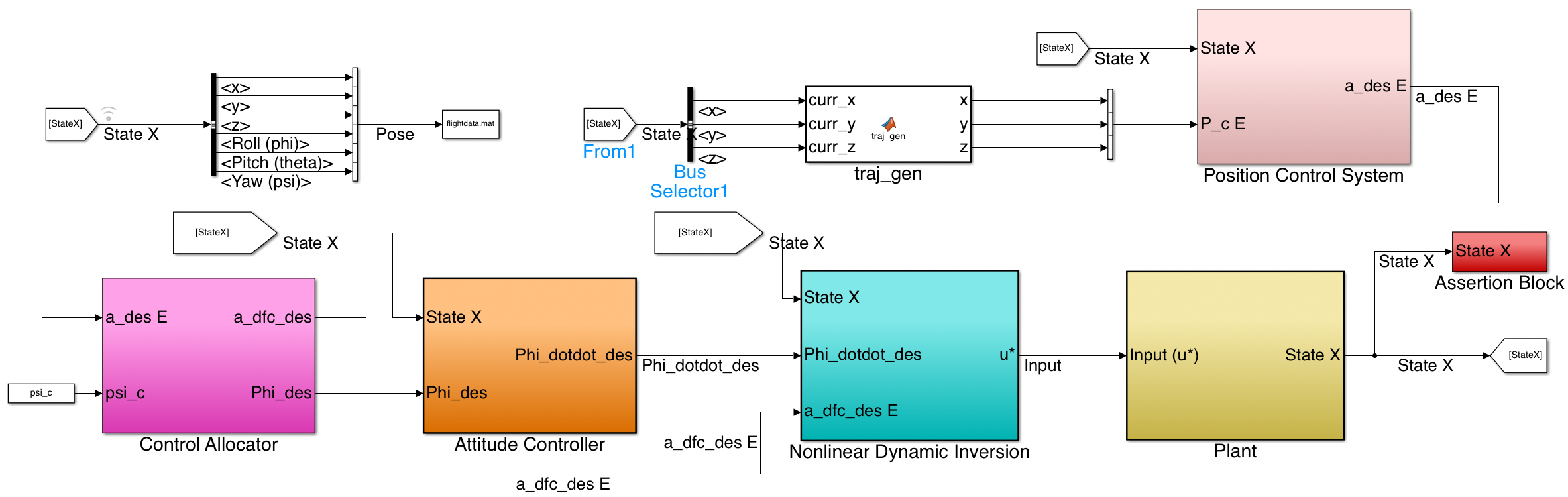}
            \caption{~}
            \label{fig:control:tests:sim-simulink-sample-c}
        \end{subfigure}

        \caption[The Simulink simulation for our hexarotor]{The Simulink simulation developed for testing the controller of the fully-actuated hexarotor used in our experiments. (a) A snapshot of the visualization of a trajectory following simulation. (b) The overall model of our controller and simulation.}
        
        \label{fig:control:tests:sim-simulink-sample}
    \end{figure}

\begin{figure}[!htb]
        \centering
        \begin{subfigure}[b]{0.8\textwidth}
            \includegraphics[width=\textwidth]{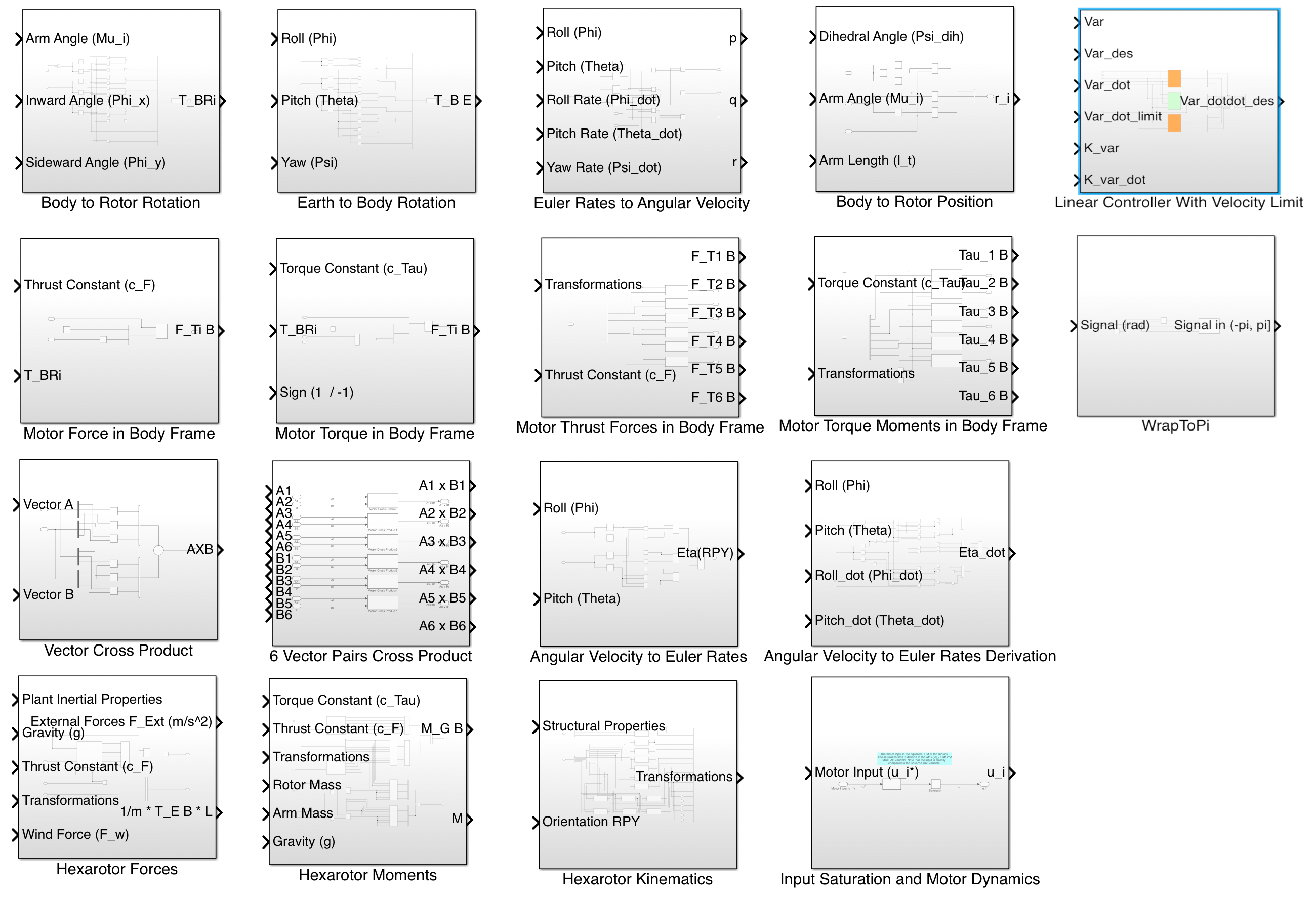}
            \caption{~}
            \label{fig:control:tests:sim-simulink-sample-b}
        \end{subfigure}

        \medskip

        \begin{subfigure}[b]{\textwidth}
            \includegraphics[width=\textwidth]{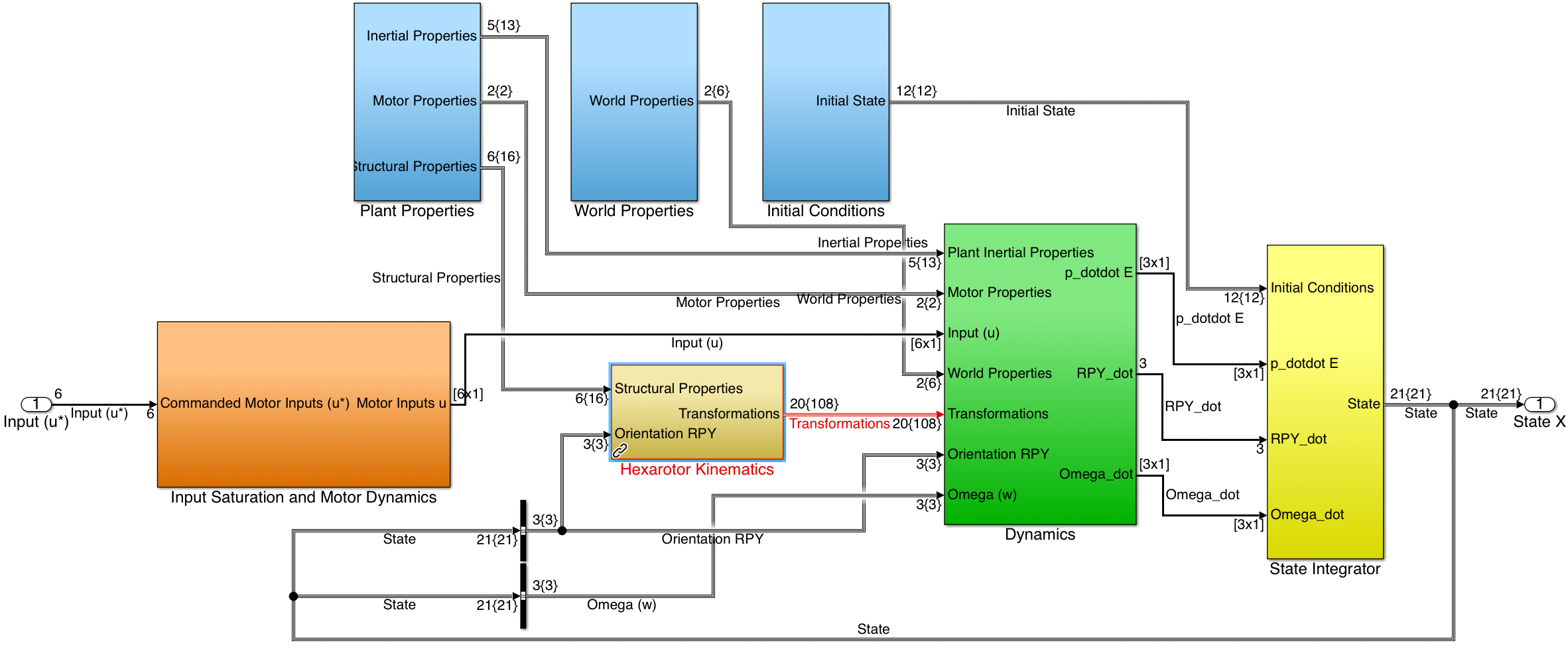}
            \caption{~}
            \label{fig:control:tests:sim-simulink-sample-d}
        \end{subfigure}
        
        \caption[Our developed Simulink library]{The Simulink library developed for testing the controller of the fully-actuated hexarotor used in our experiments. (a) The library implemented to enable rapid testing of different architectures and controllers. (b) The  fully-actuated hexarotor model made with our library. }
        
        \label{fig:control:tests:sim-simulink-library}
    \end{figure}

Three simulation environments were devised for testing the new fully-actuated UAV development and the methods described in this section: 

\begin{enumerate}
    \item \textbf{MATLAB simulator:} We developed a complete simulator for fully-actuated UAV controller development in MATLAB, which allows us quickly define, analyze and visualize new architectures and control methods. Most of the analysis of different architectures and methods in this chapter is done with this simulator. Figure~\ref{fig:control:tests:sim-matlab-sample} shows a snapshot of our multirotor flying in this simulation environment.

    \item \textbf{Gazebo simulator with PX4 SITL:} Our UAV model is developed in the Gazebo simulator, and the SITL simulation provided by the base PX4 firmware is enhanced in our code to support fully-actuated vehicles. Considering that the code-base in this simulation is the real robot's firmware, it is used for testing the code developed on the autopilot before performing tests on the real robot. Additionally, it is used for testing the onboard software developed for different missions and tasks. Figure~\ref{fig:control:tests:sim-gazebo-sample} shows the screenshot of this simulation environment. 
    
    \item \textbf{MATLAB Simulink model:} A Simulink model of the controller for our hexarotor is designed for faster simulation when required. A library is designed to allow faster model changes to test new ideas and architectures. Figures~\ref{fig:control:tests:sim-simulink-sample} and~\ref{fig:control:tests:sim-simulink-library} show the overall multirotor and controller models, as well as the developed library and a screenshot of the visualization.
\end{enumerate}

Some fixed-tilt hexarotors with two different frame sizes but similar designs are built and tested. All rotors on the robots are rotated sideways, alternatively for 30, and -30 degrees, similar to the design described in~\cite{Mehmood2017}. 

\begin{figure}[!htb]
        \centering
        \begin{subfigure}[b]{0.48\textwidth}
            \includegraphics[width=\textwidth]{ch-control/tilthex-nea.png}
            \caption{~}
            \label{fig:control:tests:tilthex-a}
        \end{subfigure}
        ~
        \begin{subfigure}[b]{0.48\textwidth}
            \includegraphics[width=\textwidth]{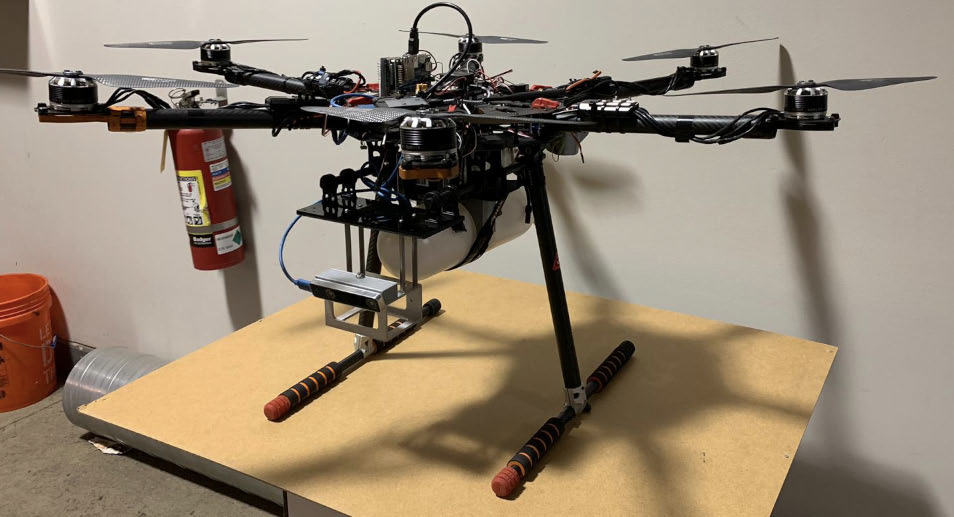}
            \caption{~}
            \label{fig:control:tests:tilthex-b}
        \end{subfigure}
            
        \caption[The fixed-pitch UAVs used in our experiments]{The fixed-pitch hexarotors used in our experiments. (a) The larger design with Tarot T960 frame. (b) The smaller design with Tarot X6 frame.}
        
        \label{fig:control:tests:tilthex}
    \end{figure}

The main body frame for the larger robot design is Tarot T960 with KDE-3510XF-475 motors, KDE-UAS35HVC electronic speed controllers (ESCs), and 14-inch propellers. It is equipped with a mRo Pixracer autopilot, an Nvidia Jetson TX2 onboard computer, a u-Blox Neo-M8N GPS module, Futaba T10J transmitter/receiver, and a 900~MHz radio for communication with the Ground Control Station computer. Figure~\ref{fig:control:tests:tilthex-a} shows the larger hexarotor.

The smaller robot is built on the Tarot X6 frame with a 0.96~$\unit{m}$ motor-to-motor diameter and a maximum payload of 7.5~$\unit{kg}$. The frame can be size-adjusted for specific tasks. Propulsion is achieved with six KDE Direct KDE-4215XF-465 brushless motors and KDEXF-UAS55HVC electronic speed controllers (ESCs). It is equipped with the same flight controller, onboard computer, and GPS module as the larger UAV. Additionally, Futaba T8J and T10J transmitters/receivers are used for the pilot's manual control. Figure~\ref{fig:control:tests:tilthex-b} shows this smaller hexarotor.

For the outdoor tests, the UAVs are also equipped with Intel T265 RealSense cameras for visual odometry and D435 RealSense cameras for RGB and depth imaging.

An Opti-Track system is used for pose estimation in indoor tests, which requires several reflective markers attached to the UAV. 

\begin{figure}[!htb]
        \centering
        \begin{subfigure}[b]{0.40\textwidth}
            \includegraphics[width=\linewidth]{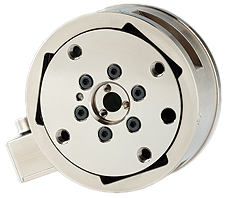}

            \caption{~}
            \label{fig:control:tests:gamma-sensor}
        \end{subfigure}
        \hfill
        \begin{subfigure}[b]{0.57\textwidth}
            \includegraphics[width=\linewidth]{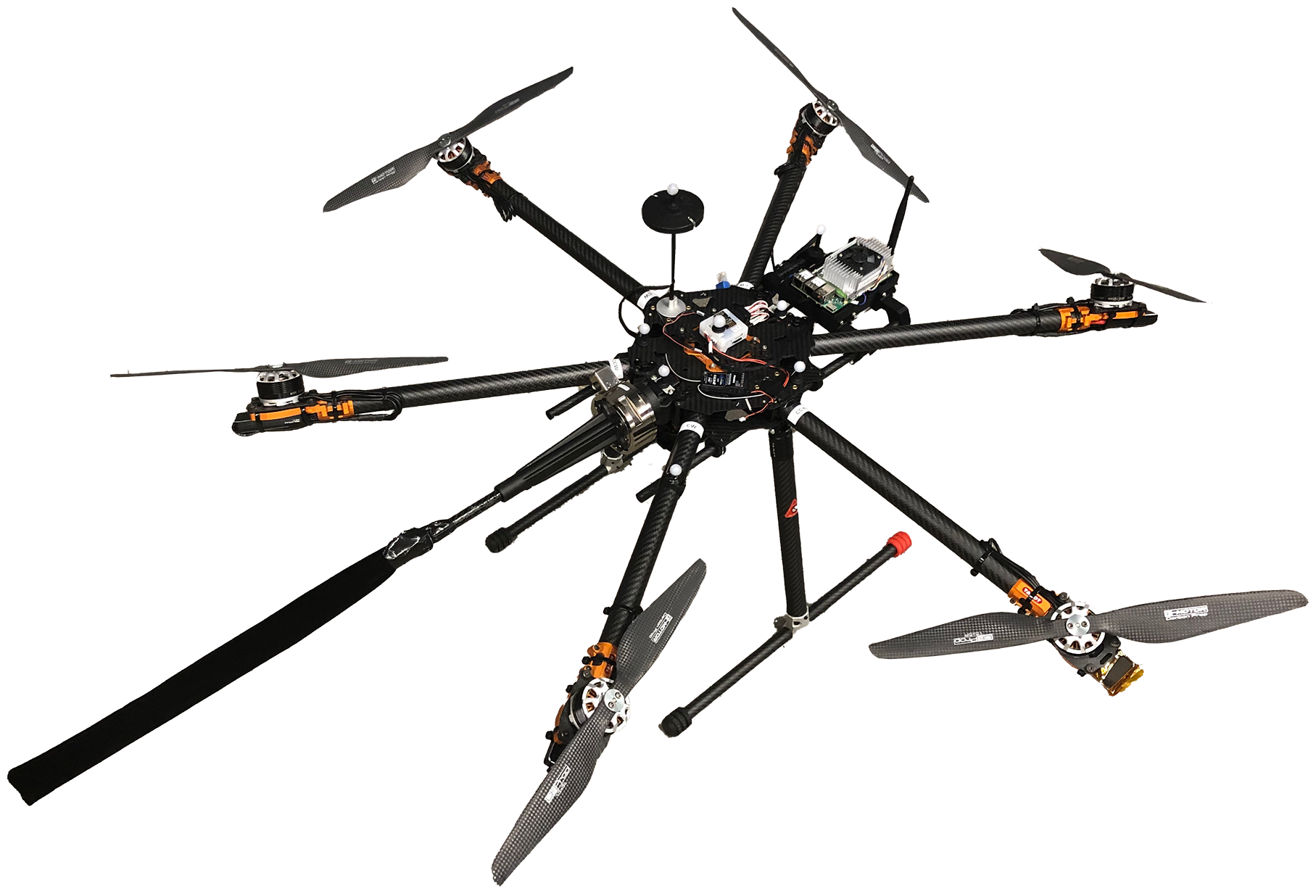}
            \caption{~}
            \label{fig:control:tests:ft-sensor-on-uav}
        \end{subfigure}
            
        \caption[ATI Gamma F/T sensor and our robot with the sensor attached]{(a) Illustration of the ATI Gamma Force/Torque Sensor used for the force control tests~\cite{atigamma}. (b) Our fully-actuated multirotor with the force/torque sensor attached.}
        
        \label{fig:control:tests:ft-sensor}
    \end{figure}

Additionally, for the force and hybrid motion-force control tests, the ATI Gamma Force/Torque sensor with the Digital Interface~\cite{atigamma} is attached to our robot to measure the forces applied at the end-effector. Tool transformation ensures the correct force/torque measurements at the end-effector instead of the sensor. Figure~\ref{fig:control:tests:gamma-sensor} shows the sensor and how it is attached to the UAV.

The onboard computers on the robots run Linux Ubuntu 18.04 (Bionic Beaver) with Robot Operating System (ROS) Melodic Morenia. Depending on the task at hand, different software packages run to plan and control the missions and trajectories.

We extended the PX4~v1.11.0 firmware to support our fully-actuated vehicles and implemented the methods presented in this section. Figure~\ref{fig:control:controller} illustrates the controller architecture of our developed firmware.

\subsection{Experiments} \label{sec:control:tests:tests}

\begin{figure}[!htb]
        \centering
        \includegraphics[width=0.6\linewidth]{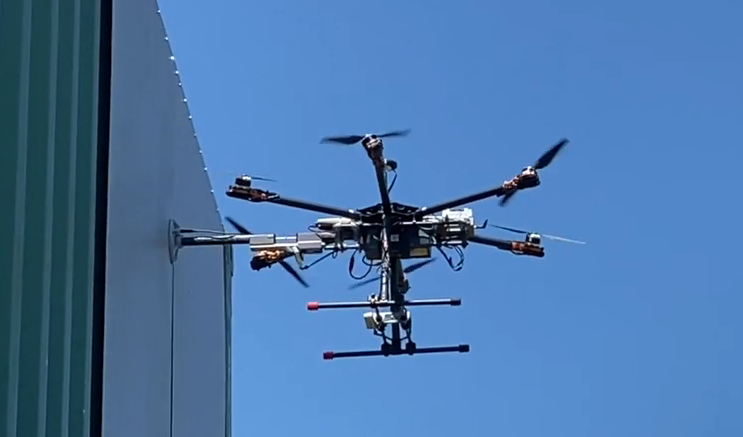}
        \caption[Unmodeled UAV contact with the wall]{A fully-actuated hexarotor using the PX4 controller extended with our proposed design making an unmodeled contact with the wall.}
        \label{fig:control:tests:contact-with-wall}
    \end{figure}

We have performed tens of indoor and outdoor flights with our fully-actuated hexarotor platform running the proposed controller and strategies. The experiments include real and simulated flights in free flight and during physical contact with the environment. Figure~\ref{fig:control:tests:contact-with-wall} shows our UAV during contact with a wall to measure the properties of the contact point using an ultrasonic sensor. The contact is unmodeled, but the UAV can keep its zero-tilt attitude and reject the disturbances. 

\subsubsection{Free Flight Experiments}

The proposed controller design is tested in both the MATLAB and Gazebo simulations and on our robot.

\begin{figure}[!htb]
        \centering
        \begin{subfigure}[b]{0.49\textwidth}
            \includegraphics[width=\textwidth]{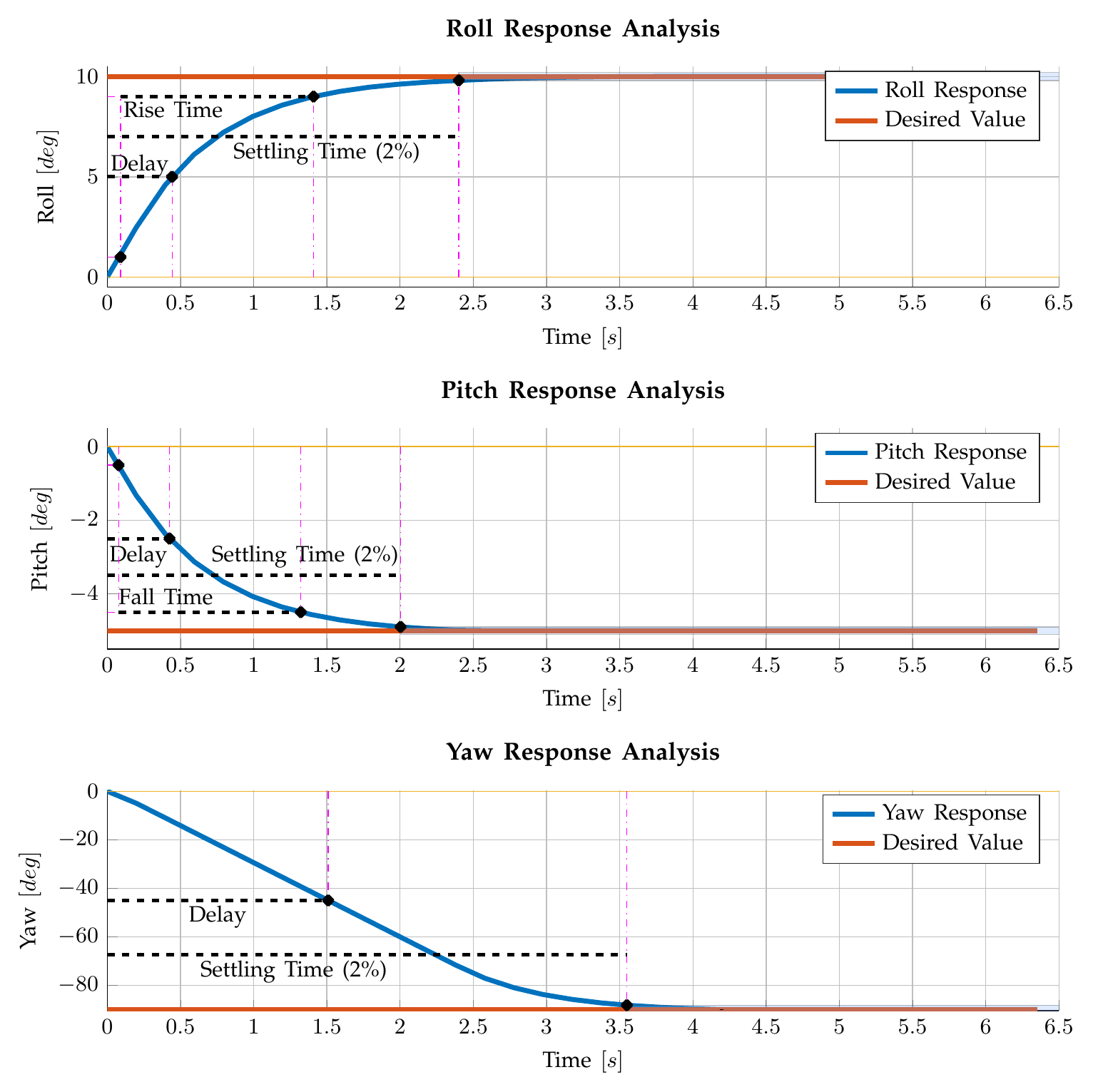}
            \caption{~}
            \label{fig:control:tests:tilthex-response-a-attitude}
        \end{subfigure}
        \hfill
        \begin{subfigure}[b]{0.49\textwidth}
            \includegraphics[width=\textwidth]{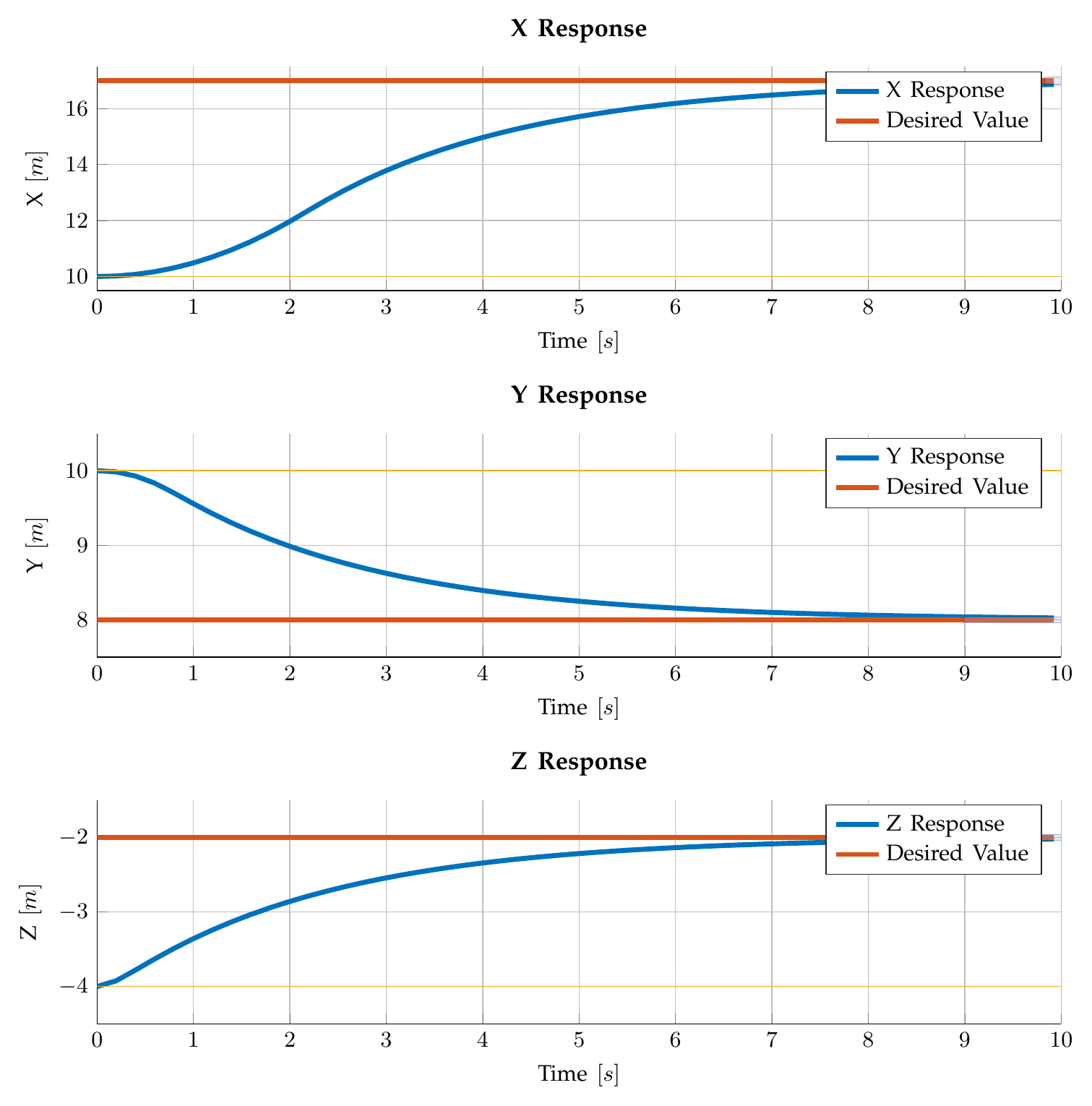}
            \caption{~}
            \label{fig:control:tests:tilthex-response-b-position}
        \end{subfigure}
            
        \caption[Attitude and position responses of the fixed-tilt hexarotor]{Responses of the fixed-tilt hexarotor used in this project to attitude and position commands simulated in the MATLAB simulator. (a) Attitude response to $\matrice{10 & -5 & -90}\T$ roll-pitch-yaw command starting from $\matrice{0 & 0 & 0}\T$. (b) Position response to $\matrice{17 & 8 & -2 & -45}\T$ x-y-z-yaw command starting from $\matrice{10 & 10 & -4 & 0}\T$.}
        
        \label{fig:control:tests:tilthex-response}
    \end{figure}

Figure~\ref{fig:control:tests:tilthex-response} shows the attitude and position responses of our fixed-pitch hexarotor model in the developed MATLAB simulator. After tuning the underlying PID controller gains, we were able to get good responses to the commands.

\begin{figure}[!htb]
        \centering
        \begin{subfigure}[b]{0.3\textwidth}
            \includegraphics[width=\textwidth]{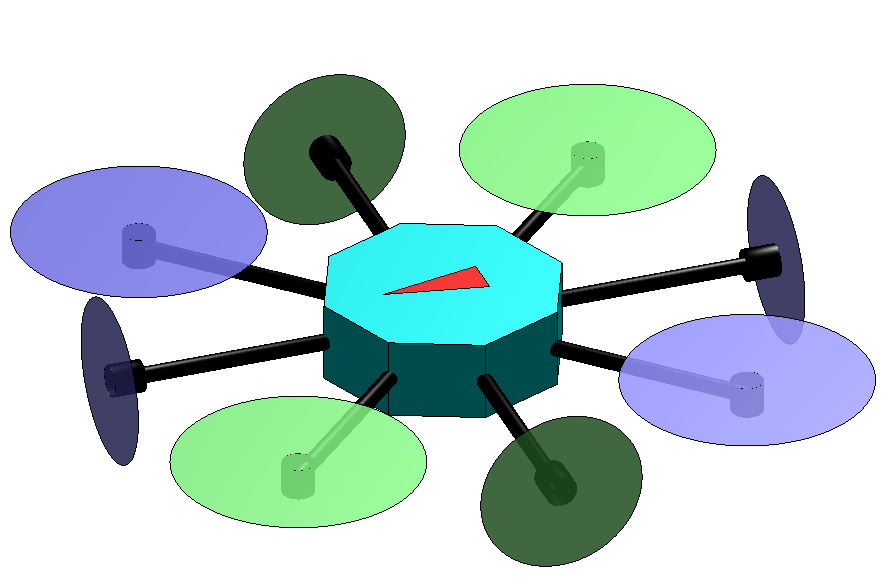}
            \caption{~}
            \label{fig:control:tests:octorotor-response-a-octorotor}
        \end{subfigure}
        \hfill
        \begin{subfigure}[b]{0.67\textwidth}
            \includegraphics[width=\textwidth, height=10cm]{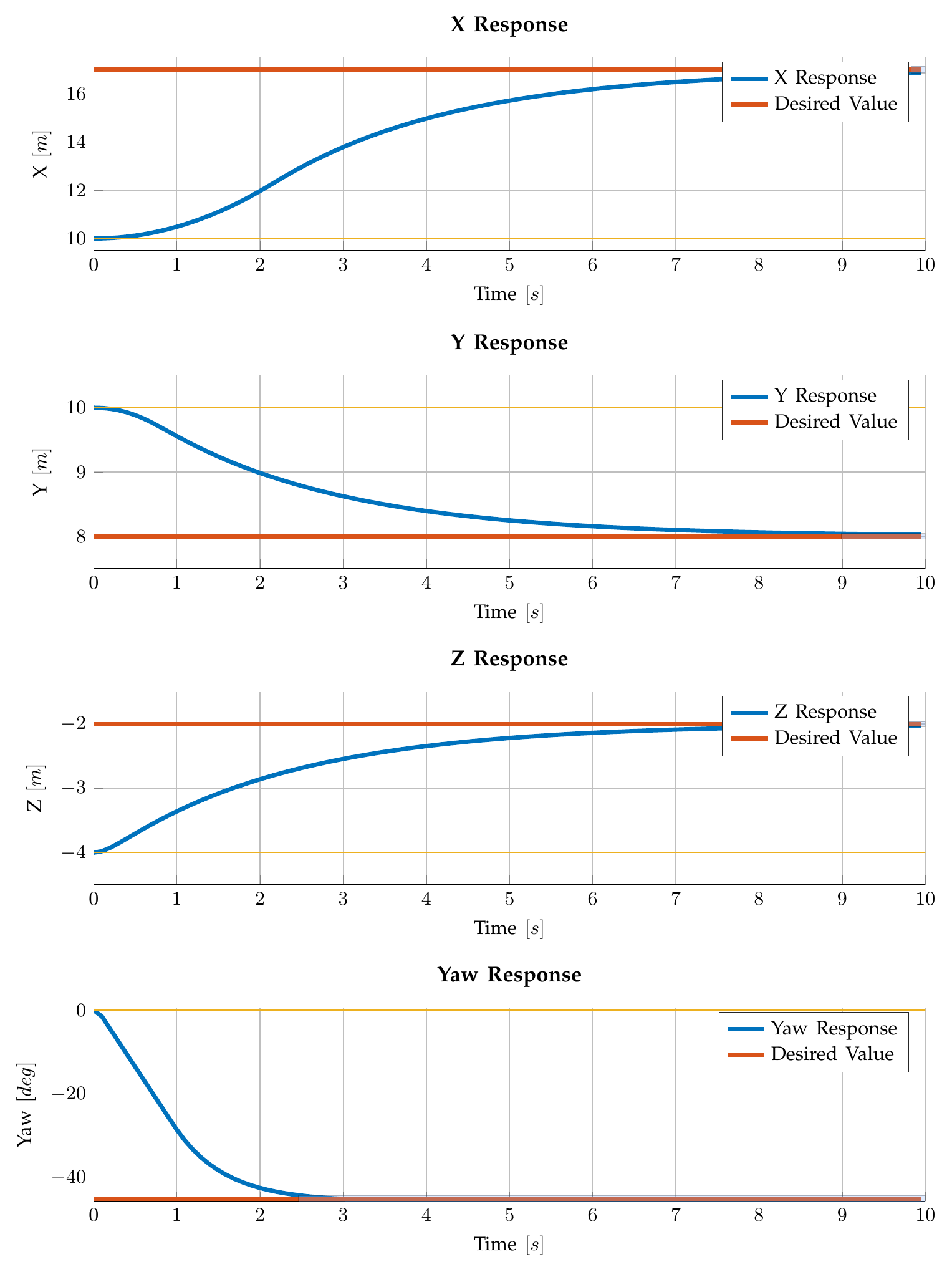}
            \caption{~}
            \label{fig:control:tests:octorotor-response-b-position}
        \end{subfigure}
            
        \caption[Position-yaw response of the fully-actuated octorotor]{(a) A fully-actuated octorotor with four co-planar upward rotors and four auxiliary motors perpendicular to the main rotors. (b) The position-yaw response to $\matrice{17 & 8 & -2 & -45}\T$ x-y-z-yaw command starting from $\matrice{10 & 10 & -4 & 0}\T$.}
        
        \label{fig:control:tests:octorotor-response}
    \end{figure}

Figure~\ref{fig:control:tests:octorotor-response} shows the position responses of a fully-actuated octorotor with four co-planar upward rotors and four auxiliary motors perpendicular to the main rotors modeled in our MATLAB simulator. The response shows similar results to the fixed-pitch hexarotor of our project.

All the proposed attitude strategies have been implemented in simulation and for the real robot.

We used the same trajectory with two waypoints to simulate all five attitude strategies in our MATLAB simulator to compare the results. The starting point is inertial zero with a zero attitude. The waypoints are $\matrice{2 & 2 & -4 & 0}\T$ and $\matrice{2 & 6 & -3 & 30}\T$, respectively. The waypoints' elements are $x$, $y$, $z$, and $yaw$ in degrees. Our project's hexarotor with fixed-pitch arms is used in all the trials.

\begin{figure}[!htb]
        \centering
        \begin{subfigure}[b]{0.49\linewidth}
            \includegraphics[width=\textwidth]{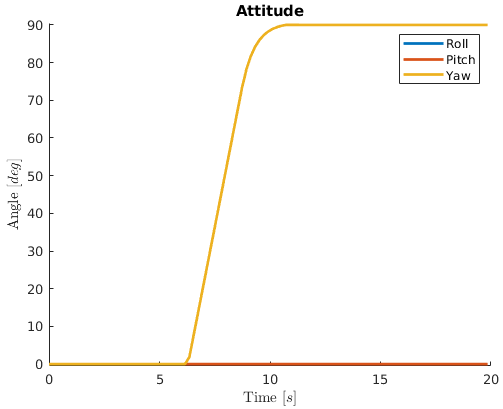}
            \caption{~}
            \label{fig:control:tests:tilthex-zero-tilt-strategy-matlab-rpy}
        \end{subfigure}
        \hfill
        \begin{subfigure}[b]{0.49\linewidth}
            \includegraphics[width=\textwidth]{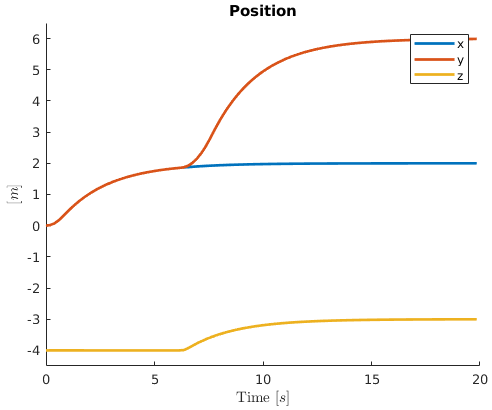}
            \caption{~}
            \label{fig:control:tests:tilthex-zero-tilt-strategy-matlab-pos}
        \end{subfigure}
        
        \caption[Trajectory simulation with zero-tilt attitude strategy]{The MATLAB simulation of the trajectory followed by the fixed-pitch hexarotor of our project using the zero-tilt attitude strategy. The tilt in this strategy is always zero to keep the robot horizontally level.}
        
        \label{fig:control:tests:tilthex-zero-tilt-strategy-matlab}
    \end{figure}

Figure~\ref{fig:control:tests:tilthex-zero-tilt-strategy-matlab} presents the attitude and position plots for zero-tilt strategy. The tilt in this strategy is always zero to keep the robot horizontally level. As a result, when the desired acceleration is high, it cannot be achieved.

\begin{figure}[!htb]
        \centering
        \begin{subfigure}[b]{0.49\linewidth}
            \includegraphics[width=\textwidth]{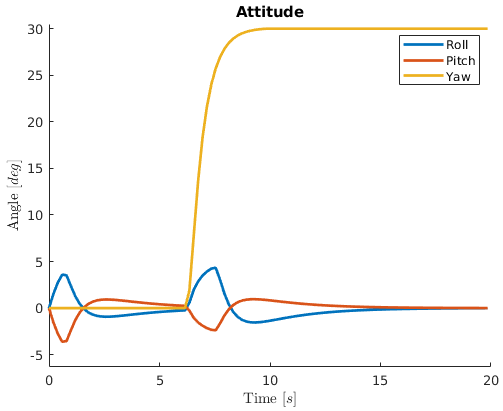}
            \caption{~}
            \label{fig:control:tests:tilthex-full-tilt-strategy-matlab-rpy}
        \end{subfigure}
        \hfill
        \begin{subfigure}[b]{0.49\linewidth}
            \includegraphics[width=\textwidth]{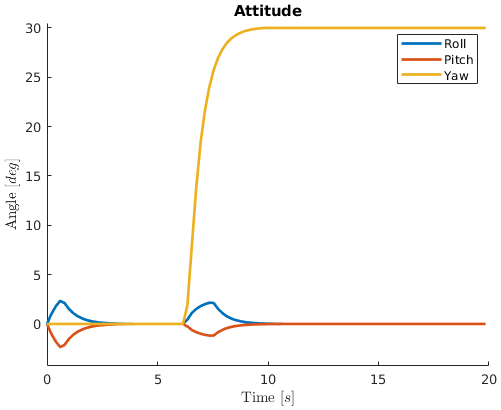}
            \caption{~}
            \label{fig:control:tests:tilthex-min-tilt-strategy-matlab-rpy}
        \end{subfigure}
        
        \caption[Trajectory simulation with full-tilt and minimum-tilt strategies]{The MATLAB simulation of the trajectory followed by the fixed-pitch hexarotor of our project using the (a) full-tilt attitude strategy. (b) The minimum-tilt attitude strategy.}
        
        \label{fig:control:tests:tilthex-full-min-tilt-strategy-matlab}
    \end{figure}

Figure~\ref{fig:control:tests:tilthex-full-min-tilt-strategy-matlab} shows the attitude plots for full-tilt and minimum-tilt strategies for the same trajectory. The tilt direction in the full-tilt strategy is always towards the direction of acceleration. The minimum-tilt strategy tends to keep the tilt at zero when the acceleration is low, but the robot starts tilting towards the acceleration direction when the command is higher. Therefore, the tilt is generally lower than in the full-tilt strategy. However, the acceleration is higher than in the zero-tilt strategy and is similar to the full-tilt strategy.

\begin{figure}[!htb]
        \centering
        \begin{subfigure}[b]{0.49\linewidth}
            \includegraphics[width=\textwidth]{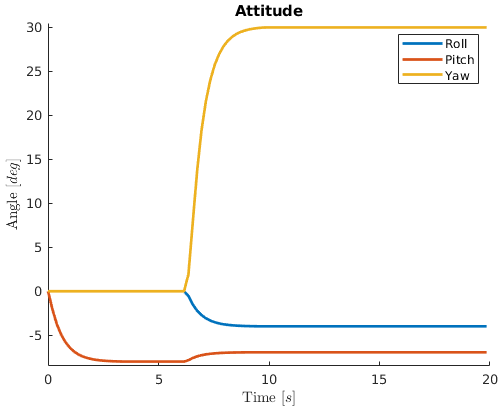}
            \caption{~}
            \label{fig:control:tests:tilthex-fixed-tilt-strategy-matlab-rpy}
        \end{subfigure}
        \hfill
        \begin{subfigure}[b]{0.49\linewidth}
            \includegraphics[width=\textwidth]{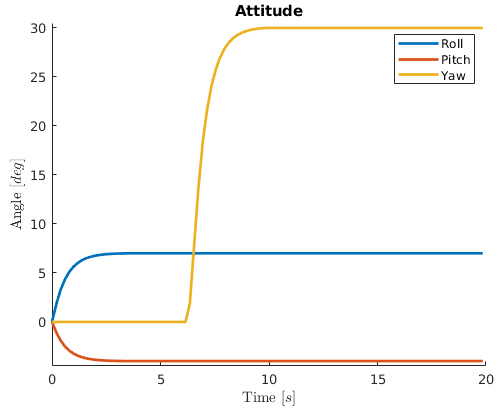}
            \caption{~}
            \label{fig:control:tests:tilthex-fixed-attitude-strategy-matlab-rpy}
        \end{subfigure}
        
        \caption[Trajectory simulation with fixed-tilt and fixed-attitude strategies]{The MATLAB simulation of the trajectory followed by the fixed-pitch hexarotor of our project using (a) The fixed-tilt attitude strategy. (b) The fixed-attitude strategy.}
        
        \label{fig:control:tests:tilthex-fixed-strategies-matlab}
    \end{figure}

Figure~\ref{fig:control:tests:tilthex-fixed-strategies-matlab} shows the attitude plots for fixed-tilt and fixed-attitude strategies for the same trajectory. The tilt's direction is set to the north in the fixed-tilt strategy, with an $8$-degree tilt. The UAV tends to keep the tilt angle and direction the same, even during and after the turn. The fixed-attitude strategy has $7$ degrees of roll and $-4$ degrees of pitch.

\begin{figure}[!htb]
        \centering
        \begin{subfigure}[b]{0.49\linewidth}
            \includegraphics[width=\textwidth]{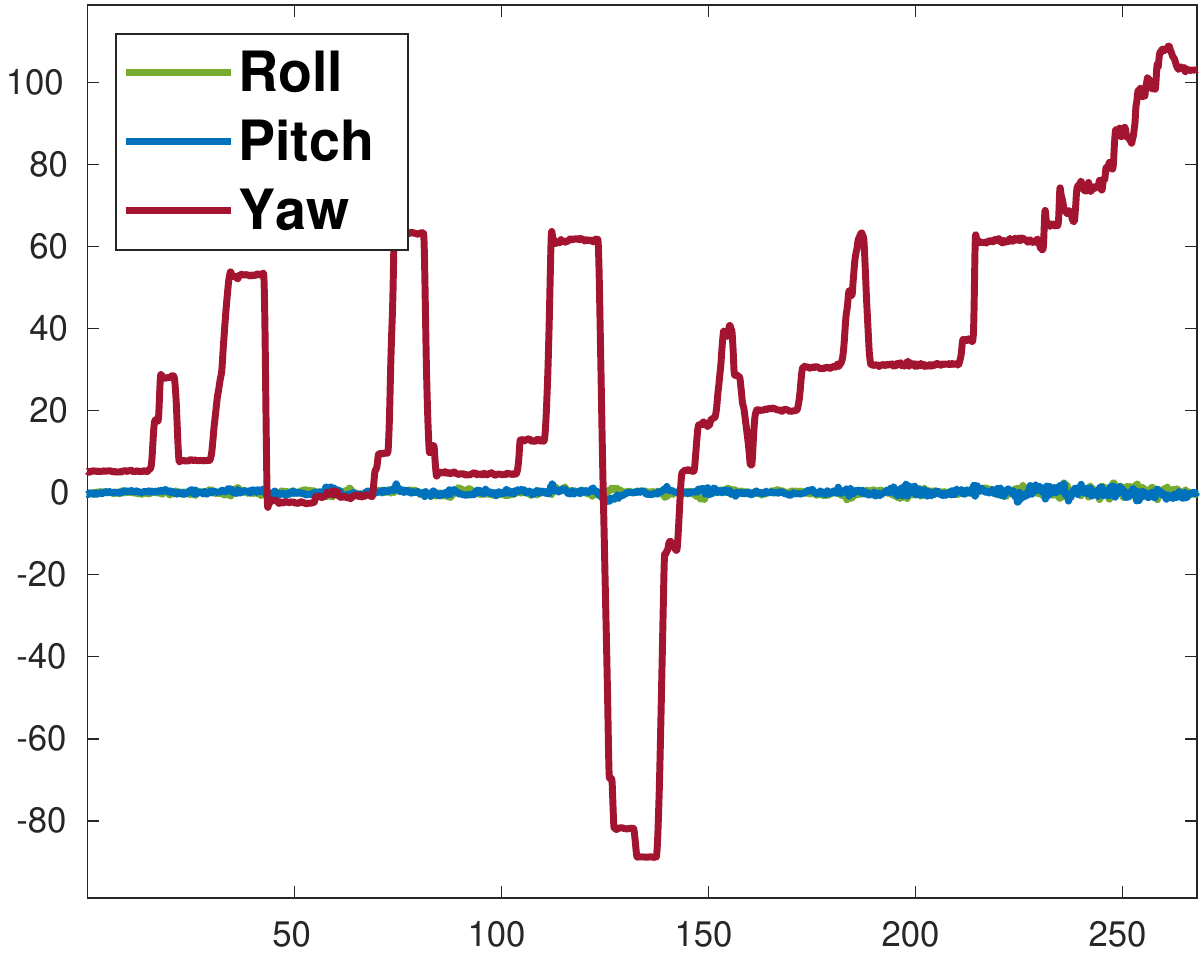}
            \caption{~}
            \label{fig:control:tests:zero-tilt-results-rpy}
        \end{subfigure}
        \hfill
        \begin{subfigure}[b]{0.49\linewidth}
            \includegraphics[width=\textwidth]{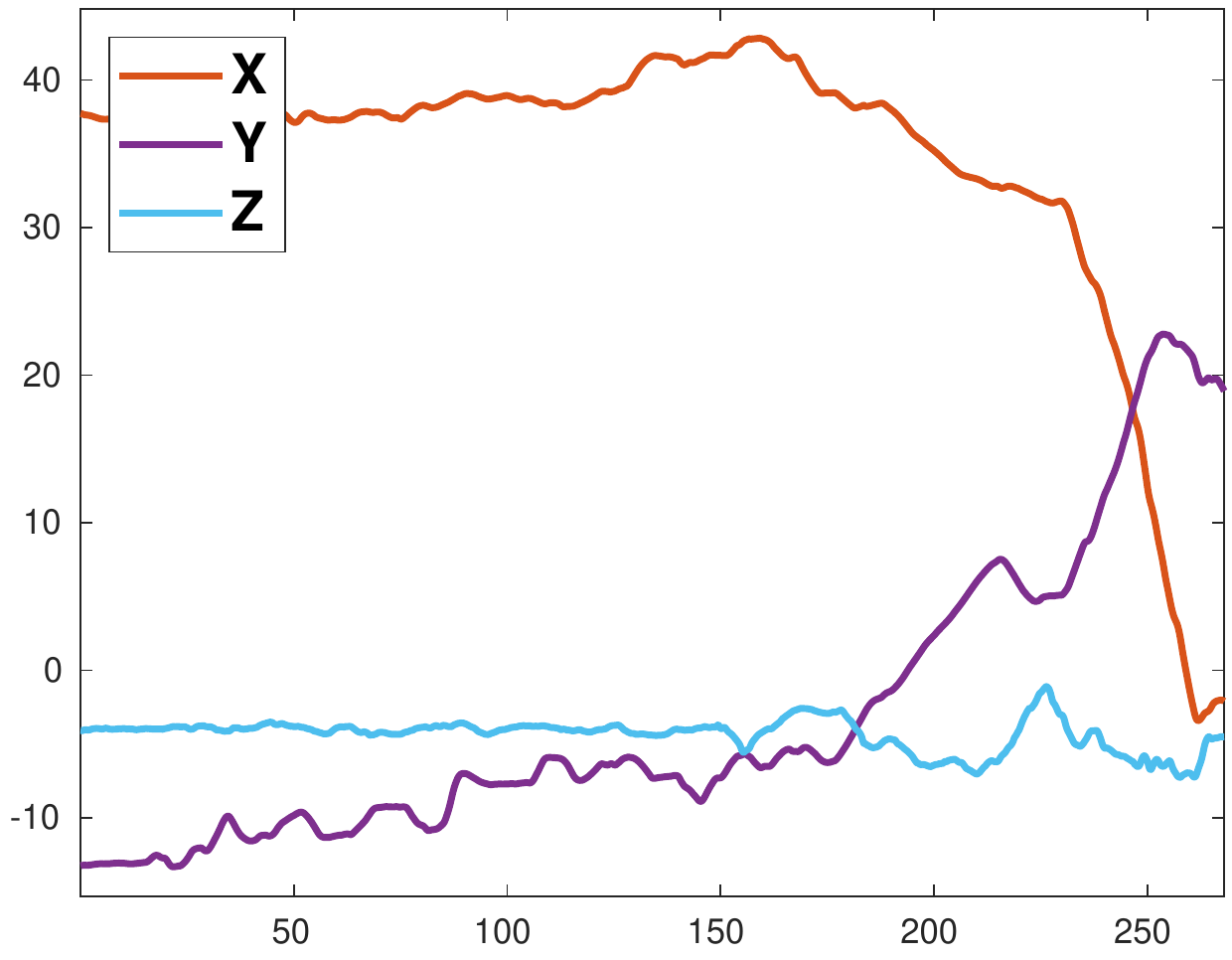}
            \caption{~}
            \label{fig:control:tests:zero-tilt-results-pos}
        \end{subfigure}
        
        \caption[Outdoor flight with zero-tilt strategy]{An outdoor flight segment with the zero-tilt strategy in the presence of winds and gusts. The yaw changes aggressively, and the multirotor is flying around while the roll and pitch stay close to zero. }
        
        \label{fig:control:tests:zero-tilt-results}
    \end{figure}

Three of the strategies have been tested on the real robot as well. Figure~\ref{fig:control:tests:zero-tilt-results} shows the robot keeping its zero-tilt attitude while aggressively flying and turning. 

\begin{figure}[!htb]
        \centering
        \begin{subfigure}[b]{0.49\linewidth}
            \includegraphics[width=\textwidth]{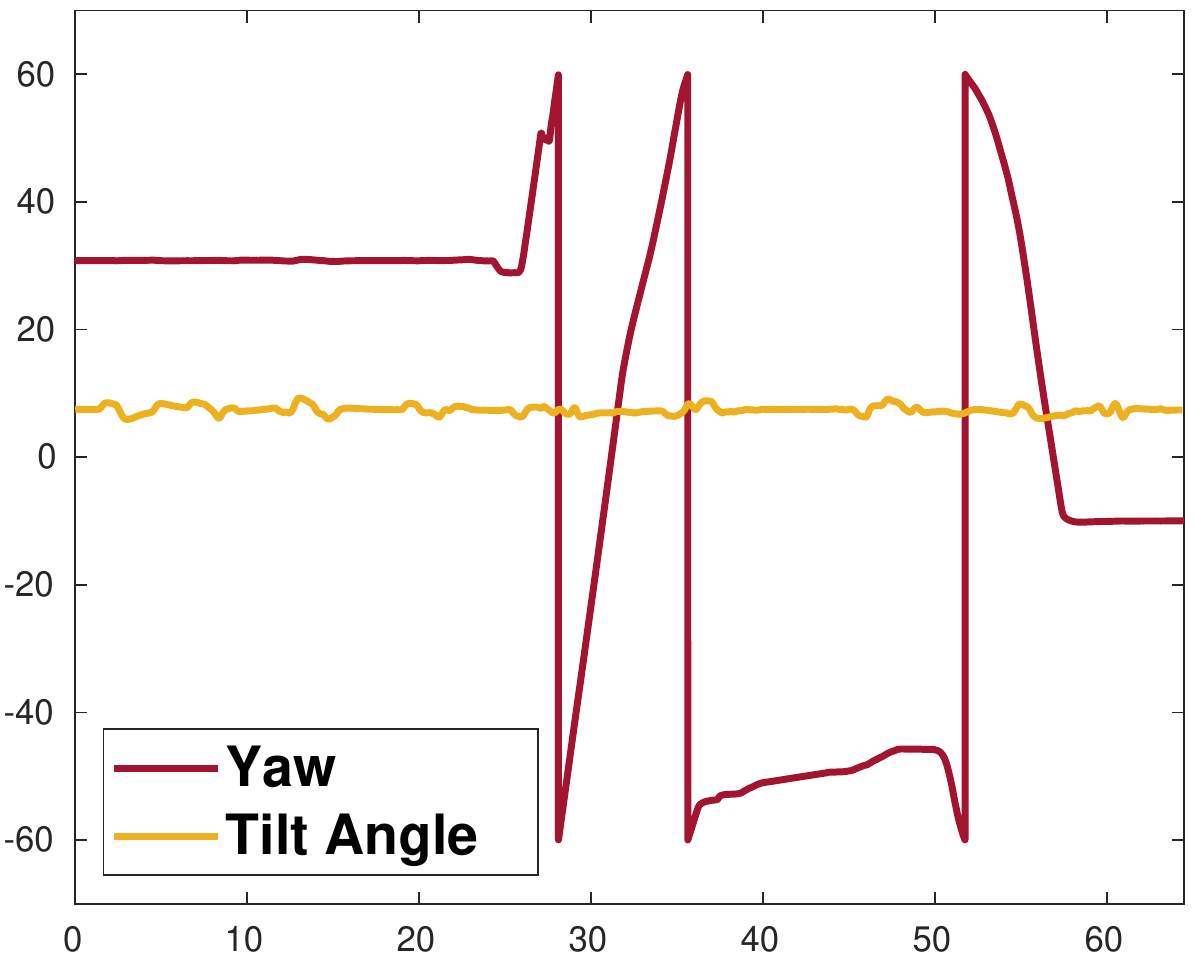}
            \caption{~}
            \label{fig:control:tests:fixed-tilt-results-y}
        \end{subfigure}
        \hfill
        \begin{subfigure}[b]{0.49\linewidth}
            \includegraphics[width=\textwidth]{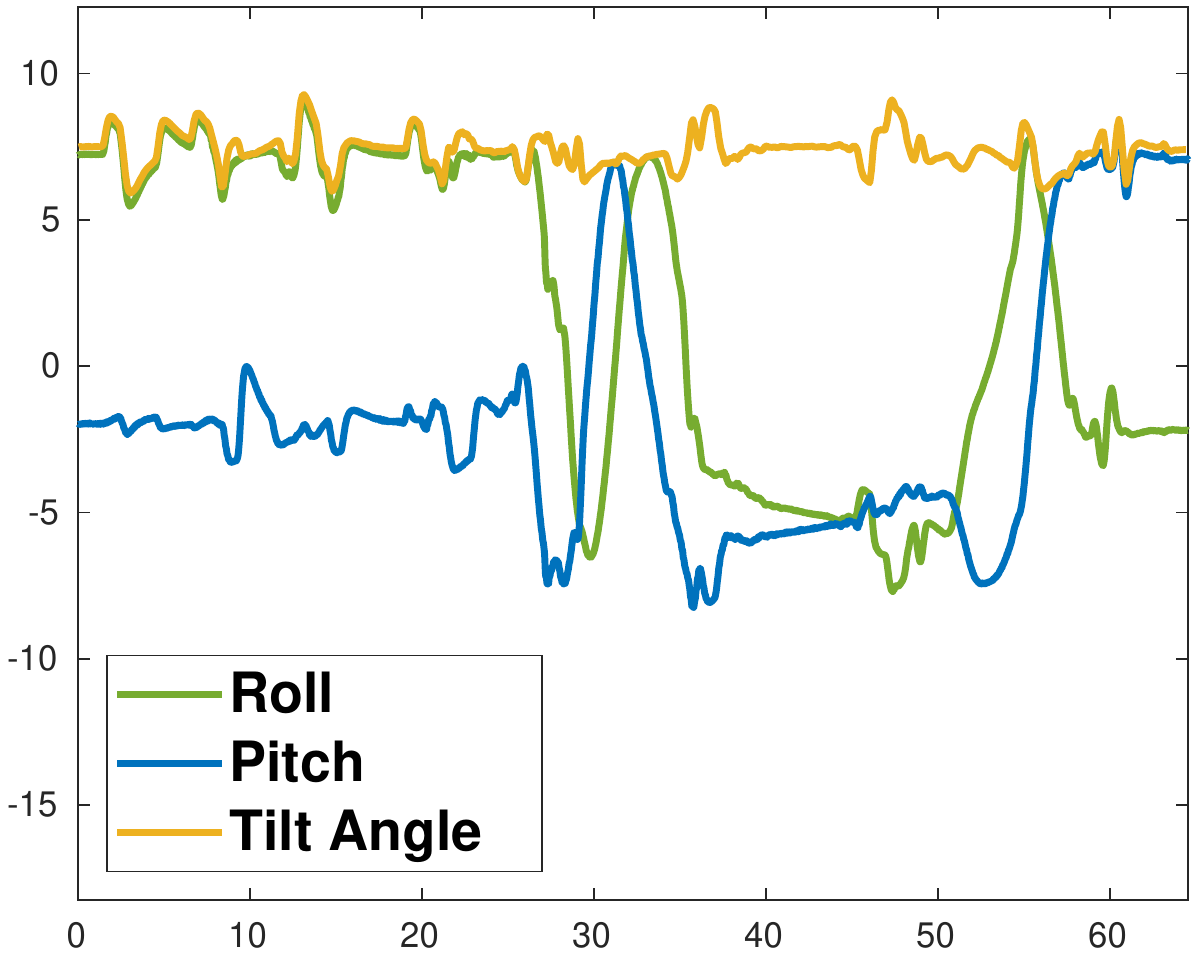}
            \caption{~}
            \label{fig:control:tests:fixed-tilt-results-rp}
        \end{subfigure}
        
        \caption[Outdoor flight with fixed-tilt strategy]{An outdoor flight segment with the fixed-tilt strategy in the presence of winds and gusts. The tilt is locked around 7.5 degrees, while the yaw changes aggressively. The right plot shows that while the tilt angle is constant, the roll and pitch change when the yaw changes. The yaw is scaled by $1/3$ in the plot.}
        
        \label{fig:control:tests:fixed-tilt-results}
    \end{figure}

    Figure~\ref{fig:control:tests:fixed-tilt-results} plots another flight segment flying with the fixed-tilt attitude strategy in a strong and gusty wind. The figure illustrates the tilt angle staying almost constant while the multirotor performs aggressive motions.
    
\subsubsection{Physical Interaction Experiments}

We used the MATLAB and Gazebo simulators and our hexarotor UAV with tilted arms (see Figure~\ref{sec:control:tests:hardware-software}) to perform the tests for the proposed position-force controller. 

For the MATLAB simulations, a scalar constant $\mu$ defines the friction at the surface. The friction force on the $\PlaneC$ plane is $\mu \left(\vec{F}_a^\frm{C} \cdot \kCv\right)$ in the opposite direction of motion, where $\vec{F}_a^\frm{C}$ is the applied force by the end effector in contact frame and $\kCv$ is the unit vector in the $\ZC$ direction.

We experimented with multiple attitude strategies for our fixed-pitch multirotor interacting with a straight wall. 

\begin{figure}[!htb]
        \centering
        \includegraphics[width=\linewidth]{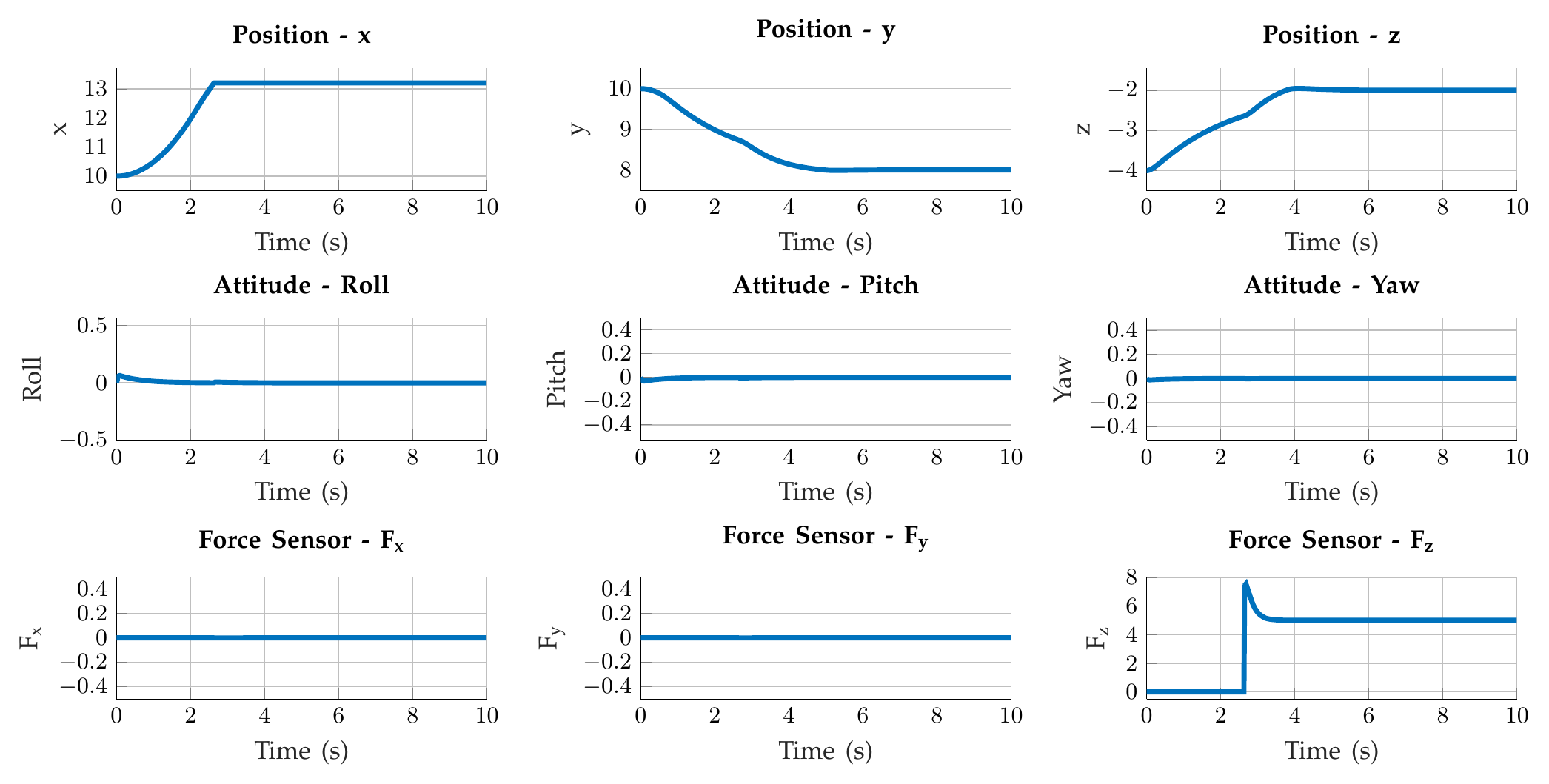}
        
        \caption[Hybrid Position-Force Controller with zero-tilt attitude strategy]{The Hybrid Position-Force controller with the zero-tilt attitude strategy applying 5~$\unit{N}$ normal force to the wall at point $p = \matrice{14 & 8 & -2}\T$.}
        
        \label{fig:control:tests:tests-tilthex-zero-tilt}
    \end{figure}

\begin{figure}[!htb]
        \centering
        \includegraphics[width=0.95\linewidth]{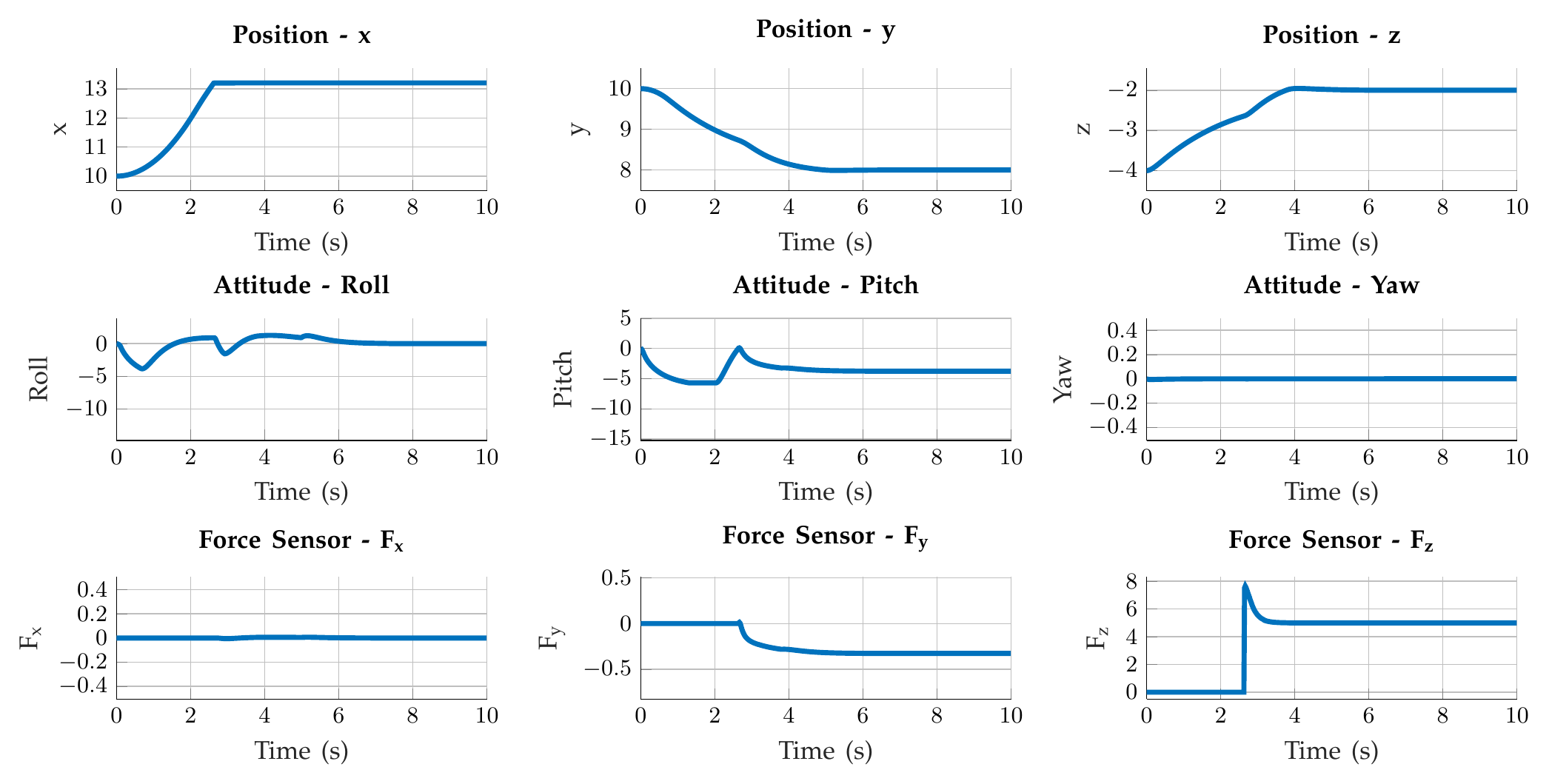}
        
        \caption[Hybrid Position-Force Controller with full-tilt attitude strategy]{The Hybrid Position-Force controller with the full-tilt attitude strategy applying 5~$\unit{N}$ normal force to the wall at point $p = \matrice{14 & 8 & -2}\T$.}
        
        \label{fig:control:tests:tests-tilthex-full-tilt}
    \end{figure}

\begin{figure}[!htb]
        \centering
        \includegraphics[width=0.95\linewidth]{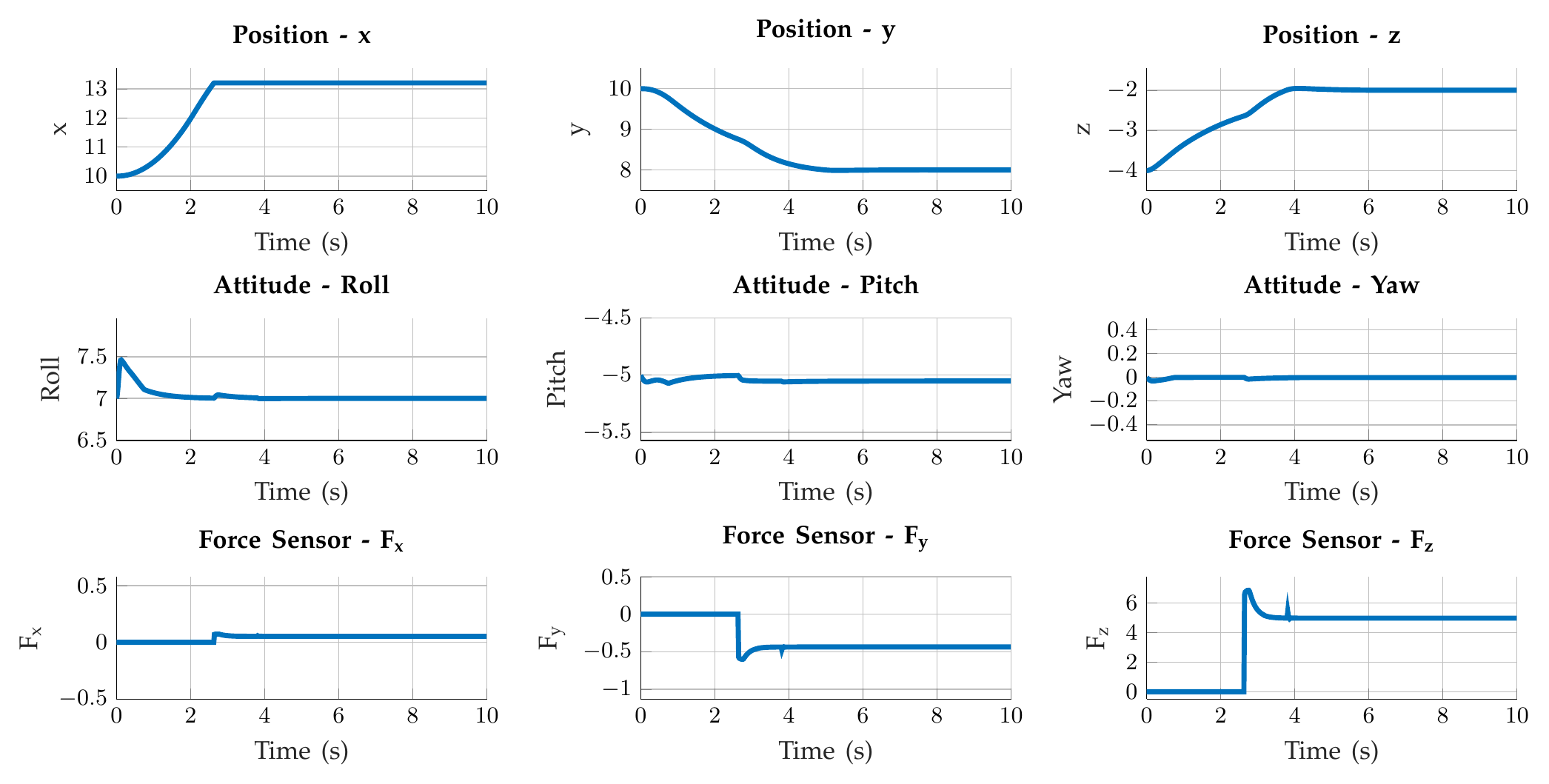}
        
        \caption[Hybrid Position-Force Controller with fixed-attitude strategy]{The Hybrid Position-Force controller with the fixed-attitude strategy applying 5~$\unit{N}$ normal force to the wall at point $p = \matrice{14 & 8 & -2}\T$. The robot's roll and pitch are set to $7$ and $-5$ degrees, respectively.}
        
        \label{fig:control:tests:tests-tilthex-fixed-attitude}
    \end{figure}

Figures~\ref{fig:control:tests:tests-tilthex-zero-tilt},~\ref{fig:control:tests:tests-tilthex-full-tilt} and~\ref{fig:control:tests:tests-tilthex-fixed-attitude} show the simulation of the fixed-pitch hexarotor with the proposed hybrid position-force controller applying a 5~$\unit{N}$ normal force to a straight wall using zero-tilt, full-tilt and fixed-orientation attitude strategies (see Section~\ref{sec:control:attitude}). 

\begin{figure}[!htb]
        \centering
        \begin{subfigure}[b]{\linewidth}
            \includegraphics[width=0.95\textwidth]{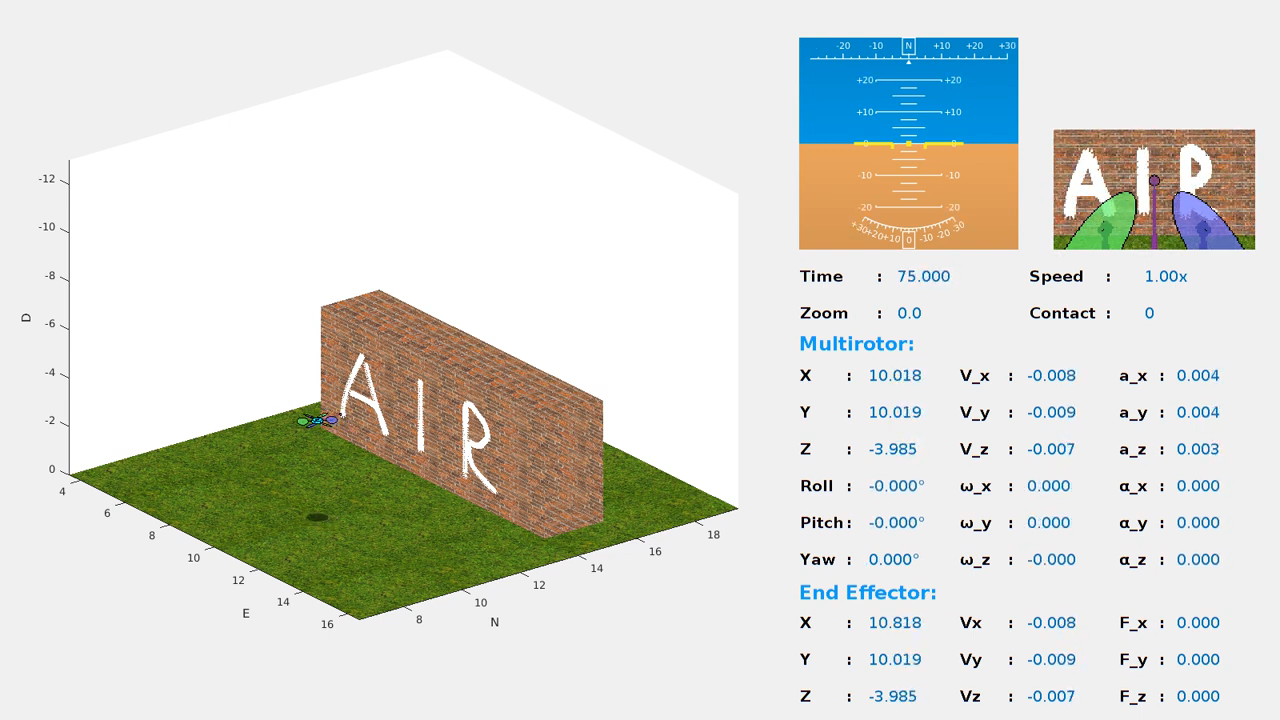}
            \caption{~}
            \label{fig:control:tests:tests-paint-on-wall-a-screenshot}
        \end{subfigure}
        \medskip
        \begin{subfigure}[b]{\linewidth}
            \includegraphics[width=\textwidth]{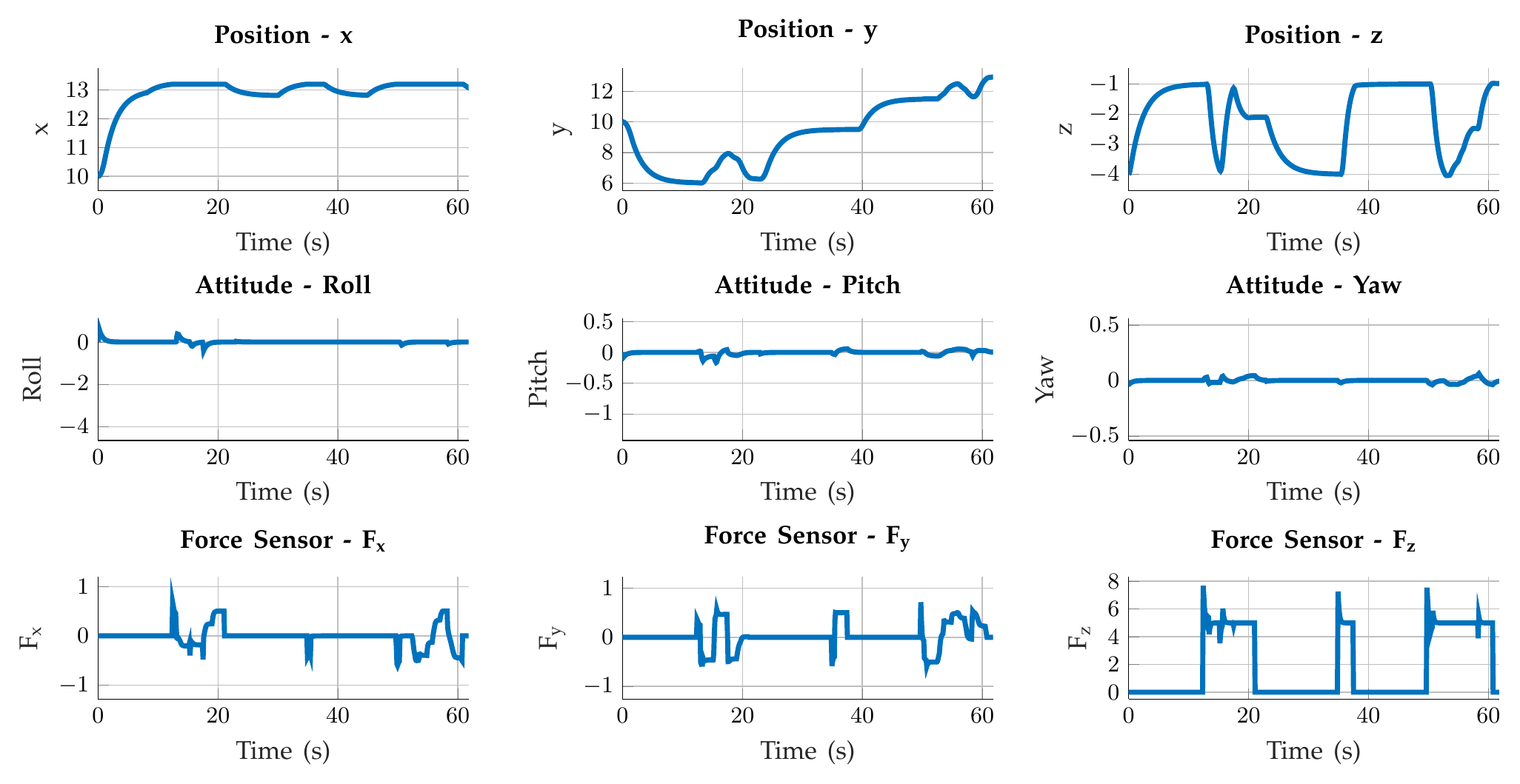}
            \caption{~}
            \label{fig:control:tests:tests-paint-on-wall-b-plots}
        \end{subfigure}
        \caption[Painting on the wall with HPF controller in MATLAB]{The Hybrid Position-Force controller with the zero-tilt strategy applying 5~$\unit{N}$ normal force to the wall to paint characters 'AIR'. (a) The screenshot from the MATLAB simulator. (b) The position, attitude and applied force plots.}
        
        \label{fig:control:tests:tests-paint-on-wall}
    \end{figure}

Figure~\ref{fig:control:tests:tests-paint-on-wall} shows the same fixed-pitch hexarotor painting on the wall. The paint is only released on the wall when the multirotor's applied force is within 0.1~$\unit{N}$ error from 5~$\unit{N}$. The wall friction is $\mu = 0.1$.
    
\begin{figure}[!htb]
        \centering
        \includegraphics[width=0.5\linewidth]{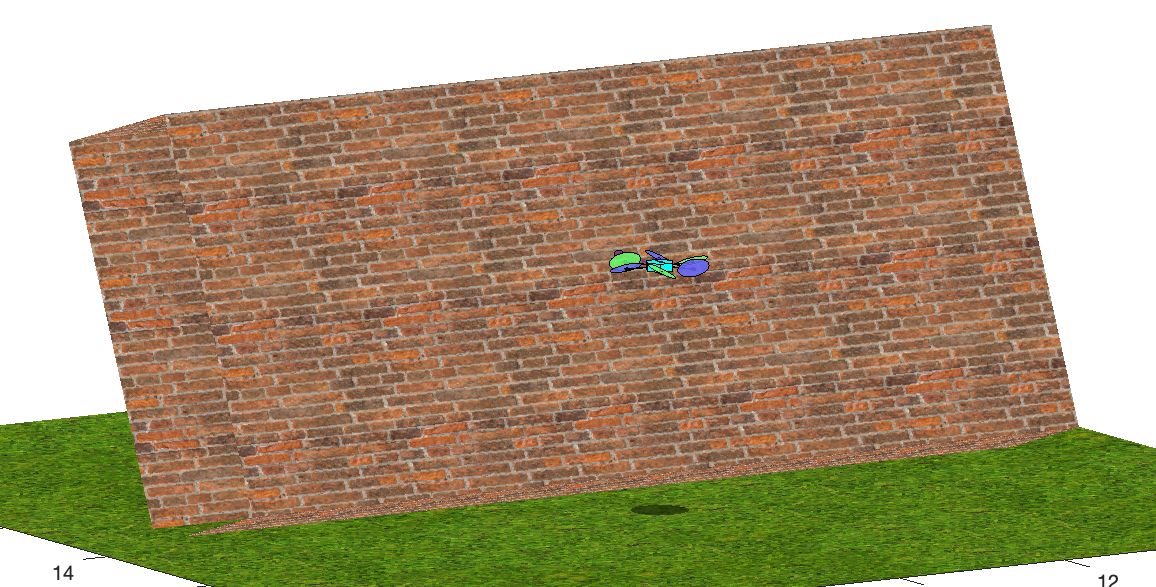}
        
        \caption[Sloped wall for testing robustness of the HPF controller]{The sloped wall pitched back 20 degrees with the multirotor contacting it to regulate 5~$\unit{N}$ force, assuming that it is a straight wall.}
        
        \label{fig:control:tests:sloped-wall-matlab}
    \end{figure}

To test the robustness of the HPF controller of Section~\ref{sec:control:hpfc} with respect to imperfect knowledge of the environment, we numerically tested contact with walls with different slopes and angles while the controller thinks that it is contacting a straight wall. Figure~\ref{fig:control:tests:sloped-wall-matlab} shows the setup for the tests.

\begin{figure}[!htb]
        \centering
        \begin{subfigure}[b]{0.3\linewidth}
            \includegraphics[width=\textwidth]{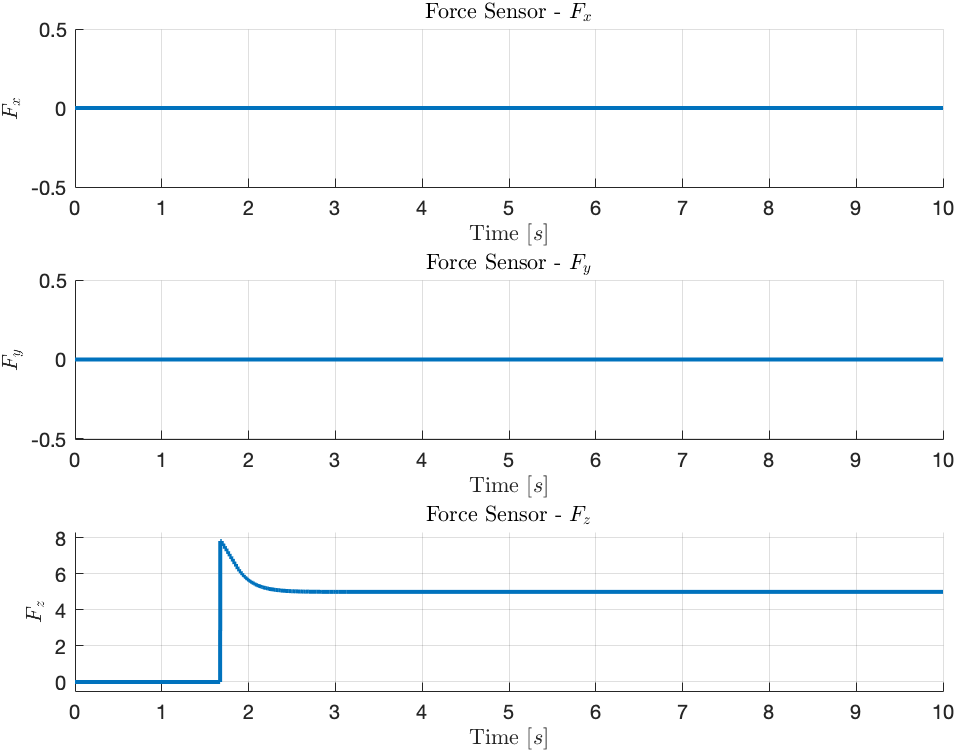}
            \caption{~}
            \label{fig:control:tests:pitched-wall-response-0deg}
        \end{subfigure}
        \hfill
        \begin{subfigure}[b]{0.3\linewidth}
            \includegraphics[width=\textwidth]{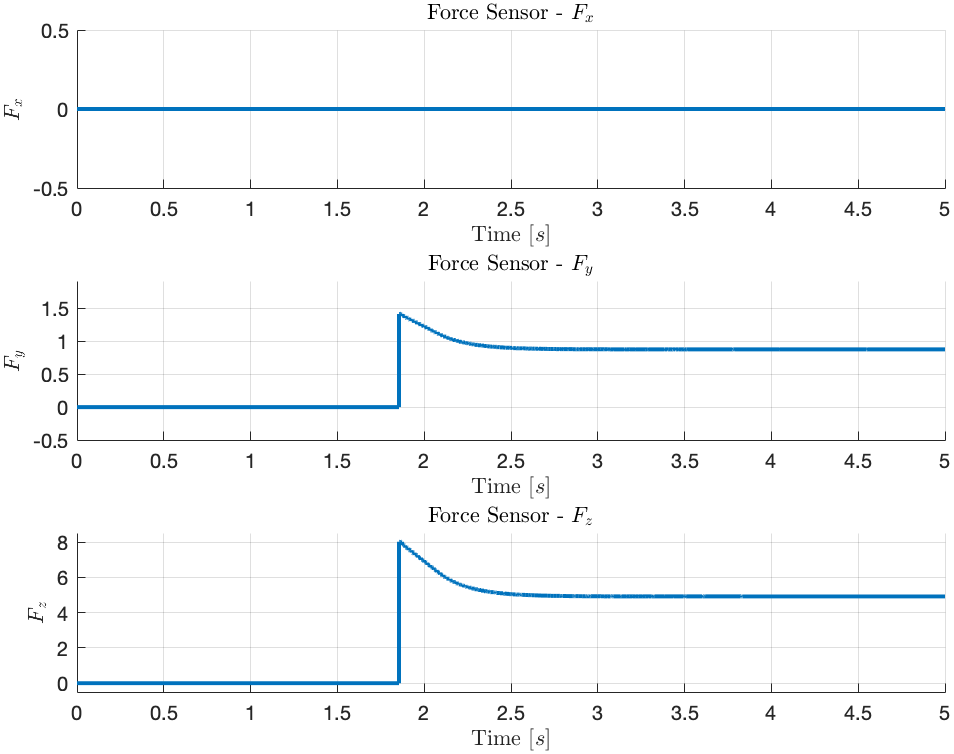}
            \caption{~}
            \label{fig:control:tests:pitched-wall-response-10deg}
        \end{subfigure}
        \hfill
        \begin{subfigure}[b]{0.3\linewidth}
            \includegraphics[width=\textwidth]{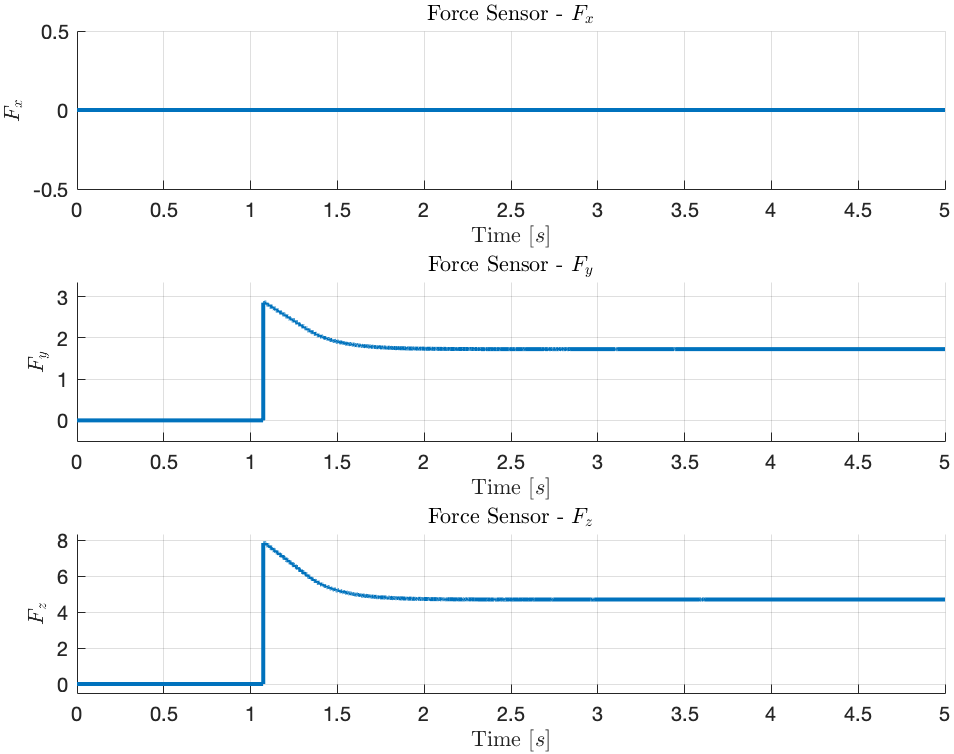}
            \caption{~}
            \label{fig:control:tests:pitched-wall-response-20deg}
        \end{subfigure}
            
        \caption[HPFC acting on a sloped wall with imperfect contact knowledge]{Hybrid Position-Force controller applying 5~$\unit{N}$ force in contact frame's $\ZC$ direction on walls with different pitch slopes with imperfect contact knowledge assuming that the walls are straight. (a) Straight wall (as baseline). (b) Wall with -10 degrees pitch. (c) Wall with -20 degrees pitch.}
        
        \label{fig:control:tests:pitched-wall-response}
    \end{figure}

In all experiments, the assumption given to the controller is that the wall normal (i.e., the $\ZC$ of contact frame) is parallel to the ground, and it is aligned with the current robot's yaw. The robot is asked to apply a steady 5~$\unit{N}$ force at a specific point on the wall in the contact frame's $\ZC$ direction. However, in each experiment, the wall is either pitched (forward or backward) or has a different yaw angle without the controller's knowledge. Figure~\ref{fig:control:tests:pitched-wall-response} shows the force response of the HPF controller to walls with 10 and 20 degrees pitch back. As can be seen, the controller can still regulate the 5~$\unit{N}$ force along its assumed $\ZE$ (which it assumes is almost the opposite of $\ZE$). However, the application of force along that direction results in forces being inadvertently applied in its $\YE$ direction as well.

\begin{figure}[!htb]
        \centering
        \begin{subfigure}[b]{0.3\textwidth}
            \includegraphics[width=\textwidth]{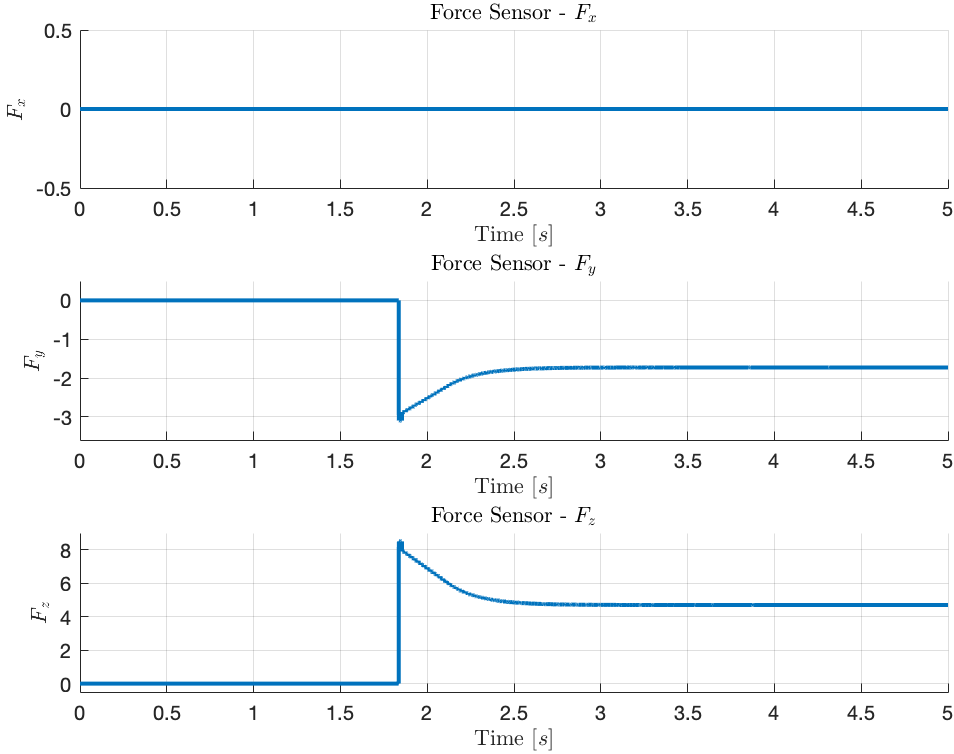}
            \caption{~}
            \label{fig:control:tests:tilted-wall-response-0deg}
        \end{subfigure}
        \hfill
        \begin{subfigure}[b]{0.3\textwidth}
            \includegraphics[width=\textwidth]{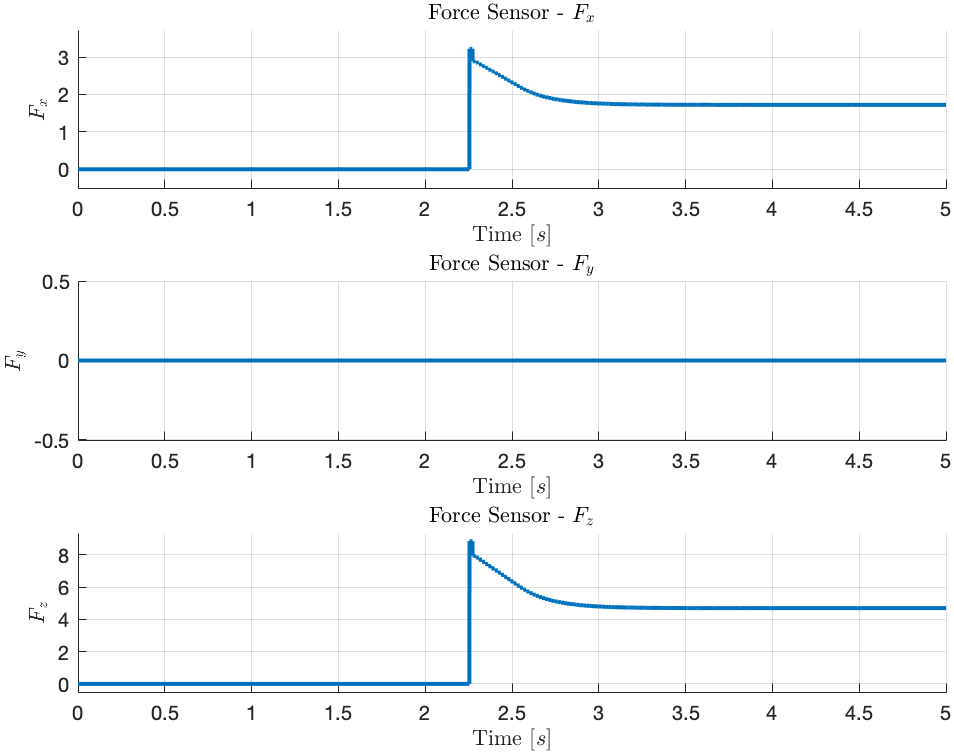}
            \caption{~}
            \label{fig:control:tests:tilted-wall-response-10deg}
        \end{subfigure}
        \hfill
        \begin{subfigure}[b]{0.3\textwidth}
            \includegraphics[width=\textwidth]{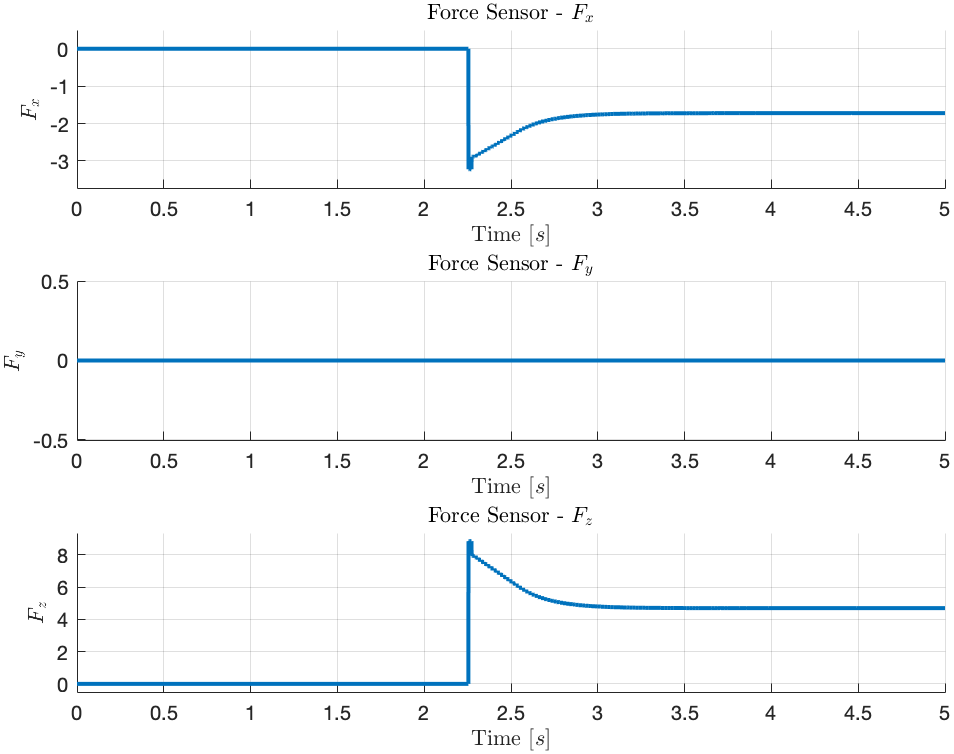}
            \caption{~}
            \label{fig:control:tests:tilted-wall-response-20deg}
        \end{subfigure}
            
        \caption[HPFC acting at an angle with imperfect contact knowledge]{Hybrid Position-Force controller applying 5~$\unit{N}$ force in contact frame's $\ZC$ direction on walls with different pitch slopes and yaw rotations with imperfect contact knowledge assuming that the walls are straight. (a) Wall with +20 degrees pitch. (b) Wall with -20 degrees yaw rotation. (c) Wall with +20 degrees yaw rotation.}
        
        \label{fig:control:tests:tilted-wall-response}
    \end{figure}

Figure~\ref{fig:control:tests:tilted-wall-response} shows the force response of the HPF controller to a wall with 20 degrees pitch forward and walls rotated at 20 degrees to the left and right. As can be seen, the controller can still regulate the 5~$\unit{N}$ force along its assumed $\ZC$ (which it assumes is almost the opposite of $\ZE$). However, the application of force along that direction results in forces being inadvertently applied in its $\YE$ and $\XE$ directions as well.

\begin{figure}[!htb]
        \centering
        \begin{subfigure}[b]{0.355\textwidth}
            \includegraphics[width=\textwidth]{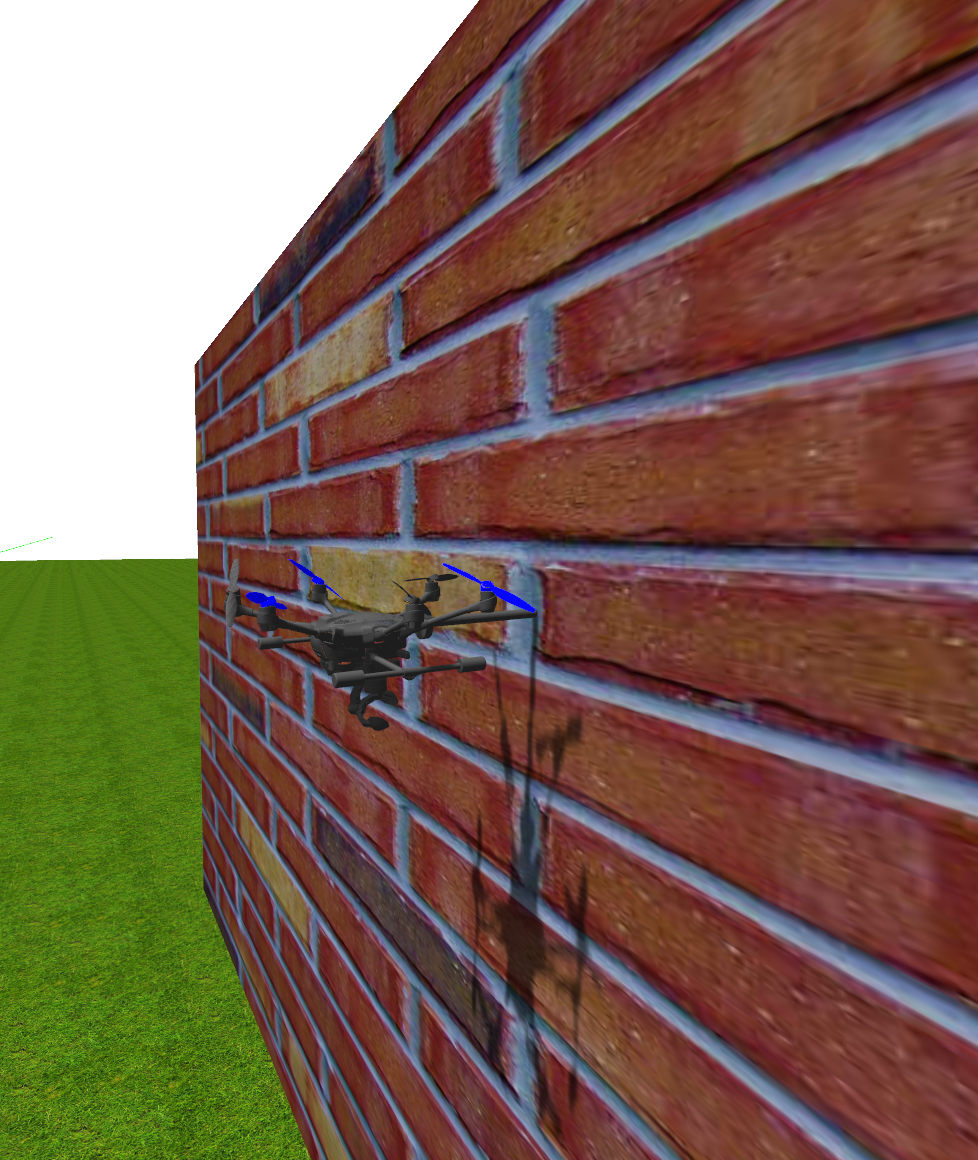}
            \caption{~}
            \label{fig:control:tests:hpfc-gazebo-setup}
        \end{subfigure}
        \hfill
        \begin{subfigure}[b]{0.615\textwidth}
            \includegraphics[width=\textwidth]{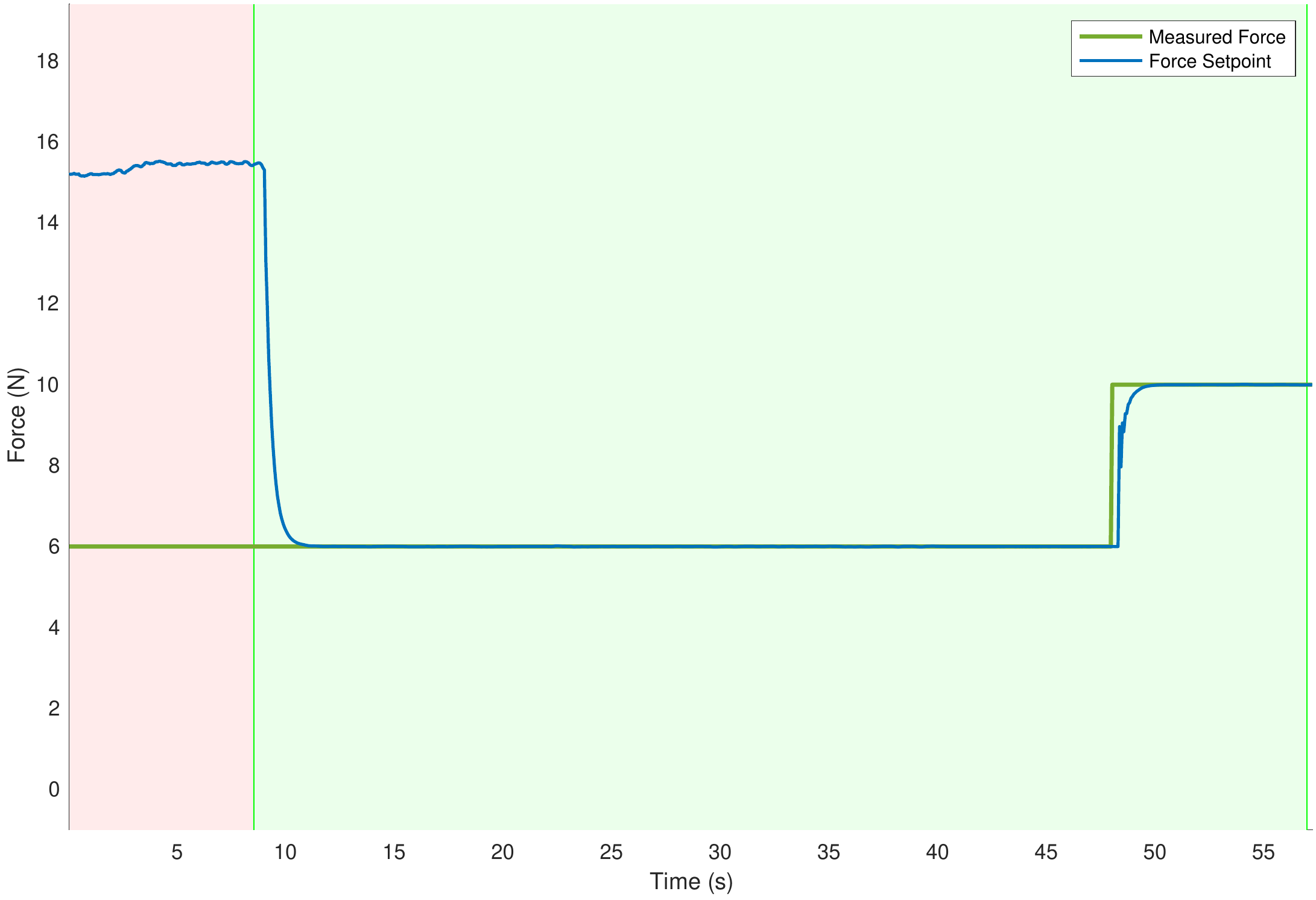}
            \caption{~}
            \label{fig:control:tests:hpfc-gazebo-response}
        \end{subfigure}

        \caption[Testing HPFC in Gazebo simulator]{Hybrid Position-Force controller applying 6~$\unit{N}$ and 10~$\unit{N}$ forces in contact frame's $\ZC$ direction on a straight wall. (a) A screenshot of the setup. (b) In response to the controller switching on, the measured force initially controls the 6~$\unit{N}$ force and then switches to 10~$\unit{N}$. The area shaded in red shows when the end-effector is in contact with the wall, but the force controller is inactive. The area shaded in green shows the times when the hybrid force-position controller is active.}
        
        \label{fig:control:tests:hpfc-gazebo}
    \end{figure}

To test our HPF controller implementation on the real robot, we first tested it in Gazebo. We equipped our hexarotor Gazebo model with a rigidly-attached arm and tested it with a straight wall. Figure~\ref{fig:control:tests:hpfc-gazebo} shows the setup and the response of the applied force to the setpoints when the controller is turned on.

We tested the implemented force controller on our UAV. The experiments included testing different force setpoints during contact with a whiteboard while the position is fixed.

\begin{figure}[!htb]
    \centering
    \includegraphics[width=\textwidth]{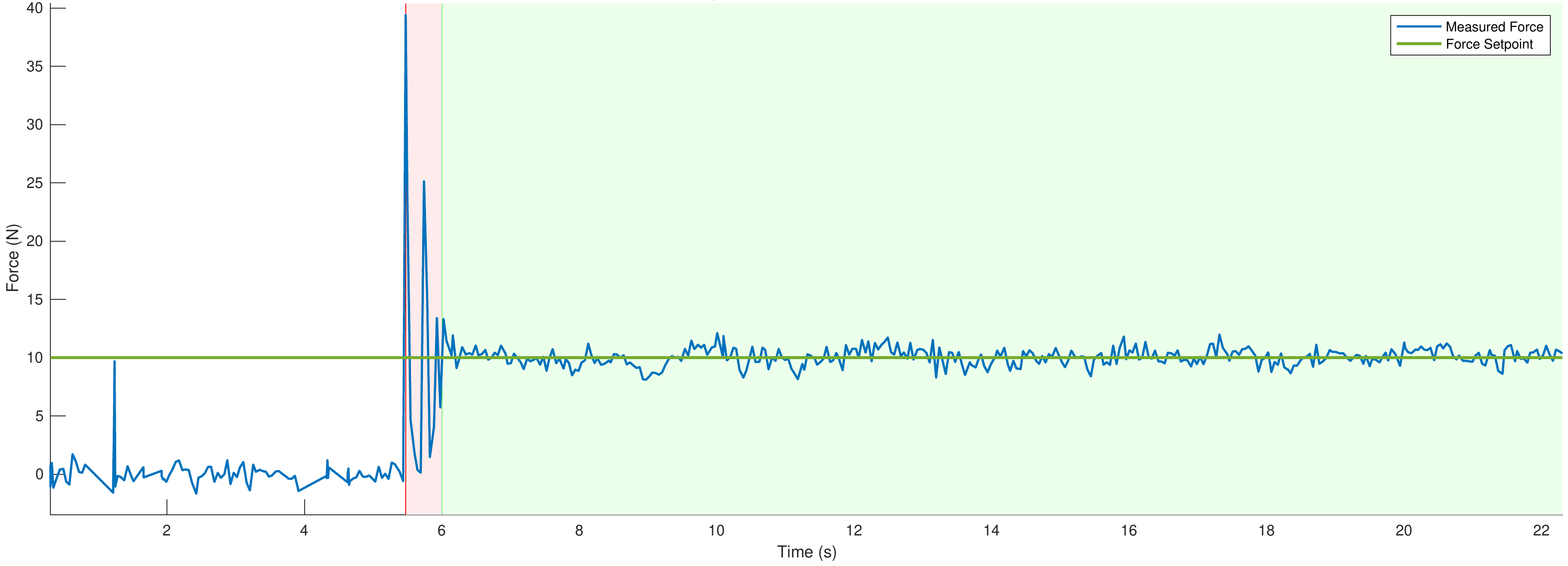}
    \caption[Force controller test on the UAV]{A force control experiment on the UAV during physical interaction with a whiteboard. The plot shows the measured force in the $\ZE$ direction compared to the force setpoint in $\ZE$. The area shaded in red shows the time frame when the end-effector is in contact with the whiteboard, but the force controller is inactive. The area shaded in green shows how the force controller follows the desired setpoint of 10~$\unit{N}$.}
    \label{fig:control:tests:force-uav}
\end{figure}

Figure~\ref{fig:control:tests:force-uav} shows an example experiment with the desired force setpoint of 10~$\unit{N}$ to be applied in the $\ZE$ direction. The plot illustrates the difference between when the force is not controlled (area shaded in red) and when it is controlled (area shaded in green). Table~\ref{tbl:control:force-control-stats} shows the statistical information for the measured force in free flight, uncontrolled contact and force-controlled contact. It signifies that during the controlled contact, the force error is similar to the sensor's natural noise (measured during free-flight), illustrating the effectiveness of the force controller in controlling the applied force as exactly as possible given the sensor's characteristics.

\begin{table}[!htb]
\centering
\caption[Measured force statistics for force controller]{Statistical comparison of the measured force (in ~$\unit{N}$) during the point contact example of Figure~\ref{fig:control:tests:force-uav}.}
\label{tbl:control:force-control-stats}
\begin{tabular}{|l|c|c|c|c|c|c|c|c|c|}
\hline
\rowcolor[HTML]{EFEFEF} 
~& Mean & Std. Dev. & Min & Max \\ \hline
Free Flight         &  0.07 & 0.72 & -2.26 & 9.70 \\ \hline
Uncontrolled Contact & 11.25 & 12.48 & 0.14 & 39.40 \\ \hline
Force-Controlled Contact & 10.07 & 0.82 & 8.12 & 13.32\\ \hline
\end{tabular}
\end{table}

\begin{figure}[!htb]
        \centering
        \begin{subfigure}[b]{0.45\linewidth}
            \includegraphics[width=\textwidth, height=5cm]{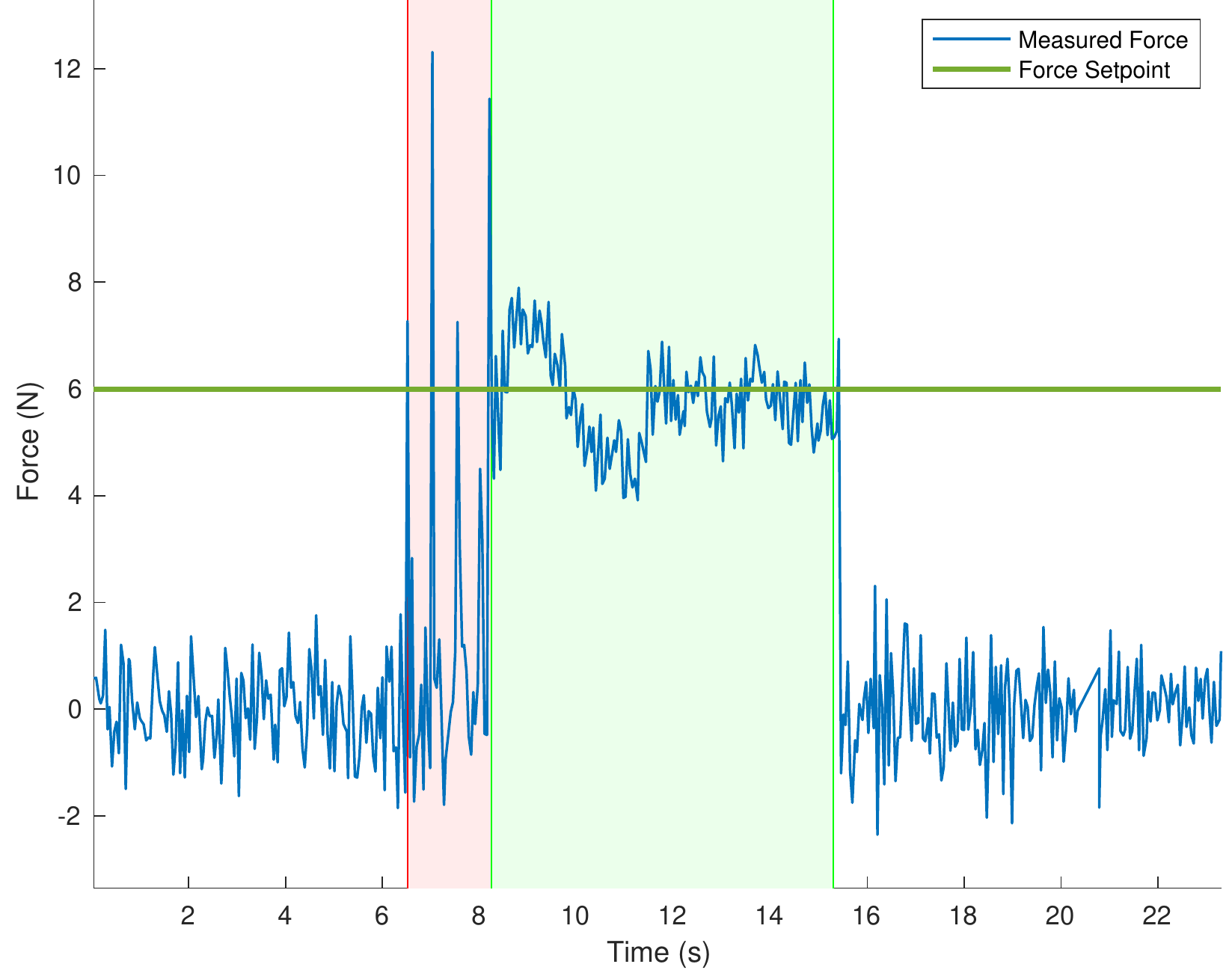}
            \caption{~}
            \label{fig:control:tests:hpfc-uav-plot}
        \end{subfigure}
        \hfill
        \begin{subfigure}[b]{0.52\linewidth}
            \includegraphics[width=\textwidth, height=5cm]{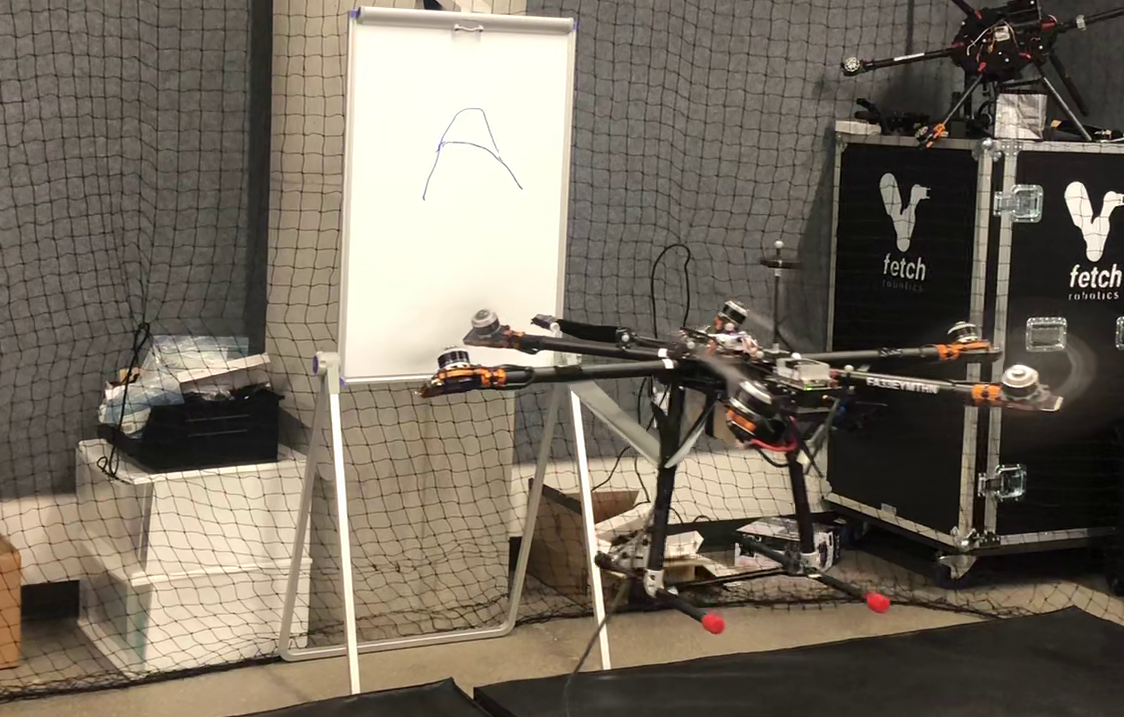}
            \caption{~}
            \label{fig:control:tests:hpfc-uav-screenshot}
        \end{subfigure}
        \caption[HPF controller test on the UAV writing "A"]{A hybrid force-position control experiment with the UAV writing the letter "A" on the whiteboard. (a) The plot shows the measured force in the $\ZE$ direction compared to the force setpoint given in $\ZE$. The area shaded in red shows the time frame when the end-effector is in contact with the whiteboard but the force controller is inactive. The area shaded in green shows how the force controller follows the desired setpoint of 6~$\unit{N}$ while writing the letter. (b) A screenshot of the experiment right after finishing writing the letter.}
        \label{fig:control:tests:hpfc-uav}
    \end{figure}

We further performed experiments to test the complete hybrid force-position controller described in Section~\ref{sec:control:hpfc}. Figure~\ref{fig:control:tests:hpfc-uav} shows an example experiment with the desired force setpoint of 6~$\unit{N}$ to be applied in the $\ZE$ direction while the robot is writing the letter "A" on the whiteboard. The plot illustrates the difference between when the force is not controlled (area shaded in red) and when it is controlled (area shaded in green). Table~\ref{tbl:control:hpfc-stats} shows the statistical information for the measured force in free flight, uncontrolled contact and force-position controlled contact. It signifies that during the controlled contact, the force error is similar to the sensor's natural noise (measured during free-flight), illustrating the effectiveness of the hybrid force-position controller in regulating the applied force as exactly as possible given the sensor's characteristics while following the desired position setpoints.

\begin{table}[!htb]
\centering
\caption[Measured force statistics for HPF controller]{Statistical comparison of the measured force (in~$\unit{N}$) during the example of Figure~\ref{fig:control:tests:hpfc-uav} writing the letter "A" on the whiteboard.}
\label{tbl:control:hpfc-stats}
\begin{tabular}{|l|c|c|c|c|c|c|c|c|c|}
\hline
\rowcolor[HTML]{EFEFEF} 
~& Mean & Std. Dev. & Min & Max \\ \hline
Free Flight         &  -0.03 & 0.68 & -2.35 & 2.31 \\ \hline
Uncontrolled Contact & 1.31 & 3.32 & -1.79 & 12.31 \\ \hline
Force-Controlled Contact & 5.77 & 0.86 & 3.91 & 7.90\\ \hline
\end{tabular}
\end{table}

\begin{figure}[!htb]
    \centering
    \includegraphics[width=0.6\textwidth]{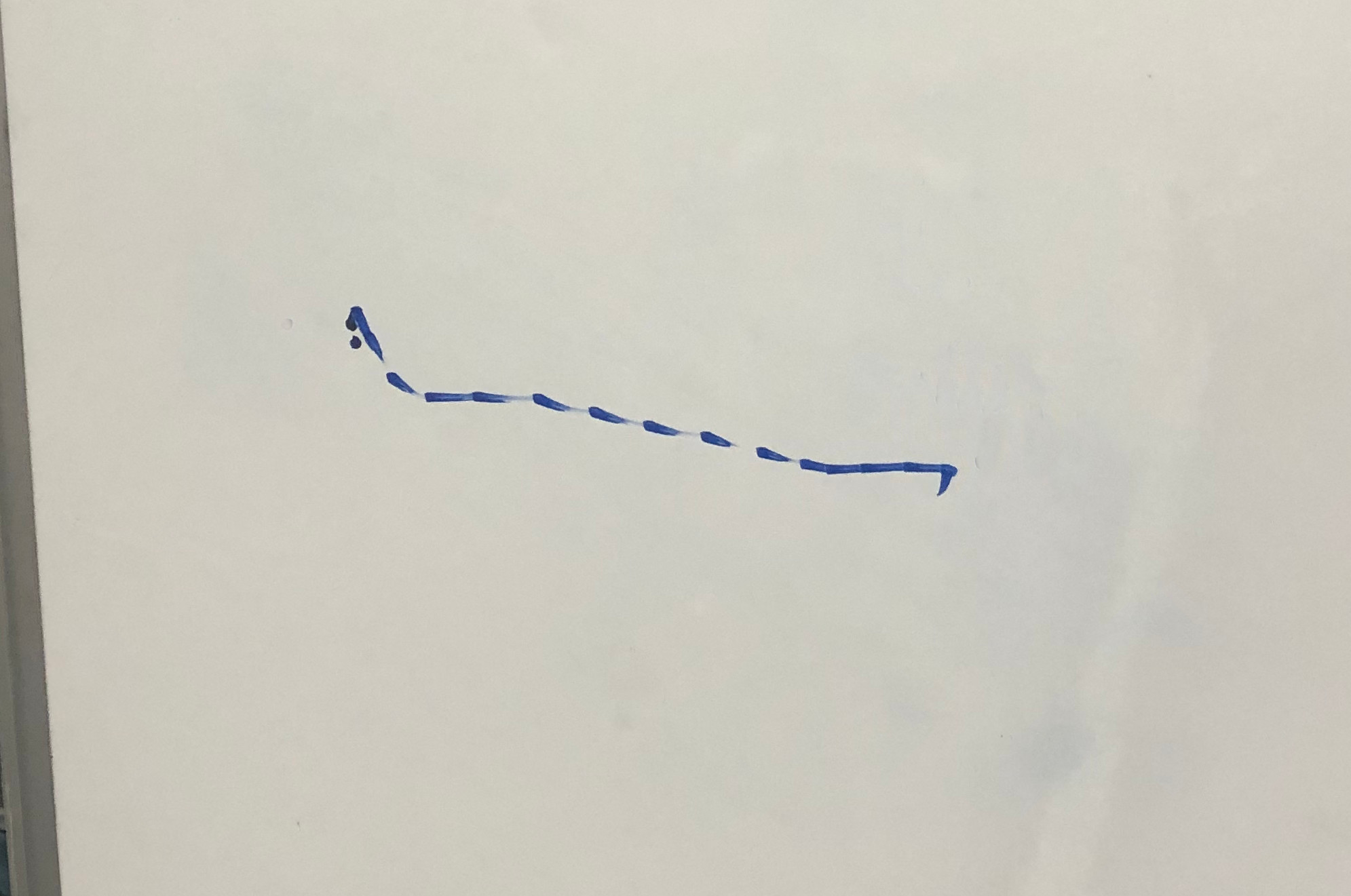}
    \caption[Effect of high force setpoint on horizontal motion]{A hybrid force-motion control experiment to draw a 20~$\unit{cm}$ horizontal line with a desired applied force of 10~$\unit{N}$. The horizontal motion of the robot is affected by the high force setpoint, resulting in a jittery motion.}
    \label{fig:control:tests:hpfc-uav-too-much-force}
\end{figure}

During the hybrid force-position experiments, we noticed that if the given force setpoint is high, it severely affects the horizontal motion of the robot during the contact. Figure~\ref{fig:control:tests:hpfc-uav-too-much-force} shows the scenario where the drawing of a 20~$\unit{cm}$ line on the whiteboard is affected by the desired 10~$\unit{N}$ applied force. The figure illustrates the jittery motion caused by the higher friction and the limited available horizontal thrust. Chapter~\ref{ch:wrench} introduces methods to estimate the available thrusts, which would allow planning for such physical interaction tasks with the appropriate force setpoints.

\begin{figure}[!htb]
    \centering
    \begin{subfigure}[b]{0.48\linewidth}
        \includegraphics[width=\textwidth]{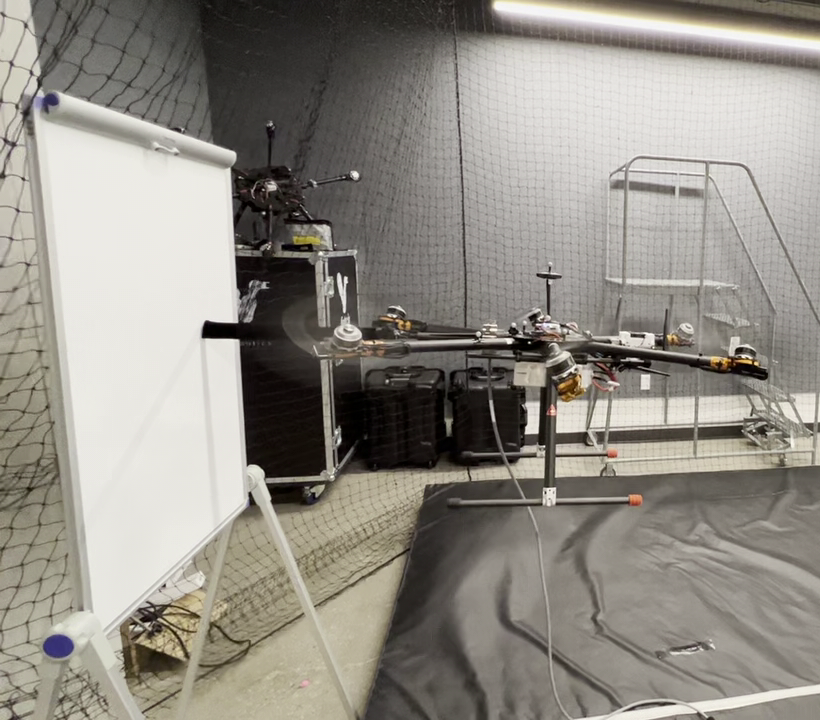}
        \caption{~}
    \end{subfigure}
    \hfill
    \begin{subfigure}[b]{0.48\linewidth}
        \includegraphics[width=\textwidth]{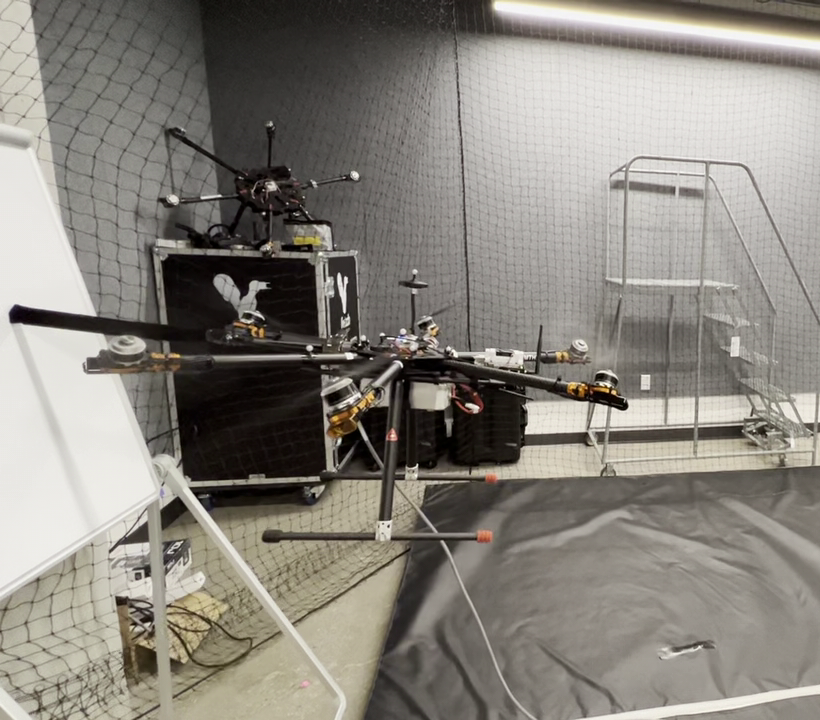}
        \caption{~}
    \end{subfigure}
    \caption[Uncontrolled contact to write on the whiteboard]{An experiment with uncontrolled force to draw on the whiteboard. The uncontrolled force results in pushing the whiteboard until it falls.}
    \label{fig:control:uncontrolled-contact-topple-whiteboard}
\end{figure}

The importance of controlling the force as opposed to the uncontrolled contact with lateral motion is emphasized in the sequences of Figure~\ref{fig:control:uncontrolled-contact-topple-whiteboard}. We observed two behaviors: the excessive force applied during the uncontrolled contact may result in damaging the environment (in this scenario, toppling the whiteboard), and the uncontrolled contact results in repeated "banging" of the robot on the surface (see Figures~\ref{fig:control:tests:force-uav} and~\ref{fig:control:tests:hpfc-uav-plot}), which is undesirable in many physical interaction applications.

\section{Conclusion and Discussion} \label{sec:completed:controller:conclusion}

This section described the development of the fully-actuated controller for our fixed-pitch hexarotor robots of different sizes. We described how the existing flight controllers for underactuated multirotors can be extended to the fully-actuated vehicles to save time and effort and reduce application development time and costs. 

We explained our controller's architecture, which is implemented on top of the existing PX4 flight stack and can accept the inputs with either complete orientation or just yaw. We proposed a set of attitude strategies that compute the multirotor's entire attitude setpoint (desired orientation) when the full attitude is not given as the input to the system. The concept of attitude strategies allows the controller developed in Section~\ref{sec:control:design} to interact with the same motion control and trajectory planner tools developed for underactuated UAVs while allowing the UAV to devise its full actuation capabilities. We proposed five such strategies and explained how the complete attitude setpoint could be computed based on the strategy.

Furthermore, we proposed using the lateral thrust limits of LBF robots in the thrust setpoint calculation. Two strategies were described with different objectives. The described strategies assumed a constant lateral thrust but can easily extend to use the lateral thrust limits calculated in real-time.

The calculation of the thrust setpoint ($\FsptB$) is independent of the devised attitude strategy. However, many flight controller systems (i.e., PX4) use the calculated attitude setpoint frame $\Frm{S}$ instead of the current body-fixed attitude $\Frm{B}$ for the output thrust calculation. The choice can simplify the calculation and avoid some issues caused by delays in reading the current attitude and may not significantly affect underactuated vehicles, as the thrust setpoint is just a scalar and projecting it on the wrong frame only has a minor effect on the magnitude. However, this projection needs to be explicitly done on the current body-fixed frame for fully-actuated robots to avoid stability issues during large commands and sudden direction changes. 

Multiple simulators were developed for the new controller's development and testing, which allowed us to experiment rapidly with different methods and ideas and facilitated the whole process. 

We shared the lessons we learned from the real-world experiments that can enhance the fully-actuated vehicles' stability and performance. Changes, such as modified motor saturation strategies, are necessary for these vehicles to be useful in practical applications. 

We further expanded our proposed controller to simultaneously control the UAV's (or the end-effector's) position and the force applied to the contact surface. The developed hybrid position-force controller devises independent control loops for the position and the force in the contact frame. The result of the loops is combined based on the orthogonal subspaces controlled by each position and force. These subspaces are defined using two selection matrices that depend on both natural constraints (the geometry of the contact) and the artificial constraints (the task's requirements). The combined result of the position and force loops is then used as the total desired thrust (or acceleration) that the UAV needs to generate.

Finally, we showed the simulated and real experiments that illustrate how the proposed controllers work. We further performed indoor and outdoor flight experiments, tested the controller's disturbance rejection using the strategies by making uncontrolled physical contact with the environment, and illustrated the force-motion controller working for multiple tasks and different attitude strategies.

\chapter{Wrench-Set Analysis for Fully-Actuated Multirotors} \label{ch:wrench}

The last decade has seen increased research focus on the physical interaction of UAVs with their environment. Such interactions range from moving boxes using a single or a group of aerial robots to UAVs' precise or compliant application of force to their surroundings. Many proposed methods for such interactions have been successfully implemented in controlled settings. However, the existing approaches use conservative limits for applied forces and moments in their interactions, which is inefficient and does not use the full potential of the aerial manipulator. This conservatism may result in some tasks being completed non-optimally, and in the worst scenario, some feasible tasks can be deemed infeasible.

In the context of our multirotor, we propose a set of real-time methods to estimate the instantaneous wrenches (i.e., forces and moments) that a robot manipulator can generate. The methods estimate the wrenches based on the multirotor's design, state, and the desired forces applied during the physical interaction and output the force and moment (wrench) polytopes. The wrench-set estimation enables the use of the full potential of the existing wrenches for optimizing the accelerations in free flight and for physical interaction tasks. Chapter~\ref{ch:applications} presents different possible applications of the real-time estimation method presented in this chapter.

In this chapter, we first introduce the problem and review the related work (Section~\ref{sec:wrench:intro}). Then we provide our solutions for decoupled thrust and moment set estimations in real-time (Sections~\ref{sec:wrench:decoupled}) and extend it to coupled lateral thrust estimation and coupled wrench set estimation(Section~\ref{sec:wrench:coupled} and~\ref{sec:wrench:lateral}). Later, we review the possible extensions to variable-pitch and more complex robots (Section~\ref{sec:wrench:extensions}). Finally, we present our experiments and test results (Section~\ref{sec:wrench:tests}).

\section{Introduction and Related Work} \label{sec:wrench:intro}

The progress in research on the physical interaction of UAVs in the last decade has provided novel tools to enable new applications ranging from package delivery to contact inspection and aerial manipulation. Different research groups and companies have shown successful results in controlled and carefully-crafted settings. However, so far, these approaches have been ignoring the robot's capabilities by using conservative force and moment limits in their interactions. These limits result in sub-optimal actions and underuse the aerial manipulator's full potential. For some tasks, the conservative limits may make a feasible task infeasible. On the other hand, if the limits are set high, other additional precautions need to be taken to monitor and correct the controller's output; otherwise, the result can be a disastrous crash.

\begin{figure}[!htb]
\centering
\includegraphics[width=0.8\linewidth]{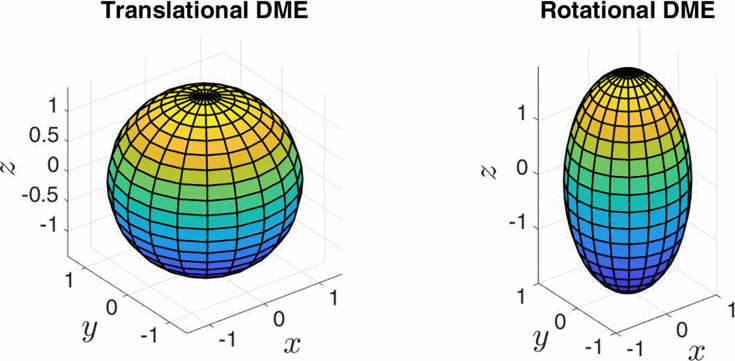}
\caption[An example dynamic manipulability ellipsoid]{An example dynamic manipulability ellipsoid of the system developed in~\cite{Kaufman2014} (figure from~\cite{Tadokoro2018a}).}
\label{fig:background:control:dynamic-manipulability-ellipsoid}
\end{figure}

The effectiveness of a specific fully-actuated multirotor can be described by analyzing its \textit{dynamic manipulability}. This concept has been used for ground robots since Yoshikawa introduced it in 1985~\cite{doi:10.1177/027836498500400201}. However, it has entered the UAV research only recently, in 2015, primarily for optimization and classification of the fully-actuated multirotors~\cite{7320711, Tadokoro2017, Tadokoro2018, Tadokoro2018a}. \textit{Dynamic manipulability ellipsoid} is the set of 6-D accelerations of the robot's CoM with unit-spherical input thrust force (rotor thrusts), represented as a six-dimensional ellipsoid. This ellipsoid illustrates how multirotor's thrusts and moments can be generated in different translation and rotation directions. Figure~\ref{fig:background:control:dynamic-manipulability-ellipsoid} shows an example of the dynamic manipulability ellipsoid of the system developed in~\cite{Kaufman2014}.

Dynamic manipulability ellipsoids have been shown to work fine in practice and are fast to compute but still fail to capture the entire space of available forces and moments. A more advanced descriptor, called \textit{dynamic manipulability polytopes}, has been introduced by Chiacchio et al.~\cite{forcepolytope, 509249}, which can represent the entire wrench space. We compute and use this latter descriptor to capture the whole space of available forces and moments in this work. 

From a geometrical perspective, the thrust vectors generated by the UAV rotors can be seen as a set of force vectors in 3-D space with lower and upper limits for each vector's magnitude. In literature, an analogous example is a multi-fingered grasp of an object, where there is a set of force vectors in 3-D (at the contact points) with lower and upper limits defined by the forces each finger can apply to its contact point~\cite{Roa2015}. Therefore, the wrench polytopes for multirotors have geometrical similarities with the \textit{grasp polytopes} used in grasp analysis, and many of the methods related to grasp polytopes can be adapted for use in the estimation of the wrench set for aerial robots. 

In 1992, Avis and Fukuda~\cite{Avis1992} introduced a pivoting algorithm that has been widely used for offline computation of the wrench set. In 2020, we introduced an online method for computation of the wrench set~\cite{Keipour:2020:cmu:proposal} in multirotors, which uses the geometrical properties and the convexity of the wrench set to compute the force and moment polytopes in real-time. Recently, our real-time thrust set estimation has been re-introduced by~\cite{9562050} in the context of general manipulation systems (including ground and humanoid robots). However, the authors have not expanded the method further to full wrench with desired (fixed) components.

The thrust and moment limits of fully-actuated multirotors have been generally treated as static values for each multirotor architecture. Franchi et al.~\cite{Franchi2018} have proposed a fast optimization-based solution for lateral thrust estimation in real-time from the thrust set, but it only uses the static thrust set computed for the hovering. They estimate only the thrust set (ignoring the wrench set) using an analytical solution, which is only possible due to the symmetric nature of their tilted hexarotor and cannot be easily generalized to other architectures. On the other hand, in reality, the sets of feasible wrenches and the wrench limits depend not only on the architectural design but also on the current state and the instantaneous wrenches generated at each time. On the other hand, considering the wrench limits as static has resulted in thrust and moment limits being assumed decoupled, despite the strong dependency between them. 

This chapter comprehensively discusses estimating the thrust and moment limits and proposes real-time methods to estimate the instantaneous wrench limits. We describe our real-time thrust set estimation method and expand it by taking into account the current state of the robot and the coupling between the thrusts and moments. 

To continue with the chapter, we need to define the terminology used here:

\textit{Thrust or Force Set} ($\SetT$): The set of all the feasible 3-D thrust setpoints at the current instance of time. In other words, it is the set of all the thrusts that the multirotor will be able to generate almost instantaneously (i.e., in a short time period $\delta t$ from the current time $t$).

\textit{Moment or Torque Set} ($\SetM$): The set of all the feasible 3-D moment setpoints at the current instance of time. In other words, it is the set of all the moments that the multirotor will be able to generate almost instantaneously (i.e., in a short time period $\delta t$ from the current time $t$).

\textit{Wrench Set} ($\SetW$): The set of all the feasible 6-D wrench (i.e., thrust and moment) setpoints at the current instance of time. In other words, it is the set of all the wrenches that the multirotor will be able to generate almost instantaneously (i.e., in a short time period $\delta t$ from the current time $t$).

\section{Decoupled Thrust and Moment Set Estimation} \label{sec:wrench:decoupled}

This section discusses our method for constructing the Thrust Set $\SetT$ and then explains how the Moment Set $\SetM$ can be constructed similarly.

Let us assume $\SetT(t)$ is the set of all feasible thrust vectors that our multirotor can generate at time $t$. We will omit the time dependency notation $(t)$ from now on when the context is clear. The set $\SetT$ depends on the UAV's architectural design and hardware, the current state, and the environment. In this section, when we mention the current state, we also account for the external forces (such as wind and gravity).

In reality, due to the imperfect motors and motor controllers, delays in the system, and other uncertainties, calculating the exact set of feasible thrusts is impossible. Moreover, the actual thrust set would be an infinitesimally small volume around the current thrust due to the motor and system delays.

In practice, our main interest in $\SetT$ is to gain the ability to modify the thrust setpoint so that it lies within the set. Therefore, the goal is calculating an estimation $\SetTp$ to the union of all possible sets $\SetT(t + \delta t)$ computed for a short time period $\delta t$ from time $t$. $\delta t$ is the shortest time that allows the thrust and moment setpoints to realize (i.e., $\delta t$ is assumed to be larger than system and motor delays but too small for any other state or environment variables to change meaningfully).

An approximation to the $\SetT$ set can be mathematically estimated directly from the UAV structure using numerical or exact methods. The resulting thrust set ignores the effects such as aerodynamic interference between the rotors, imperfections in the model, the current state of the system, and the current moment setpoint. Tadokoro et al.~\cite{Tadokoro2017, Tadokoro2018a} define Dynamic Manipulability Measure for UAVs to quantify the relationship between the structure of the UAVs and their feasible thrust and moment sets. These measures are beneficial for optimizing the UAV design but have limited use in estimating the feasible thrust set in practice. 

For some special (mostly symmetric) structures, it is possible to formulate the $\SetT$ set mathematically~\cite{Franchi2018}. While these formulas can be helpful due to their simplicity, they completely ignore the vehicle's state, which affects the thrust set. In our experiments (see Section~\ref{sec:wrench:tests}), we demonstrate how the thrust set changes with the change in the architecture and the state of the UAVs.

We propose a real-time method to numerically estimate the $\SetT$ set for a general UAV system from the mathematical model of the system that can be used in real-time. The choice of the input $\mat{u}$ to the system (see Figure~\ref{fig:control:typical-controller-architecture}) is flexible, as long as each element $\mat{u}$ is the command for one of the motors, and each element is a monotonically increasing or decreasing function of the generated motor thrust. For example, $\mat{u}$ can be the vector of rotor thrusts, the vector of motor speeds (usually squared), or the vector of motor PWM signals. We also assume that the input elements' upper and lower bounds are known.

Algorithm~\ref{alg:wrench:decoupled:thrust-set-estimation} shows our proposed algorithm for real-time feasible thrust set calculation. 

\begin{algorithm}[!htb]
\caption{Proposed approach for thrust set estimation}
\label{alg:wrench:decoupled:thrust-set-estimation}
\begin{algorithmic}[1]
\LineComment {This function estimates the thrust set for the UAV model and state}
\Function{EstimateThrustSet}{\var{model}, \var{state}}
    \LineComment {Update the UAV model state to the input state}
    \State $\var{model} \gets \func{SetState}(\var{model}, \var{state})$
    \LineComment {Get the number of rotors in the UAV}
    \State $\var{r} \gets \func{GetNumOfRotors}(\var{model})$
    \LineComment {Extract system's min and max input values as $r\times1$ arrays, with each cell corresponding to a motor/rotor}
    \State $(\var{minU}, \var{maxU}) \gets \func{GetInputRange}(\var{model})$
    \LineComment {Define an empty set for the thrusts}
    \State $\var{thrusts} \gets \emptyset$
    \LineComment {Generate thrusts from all min/max combinations of motor commands}
    \For {$\var{i} = 0$ \textbf{to} $2^r-1$} 
        \LineComment {Convert the iterator $\var{i}$ into an $r\times1$ binary array}
        \State $\var{iArray} \gets \func{NumToBinaryArray}(\var{i}, \var{r})$
        \State $\var{iArrayNegated} \gets \func{Not}(\var{iArray})$
        \LineComment {Calculate the input array from the binary array}
        \LineComment {$\odot$, $\oplus$: element-wise multiplication and addition}
        \State $\var{u} \gets (\var{maxU} \odot \var{iArray}) \oplus (\var{minU} \odot \var{iArrayNegated})$
        \LineComment {Calculate the thrusts generated by the model}
        \State $\var{t} \gets \func{CalculateTotalThrust}(model, u)$
        \LineComment {Add the calculated thrust to the thrust set}
        \State $\var{thrusts} \gets \func{AddToSet}(\var{thrusts}, \var{t})$
    \EndFor
    \LineComment {Calculate the convex hull of the thrusts}
    \State $\var{feasibleThrusts} \gets \func{ConvexHull}({\var{thrusts})}$
    \LineComment {Return the result}
    \State \Return $\var{feasibleThrusts}$
\EndFunction
\end{algorithmic}
\end{algorithm}

The proposed algorithm calculates the convex hull of all the thrusts resulting from the model's combinations of minimum and maximum rotor/motor inputs. The algorithm's expected time complexity is $O(2^{1.5r})$, where $r$ is the number of rotors. However, considering that the number of rotors in common UAV architectures is small, an efficient implementation of the algorithm can run on standard autopilot systems to approximate the feasible thrust set for the system's current state.

~

\begin{prop}
The feasible thrust set approximated by Algorithm~\ref{alg:wrench:decoupled:thrust-set-estimation} ($\SetTp$) is the complete thrust set that can be generated by the model and the robot state given to the algorithm as inputs.
\end{prop}

\begin{proof} 
To restate the problem, the goal is to find the vector span of $r$ vectors (thrusts normal to the rotors), having the minimum and maximum magnitudes of the vectors. 

The proposition claims that the vector span is the convex hull of all the vertices created by combinations of minimums and maximums of all the vectors. In other words, the sum of the vectors (each with a magnitude between their minimum and maximum) is within the convex hull, and each point within the convex hull can be written as a sum of the vectors with valid magnitudes. 

Let us define the unit vector in the positive thrust direction of the $\nth{i}$ rotor as $\vec{v}_i$ and the set of all $r$ unit vectors as $\SetV$. Assuming $min_i$ as the minimum and $max_i$ as the maximum magnitude of the $\nth{i}$ rotor's thrust, we can define the set of all the thrust intervals as $K$. The span of $\SetV$ over $K$ (with its vectors bounded by their maximums and minimums) can be defined as:

\begin{equation}
    \Span_K(\SetV) = \left\{\sum_{i=1}^r \lambda_i \vec{v}_i \ | \ \vec{v}_i \in \Set{V},\ \lambda_i \in [min_i, max_i] \right\}
\end{equation}

First, we define the set of points resulting from the sum of maximum and minimum values of the vectors as:

\begin{equation}
    \Set{P} = \left\{ \sum_{i = 1}^r \lambda_i \vec{v}_i  \ | \ \vec{v}_i \in \Set{V},\ \lambda_i \in \left\{ min_i, max_i \right\} \right\}
\end{equation}

The proposition claims that the span of $r$ vectors with their corresponding bounds is the convex hull of all the points generated by the combinations of those bounds. In other words:

\begin{equation}
    \Span_K(\SetV) = \Conv\left(\Set{P}\right)
\end{equation}

To prove the proposition, we use induction. For simplicity, we are dropping $K$ out of the notation.

\textbf{\textit{Base Case:}} With only one rotor ($r = 1$), the set $\SetUi$ of all the possible 3-D thrusts (represented as 3-D points) is a segment of the line passing the center in the direction of $\vec{v}_1$ spanning from $min_1$ to $max_i$. In this case, $\Span(\SetV)$ is obviously the convex hull created only by $2^1$ points: the minimum and maximum thrusts. 

\textbf{\textit{Induction Step:}} Assuming the proposition holds for $r = n$, we want to show that it also holds for $r = n+1$. In other words, if we add a vector $\vec{v}_{n+1}$ to our set of $n$ vectors, the resulting span is still a convex hull created only from the $2^{r+1}$ combinations of minimum and maximum thrust magnitudes. 

Considering that both the $\Span(\Seti{V}{1\dots n})$ and $\Set{U}{n+1}$ are convex, the new vector span is the Minkowski sum of the two sets, which is proven to be convex:

\begin{equation}
    \Span(\Seti{V}{1\dots (n + 1)}) = \Span(\Seti{V}{1\dots n}) +^M \Set{U}{n+1}
\end{equation}

\noindent where $+^M$ is the Minkowski sum of the two sets. 

The resulting $\Span(\Seti{V}{1\dots (n+1)})$ can be obtained by the sum of infinite sets resulting from shifting the $\Span(\Seti{V}{1\dots n})$ by all points in $[min_{n+1}, max_{n+1}]$ interval in the direction of $\vec{v}_{n+1}$ vector. The result is equivalent to shifting $\Span(\Seti{V}{1\dots n})$ in the direction of $\vec{v}_{n+1}$ vector by $min_{n+1}$, then shifting it by $max_{n+1}$ and taking the convex hull of the two sets. Note that the $\Span(\Seti{V}{1\dots n})$ is already assumed to be the convex hull created from only the $2^r$ combinations of minimum and maximum thrust magnitudes. 

Now, considering that $\Span(\Seti{V}{1\dots (n + 1)})$ is the convex hull created only by the vertices of $\Span(\Seti{V}{1\dots n})$ shifted once by $min_{n+1}$ and once by $max_{n+1}$, the vertices of the convex hull can only be a subset of the vertices of $\Span(\Seti{V}{1\dots n})$ shifted by $min_{n+1}$ or $max_{n+1}$ in $\vec{v}_{n+1}$ direction.

\end{proof}

Noting that moments are also fundamentally vectors and have vector properties, a similar algorithm can be used to compute the moment set $\SetM(t)$. The only difference would be using the computed moments instead of thrusts at each iteration (i.e., using $CalculateTotalMoment$ instead of $CalculateTotalThrust$) in Algorithm~\ref{alg:wrench:decoupled:thrust-set-estimation}.

\begin{figure}[!htb]
        \centering
        
        \begin{subfigure}{0.40\textwidth}
            \includegraphics[width=\textwidth]{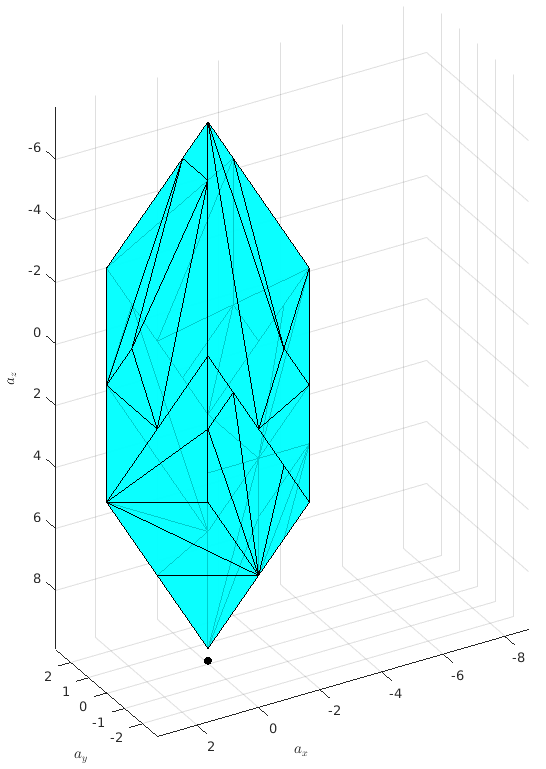}
            \caption{~}
            \label{fig:wrench:decoupled:thrust-set-tilted-hex-rotation-a}
        \end{subfigure}
        \hspace{1cm}
        \begin{subfigure}{0.40\textwidth}
            \includegraphics[width=\textwidth]{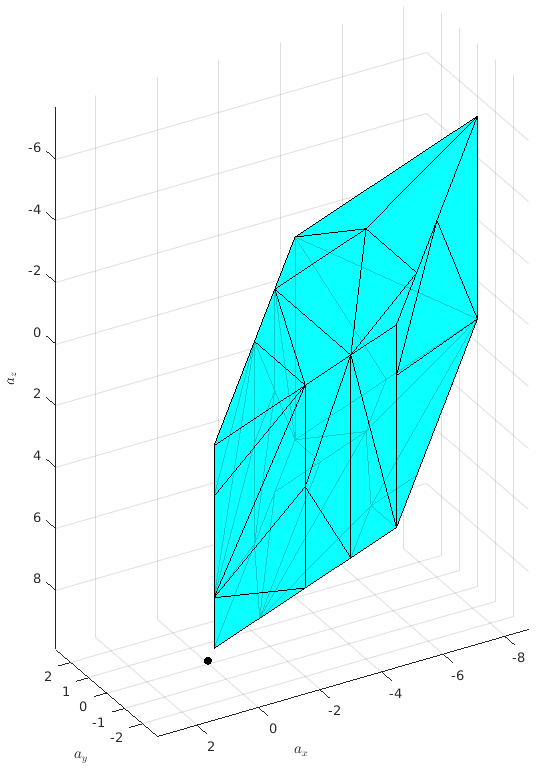}
            \caption{~}
            \label{fig:wrench:decoupled:thrust-set-tilted-hex-rotation-b}
        \end{subfigure}

        \caption[Thrust sets for different attitudes]{Thrust sets for architectures with fixed rotor angles rotate around the point of the total external force, shown as a black dot in the plots. The thrust sets are shown for our fixed-pitch hexarotor architecture at (a) zero roll, pitch, and yaw. (b) zero roll and yaw, but $30^\circ$ pitch.}
        
        \label{fig:wrench:decoupled:thrust-set-tilted-hex-rotation}
    \end{figure}

For multirotors with fixed rotor angles (i.e., fixed-pitch multirotors) and with a second-order model, the \textit{shapes} of the feasible thrust and moment sets approximated by Algorithm~\ref{alg:wrench:decoupled:thrust-set-estimation} ($\SetTp$) are independent of the current state and the external wrenches. However, the set's orientation depends on the current robot's attitude, and its location in the body-fixed frame $\FB$ translates with the external forces (e.g., gravity and wind). This observation for these architectures can be leveraged to speed up the computation significantly. A base set (e.g., $\SetTb$) can be computed only once in the body-fixed frame, ignoring the external forces. Then, to obtain the thrust set $\SetTp$ for any external force and robot attitude, the base set $\SetTb$ can be rotated to the inertial frame and then shifted with the external forces. Figure~\ref{fig:wrench:decoupled:thrust-set-tilted-hex-rotation} shows how the same thrust set rotates around when the UAV attitude changes. 

Our experiments (Section~\ref{sec:wrench:tests}) show that the time complexity of this rotation method is still exponential to the number of rotors (similar to Algorithm~\ref{alg:wrench:decoupled:thrust-set-estimation}). However, utilizing it is still a few orders of magnitude faster than recalculating the entire wrench space at each iteration. Note that, even though this approach can significantly improve the execution speed, its use is only limited to the cases where all the following conditions are met: the robot is a fixed-pitch multirotor, the robot model is second-order, and the coupling of thrusts and moments is not important for the application.

The following section will present a solution to capturing the coupling between thrusts and moments when the goal is to estimate the lateral thrusts available to the UAV.

\section{Coupled Lateral Thrust Estimation} \label{sec:wrench:lateral}

The thrust set approximated by Algorithm~\ref{alg:wrench:decoupled:thrust-set-estimation} is more accurate compared to the fixed thrust set assumption of the current literature. However, it does not take into account the required moments when the thrust is generated. The actual thrust set with a nonzero moment is usually an even smaller set. In fully actuated multirotors, the moments should have priority over the thrusts (see Section~\ref{sec:control:design}). If we assume that the multirotor can achieve the moments without delay (a reasonable assumption for multirotors), the generated moments that affect the thrust set would be the same as the current moment setpoint.

If the goal is only to estimate the UAV's lateral thrust bound, a Monte Carlo method can be devised that calculates the lateral thrust set for a desired normal thrust $\vecelem{F\spt}{z}$. The method devises the available Control Allocation module and takes the current system state and the current moment setpoint into account. Similar to Algorithm~\ref{alg:wrench:decoupled:thrust-set-estimation}, it requires knowledge about the input limits of the motors. 

Algorithm~\ref{alg:wrench:lateral:lateral-thrust-estimation} illustrates the proposed lateral thrust estimation method. Note that the algorithm requires an additional moment setpoint (or desired angular acceleration) input to the Thrust Setpoint Generator module, which is illustrated in Figures~\ref{fig:control:controller} and~\ref{fig:control:thrust:thrust-setpoint-generation-module} in dashed boxes.

\begin{algorithm}[!htb]
\caption{Proposed approach for lateral thrust limits estimation}
\label{alg:wrench:lateral:lateral-thrust-estimation}
\begin{algorithmic}[1]
\LineComment {This function estimates the lateral thrust for input UAV model, current state, desired normal thrust and desired angular acceleration (alpha)}
\Function{EstimateLateralThrust}{\var{ctrlAlloc}, \var{model}, \var{state}, \var{zd}, \var{alphad}}
    \LineComment {Initialize the control allocation}
    \State $\var{ctrlAlloc} \gets \func{Init}(\var{ctrlAlloc}, \var{model}, \var{state})$
    \LineComment {Extract system's min and max input values as $r\times1$ arrays, with each cell corresponding to a motor/rotor}
    \State $(\var{minU}, \var{maxU}) \gets \func{GetInputRange}(\var{model})$
    \LineComment {Get the possible range for the UAV lateral thrust}
    \LineComment {This can be guessed or calculated from Algorithm~\ref{alg:wrench:decoupled:thrust-set-estimation}}
    \State $(\var{xRange}, \var{yRange}) \gets \func{GetLateralRange}(model)$
    \LineComment {Define an empty set for the thrusts}
    \State $\var{thrusts} \gets \emptyset$
    \LineComment {Generate and test random samples}
    \LineComment {$K$ is the desired number of iterations}
    \For {$\var{i} = 1$ \textbf{to} $\var{K}$} 
        \LineComment {Choose random numbers for $x$ and $y$ thrusts}
        \State $(\var{xd}, \var{yd}) \gets \func{Rand}(\var{xRange}, \var{yRange})$
        \LineComment {Form the desired thrust vector}
        \State $\var{thrustd} \gets \func{MakeVector}(\var{xd}, \var{yd}, \var{zd})$
        \LineComment {Check if the inputs are in the valid range}
        \State $\var{u} \gets \func{CalcInput}(\var{ctrlAlloc}, \var{thrustd}, \var{alphad}))$
        \LineComment {Form the desired thrust vector}
        \State $\var{thrustd} \gets \func{MakeVector}(\var{xd}, \var{yd}, \var{zd})$
        \LineComment {Check if the inputs are in the valid range}
        \If {\func{IsInRange}(\var{u}, \var{minU}, \var{maxU})}
            \LineComment {Add the valid thrust to the thrust set}
            \State $\var{thrusts} \gets \func{AddToSet}(\var{thrusts}, \var{thrustd})$
        \EndIf
    \EndFor
    \LineComment {Calculate the 2-D convex hull of the thrusts}
    \State $\var{latheralThrusts} \gets \func{ConvexHull}({\var{thrusts})}$
    \LineComment {Return the result}
    \State \Return $\var{lateralThrusts}$
\EndFunction
\end{algorithmic}
\end{algorithm}

The execution time depends on the speed of the Control Allocation's input calculation function (e.g., the mixer function) and the selected number of iterations $K$. The choice of $K$ is a trade-off between the execution time (keeping it real-time) and the precision required in estimating the lateral thrust. 

The following section will show a more general solution to capturing the coupling between thrusts and moments.

\section{Coupled Wrench Set Estimation} \label{sec:wrench:coupled}

Algorithm~\ref{alg:wrench:decoupled:thrust-set-estimation} proposed in Section~\ref{sec:wrench:decoupled} can provide a real-time estimation of the thrust and moment sets (i.e., $\SetT$ and $\SetM$), which depends on the design of the multirotor and its current state. However, it does not capture the coupling between the thrusts and moments, which is especially necessary for physical interaction applications. This section describes our real-time wrench set estimation method that considers the desired forces and moments in its calculation.

Let us assume $\SetW(t)$ is the set of all feasible 6-D wrenches that our multirotor can generate at time $t$. Similar to the previous section, we omit the time dependency notation $(t)$ from now on when the context is clear. The set $\SetW$ not only depends on the UAV's design, the current state, and the environment but also on the desired moments and thrusts for the physical interaction with the environment. In this section, when we mention the current state, we consider the external forces (such as wind and gravity) also included in the current state.

In practice, our main interest in $\SetW$ is to modify the thrust and moment setpoints, so they lie within the wrench set when some desired thrust and moment components are already fixed (e.g., when applying a 2~$\unit{N}$ force to the wall). Therefore, the goal is to calculate an estimation $\SetWp$ to the union of all possible sets $\SetW(t + \delta t)$ computed for a short time period $\delta t$ from time $t$. The $\delta t$ is the shortest time that allows the thrust and moment setpoints to realize (i.e., $\delta t$ is assumed to be larger than system and motor delays but too small for any other state or environment variables to change meaningfully).

The proposed real-time method for estimating the wrench set $\SetW$ first calculates the 6-D full wrench sets similar to Algorithm~\ref{alg:wrench:decoupled:thrust-set-estimation}, ignoring the knowledge about the desired moment and thrust components. Algorithm~\ref{alg:wrench:coupled:6d-wrench-set-estimation} shows the steps to calculate the full 6-D wrench set $\SetWf$.

\begin{algorithm}[!htb]
\caption{Proposed approach for 6-D wrench set estimation}
\label{alg:wrench:coupled:6d-wrench-set-estimation}
\begin{algorithmic}[1]
\LineComment {This function estimates the 6-D wrench set for the input UAV model and the current state}
\Function{Estimate6DWrenchSet}{\var{model}, \var{state}}
    \LineComment {Update the UAV model state to the input state}
    \State $\var{model} \gets \func{SetState}(\var{model}, \var{state})$
    \LineComment {Get the number of rotors in the UAV}
    \State $\var{r} \gets \func{GetNumOfRotors}(\var{model})$
    \LineComment {Extract system's min and max input values as $r\times1$ arrays, with each cell corresponding to a motor/rotor}
    \State $(\var{minU}, \var{maxU}) \gets \func{GetInputRange}(\var{model})$
    \LineComment {Define an empty set for the 6-D wrenches}
    \State $\var{wrenches} \gets \emptyset$
    \LineComment {Generate wrenches from all min/max combinations of motor commands}
    \For {$\var{i} = 0$ \textbf{to} $2^r-1$} 
        \LineComment {Convert the iterator $\var{i}$ into an $r\times1$ binary array}
        \State $\var{iArray} \gets \func{NumToBinaryArray}(\var{i}, \var{r})$
        \State $\var{iArrayNegated} \gets \func{Not}(\var{iArray})$
        \LineComment {Calculate the input array from the binary array}
        \LineComment {$\odot$, $\oplus$: element-wise multiplication and addition}
        \State $\var{u} \gets (\var{maxU} \odot \var{iArray}) \oplus (\var{minU} \odot \var{iArrayNegated})$
        \LineComment {Calculate the wrench generated by the model}
        \State $\var{w} \gets \func{CalculateTotal6DWrench}(model, u)$
        \LineComment {Add the calculated wrench to the wrench set}
        \State $\var{wrenches} \gets \func{AddToSet}(\var{wrenches}, \var{w})$
    \EndFor
    \LineComment {Calculate the convex hull of the wrenches}
    \State $\var{feasibleWrenches} \gets \func{ConvexHull}({\var{wrenches})}$
    \LineComment {Return the result}
    \State \Return $\var{feasibleWrenches}$
\EndFunction
\end{algorithmic}
\end{algorithm}

Now, assume that the 6-D wrench set $\SetWf$ is computed, and we would like to know the feasible wrenches when we are applying (or desiring to apply) a specific force or moment to the environment during the physical interaction. The result is all the forces and moments inside $\SetWf$ that lie on the hyperplane defined by the fixed force or moment component. In other words, to compute the set of feasible wrenches when one of the 6-D dimensions is fixed, we can simply intersect the hyperplane of the fixed dimension with the convex hull of the wrench set $\SetWf$. The intersection represents the feasible set of wrenches (i.e., forces and moments) where the given wrench dimension is fixed to the desired value. Note that the intersection of a hyperplane with a convex shape is a convex shape with a smaller dimension. Therefore, the resulting wrench set after the intersection is also convex.

To take it further, for each desired dimension of forces and moments, we can iteratively intersect the wrench set and get the new wrench set with the desired constraints. Algorithm~\ref{alg:wrench:coupled:wrench-set-estimation-with-desired-components} shows the steps for estimating the wrench set $\SetW$ when one or more components of the forces and moments are fixed to the desired value.

\begin{algorithm}[!htb]
\caption{Wrench set estimation with desired (fixed) components}
\label{alg:wrench:coupled:wrench-set-estimation-with-desired-components}
\begin{algorithmic}[1]
\LineComment {This function estimates the wrench set for the input UAV model, the current state and the desired wrench components}
\Function{EstimateWrenchSet}{\var{model}, \var{state}, \var{desiredWrenches}}
    \LineComment {Estimate the full 6-D wrenches first}
    \State $\var{feasibleWrenches} \gets \func{Estimate6DWrenchSet}(\var{model}, \var{state})$
    \LineComment {Iterate through the fixed (desired) wrench dimensions}
    \For {$\var{i} = 1$ \textbf{to} \func{Size}(\var{desiredWrenches})} 
        \LineComment {Get the hyperplane for the fixed dimension}
        \State $\var{hyperplane} \gets \func{ConstructHyperplane}(desiredWrenches[\var{i}])$
        \LineComment {Intersect the hyperplane with the convex region of wrenches}
        \State $\var{feasibleWrenches} \gets \func{Intersect}(\var{feasibleWrenches}, \var{hyperplane})$
    \EndFor
    \LineComment {Return the result}
    \State \Return $\var{feasibleWrenches}$
\EndFunction
\end{algorithmic}
\end{algorithm}

The final result of the proposed approach is a convex (can be empty) set of all the feasible wrenches with the desired components fixed. Note that if $n$ fixed dimensions are desired, the final convex set $\SetW$ can have a dimension of at most $6-n$, and an empty set means that the UAV cannot achieve all the desired forces and moments at the same time. 

Section~\ref{sec:wrench:tests} shows the results of our implementation of the algorithm and illustrates how different desired wrench components can affect the wrench set.

The following section briefly describes how this method can be further extended to multirotors with variable-pitch rotors.
\section{Additional Extensions of the Method} \label{sec:wrench:extensions}

The proposed Algorithm~\ref{alg:wrench:coupled:wrench-set-estimation-with-desired-components} presented in Section~\ref{sec:wrench:coupled} estimates the instantaneous wrench set $\SetW$ for multirotors with fixed-pitch rotors. For multirotors with variable-pitch rotors, the same algorithm can estimate the instantaneous wrench set $SetW$ given the following two conditions:
\begin{enumerate}
    \item The multirotor state also includes the current pitch of the rotors (i.e., the angle of the servos controlling the rotor pitch),
    \item For the purposes of the wrench set estimation, the change rates for the rotor pitch angles are assumed to be low, allowing us to ignore the rotor pitch changes for the short future time $\delta t$, which is used for wrench set definition (see Section~\ref{sec:wrench:decoupled}).
\end{enumerate}

\begin{figure}[!htb]
        \centering
        \includegraphics[width=0.5\linewidth]{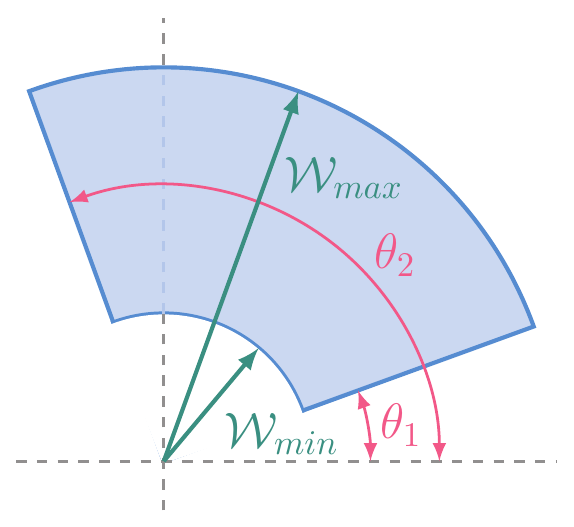}
        
        \caption[Space of possible wrenches for a variable-pitch rotor]{The space of possible generated wrenches for a single rotor connected to a servo pitching with the angle limit between $\theta_1$ to $\theta_2$.}
        
        \label{fig:wrench:variable-pitch-sector}
    \end{figure}

However, there are applications where estimating the set of all possible wrenches is desired, assuming that the rotor pitches can take any angle within their limits. For example, when a single or more motors fail, knowing the whole wrench space for any rotor pitch angle is desired to recover from the failure. Remember that each rotor can generate wrenches in a single direction for fixed-pitch rotors and with the magnitude between its minimum and maximum wrench. In the variable-pitch rotor scenario, each servo motor connected to the rotor can create a sweep over the space, resulting in the space of possible wrenches looking like a sector of a circle between the angles of the servo rotor and filling the radius between the minimum and maximum wrench of the rotor itself. Figure~\ref{fig:wrench:variable-pitch-sector} illustrates the range of possible wrenches for a single rotor that is connected to a servo pitching from $\theta_1$ to $\theta_2$.

With such a non-linear space of possible wrenches for each rotor, quickly combining two or more spaces becomes intractable and difficult to compute analytically. However, it is possible to estimate the wrench space by sampling different pitch angles. Section~\ref{sec:wrench:tests} illustrates this idea and Chapter~\ref{ch:applications} shows some of the applications of these analysis.

\section{Experiments and Results} \label{sec:wrench:tests}

\begin{figure}[!htb]
        \centering
        \begin{minipage}[b]{.48\linewidth}
            \begin{minipage}[b]{.54\linewidth}
                \begin{subfigure}[b]{\linewidth}
                    \includegraphics[width=\textwidth]{ch-wrench/thrust-set-tilted-hex}
                    \caption{~}
                    \label{fig:wrench:tests:thrust-set-tilted-hex-a}
                \end{subfigure}
                
                \medskip
                \begin{subfigure}[b]{\linewidth}
                    \includegraphics[width=0.95\textwidth]{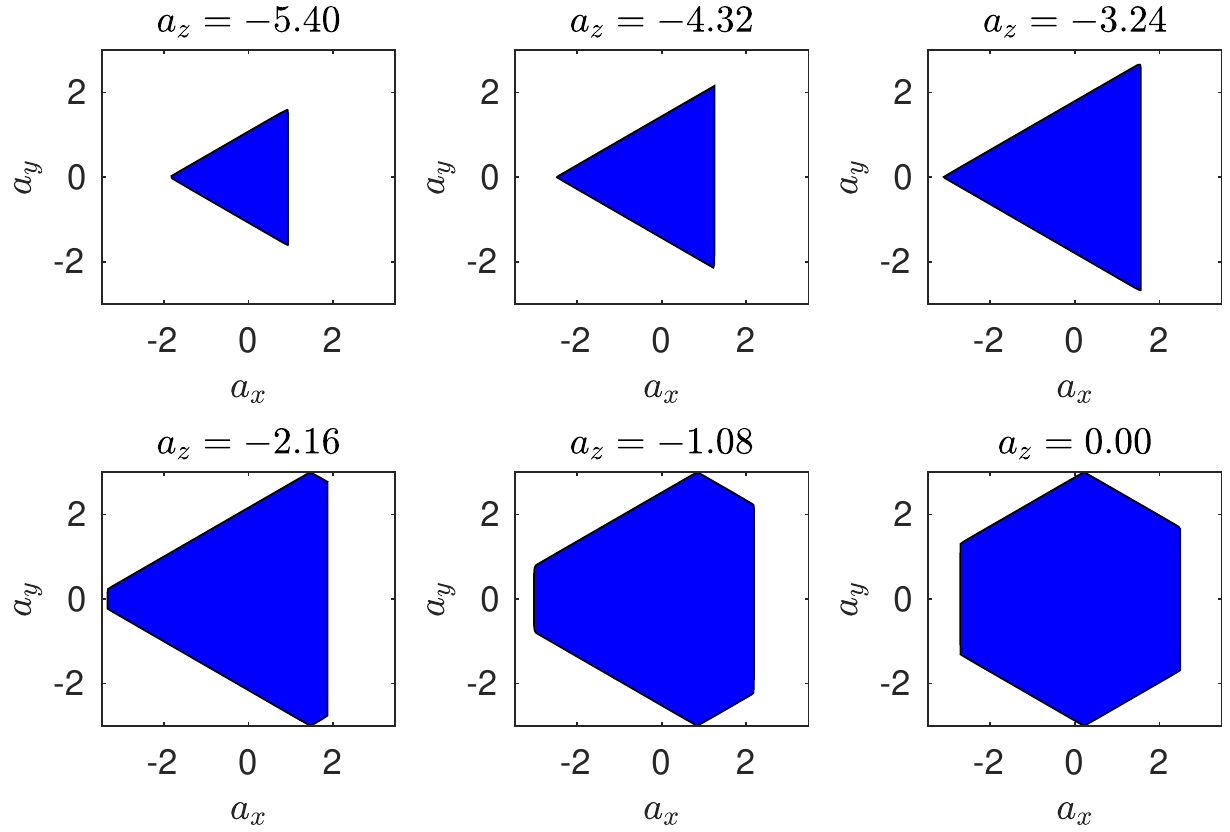}
                    \caption{~}
                    \label{fig:wrench:tests:thrust-set-tilted-hex-b}
                \end{subfigure}
            \end{minipage}
            \hfill
            \begin{minipage}[t]{.42\linewidth}
                \begin{subfigure}[b]{\linewidth}
                    \includegraphics[width=\textwidth, height=6cm]{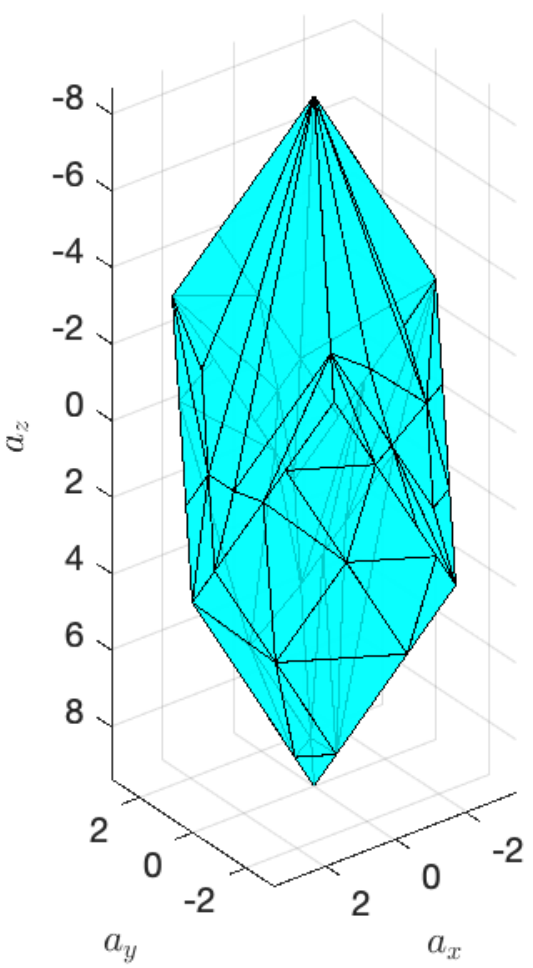}
                    \caption{~}
                    \label{fig:wrench:tests:thrust-set-tilted-hex-c}
                \end{subfigure}
            \end{minipage}

            \medskip
            
            \begin{subfigure}[b]{0.96\linewidth}
                \includegraphics[width=\textwidth, height=4cm]{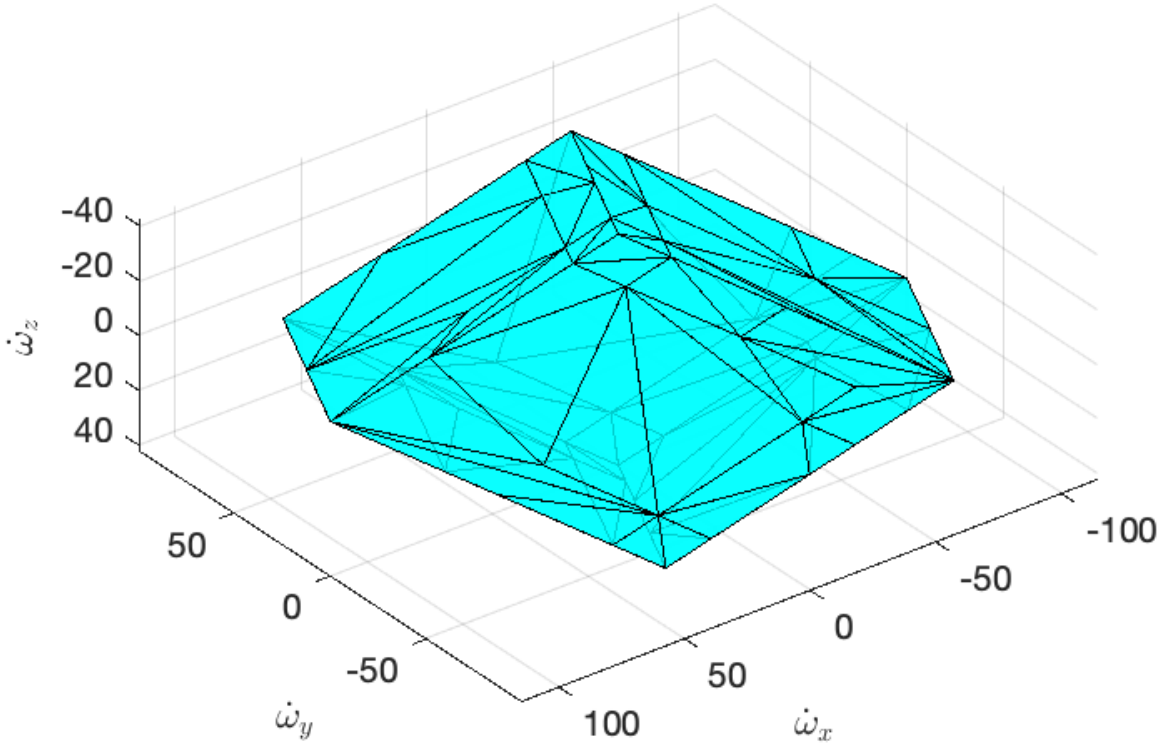}
                \caption{~}
                \label{fig:wrench:tests:thrust-set-tilted-hex-d}
            \end{subfigure}
        \end{minipage}
        \hfill
        \begin{minipage}[b]{.48\linewidth}
            \begin{minipage}[b]{.54\linewidth}
                \begin{subfigure}[b]{0.95\linewidth}
                    \includegraphics[width=\textwidth]{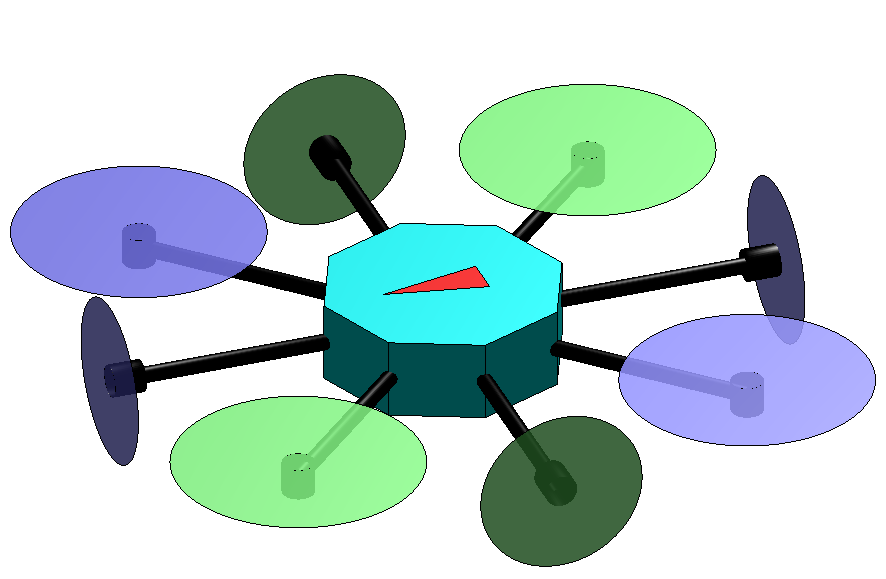}
                    \caption{~}
                    \label{fig:wrench:tests:thrust-set-octorotor-e}
                \end{subfigure}
                
                \medskip
                \begin{subfigure}[b]{0.9\linewidth}
                    \includegraphics[width=\textwidth]{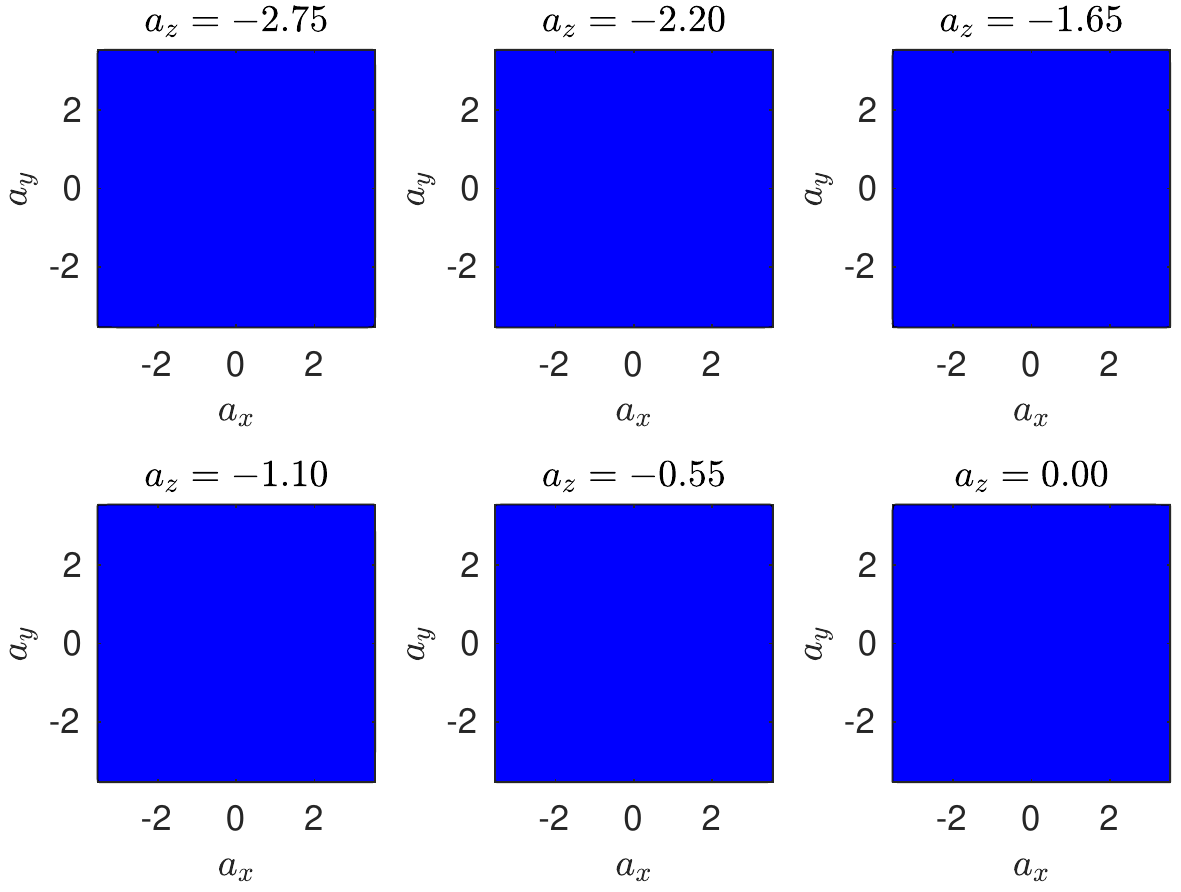}
                    \caption{~}
                    \label{fig:wrench:tests:thrust-set-octorotor-f}
                \end{subfigure}
            \end{minipage}
            \hfill
            \begin{minipage}[t]{.42\linewidth}
                \begin{subfigure}[b]{0.8\linewidth}
                    \includegraphics[width=\textwidth, height=6cm]{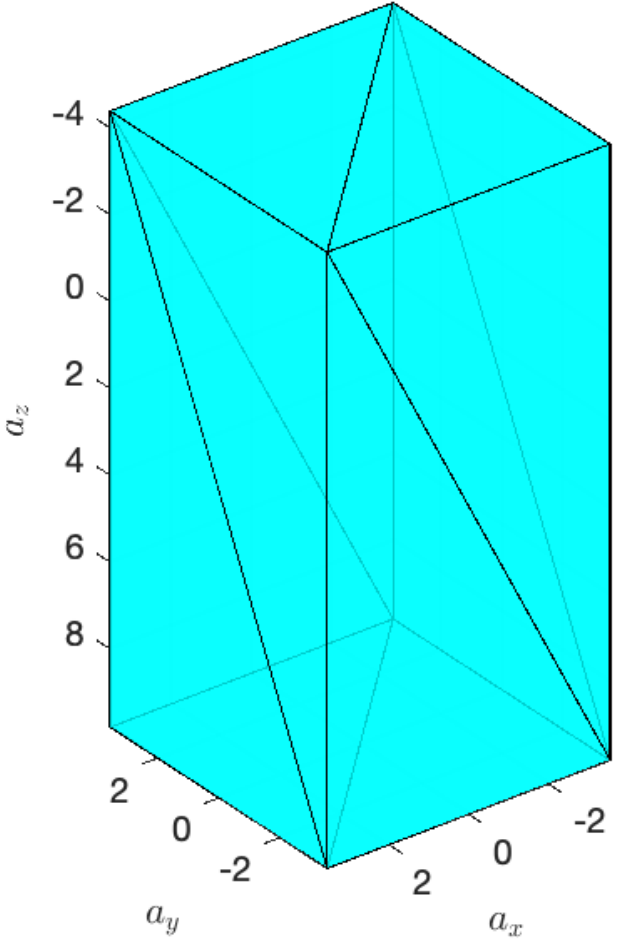}
                    \caption{~}
                    \label{fig:wrench:tests:thrust-set-octorotor-g}
                \end{subfigure}
            \end{minipage}

            \medskip
            
            \begin{subfigure}[b]{0.96\linewidth}
                \includegraphics[width=\textwidth, height=4cm]{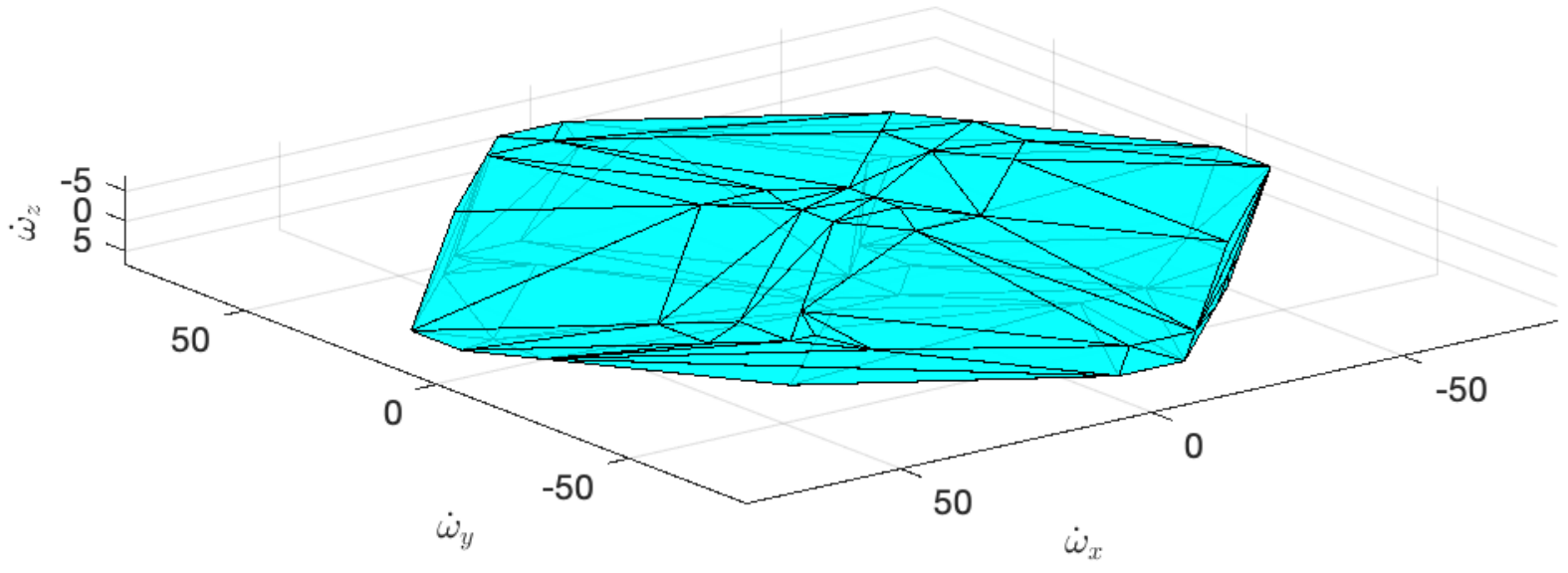}
                \caption{~}
                \label{fig:wrench:tests:thrust-set-octorotor-h}
            \end{subfigure}
        \end{minipage}
        
        \caption[Thrust and moment sets for different hardware architectures]{Thrust and moment sets have different shapes depending on the architecture of the fully-actuated UAV. (a) A hexarotor with rotors tilted sideways along with the (d) moment set, (b, c) thrust set, and its cross-sections along the $\ZI$ axis. The larger $Z$-thrust reduces the available lateral thrust. (e) An octorotor with four co-planar upward rotors and four auxiliary motors perpendicular to the main rotors along with the (h) moment set, (f, g) thrust set and its cross-sections along the $\ZI$ axis. The lateral thrust in this architecture is entirely independent of the normal thrust.}
        
        \label{fig:wrench:tests:thrust-set-architectures}
    \end{figure}

\begin{figure}[!htb]
    \centering
    \begin{subfigure}{0.3\textwidth}
        \includegraphics[width=\textwidth]{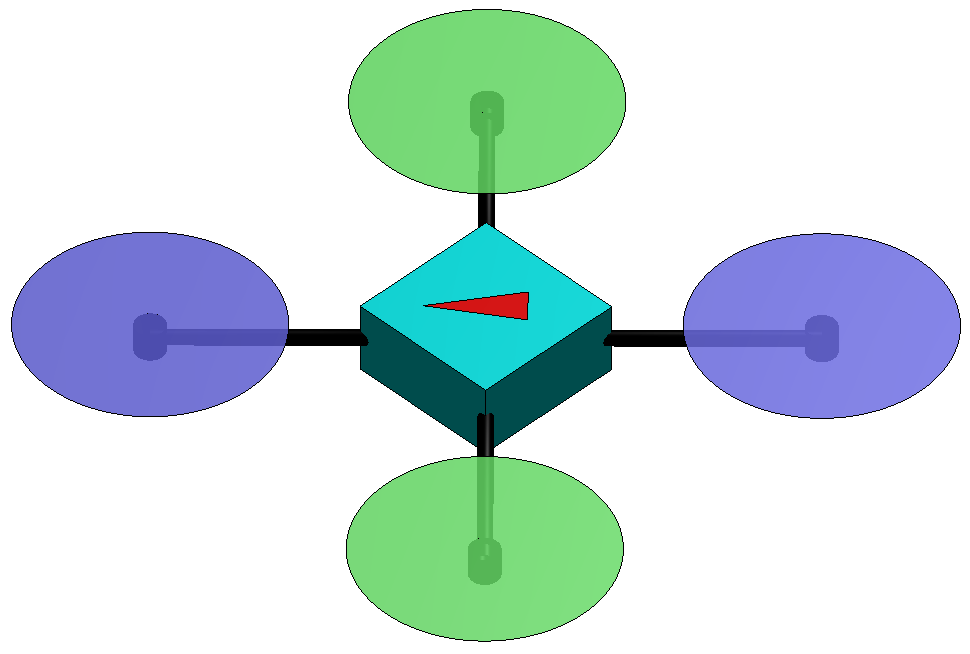}
        \caption{~}
        \label{fig:wrench:tests:thrust-set-states-a}
    \end{subfigure}
    ~
    \begin{subfigure}{0.15\textwidth}
        \includegraphics[width=\textwidth, height=4.5cm]{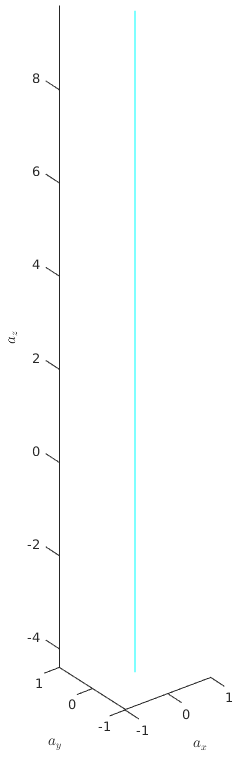}
        \caption{~}
        \label{fig:wrench:tests:thrust-set-states-c}
    \end{subfigure}
    \hfill
    \begin{subfigure}{0.3\textwidth}
        \includegraphics[width=\textwidth]{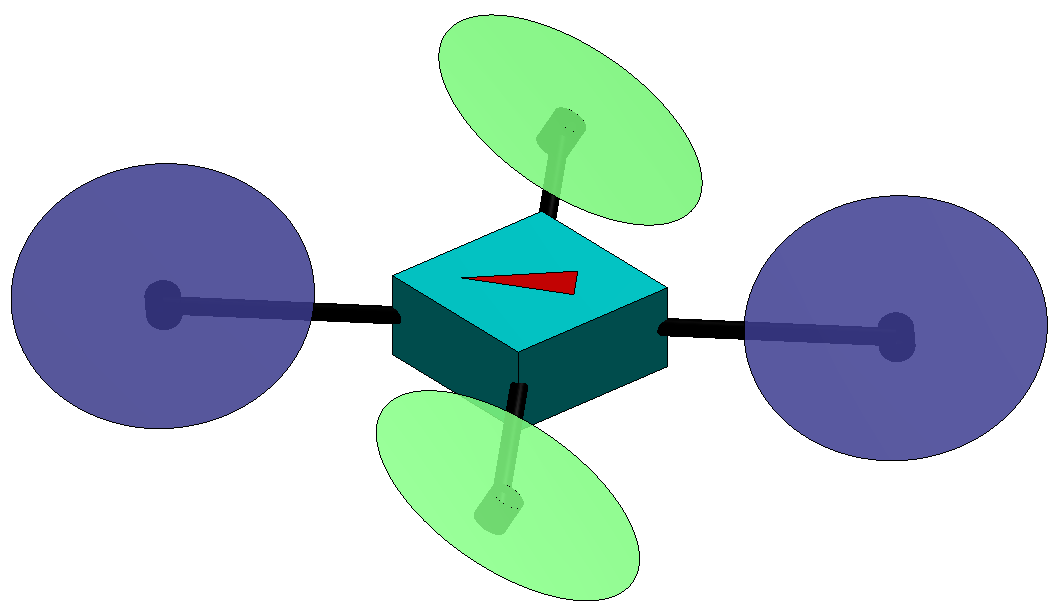}
        \caption{~}
        \label{fig:wrench:tests:thrust-set-states-b}
    \end{subfigure}
    ~
    \begin{subfigure}{0.15\textwidth}
        \includegraphics[width=\textwidth]{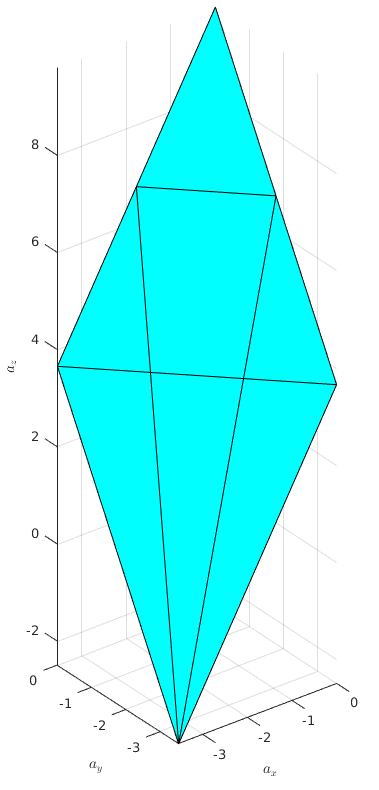}
        \caption{~}
        \label{fig:wrench:tests:thrust-set-states-d}
    \end{subfigure}

    \caption[Thrust sets for different robot states]{Thrust and moment sets have different shapes depending on the state of the UAV. A quadrotor with variable-pitch thrusters is shown along with its thrust set at different states: (a, c) When all the rotors are upward and parallel, the thrust set is a line. (b, d) Front and back rotors are tilted to the left, and side rotors are tilted to the back. The thrust set is a planar subspace.}
    \label{fig:wrench:tests:thrust-set-states}
    \end{figure}

All the methods proposed in this section have been implemented and tested. Figures~\ref{fig:wrench:tests:thrust-set-architectures} and~\ref{fig:wrench:tests:thrust-set-states} illustrate how Algorithm~\ref{alg:wrench:decoupled:thrust-set-estimation} can estimate the thrust and moment sets of different multirotor architectures. All the thrust sets are estimated for zero attitude. The cross-sections calculated using Algorithm~\ref{alg:wrench:lateral:lateral-thrust-estimation} show how the lateral thrust changes with the change in the desired $Z$ thrust.

Figure~\ref{fig:wrench:tests:thrust-set-architectures} demonstrates how the thrust and moment sets can significantly differ for different multirotor architectures. On the other hand Figure~\ref{fig:wrench:tests:thrust-set-states} highlights how different UAV states can also result in very different thrust and moment sets.

\begin{figure}[!htb]
    \centering
    \begin{subfigure}{0.15\textwidth}
        \includegraphics[width=\textwidth, height=4.5cm]{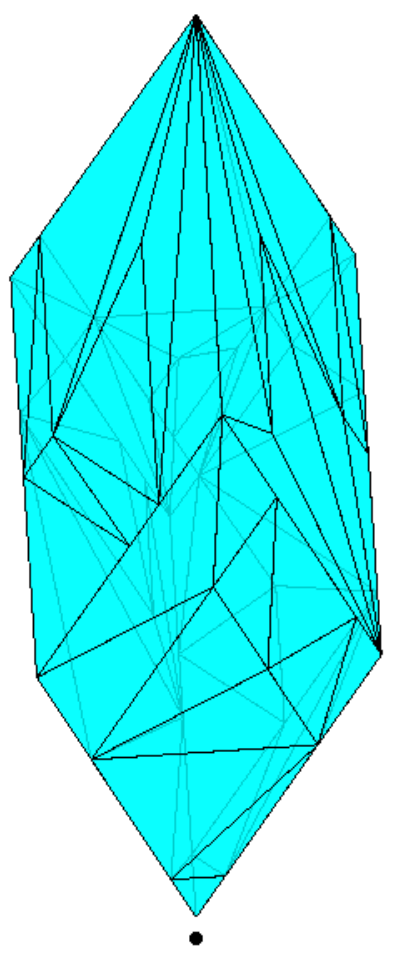}
        \caption{~}
    \end{subfigure}
    ~
    \begin{subfigure}{0.17\textwidth}
        \includegraphics[width=\textwidth,height=4.5cm]{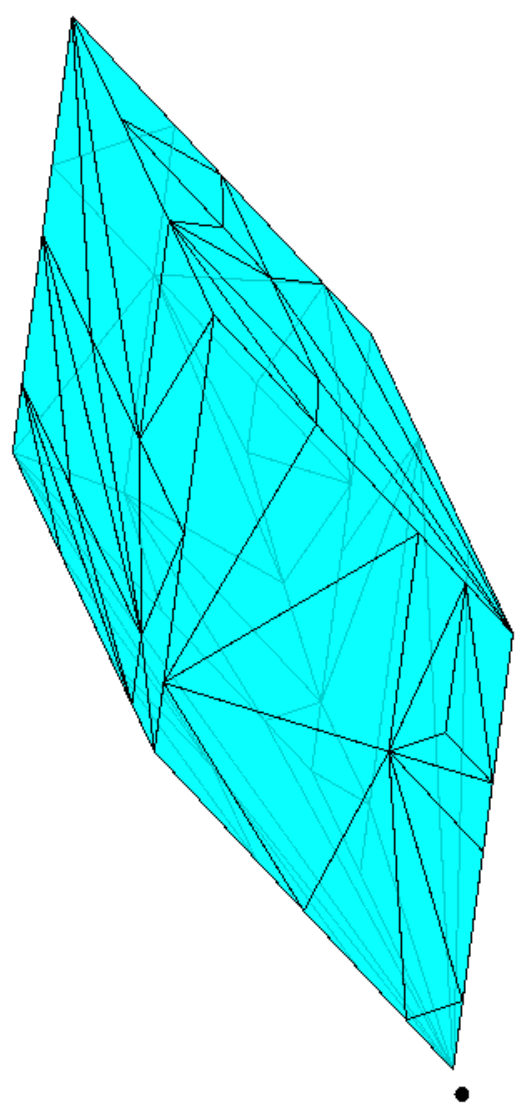}
        \caption{~}
    \end{subfigure}
    ~~~~~~
    \begin{subfigure}{0.15\textwidth}
        \includegraphics[width=\textwidth,height=4.5cm]{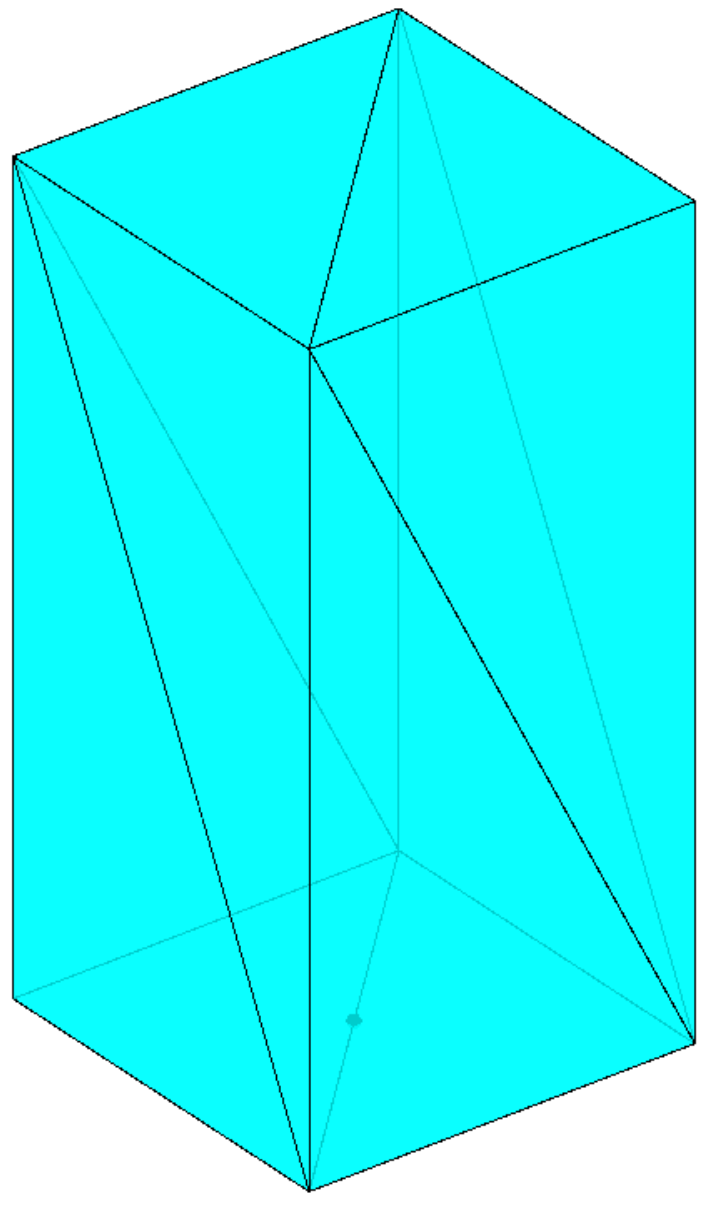}
        \caption{~}
    \end{subfigure}
    ~
    \begin{subfigure}{0.19\textwidth}
        \includegraphics[width=\textwidth,height=4.5cm]{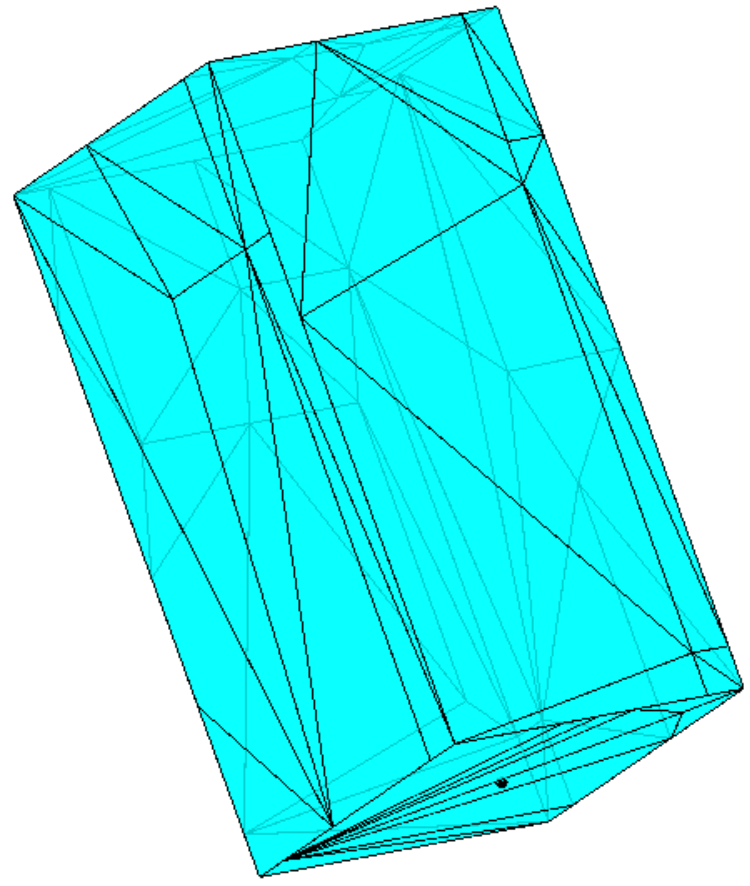}
        \caption{~}
    \end{subfigure}

    \caption[Thrust sets for UAV rotations]{In fixed-pitch multirotors, the thrust sets have a fixed shape and are fixed in the body-fixed frame $\FB$. Illustrated are thrust sets at different robot orientations for (a, b) the hexarotor in Figure~\ref{fig:wrench:tests:thrust-set-tilted-hex-a}, and (c, d) the octorotor in Figure~\ref{fig:wrench:tests:thrust-set-octorotor-e}. The center of rotation is shown with a dot at the bottom of each shape.}
    \label{fig:wrench:tests:thrust-set-rotations}
    \end{figure}

The shapes of the thrust and moment sets computed using Algorithm~\ref{alg:wrench:decoupled:thrust-set-estimation} depend on the architecture as well as the robot's state. However, as discussed in Section~\ref{sec:wrench:decoupled}, for the specific class of multirotors with fixed-pitch rotors, the shapes of the decoupled thrust and moment sets do not change, i.e., only rotating with the body-fixed frame in the inertial frame $\FI$ and shifting with the external forces. Figure~\ref{fig:wrench:tests:thrust-set-rotations} shows thrust sets computed for different orientations of two robot architectures with fixed-pitch rotors.

\begin{figure}[!htb]
        \centering
        
        \begin{subfigure}{0.48\textwidth}
            \includegraphics[width=\textwidth, height=5cm]{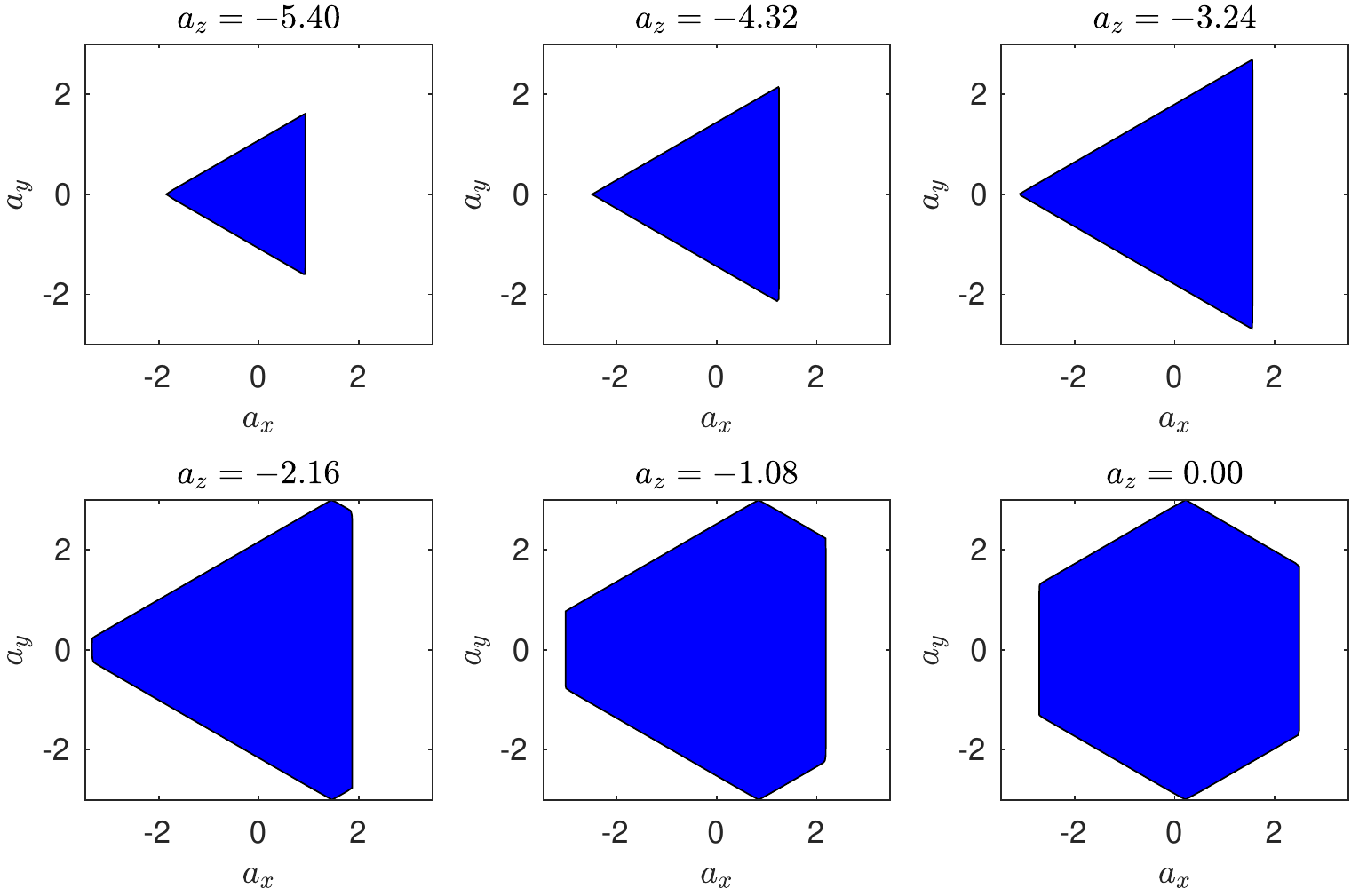}
            \caption{~}
            \label{fig:wrench:tests:lateral-thrust-estimation-a}
        \end{subfigure}
        ~
        \begin{subfigure}{0.48\textwidth}
            \includegraphics[width=\textwidth, height=5cm]{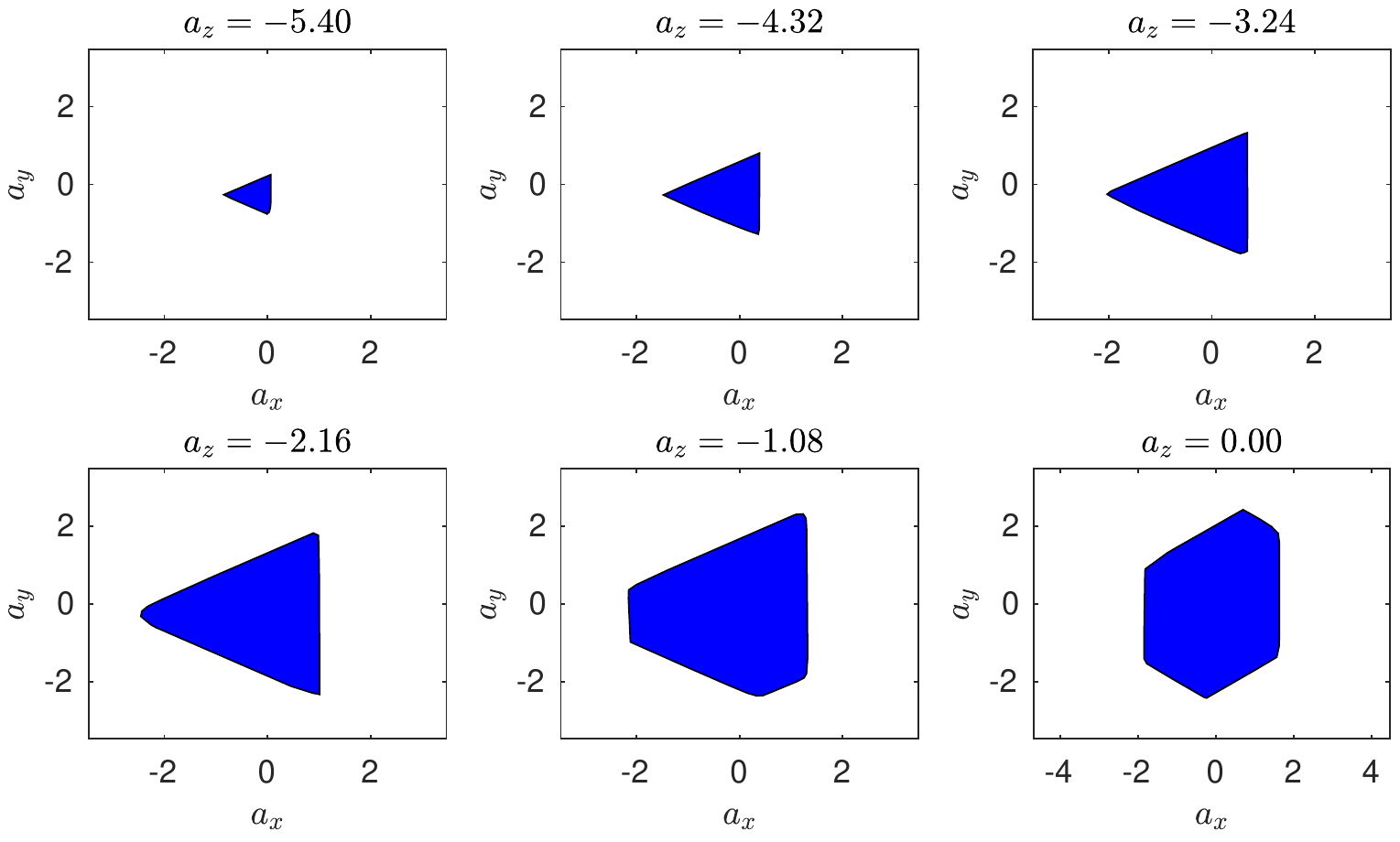}
            \caption{~}
            \label{fig:wrench:tests:lateral-thrust-estimation-b}
        \end{subfigure}
        
        \caption[Thrust sets for different moment setpoints]{Thrust set depends on the desired angular accelerations (moments) as well as the UAV architecture and state. The lateral thrust set calculated for different normal thrusts by Algorithm~\ref{alg:wrench:lateral:lateral-thrust-estimation} for the hexarotor of Figure~\ref{fig:wrench:tests:thrust-set-tilted-hex-a} at the same state. Each figure (a) and (b) shows the thrust sets calculated with the same desired angular acceleration but different desired normal thrusts. The two figures are plotted for the same set of desired normal thrusts and only differ in their desired angular accelerations.}
        
        \label{fig:wrench:tests:lateral-thrust-estimation}
    \end{figure}

Figure~\ref{fig:wrench:tests:lateral-thrust-estimation} shows the lateral thrust sets calculated using Algorithm~\ref{alg:wrench:lateral:lateral-thrust-estimation} for the hexarotor with tilted rotors (in Figure~\ref{fig:wrench:tests:thrust-set-tilted-hex-a}) with different desired normal thrusts and angular accelerations. All the thrust sets are calculated for the same architecture and state, but the resulting limits are very different.

\begin{figure}[!htb]
    \centering
    \begin{subfigure}{0.32\linewidth}
        \includegraphics[width=\textwidth]{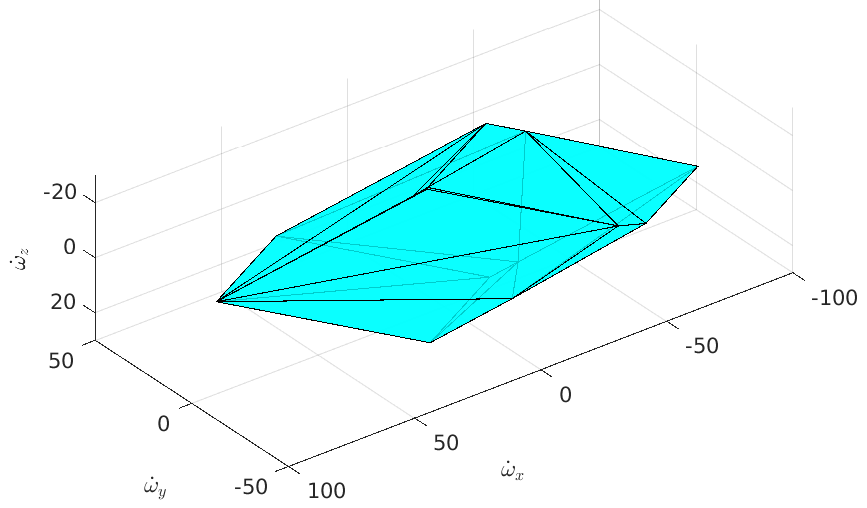}
        \caption{~}
        \label{fig:wrench:tests:moment-set-with-y-thrust-a}
    \end{subfigure}
    \hfill
    \begin{subfigure}{0.32\linewidth}
        \includegraphics[width=\textwidth]{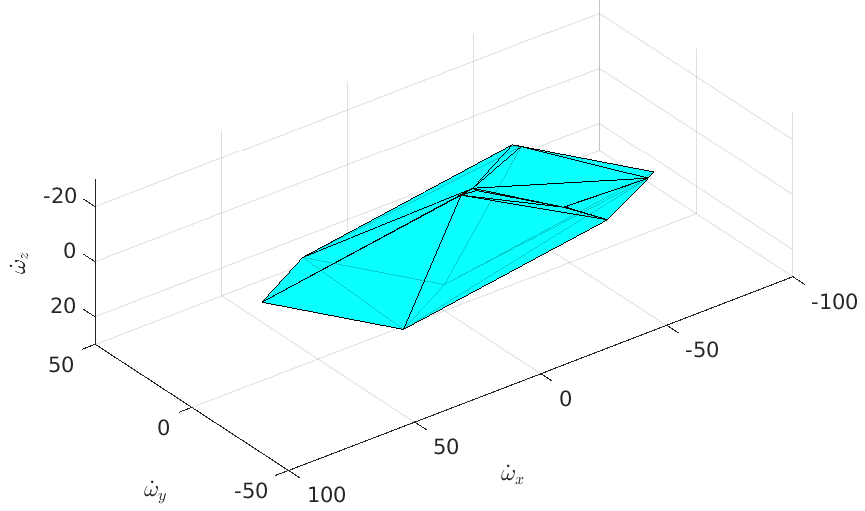}
        \caption{~}
        \label{fig:wrench:tests:moment-set-with-y-thrust-b}
    \end{subfigure}
    \hfill
    \begin{subfigure}{0.32\linewidth}
        \includegraphics[width=\textwidth]{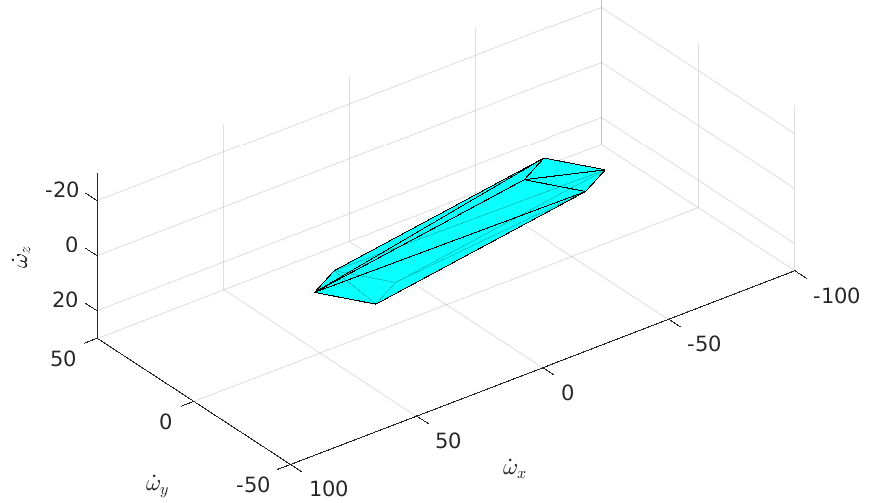}
        \caption{~}
        \label{fig:wrench:tests:moment-set-with-y-thrust-c}
    \end{subfigure}

    \caption[Moment sets for different applied forces]{Moment sets have different shapes and limits depending on fixed (i.e., desired) thrusts. A hexarotor with tilted arms applying forces to the environment. Here, the forces on axes $\XB$ and $\ZB$ are the same, but the forces on $\YB$ are (a) $0$, (b) $1$ and (c) $2$ Newtons.}
    \label{fig:wrench:tests:moment-set-with-y-thrust}
    \end{figure}

Figure~\ref{fig:wrench:tests:moment-set-with-y-thrust} illustrates how the wrench set can be estimated using Algorithm~\ref{alg:wrench:coupled:wrench-set-estimation-with-desired-components} when some of the wrench components already have assigned (desired) values. This example shows how the decoupled moment set of Figure~\ref{fig:wrench:tests:thrust-set-tilted-hex-d} shrinks in Figure~\ref{fig:wrench:tests:moment-set-with-y-thrust} when the desired thrusts are generated during the physical interaction.

\begin{figure}[!htb]
    \centering
    \begin{subfigure}{0.31\linewidth}
        \includegraphics[width=\textwidth]{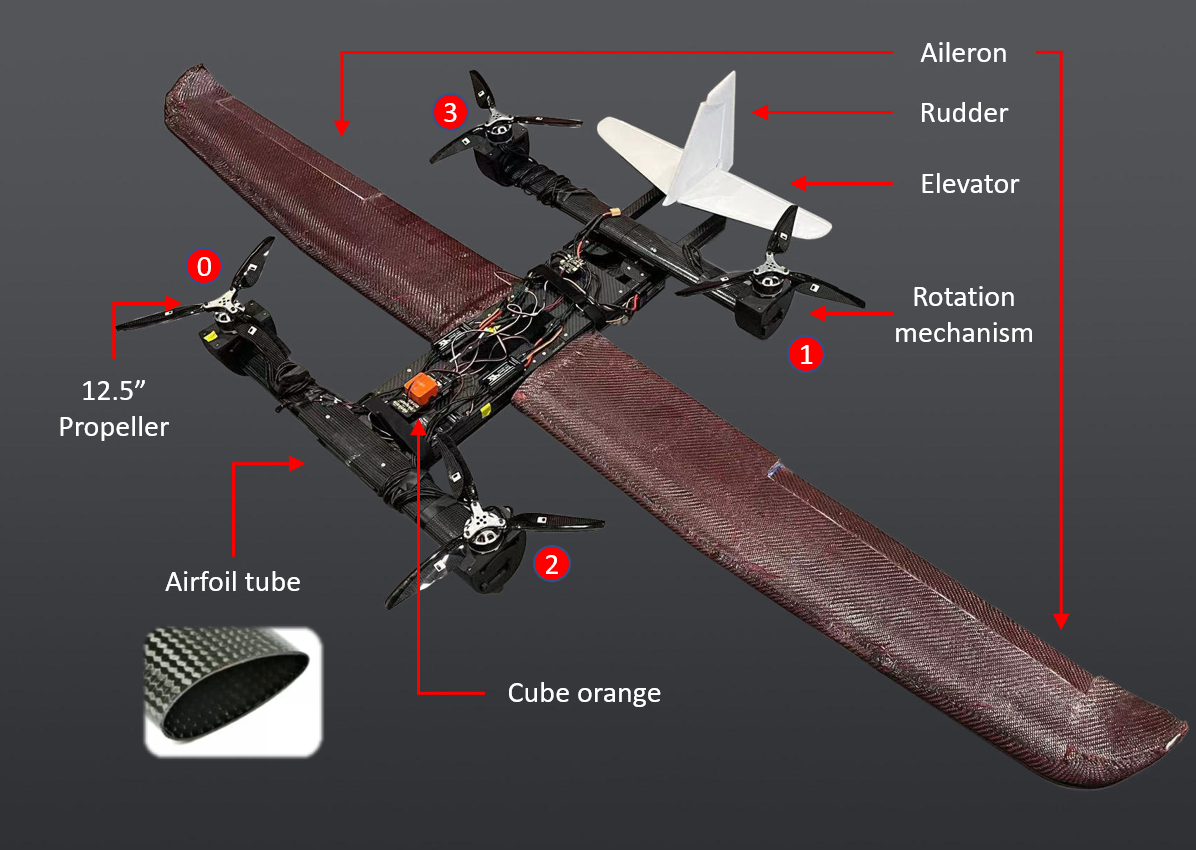}
        \label{fig:wrench:vtol-variable-pitch}
        \caption{~}
    \end{subfigure}
    \hfill
    \begin{subfigure}{0.31\linewidth}
        \includegraphics[width=\textwidth]{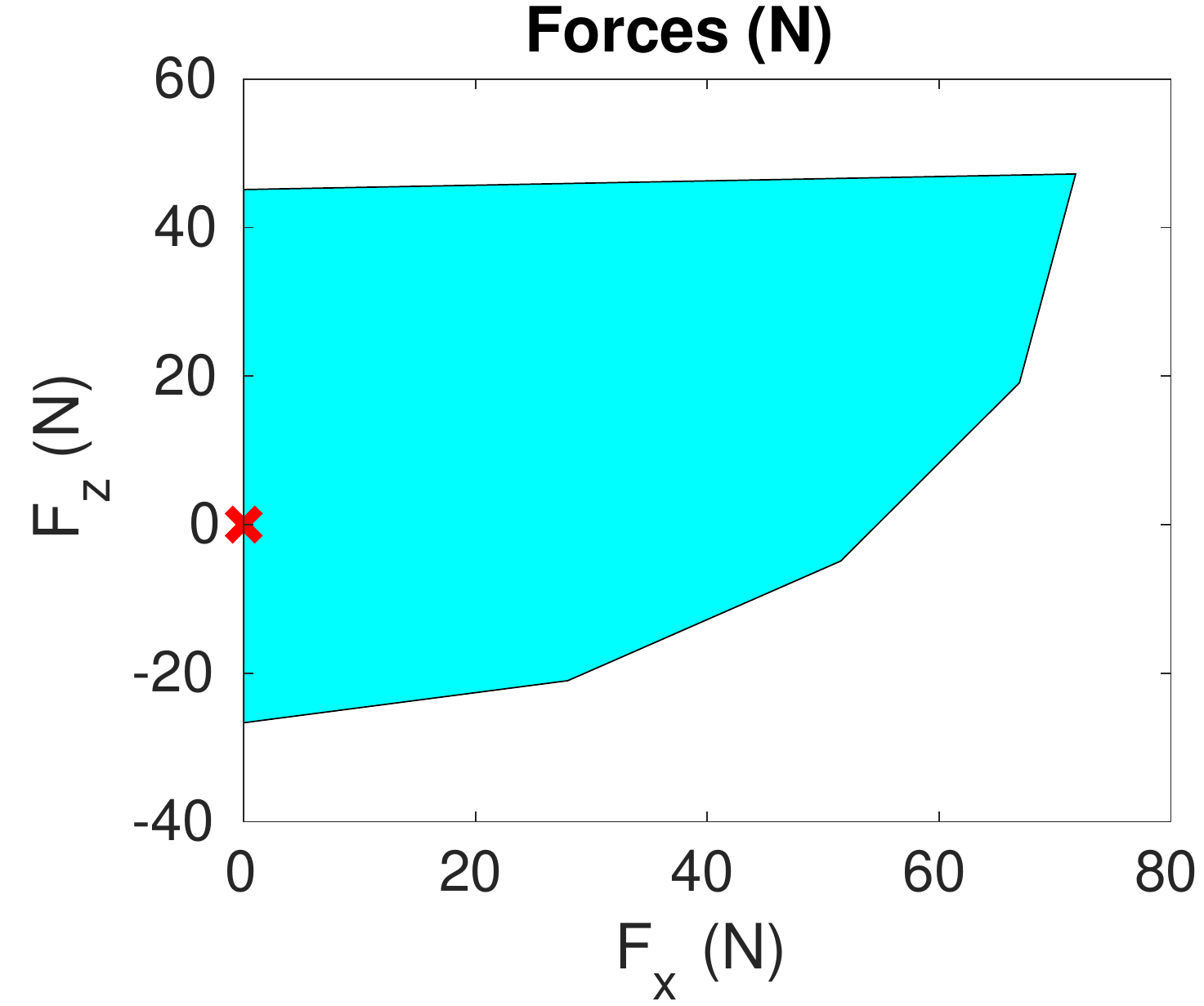}
        \caption{~}
    \end{subfigure}
    \begin{subfigure}{0.35\linewidth}
        \includegraphics[width=\textwidth]{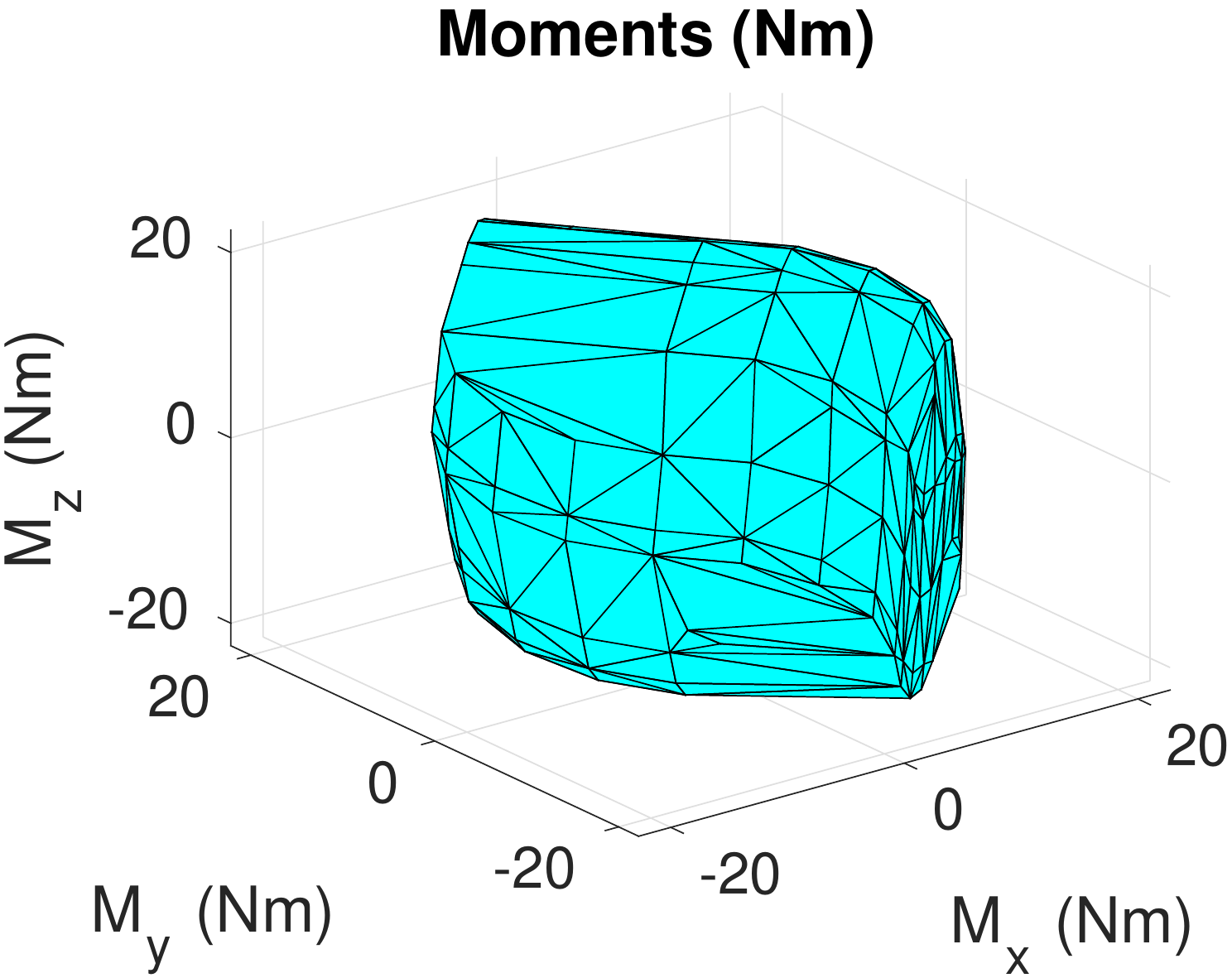}
        \caption{~}
    \end{subfigure}
    \caption[Wrench set estimation for variable-pitch rotors]{The wrench set of a VTOL with four variable-pitch rotors estimated using the sampling method discussed in Section~\ref{sec:wrench:extensions} and Algorithm~\ref{alg:wrench:coupled:wrench-set-estimation-with-desired-components}. (a) VTOL UAV. (b) Thrust set. (c) Moment set. Figure taken from our work in~\cite{Mousaei:2022:iros:vtol, Mousaei:2022:icra-workshop:vtol}.}
    \label{fig:wrench:variable-pitch-wrench-set}
\end{figure}

Figure~\ref{fig:wrench:variable-pitch-wrench-set} shows the wrench sets for a Vertical Take-Off and Landing (VTOL) aerial robot estimated for different actuator failures from our work in~\cite{Mousaei:2022:iros:vtol, Mousaei:2022:icra-workshop:vtol}. The wrench sets are calculated by sampling different rotor pitch angles as discussed in Section~\ref{sec:wrench:extensions}. For each sample of pitch angles, Algorithm~\ref{alg:wrench:coupled:wrench-set-estimation-with-desired-components} is utilized to estimate the wrench set, and all the points of the wrench sets are used at the end to estimate the new convex hull for the complete wrench set.

We measured the execution speeds for different real-time wrench set estimation methods introduced in this chapter. Table~\ref{tbl:wrench:wrench-methods-times} presents the execution times and frequencies for our hexarotor with tilted arms (see Figure~\ref{fig:wrench:tests:thrust-set-tilted-hex-a}) and the fully-actuated octorotor shown in Figure~\ref{fig:wrench:tests:thrust-set-octorotor-e}. The following scenarios were measured in the table:
\begin{itemize}[leftmargin=*]
    \item Decoupled thrust and moment set estimation using Algorithm~\ref{alg:wrench:decoupled:thrust-set-estimation}.
    \item Decoupled thrust and moment set estimation by rotation of the initial sets pre-computed using Algorithm~\ref{alg:wrench:decoupled:thrust-set-estimation}.
    \item Lateral thrust estimation using Algorithm~\ref{alg:wrench:lateral:lateral-thrust-estimation} with 500 query points.
    \item Coupled wrench set estimation, by first using Algorithm~\ref{alg:wrench:coupled:6d-wrench-set-estimation} to estimate the decoupled 6-D wrench, then using Algorithm~\ref{alg:wrench:coupled:wrench-set-estimation-with-desired-components} to estimate the coupled wrench set with three fixed wrench components.
    \item Coupled wrench set estimation, by first rotating the initial 6-D wrench (pre-computed using Algorithm~\ref{alg:wrench:coupled:6d-wrench-set-estimation}), then using Algorithm~\ref{alg:wrench:coupled:wrench-set-estimation-with-desired-components} to estimate the coupled wrench set with three fixed wrench components.
\end{itemize}

All tests were performed in our MATLAB simulator on a system with Intel® Core™ i9-10885H CPU and 64 GB DDR4 RAM.

\begin{table}[!htb]
\centering
\caption[Execution speeds for wrench set estimation methods]{Execution times and frequencies for different wrench set estimation methods for the hexarotor with tilted arms in Figure~\ref{fig:wrench:tests:thrust-set-tilted-hex-a} and the octorotor in Figure~\ref{fig:wrench:tests:thrust-set-octorotor-e}, measured in our MATLAB simulator.}
\label{tbl:wrench:wrench-methods-times}
\begin{tabular}{|l|c|c|c|c|c|c|c|c|c|}
\hline
\rowcolor[HTML]{EFEFEF} 
\multirow{2}{*}{~} & \multicolumn{2}{c|}{Hexarotor} & \multicolumn{2}{c|}{Octorotor} \\
\rowcolor[HTML]{EFEFEF} 
& Time (ms) & Hz & Time (ms) & Hz \\ \hline
Decoupled (Algorithm~\ref{alg:wrench:decoupled:thrust-set-estimation}) &  22.90 & 43.67 & 60.35 & 16.57 \\ \hline
Decoupled (Rotation) & 0.11 & 8813 & 0.45 & 2209 \\ \hline
Lateral (Algorithm~\ref{alg:wrench:lateral:lateral-thrust-estimation}) &  25.09 & 39.86 & 21.17 & 47.25 \\ \hline
Coupled (Algorithms~\ref{alg:wrench:coupled:6d-wrench-set-estimation} + \ref{alg:wrench:coupled:wrench-set-estimation-with-desired-components}) & 51.99 & 19.23 & 153.13 & 6.53\\ \hline
Coupled (Rotation + Algorithm~\ref{alg:wrench:coupled:wrench-set-estimation-with-desired-components}) & 34.09 & 29.34 & 106.60 & 9.38\\ \hline
\end{tabular}
\end{table}

The results from Table~\ref{tbl:wrench:wrench-methods-times} emphasize that the time complexity of the wrench set estimation algorithms is exponential on the number of rotors, regardless of being coupled or decoupled, and with or without leveraging the rotation (in fixed-pitch robots). 
Another observation is that rotation of the pre-computed initial force and moment sets can significantly speed up the thrust and moment set estimation, even though rotation also has exponential time complexity. The effect of rotating the pre-computed wrench space in coupled wrench set estimation is still positive; however, the resulting execution time is in the same order of magnitude as re-computing the 6-D wrench set at each iteration using Algorithm~\ref{alg:wrench:coupled:6d-wrench-set-estimation}. Note that the rotation method can only be used for a limited class of fixed-pitch multirotors. However, given the provided speed improvements, it should be utilized whenever all of the rotors in the UAV have fixed pitches.

Finally, the execution times for the lateral thrust estimation of Algorithm~\ref{alg:wrench:lateral:lateral-thrust-estimation} show that, unlike the other methods, the time complexity of the lateral thrust estimation does not have an exponential dependency on the number of rotors, which is an advantage over the other methods when only the coupled lateral thrust is required for a multirotor with many rotors. 

\section{Conclusion and Discussion} \label{sec:wrench:conclusion}

This chapter proposed several real-time algorithms to calculate the decoupled and coupled moment and thrust sets for the aerial robot in the current state and with the desired applied wrenches. The algorithms can be used for many purposes ranging from planning and control allocation to failure recovery and design optimization to improve the physical interaction of UAVs with their environment. Some of the applications of the methods are presented and discussed in Chapter~\ref{ch:applications}.

First, we introduced a real-time method to independently estimate the forces (thrusts) and moments. This geometry-based method creates the convex set of forces and moments by considering the lower and upper limits of each rotor's possible thrusts and torques. 

Then a random sampling-based real-time method was described to obtain the lateral forces considering the desired moments. This method directly computes the lateral thrust without computing the entire thrust and moment sets, which allows it to run without the exponential time complexity of the full set computation.

Then we extended the force and moment set estimation into a real-time wrench set computation method that captures the coupling between the forces and moments and can provide a much more accurate estimation of the wrenches during physical interaction. We further explored how these methods can be extended to more complex robotic configurations such as aerial robots with variable-pitch rotors (i.e., thrusters).

The experiments on different multirotor architectures and conditions illustrated how the methods described here work in practice. 

The following chapter presents some of the applications of the methods proposed in this chapter.

\chapter{Wrench-Set Applications for Fully-Actuated Multirotors} \label{ch:applications}

Dynamic manipulability analysis and the wrench set estimation for multirotors have opened the door to many possible improvements for different applications. Offline decoupled analysis was introduced around seven years ago and has since been devised to optimize UAV designs and improve the controller. With our real-time coupled wrench set estimation method proposed in Chapter~\ref{ch:wrench}, not only the existing applications can improve further, but also many new applications become possible. 

This chapter briefly describes some example applications for our proposed real-time wrench set estimation method. However, we believe that this powerful tool enables many more enhancements from hardware to different parts of the flight software to improve the flight quality and boost the abilities of UAVs to have physical interaction with their environment.

Section~\ref{sec:applications:intro} reviews the wrench-set applications presented in this chapter. Section~\ref{sec:applications:control-allocation} illustrates how the wrench set estimation method can be utilized to improve control allocation for fully-actuated multirotors. Section~\ref{sec:applications:wind} shows how the performance can be optimized when an external force is applied to the UAV. Finally, Section~\ref{sec:applications:planning} explains the ways that wrench set estimation can optimize planning for physical interaction tasks.
\section{Introduction} \label{sec:applications:intro}

The introduction of dynamic manipulability analysis in the robotics community enabled improvements in many applications, optimizing the physical interaction tasks for ground manipulators by providing information about their instantaneous and total forces and moments. As discussed in Chapter~\ref{ch:wrench}, real-time applications have been constrained to using the limited dynamic manipulability ellipsoids due to the high computational cost of dynamic manipulability polytopes, which has forced their use mainly to offline scenarios. Our work in Chapter~\ref{ch:wrench} has provided a novel solution for the computation of dynamic manipulability polytopes (i.e., wrench set) that can be used in many manipulator robots in real-time applications. However, due to the scope of our work in this thesis, this chapter illustrates example applications of this powerful tool for aerial robots.

Offline decoupled computation of the force and moment sets for aerial robots was introduced five years ago and has since been devised to optimize UAV designs and improve the controller. 

Control allocation for multirotors is a vital module that computes the actuator commands (i.e., rotor speeds) based on the desired forces and moments computed by the controller. A simple control allocation system is just a function that uses the inverted dynamics of the system to compute the actuator commands. However, the output of such a system is not protected from actuator commands being computed outside the physical range of the actuators. Therefore, when the desired forces and moments given to the control allocation module are outside the feasible range, the physical saturation of the commands by the actuator may make the UAV unstable or degrade its performance. The physical interaction tasks are even more critical, and the UAV needs to be very precise at controlling its poses and wrenches. Section~\ref{sec:applications:control-allocation} explains how our real-time wrench set estimation method can improve the control allocation module in multirotors.

The ambient wind or physical contact with the environment can affect the UAV's performance by exerting forces on the robot and changing the aerodynamic properties. When LBF fully-actuated vehicles try to keep the desired attitude, they may lose a significant portion of their lateral thrust to oppose these forces. In extreme cases, they may consume all their lateral thrust and drift, losing their tracking ability. Section~\ref{sec:applications:wind} describes our solution to estimate the optimal tilt, which reduces the consumption of the lateral thrust when constant forces are applied to the UAV.

Planning controlled physical interaction tasks for multirotors requires accounting for the required wrenches to perform the task in addition to the desired motions. With conservative assumptions for the available wrench limits at each step of the planning, the planner will not utilize the full potential of the robot, and the resulting plan may be suboptimal. Knowing the possible wrenches at each state allows optimizing the time, energy, or other parameters of the physical interaction task. In extreme examples, the conservative wrench limit assumptions may result in the planner failing to find a viable plan for finishing the task that could be otherwise performed with the knowledge about the complete wrench set. Section~\ref{sec:applications:planning} expands on this idea and describes the benefits of the real-time wrench set in planning in more detail.

Although we showcase only the mentioned applications of the wrench set, there are many other ways that the real-time wrench set estimation can benefit aerial robots. 

For example, it has been used for multirotor design optimization and finding the optimal design parameters for the desired objective~\cite{Park2016, Rashad2017, 7320711, Tadokoro2018}. 

\begin{figure}[!htb]
    \centering
    \begin{subfigure}{0.28\textwidth}
        \includegraphics[width=\textwidth]{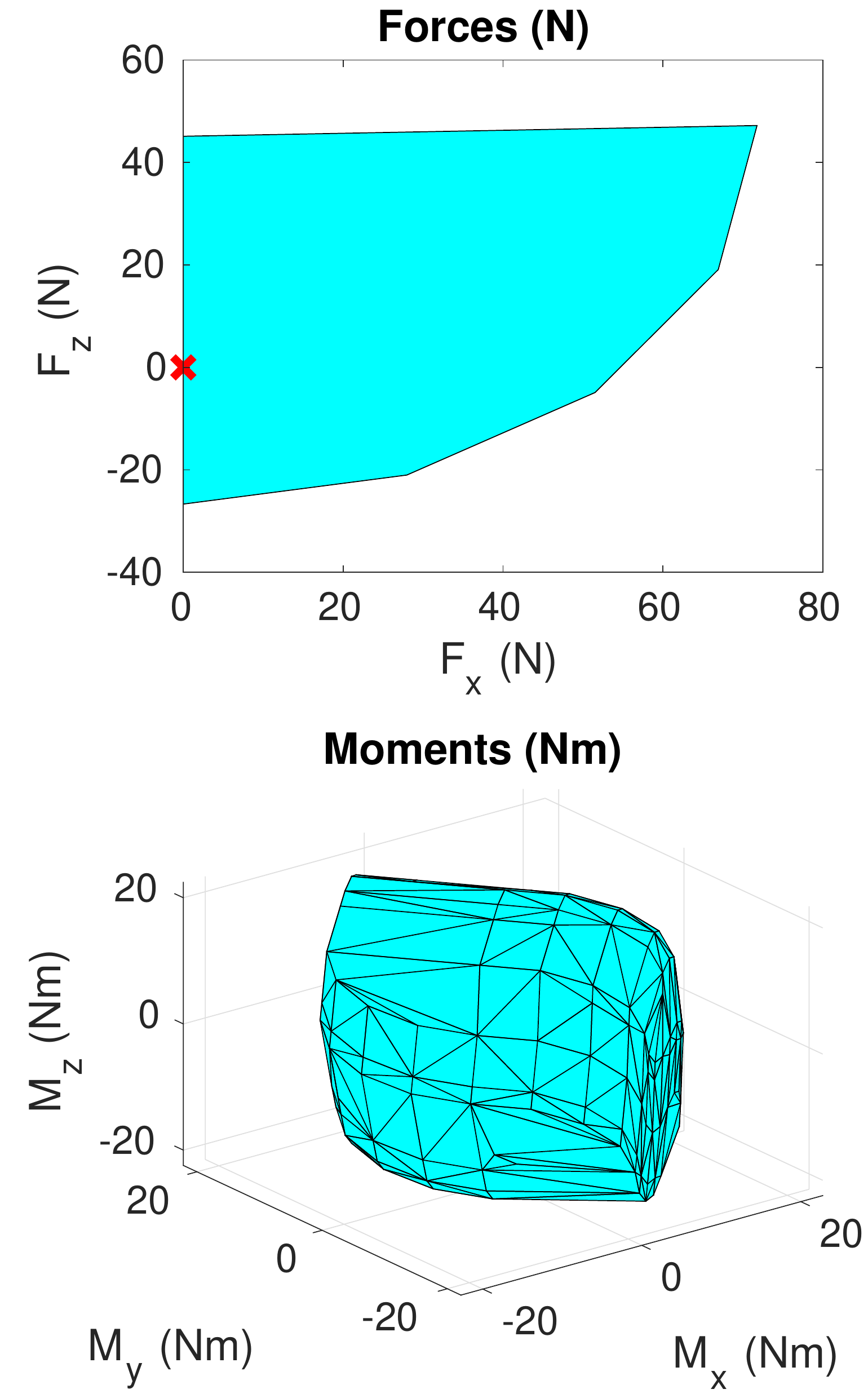}
        \caption{Healthy MC}
    \end{subfigure}
    \hfill
    \begin{subfigure}{0.28\textwidth}
        \includegraphics[width=\textwidth]{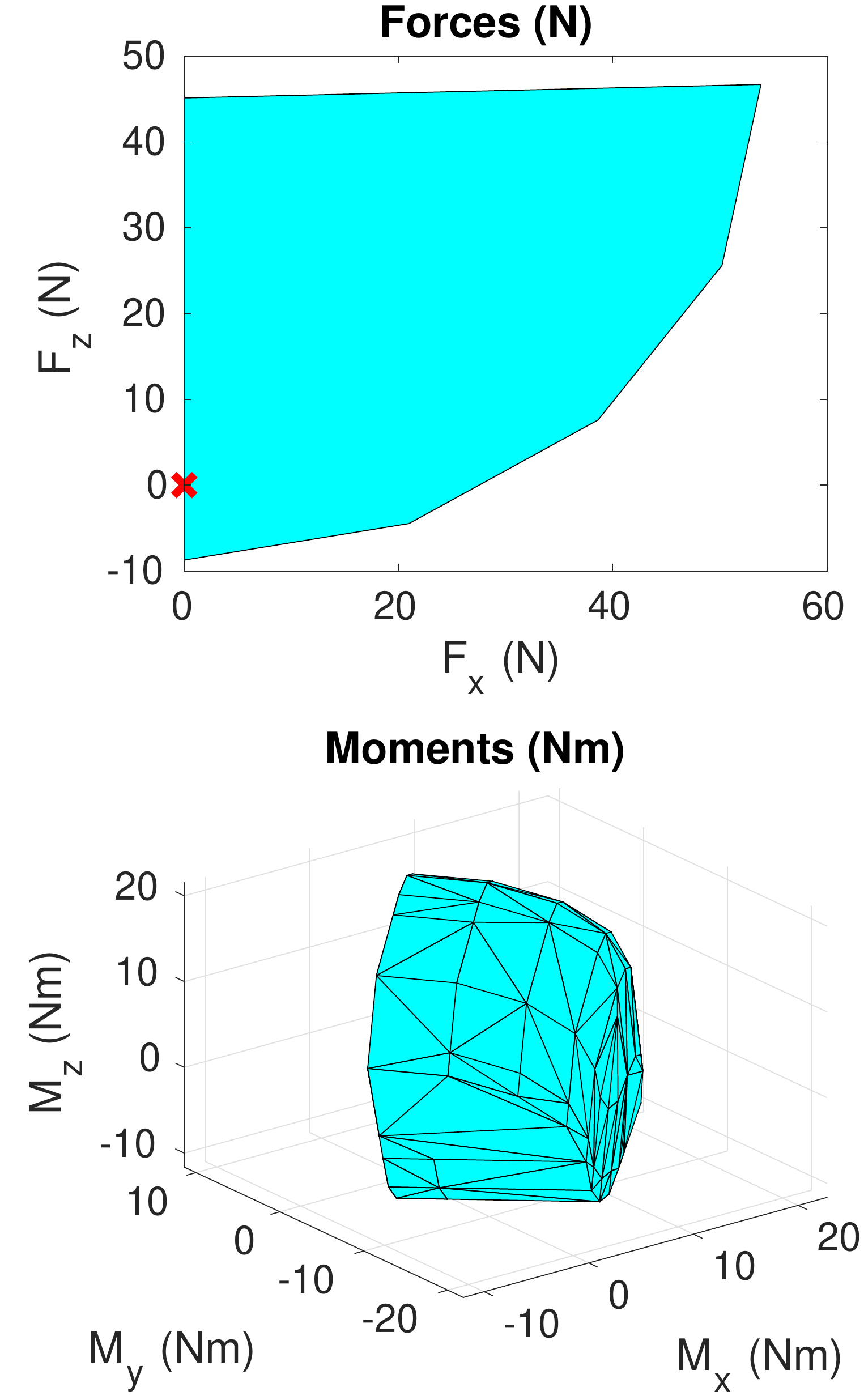}
        \caption{MC with 1 motor failed}
    \end{subfigure}
    \hfill
    \begin{subfigure}{0.28\textwidth}
        \includegraphics[width=\textwidth]{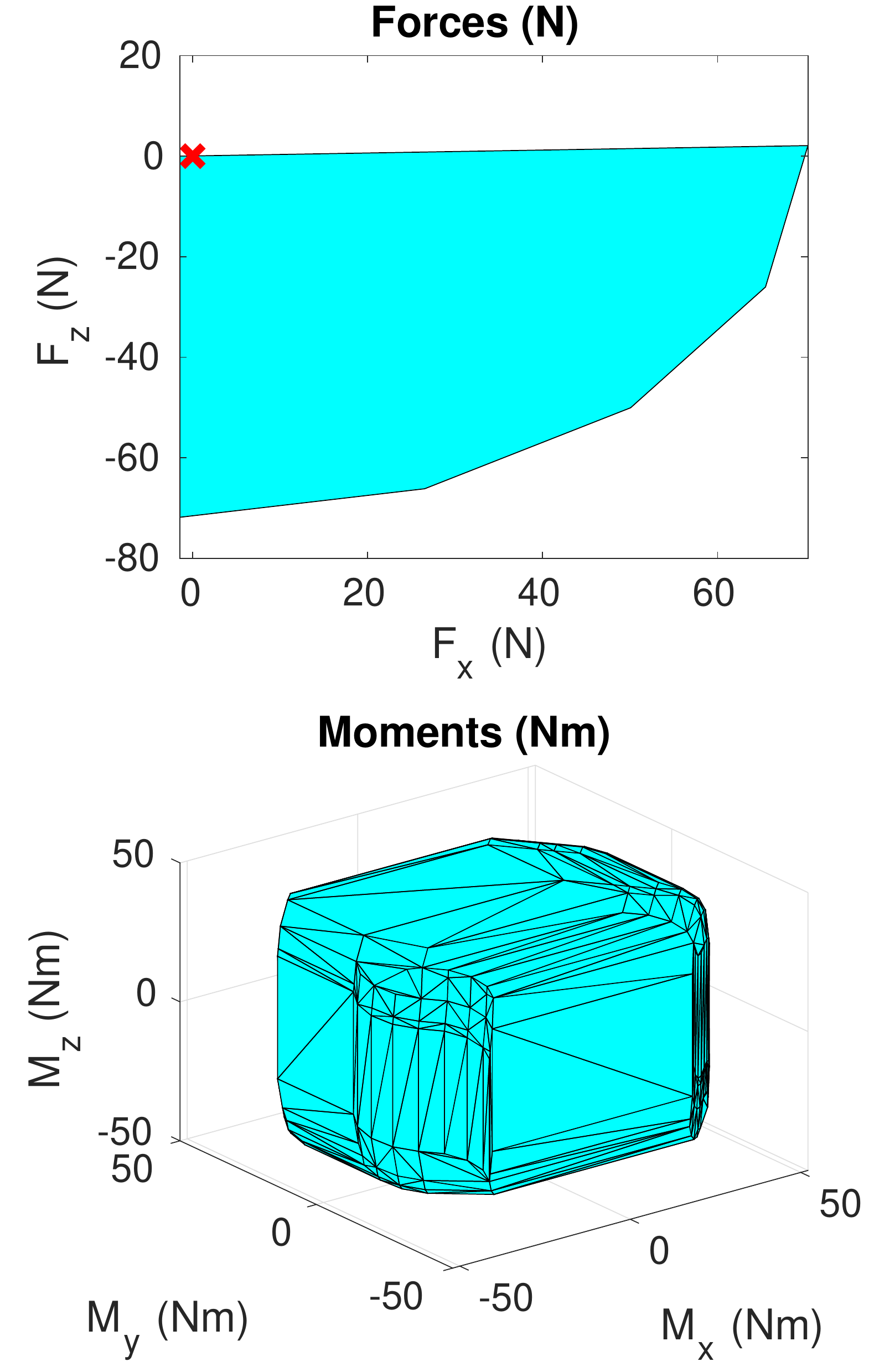}
        \caption{Healthy FW}
    \end{subfigure}

    \medskip
    
    \begin{subfigure}{0.28\textwidth}
        \includegraphics[width=\textwidth]{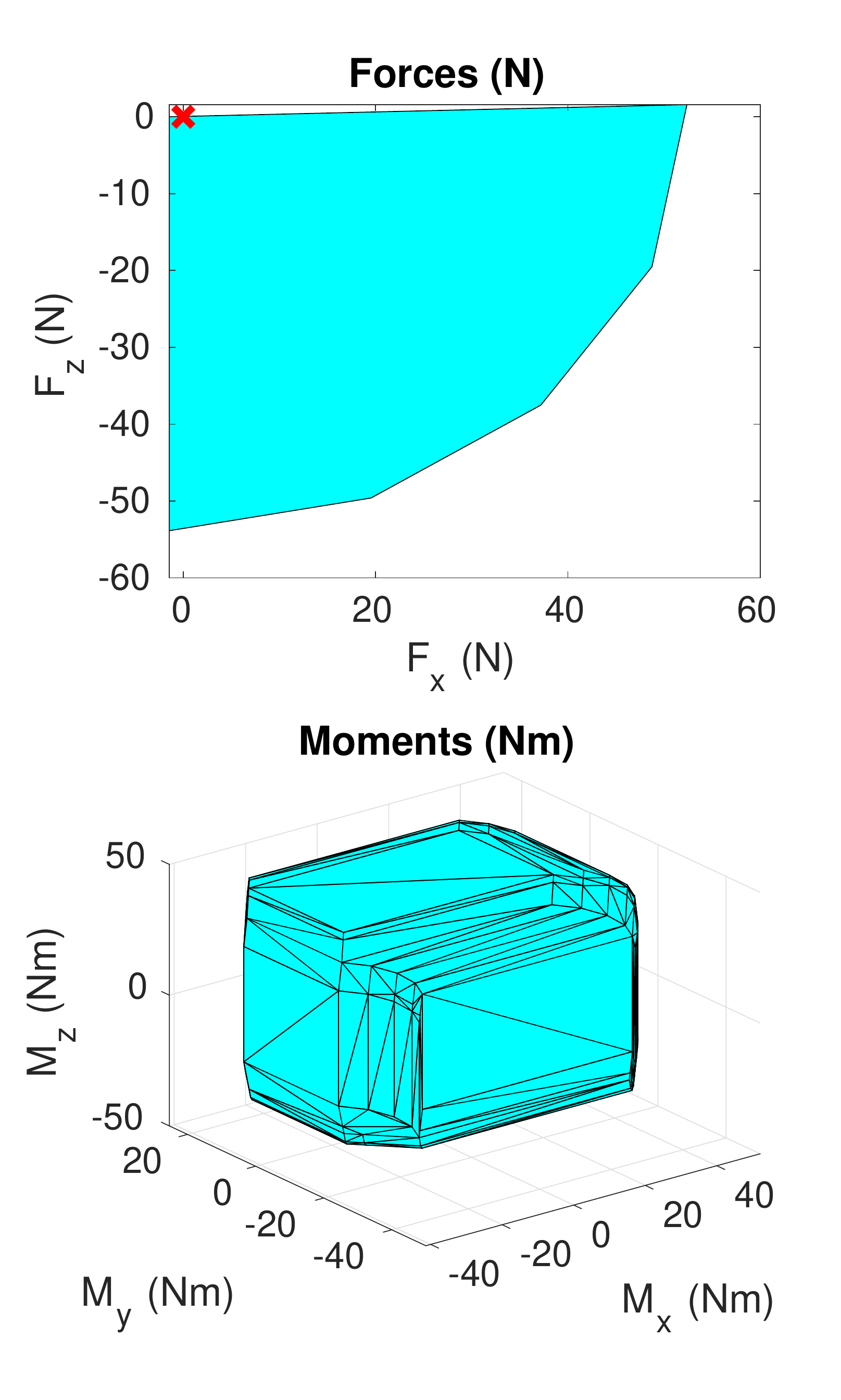}
        \caption{FW with 1 motor failed}
    \end{subfigure}
    \hfill
    \begin{subfigure}{0.28\textwidth}
        \includegraphics[width=\textwidth]{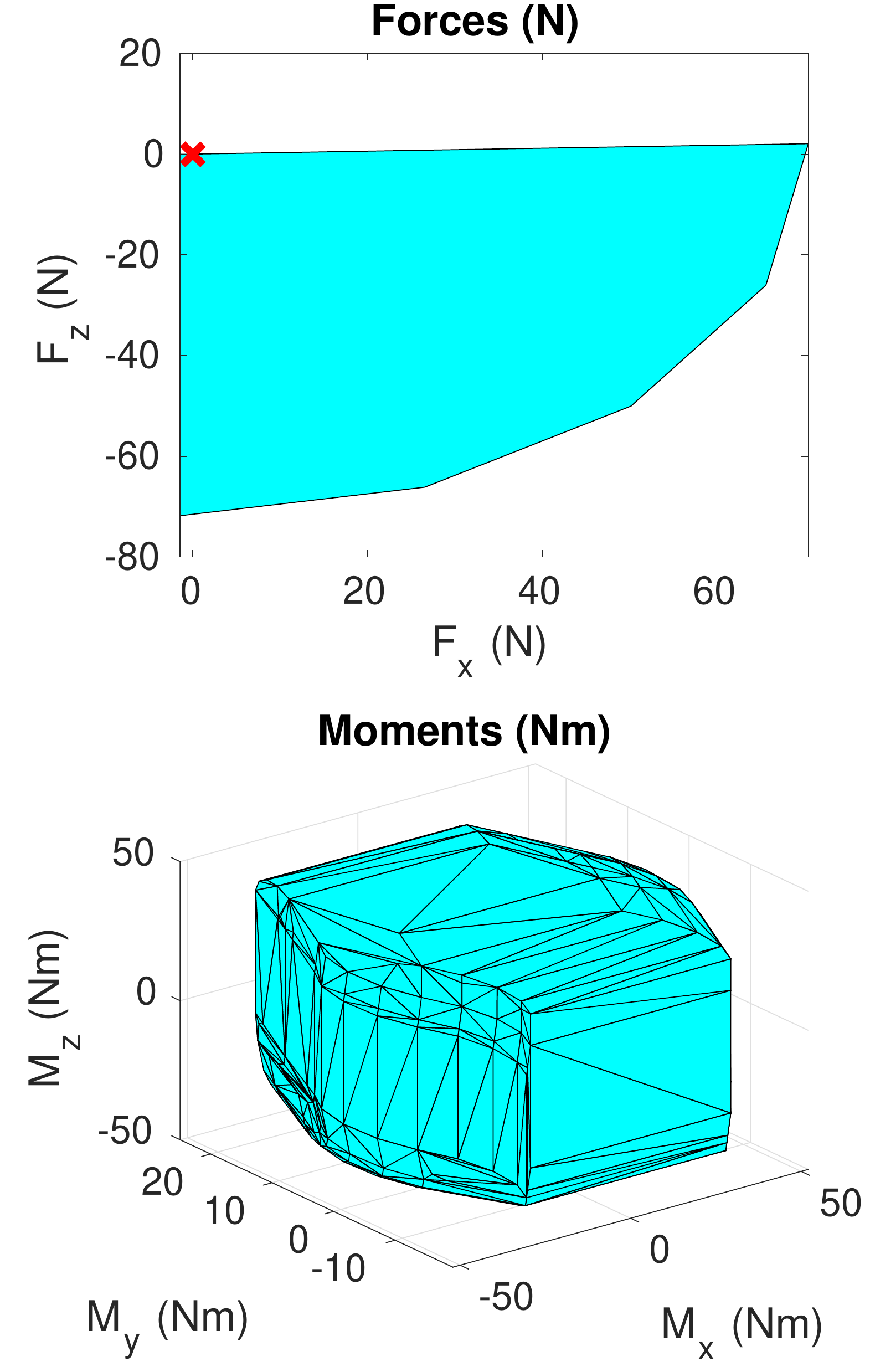}
        \caption{FW with 1 elevator failed}
    \end{subfigure}
    \hfill
    \begin{subfigure}{0.28\textwidth}
        \includegraphics[width=\textwidth]{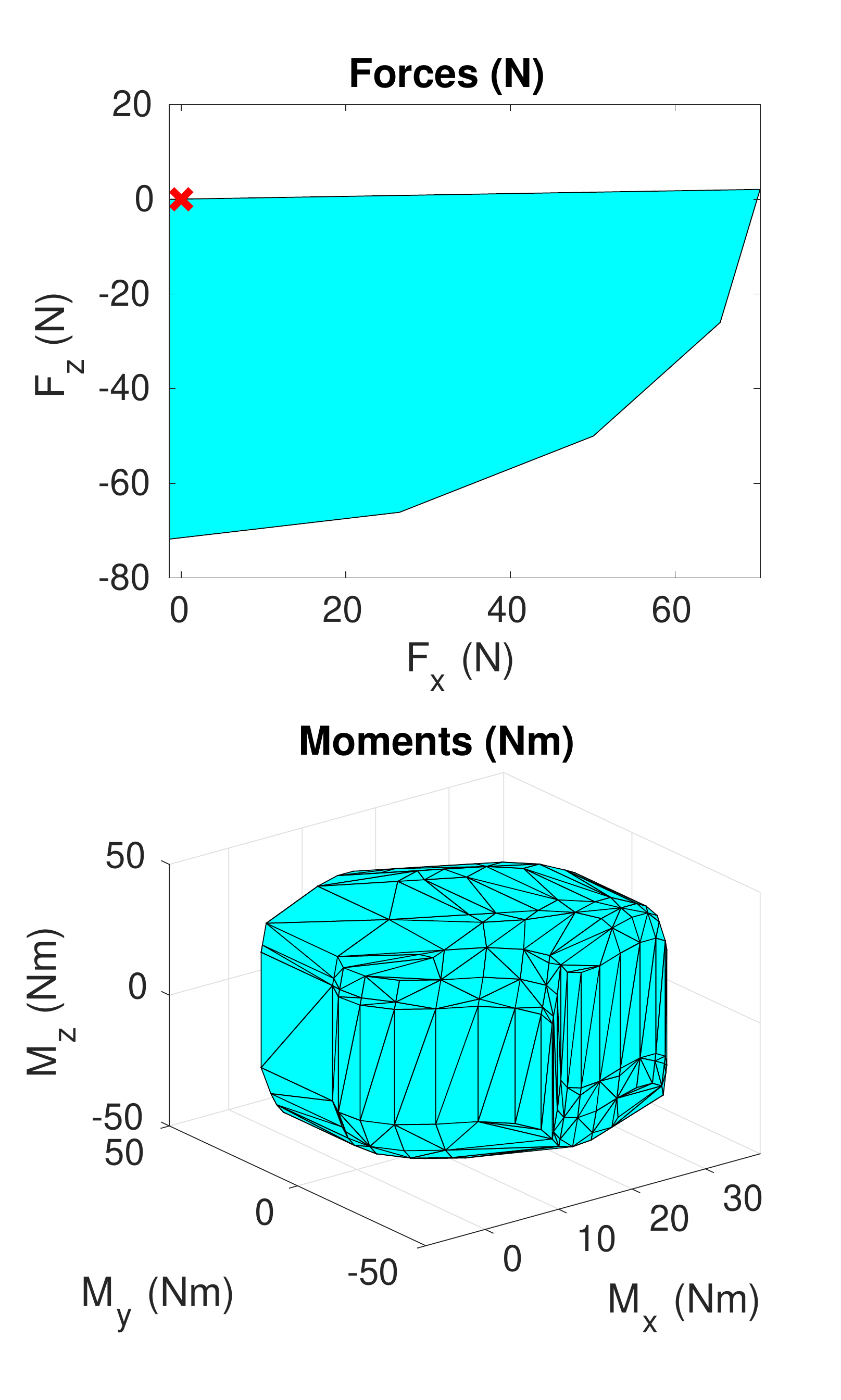}
        \caption{FW with 1 aileron failed}
    \end{subfigure}

    \caption[Wrench set estimation for variable-pitch rotors]{The wrench set of a VTOL with four variable-pitch rotors estimated for different actuator failure conditions using the sampling method discussed in Section~\ref{sec:wrench:extensions} and Algorithm~\ref{alg:wrench:coupled:wrench-set-estimation-with-desired-components}. Figure taken from our work in~\cite{Mousaei:2022:iros:vtol, Mousaei:2022:icra-workshop:vtol}.}
        
    \label{fig:applications:vtol-wrench-set}
\end{figure}

Mochida et al.~\cite{MOCHIDA20209334, 9393789} have used the idea of thrust-set estimation for optimization of the multirotor design to make it robust to failure, as well as for hoverability and failure analysis. We have also recently used it in our work on VTOL failure recovery to understand how the wrench set changes when a failure happens~\cite{Mousaei:2022:iros:vtol, Mousaei:2022:icra-workshop:vtol}. Figure~\ref{fig:applications:vtol-wrench-set} shows the computed thrust and moment sets when different actuator failures happen. 

\section{Improving Control Allocation Performance} \label{sec:applications:control-allocation}

\begin{figure}[!htb]
        \centering
        \includegraphics[width=0.6\linewidth]{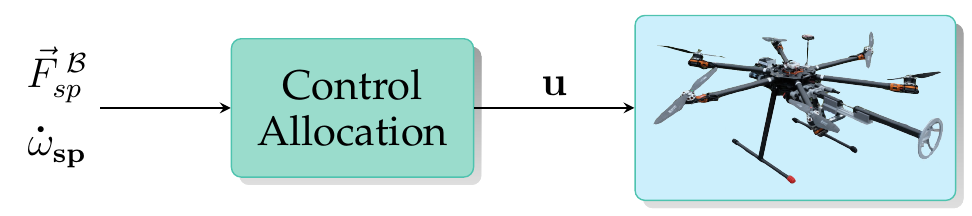}
        
        \caption[Control allocation module with inputs and outputs]{A high-level illustration of a typical control allocation module with its inputs and outputs.}
        
        \label{fig:applications:control-allocation:typical-module}
    \end{figure}

In Chapters~\ref{ch:background} and~\ref{ch:control} we have shown the conventional control design for a fully-actuated controller (see Figure~\ref{fig:control:typical-controller-architecture}). This design has a Control Allocation module tasked with converting the desired forces and moments (or linear and angular accelerations) computed by the previous parts of the controller and converting them to the required rotor inputs $\mat{u}$ that can result in the desired forces and moments (or accelerations). Figure~\ref{fig:applications:control-allocation:typical-module} shows this module with its inputs and outputs.

Section~\ref{sec:background:control:control-allocation} derived the most common control allocation method for fully-actuated multirotors which uses \textit{exact feedback linearization and decoupling method} (a.k.a. \textit{nonlinear dynamic inversion})~\cite{Crowther2011, Rajappa2015, Mehmood2017, Brescianini2016}. The major drawback of this method is that the resulting output (i.e., the elements of $\mat{u}$) may exceed the minimum or maximum physical limits of what the rotors can achieve. If not properly treated, the over- or under-saturated rotors can result in a different total wrench than desired, causing instability and, in the worst case resulting in a crash. The problem exacerbates when the demands for the generated forces and moments are higher and more precise control over wrenches is required, such as during the robot's interaction with the environment.

Researchers have proposed three approaches to alleviate the issue: 

\begin{itemize}
    \item Limiting the computed output $\mat{u}$ to within the limits of the motor inputs after $\mat{u}$ is computed by the control allocation module.
    \item Utilizing an optimization-based allocation method to keep the computed motor commands within the desired limits.
    \item Limiting the desired forces and moments to ensure that the computed output $\mat{u}$ is within the physical limits of the multirotor motors and what rotors can generate.
\end{itemize}

A possible solution with the first approach (limiting $\mat{u}$ after the computation) is to saturate the computed values in order to send feasible commands to the motors. However, this approach has similar results to not doing anything for the actual UAV because sending an out-of-bounds command to motors will naturally result in saturation regardless. Another solution is to change the computed commands prioritizing the stability of the robot, usually by trying to reduce the rotor commands in a way that the moments are kept intact, and the thrusts are reduced~\cite{Meier2015}. However, this approach is strongly architecture-dependent and hard to generalize.

In the second approach, i.e., devising an optimization-based allocation method, the motor limits are considered as constraints, while the dynamic inversion equation is considered a soft constraint in the cost function~\cite{Dyer2019, 7524963}. In addition to the computational cost of the nonlinear optimization, the approach makes it very difficult to tune the resulting output wrenches when the input desired wrench is infeasible. Therefore, these methods are not suitable for physical interaction applications where keeping the direction of the wrenches and prioritizing the moments over the thrusts to keep the orientation may be desirable.

For the third approach, i.e., limiting wrenches before the computation of outputs in the control allocation module, most existing solutions only consider the thrust limits, and all of them consider the set of possible thrusts and moments as static~\cite{6608749, Mehmood2017, INVERNIZZI201711565}. The primary issue with considering static bounds for wrenches is that the full potential of the robot is not utilized if the limits are set too small. On the other hand, even if the smallest limits are chosen based on the desired operation of the robot, any unpredicted situation causing an even smaller limit for the wrenches may cause instability due to the control allocation exceeding the limits and saturating the motors. 

One of the most advanced methods in this class is proposed by Franchi et al.~\cite{Franchi2018}, which limits the lateral thrusts by estimating the static thrust set at hovering, ignoring the orthogonal thrust and all moment limits, and ignoring the effect of the UAV state and the desired wrenches on the thrust set. Therefore, their solution works well for near-hovering conditions but does not result in correct limits for other scenarios.

A recent improvement is proposed by Bezzera and Santos~\cite{BEZERRA2021}, which uses a method similar to our proposed decoupled algorithm for thrust set estimation (Algorithm~\ref{alg:wrench:decoupled:thrust-set-estimation}). Their work is limited to thrusts and ignores the effects of the UAV orientation on the wrenches. However, it demonstrates the improvements that can be made using wrench set estimation.

\begin{figure}[!htb]
        \centering
        \includegraphics[width=0.8\linewidth]{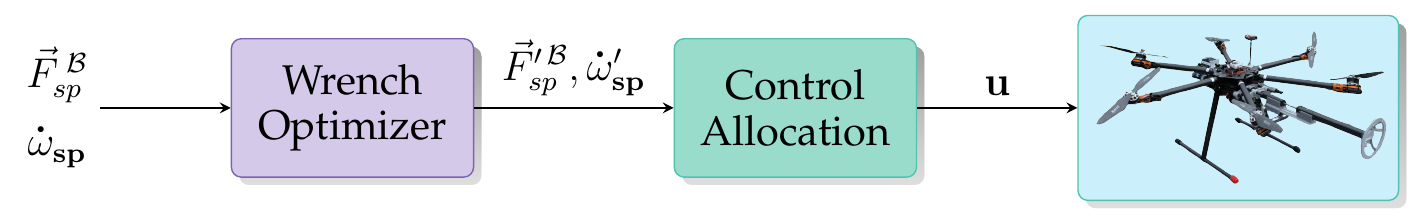}
        
        \caption[Improved control allocation with Wrench Optimizer]{A high-level illustration of the control allocation module with its inputs and outputs.}
        
        \label{fig:applications:control-allocation:improved-module}
    \end{figure}

Using the coupled wrench set estimation method of Section~\ref{sec:wrench:coupled} (i.e., Algorithm~\ref{alg:wrench:coupled:wrench-set-estimation-with-desired-components}), the control allocation can be improved further. Figure~\ref{fig:applications:control-allocation:improved-module} illustrates our proposed control allocation method.

The Wrench Optimizer works as follows:

\begin{enumerate}
    \item Estimates the wrench set using Algorithm~\ref{alg:wrench:coupled:6d-wrench-set-estimation},
    \item Checks if the input thrust and moment setpoints are feasible (i.e., fall within the wrench set):
    \begin{enumerate}
        \item If feasible, it outputs the setpoints without a change to the conventional control allocation module. 
        \item Otherwise, bring the thrust and moment setpoints to within the wrench set using the priorities set by the task and the developer.
    \end{enumerate}
\end{enumerate}

In this method, the developer has flexibility over what to prioritize when the thrust or moment setpoints are outside the feasible wrench set. Depending on the task at hand, the priority might be keeping the attitude and altitude or the force applied to the environment. In either of these cases, due to the geometric nature of the problem, the final solution can be computed by finding the intersection of a 6-D wrench vector with the boundary of the 6-D convex set, which simplifies into the intersection of a line with a hyperplane and checking if the intersection point is in the polygonal face of the set.

\section{Flight Optimization in the Presence of Constant Force} \label{sec:applications:wind}

External forces acting on the multirotor, such as forces from the physical interaction or the ambient wind, can affect the UAV's performance by exerting forces on the robot and changing the aerodynamic properties. Fully-actuated vehicles may lose a portion of their lateral thrust to oppose the external force when trying to keep the desired attitude. In LBF vehicles, the consumed lateral thrust can be a large portion of their maximum possible lateral thrust. In extreme cases, the external force may overpower the UAV, in which case it starts drifting (e.g., with the wind) and can lose its tracking ability in the direction of the external force.

Many hardware and software solutions for wind estimation have been introduced to mitigate the mentioned effects (see~\cite{Abichandani2020, Schopferer:2018:icuas:planning, Hullmann2018}). Hardware devices, such as flow sensors (pitot tubes) and ultrasonic anemometers, are generally more accurate than the software methods. However, they add to the total payload weight, increase hardware complexity, and reduce the UAV's balance and stability due to their placement requirements. Moreover, their use cannot be extended to other sources of constant force applied to the robot, such as physical contact. On the other hand, purely software approaches, such as the tilt angle and rotor speeds methods, are simple to implement and work with other sources of applied force but have lower accuracy and require extensive calibration and specific working conditions. 

Our goal, however, is to increase the performance of the LBF vehicles in the presence of the constant force (e.g., ambient wind), and directly measuring these forces may be unnecessary for this purpose. 

We define performance as the ability of the UAV to accelerate in any direction. Let us call the maximum acceleration that the UAV can achieve in all directions in its current state as \textit{omni-directional acceleration} $a_o$. Maximizing the performance of the UAV in the wind would mean maximizing the omni-directional acceleration. 

The variables controllable by the UAV controller that can affect the solution are UAV's attitude and generated thrust. Therefore, we can define the problem as finding the optimal force and attitude setpoints that maximize the omni-directional acceleration:

    \begin{equation} \label{eq:applications:wind:problem-definition}
        \begin{split}
            &(\Fspt, \mat{\Phi\spt}) = \argmax_{(\vec{F},\ \mat{\Phi})} a_o \\
            &\text{subject to}\ \vec{F}\, \in \, \SetT
        \end{split}
    \end{equation}

In our solution, we make some assumptions:

\begin{enumerate}
    \item All UAV rotors have fixed angles and positions in the body-fixed frame.
    
    \item When hovering with no external forces other than the gravity acting on the UAV, the optimal tilt is zero (i.e., the $\ZB$ and $\ZI$ axes should align). In a more general case, when the assumption is not valid (e.g., when the horizon is not leveled), we can consider the optimal (non-zero) tilt for hovering as the baseline and add the offset after the optimal tilt is calculated with respect to this baseline.
    
    \item If the force in the $ZI$ direction increases (e.g., we add the robot's weight), the optimal tilt remains zero. The assumption is valid for many UAV architectures (including the fixed-pitch hexarotor in our experiments) as long as the UAV remains an LBF robot (when the remaining normal thrust is much higher than the available lateral thrust).

\end{enumerate}

When the aerial robot is interacting with the physical world or is flying in the wind, a force $\Fwind$ is impacting its flight. Combined with the gravity force $\Fgrav$ (i.e., the weight), this force applies a total external force to the UAV. Ignoring minor aerodynamic effects, the total external force $\Fext$ is:

    \begin{equation} \label{eq:applications:wind:external-force}
        \Fext = \Fwind + \Fgrav
    \end{equation}

\begin{figure}[t]
        \centering
    
        \begin{subfigure}{0.25\textwidth}
            \includegraphics[width=\textwidth]{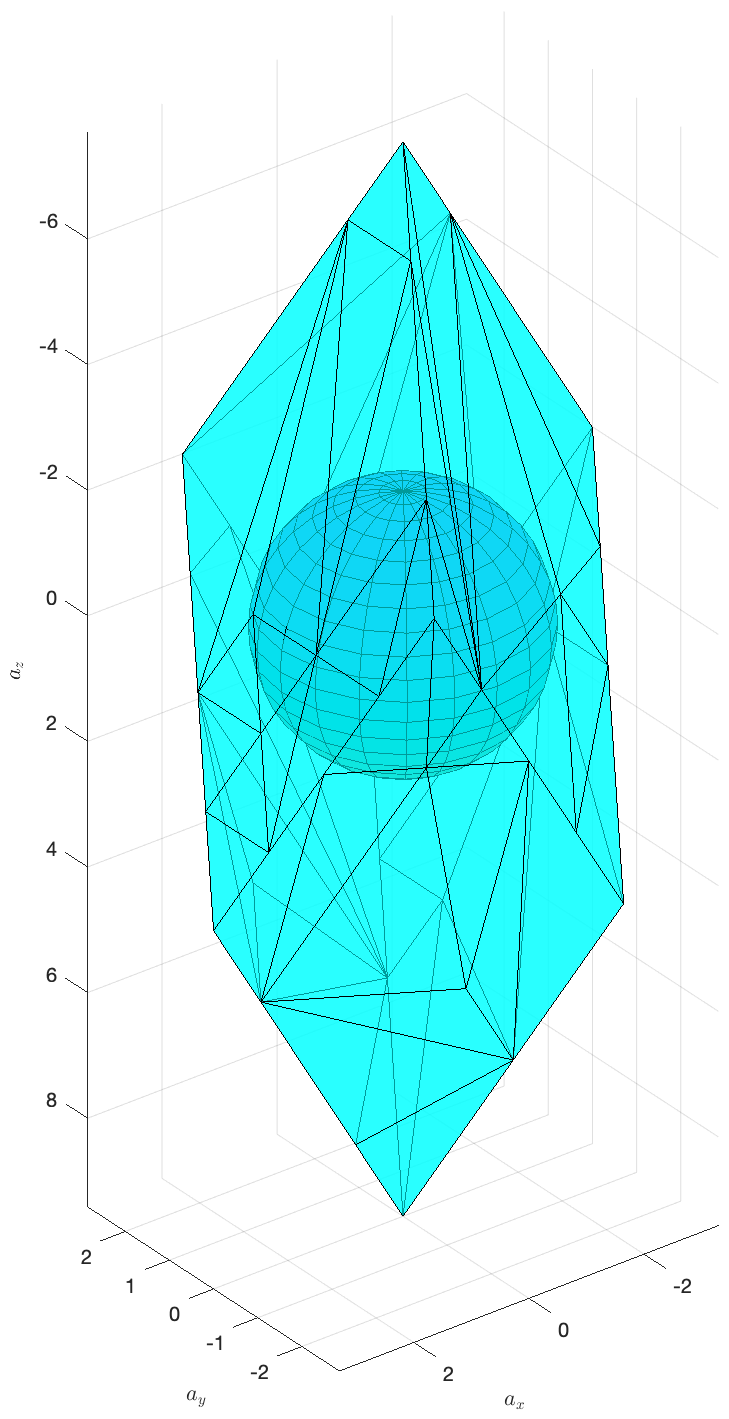}
            \caption{~}
            \label{fig:applications:wind:omni-directional-acceleration-sphere-a-convex-hull}
        \end{subfigure}
        \hfill
        \begin{subfigure}{0.61\textwidth}
            \includegraphics[width=\textwidth]{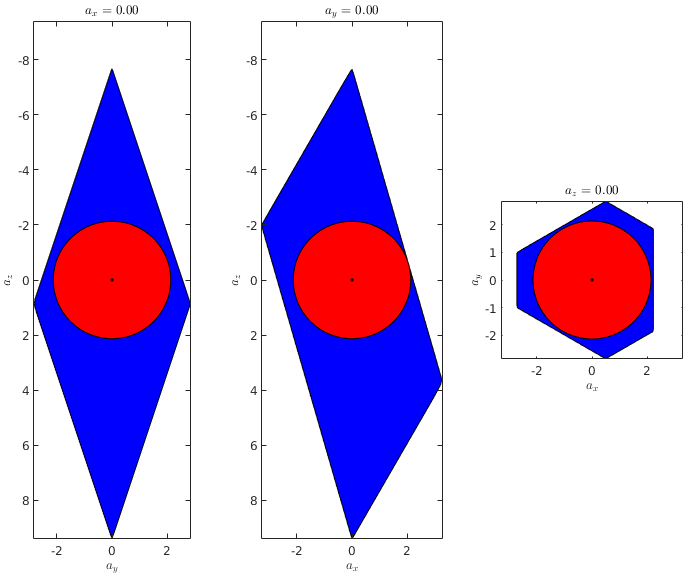}
            \caption{~}
            \label{fig:applications:wind:omni-directional-acceleration-sphere-b-cross-sections}
        \end{subfigure}
        
        \caption[Omni-directional acceleration sphere inside the acceleration set]{The illustration of the largest sphere centered around $\matrice{0&0&0}\T$ inscribed in the acceleration set of Figure~\ref{fig:wrench:tests:thrust-set-tilted-hex-c}. The sphere's radius is the largest acceleration that the UAV can generate in any desired direction at the current state called \textit{omni-directional acceleration}. (a) The full acceleration set. (b) The cross-sections passing through the center along the $\XI$, $\YI$, $\ZI$ axes.}
        
        \label{fig:applications:wind:omni-directional-acceleration-sphere}
    \end{figure}

Let us imagine the maximum inscribed sphere inside the current acceleration set (the set of all the possible accelerations at the current state) centered around our desired acceleration. Figure~\ref{fig:applications:wind:omni-directional-acceleration-sphere} illustrates this sphere on the acceleration set of our fixed-pitch hexarotor.

Without the loss of generality, we assume that the desired acceleration is zero, meaning that the UAV either intends to hover or fly at a constant speed. Thus, the center of the inscribed sphere should be $\matrice{0&0&0}\T$. Extending the method to non-zero acceleration is straightforward and will not be discussed.

An LBF multirotor has a much smaller lateral acceleration compared to the acceleration normal to its body. Therefore, the radius of the inscribed sphere will ultimately be limited due to the lateral acceleration limit. This radius is the maximum acceleration that the robot can achieve in any desired direction at its current state and is the same as the omni-directional acceleration $a_o$ defined above.

To extend the current analysis to devise the thrust set estimation methods described in Chapter~\ref{ch:wrench}, note that the shape of the acceleration set is just a linear scaling of the thrust set. However, to achieve zero acceleration (to stay in equilibrium), the UAV needs to generate the total thrust with the same magnitude but in the opposite direction of the total external force (i.e., $-\Fext$). Therefore, to find the omni-directional \textit{thrust} $F_o$, we need to find the maximum inscribed sphere inside the thrust set centered around $-\Fext$. 

~

\begin{prop}
With the assumptions of Section~\ref{sec:applications:intro}, the full-tilt attitude strategy of Section~\ref{sec:control:attitude:full-tilt}, combined with an optimal method for producing the thrust setpoint, will converge to the optimal tilt and thrust setpoints that maximize the omni-directional acceleration in the presence of the external force $\Fext$.
\end{prop}

\begin{proof}
If the devised position control method is optimal (the case for almost all popular methods), the desired thrust $\Fdes$ calculated by the position controller will eventually converge to $-\Fext$ if the external force is constant. Therefore, considering that the gravity changes are negligible, $\Fdes$ will converge to $-\Fext$ when the airspeed is constant.

From Section~\ref{sec:wrench:decoupled}, we know that for the robots with the fixed rotor angles, the tilt of the robot will only rotate the thrust set around the external force point without any change to the shape. On the other hand, the third assumption of the problem (see Section~\ref{sec:applications:intro}) means that the robot will have the largest omni-directional acceleration when the desired acceleration point falls on the $\ZB$ line. This also means that the attitude resulting in zero consumed lateral thrust is optimal.

The zero consumed lateral thrust of the UAV happens when the total generated thrust is normal to the body. Thus, during the equilibrium (when $\Fdes = -\Fext$), the full-tilt attitude strategy (Section~\ref{sec:control:attitude:full-tilt}) results in the optimal attitude setpoint. 
\end{proof}

In reality, even when the UAV is hovering, there are changes to the external force (e.g., due to the air pressure changes and gusts). Therefore, the optimal tilt of the UAV will continuously be changing. However, in practice, assuming that the average wind or external force is constant, the optimal tilt can be averaged over some time and then locked using the fixed-tilt strategy (Section~\ref{sec:control:attitude:fixed-tilt}). This way, the UAV will keep the tilt required to oppose the external force (e.g., wind) effectively and will have the largest thrust left to accelerate in any direction and to reject the unpredicted disturbances.

\begin{figure}[!htb]
        \centering
    
        \begin{subfigure}{0.30\textwidth}
            \includegraphics[width=\textwidth]{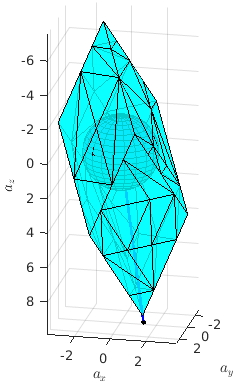}
            \caption{~}
            \label{fig:applications:wind:omni-acceleration-tilt-a-thrust-set}
        \end{subfigure}
        \hfill
        \begin{subfigure}{0.65\textwidth}
            \includegraphics[width=\textwidth]{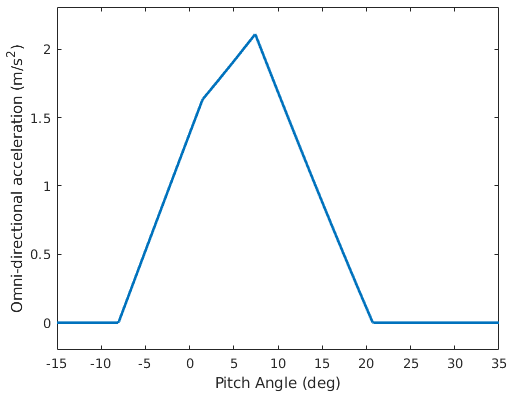}
            \caption{~}
            \label{fig:applications:wind:omni-acceleration-tilt-b-plot}
        \end{subfigure}
        
        \caption[Optimal tilt in the presence of external force]{The relation of the omni-directional acceleration with the tilt in the presence of external forces. A 10~$\unit{N}$ force is applied from south to north by the wind to the fixed-pitch hexarotor used in our project. (a) The optimal tilt is when the center of the omni-directional acceleration sphere (i.e., the desired acceleration) is placed on the UAV's $\ZB$ axis. (b) The omni-directional acceleration for different pitch angles.}
        
        \label{fig:applications:wind:omni-acceleration-tilt}
    \end{figure}

We performed tests on several architectures to experimentally check the validity of the optimal tilt obtained from the full-tilt attitude strategy. Figure~\ref{fig:applications:wind:omni-acceleration-tilt} illustrates such test performed on the architecture of our fixed-pitch hexarotors (see Figure~\ref{fig:wrench:tests:thrust-set-tilted-hex-a}) that shows the relationship between the tilt and the omni-directional acceleration.

\begin{figure}[!htb]
        \centering
    
        \includegraphics[width=0.8\textwidth]{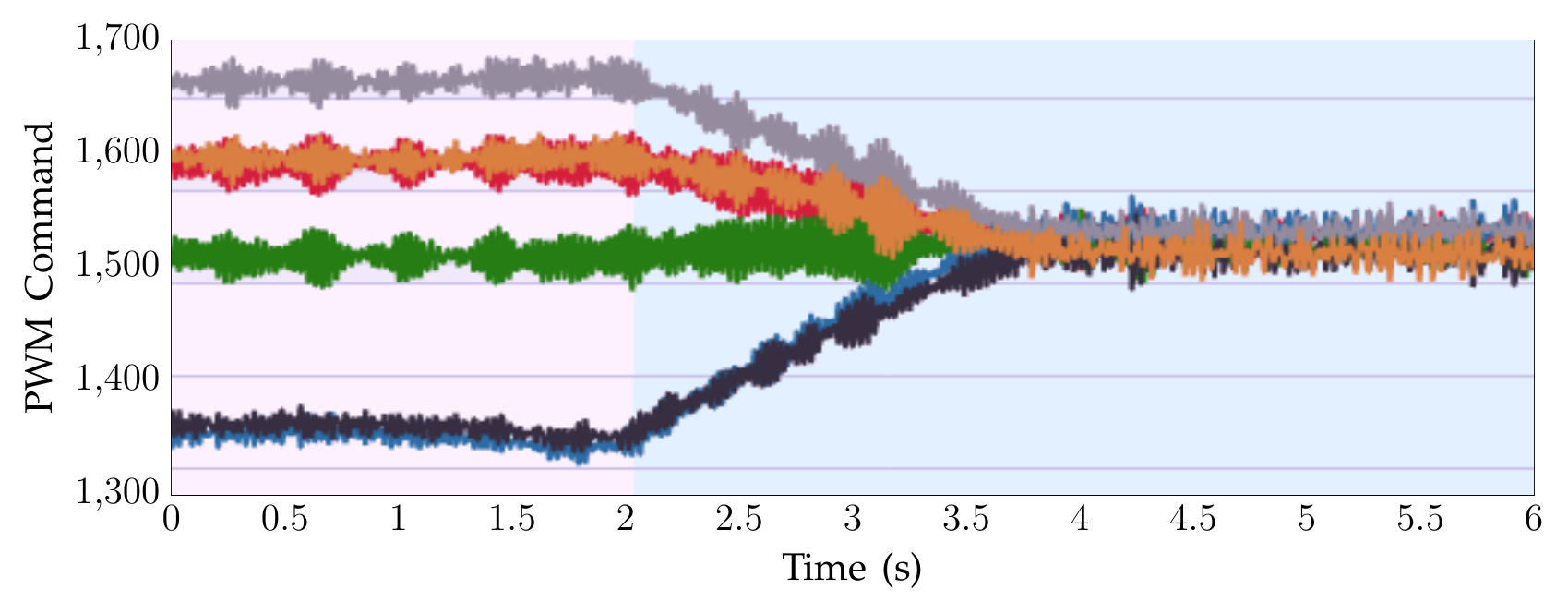}
        
        \caption[Motor inputs during optimal tilt estimation]{The motor inputs for the fixed-pitch symmetric hexarotor of Figure~\ref{fig:control:tests:tilthex-a} converge over time when the symmetric hexarotor is in optimal tilt to oppose the wind force. The plot shows the PWM commands for the motors vs. time during hovering in an almost constant wind. The pink background shows the flight with the zero-tilt attitude strategy, while the blue background shows when the wind estimation optimizer runs during the full-tilt attitude strategy.}
        
        \label{fig:applications:wind:wind-tilthex-plot}
    \end{figure}

Another interesting observation is that when the UAV structure is symmetric, the motor inputs tend to converge to the same value at the optimal tilt when the accelerations are zero. Figure~\ref{fig:applications:wind:wind-tilthex-plot} illustrates how the motor inputs converge once the tilt estimation starts for our fixed-pitch hexarotor of Figure~\ref{fig:control:tests:tilthex-a}.

\section{Planning Physical Interaction Tasks} \label{sec:applications:planning}

The planner system planning for a task involving physical interaction of the robot with its environment requires to respect not only the environment constraints (such as obstacles) and robot kinematic and dynamic constraints (such as velocity limits and maximum turn rates) but also the constraints imposed due to the physical interaction. Such limits include the contact constraints as described in Section~\ref{sec:control:hpfc}, as well as the limits on the possible wrenches than can be applied during that interaction.

The study of the feasibility of applying desired motions and wrenches for a specific task is called \textit{feasibility analysis} and was first introduced by Chiu in 1987~\cite{1087795,027836498800700502}. Feasibility and manipulability analysis have been used for ground manipulators, humanoid robots, and space manipulators in conjunction with task planning to provide feasible plans for physical interaction and manipulation of their environment and to optimize the manipulability of the plans~\cite{XU2020103548, VahrenkampAMSD12, 4791142, 1642001}. 

\begin{figure}[t]
        \centering
        \begin{subfigure}[b]{0.48\textwidth}
            \includegraphics[width=\textwidth]{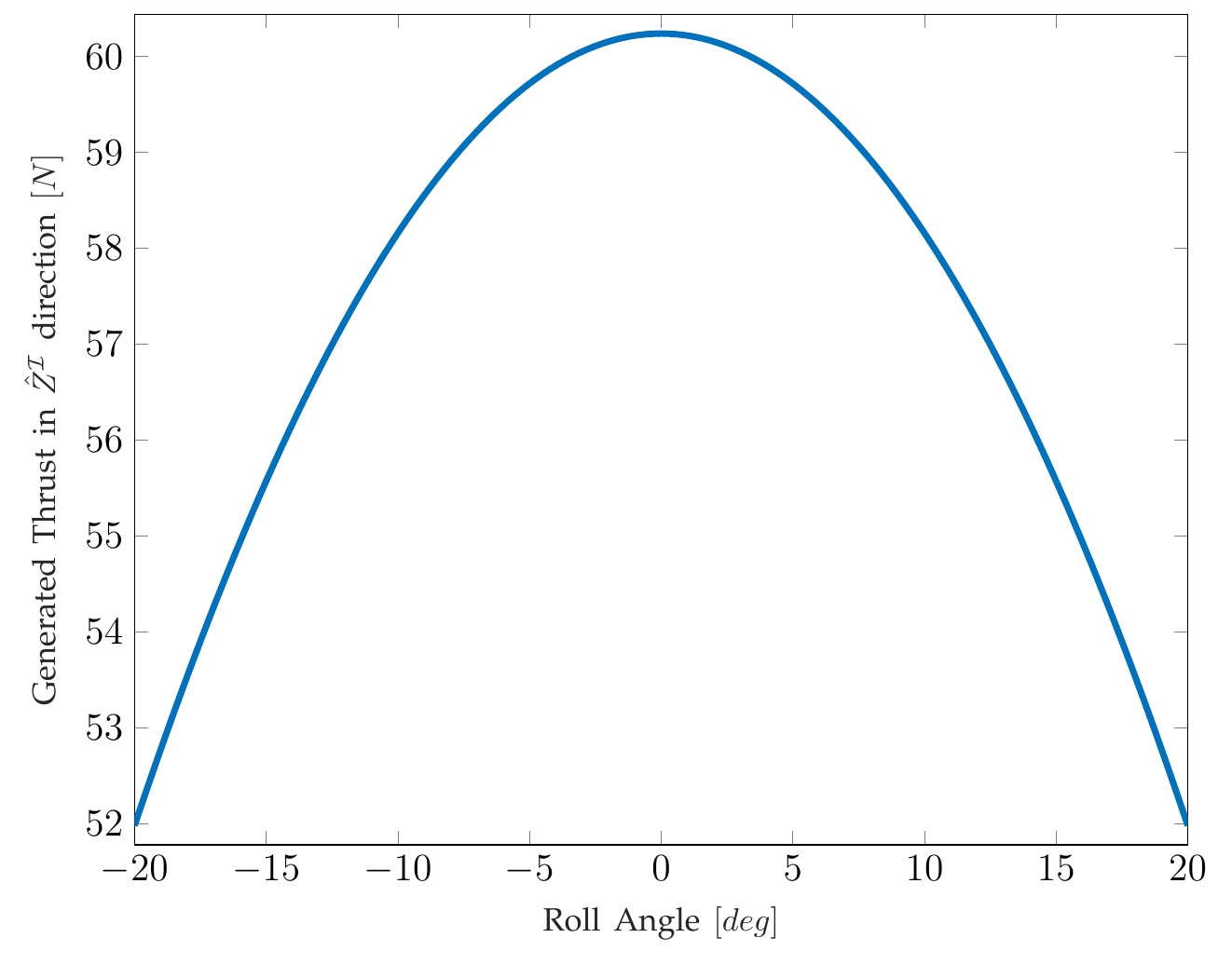}
            \caption{~}
        \end{subfigure}
        \hfill
        \begin{subfigure}[b]{0.48\textwidth}
            \includegraphics[width=\textwidth]{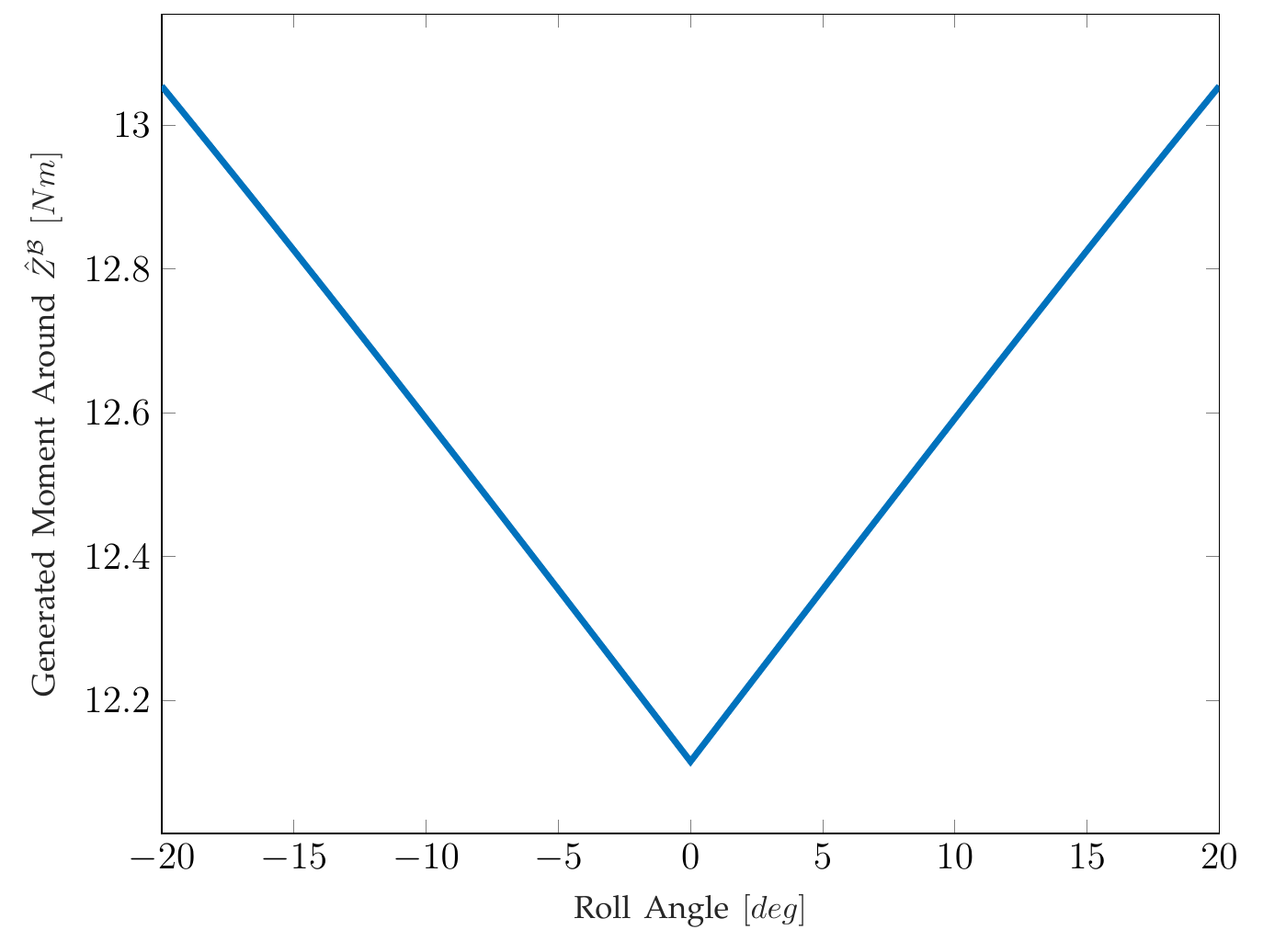}
            \caption{~}
        \end{subfigure}
        
        \caption[Operational profile example]{The maximum generated vertical thrust and the moment around the $\ZB$ axis at different roll angles for the fixed-pitch hexarotor used in this project (see Figure~\ref{fig:control:tests:tilthex-a}). }
        
        \label{fig:applications:planning:operational-profile}
    \end{figure}

In addition to pure feasibility check on the tasks and instantaneously applied wrenches, manipulability analysis can optimize the robot's approach and pose for the specific task to achieve the best manipulation. Jaquier et al.~\cite{0278364920946815} provides a planning system to optimize the robot's pose for the given task. Figure~\ref{fig:applications:planning:operational-profile} shows the thrust and moments generated at different roll angles of our hexarotor platform during contact. The plots show that the optimal outputs do not always align, and depending on the task, different trajectories should be planned to achieve the optimal goal. 

So far, planning methods have been using dynamic manipulability ellipsoids for online applications. These ellipsoids are computationally fast to compute, but they omit a significant portion of the wrench set, preventing the use of all the available wrenches.

On the other hand, the computation method for estimating the complete wrench set (dynamic manipulability polytopes) has been computationally expensive, limiting its use only to offline task and trajectory planners~\cite{0278364920946815}.

Our real-time wrench set methods proposed in Chapter~\ref{ch:wrench} allow the use of the entire wrench space in online planning methods. This extension can allow tasks that have been deemed infeasible with the limits given by dynamic manipulability ellipsoids and can further optimize the physical interaction tasks. In our previous work~\cite{Ashley:2021:riss:planning} we have shown an example of how our method can be used with the RRT*-Connect planner to plan a flight and manipulation task using our hexarotor with tilted arms. 

\section{Conclusion and Discussion} \label{sec:applications:conclusion}

This chapter discussed several aerial robotics applications where our proposed real-time wrench set estimation method is beneficial. The method's benefits can be seen in all kinds of free flight and contact scenarios, but its impact on controlled physical interaction and manipulation is much more significant. 

We first described how the control allocation module of aerial robots can utilize the real-time wrench set estimation methods to optimize the robot's behavior. Using this method ensures that the input of the control allocation is within the feasible limits and thus eliminates the need for optimization methods to refine the output of the dynamic inversion method (i.e., computed actuator commands). On the other hand, when the desired wrenches computed by the controller are infeasible, using the wrench set provides a flexible geometric method to decide the robot's behavior as opposed to the optimization-based post-processing of the output, which is difficult to formulate and implement.

Next, we used the wrench set estimation methods to prove that the full-tilt attitude strategy of our controller provides the maximum omni-directional acceleration when an external force is acting on the robot. Then we described how the optimal tilt and thrust could be computed in this scenario. The method is helpful for LBF robots where the available lateral force is limited, and the external forces can consume all the lateral force resulting in the loss of trajectory tracking performance.

We discussed how planning for physical interaction tasks can use the estimated wrench set to find feasible ways to perform the desired tasks and compute optimal plans that utilize the complete wrench set. The information on the possible wrenches in each step allows the planner to explore the entire configuration space and plan for motions and wrenches in real-time that were not possible before due to conservative assumptions for the wrench set.

Finally, we briefly mentioned some other applications of the methods for aerial robots, such as optimizing the multirotor design and recovery from failures during the flight.

Having the powerful tool of real-time wrench set estimation in hand, many other applications and improvements are possible. Due to the scope of our work, we only covered example applications for aerial robots. However, the method is general and can also be extended to other manipulator robots to optimize their physical interaction.

The following chapters focus on deformable objects and explore how aerial robots can perform physical interaction tasks on these objects.

\chapter{Deformable One-Dimensional Object Detection} \label{ch:doo-detection}

Aerial robots are increasingly being used for automated inspection and maintenance of infrastructure. While recent advances have primarily addressed the UAV interaction with rigid external bodies, the work on automated interaction and manipulation of deformable objects such as wires and cables has been minimal. In order to enable applications such as autonomous inspection, maintenance, and interaction of the infrastructure containing such deformable objects (e.g., utility poles, network interface devices, cable boxes), an effective method is required to detect these objects in a suitable manner for robotic manipulation. 

Various methods were developed across industrial and surgical robotics to segment the deformable objects. On the other hand, several methods have been proposed and implemented across academia and industry to allow the detection of power lines and bridge cables using UAVs for visual inspection or obstacle avoidance purposes. While the output of all these methods can be used for remote user operation, visual inspection, and obstacle avoidance, they are not suitable for manipulating these objects. 

On the other hand, many methods exist to model and track deformable one-dimensional objects (e.g., cables, ropes, and threads) across a stream of video frames. However, these methods depend on the existence of some initial conditions, and to the best of our knowledge, the topic of detection methods that can extract those initial conditions in non-trivial situations has hardly been addressed. The lack of detection methods limits the use of the tracking methods in real-world applications and is a bottleneck for fully autonomous applications that work with these objects.

This chapter proposes our approach for detecting deformable one-dimensional objects, which can be used for tasks such as routing and manipulation and automatically provides the initialization required by the tracking methods, effectively filling the gap between the segmentation methods and the tracking methods. It takes an image containing a deformable object and outputs a chain of fixed-length cylindrical segments connected with passive spherical joints. The chain follows the natural behavior of the deformable object and fills the gaps and occlusions in the original image. Our tests and experiments show that the method can correctly detect deformable one-dimensional objects in various complex conditions and can handle crossings and occlusions.

The proposed method moves us closer to enabling automated UAV inspection, maintenance, and manipulation of infrastructure containing cables and wires in the future. It was also a missing step for the complete automation of robotics applications involving planning and manipulation of such deformable objects in other fields as well, including industrial and surgical robotics settings.

\section{Introduction and Related Work} \label{sec:doo-detection:intro}

Manipulating non-rigid objects using robots has long been the subject of research in various contexts~\cite{frobt.2020.00082}. A specific class of non-rigid objects is called Deformable One-dimensional Objects (DOOs) or Deformable Linear Objects (DLOs) and includes objects such as ropes, cables, threads, sutures, and wires. 

In order to achieve full autonomy in physical interaction tasks using UAVs involving DOOs (e.g., inspection and maintenance of utility poles), one of the essential and challenging parts is perception. Many such applications require complete knowledge of the object's initial conditions, even in the occluded parts. Moreover, routing and manipulation applications also require a representation model of the DOO that allows simulation and computation of its dynamics.

There are many methods proposed in the medical imaging community that can find the DOOs (known as tubular structures in that context) in the image frame~\cite{KRISSIAN2000130, 28493, Wang2021, Noble2011, 32226}. These methods are considered segmentation methods and are perfected for low signal-to-noise images, and some can even work with self-crossings. However, they mainly only provide the region in the image containing the DOO, and their output is a collection of data points (e.g., pixels or point clouds). Such methods can be adapted for our robotics settings, such as physical UAV inspection. However, their output needs further processing to be useful for automated robotics tasks such as manipulation or routing.

The UAV community has also extensively worked on the segmentation of the cables and wires for visual inspection and obstacle avoidance purposes. These methods can effectively segment out the power lines, bridge cables, and other near-straight deformable objects~\cite{8206190, 8577142, 7532456, Dai2020, 7279641, 7181891, PAGNANO2013234, 6322366}. However, none of these approaches can handle true deformations that other cables and wires may have, such as turns and crossings, making these approaches unsuitable for other types of deformable one-dimensional objects used in utility poles and other robotics tasks.

On the other hand, various algorithms have been proposed for industrial settings to track a DOO across the video frames, even in the presence of occlusions and self-crossings. These methods may devise different tools such as registration methods (e.g., Coherent Point Drift~\cite{5432191}), learning, dynamics models, and simulation to predict and correct the prediction across the consecutive frames~\cite{5980431, 15711, wang2020tracking, rastegarpanah, 6630714, 8560497, af66d77e53304e6bbb64049bb46193cb, 8206058}. While some of these methods can initialize the DOO in the first frame using trivial conditions (e.g., a straight rope in camera view), the other methods require even a simple DOO configuration to be provided to them a priori. Most tracking methods will fail to initialize in even slightly complex initial conditions. 

The initial conditions are not generally provided in real-world applications, and pure segmentation is insufficient and requires further processing. For example, our application of an aerial manipulator working with wires at the top of a utility pole needs to detect the desired wire, including its occluded parts and crossings and requires a model of the detected wire to perform any task on it.

The authors of~\cite{8972568} proposed a method that trains on rendered images and fine-tunes on real images to detect and track the state of a rope. The method requires tens of thousands of rendered and real images of the same rope for training and needs re-training and tuning for a new DOO. This approach may be acceptable for an industrial application focusing on DOOs of the exact same characteristics, but it may not be practical for many other applications, including ours.

We propose a method to detect the initial conditions of a deformable one-dimensional object~\cite{Keipour:2022:ral:doo, Keipour:2022:icra-workshop:doo}. The method fills the occluded parts and works with crossings. The output is a single object represented as a discretized model useful for routing, manipulation, and simulation. The detected object can be used for initialization of other existing tracking methods to be used in applications such as manipulating DOOs into desired shapes and knots~\cite{9044123, 8403315, 0278364906064819, 4359263} and is a step toward enabling aerial manipulation of wires and cables for inspection and maintenance tasks.

\section{Problem Definition} \label{sec:doo-detection:problem}

This work addresses the detection of cable-like deformable shapes. Unlike rigid objects, the shape of deformable objects can change, and some form of a flexible model is required to represent the current state of their shape. On the other hand, to predict the reaction of the deformable objects to the applied forces and moments, the representation model should facilitate the integration of a dynamics model. 

In theory, a DOO represented by its pixels (or voxels) in the camera frame can be integrated with a dynamics model. However, the model would typically require finite element analysis and is computationally expensive, making it impractical for robotics applications. Simpler representations are commonly used in tasks such as manipulation, routing, and planning. Arriola-Rios et al.~\cite{frobt.2020.00082} provide an overview of the common representations for deformable objects.

\begin{figure}[!htb]
    \centering
    \includegraphics[width=0.6\linewidth, height=1cm]{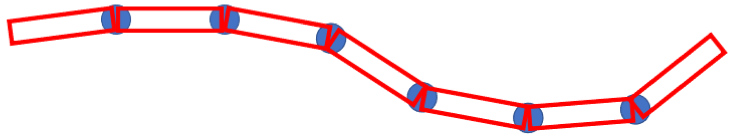}
    \caption[DOO representation as a chain of cylinders]{The representation of a DOO as a chain of fixed-length cylinders connected by spherical joints.}
    \label{fig:doo-detection:doo-representation}
\end{figure}

A commonly-used approach for representing DOOs is to model them as a chain of fixed-length cylindrical segments connected by spherical joints. This simple model can easily integrate with an efficient dynamics model to simulate or predict the object's behavior. While the chosen model does not affect the ideas described in the proposed method, our method utilizes this model. In our application, the length of each segment is represented by $l_s$, and there is no gap between the segments (i.e., each segment starts precisely where the neighbor segment ends).
Figure~\ref{fig:doo-detection:doo-representation} illustrates the fixed-length cylinder chain model used in this work.

The focus of this work is to provide the cylinder chain representation of a deformable one-dimensional object seen in the camera frame (which can be RGB or RGB-D/3-D). The output chain model should predict and fill the path taken by the DOO under the occlusions and return a single chain object. 

\section{Proposed Method} \label{sec:doo-detection:method}

\begin{figure}[!htb]
    \centering
    \includegraphics[width=0.8\linewidth]{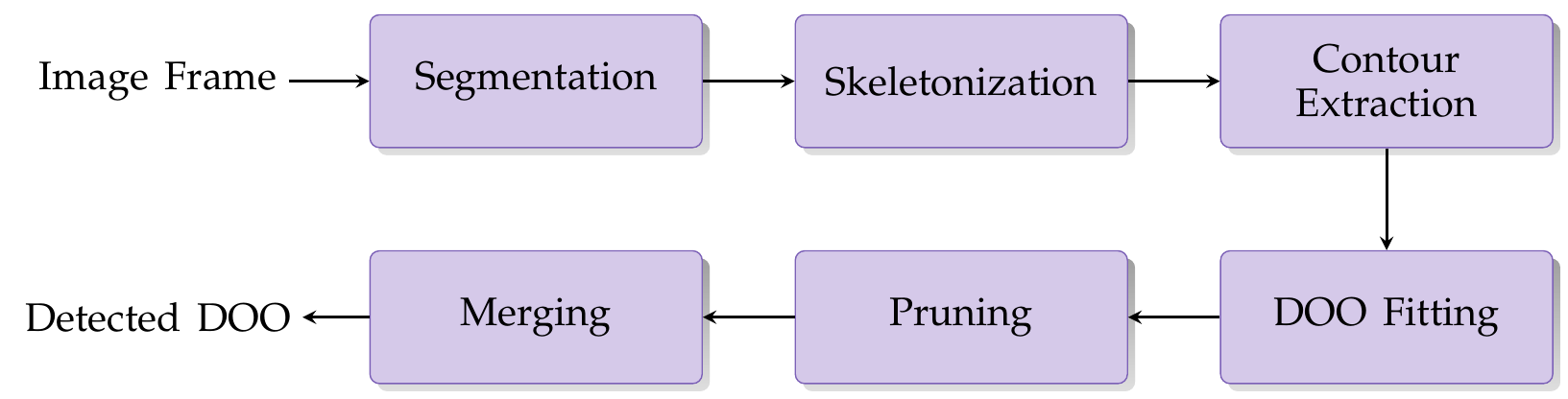}
    \caption[DOO detection high-level overview]{The high-level overview of the proposed method for detection of deformable one-dimensional objects.}
    \label{fig:doo-detection:overview}
\end{figure}

The DOO detection method in this work takes the camera frame and performs a sequence of processing steps to output a single object in chain representation (described in Section~\ref{sec:doo-detection:method}). Figure~\ref{fig:doo-detection:overview} shows a high-level overview of the method.

The first three steps in the proposed method are well-known processes. Our algorithm can work with different segmentation, skeletonization, and contour extraction approaches. The choice depends on the task at hand. The last three processing steps are the contributions of our algorithm. 

Algorithm~\ref{alg:doo-detection:doo-detection} shows the pseudo-code of our approach. This section describes each of those steps in more detail.

\begin{algorithm}[!htb]
\caption{Deformable one-dimensional object detection.}
\label{alg:doo-detection:doo-detection}
\begin{algorithmic}[1]
\LineComment {Detects and extract DOOs from the input image frame.}
\Function{DetectDOO}{frame}
    \LineComment{Segment the image to extract the DOO region}
    \State $segmented\_img \gets \textsc{Segment}(frame)$
    \LineComment{Extract the skeletons of the segmented regions}
    \State $thinned\_img \gets \textsc{Skeletonize}(segmented\_img)$
    \LineComment{Extract the contours from the skeletons}
    \State $contours \gets \textsc{ExtractContours}(thinned\_img)$
    \LineComment {Fit DOO chains to all contours}
    \State $Chains \gets \emptyset$
    \ForAll {$c \in contours$}
        \State $fitted\_chains \gets \textsc{FitDOO}(c)$
        \State $Chains$.insert$(fitted\_chains)$
    \EndFor
    \LineComment {Prune all the overlapping segments}
    \State $Chains \gets \textsc{Prune}(Chains)$
    \LineComment {Merge the DOO chains into a single DOO}
    \While {$Chains.length > 1$}
        \State $C_1, C_2 \gets \textsc{FindBestMergeMatch}(Chains)$
        \State $C_{merged} \gets \textsc{MergeChains}(C_1, C_2)$
        \State $Chains$.remove$(\{C_1, C_2\})$
        \State $Chains$.insert$(C_{merged})$
    \EndWhile
    \State \Return $Chains[0]$ \NewComment {Return the final DOO chain}
\EndFunction
\end{algorithmic}
\end{algorithm}

\subsection{Segmentation}

A vital step in extracting the complete DOO from the input camera image is segmentation. This step aims to filter the image data to extract the DOO portions and exclude all other data. More formally, defining $P_{doo}$ as the collection of all image data points (pixels in 2-D case) belonging to the DOO, the segmentation output should ideally be the collection $P_{seg}$ where:
    \begin{equation}
        P_{seg} \subseteq P_{doo}, \quad \frac{|P_{seg}|}{|P_{doo}|} \approx 1,
    \end{equation}
  
\noindent where $\|\cdot\|$ is the number of data points in the collection.

The simplest segmentation methods include color-based filtering and background subtraction, which can work in lab settings, but more complex methods are required for real-world applications.

The medical research community provides an extensive set of segmentation methods to address vein and vessel detection, which can be used here directly or with small modifications. Such methods include both model-based~\cite{BFb0056195, KRISSIAN2000130, 28493, Wang2021, 10605, Noble2011} and learning-based~\cite{72a4e1c53c, 32226} approaches and are often robust to clutter in the input image and can work in low signal-to-noise conditions.

\begin{figure}[!htb]
\centering
    \begin{subfigure}[b]{0.48\linewidth}
        \includegraphics[width=\textwidth]{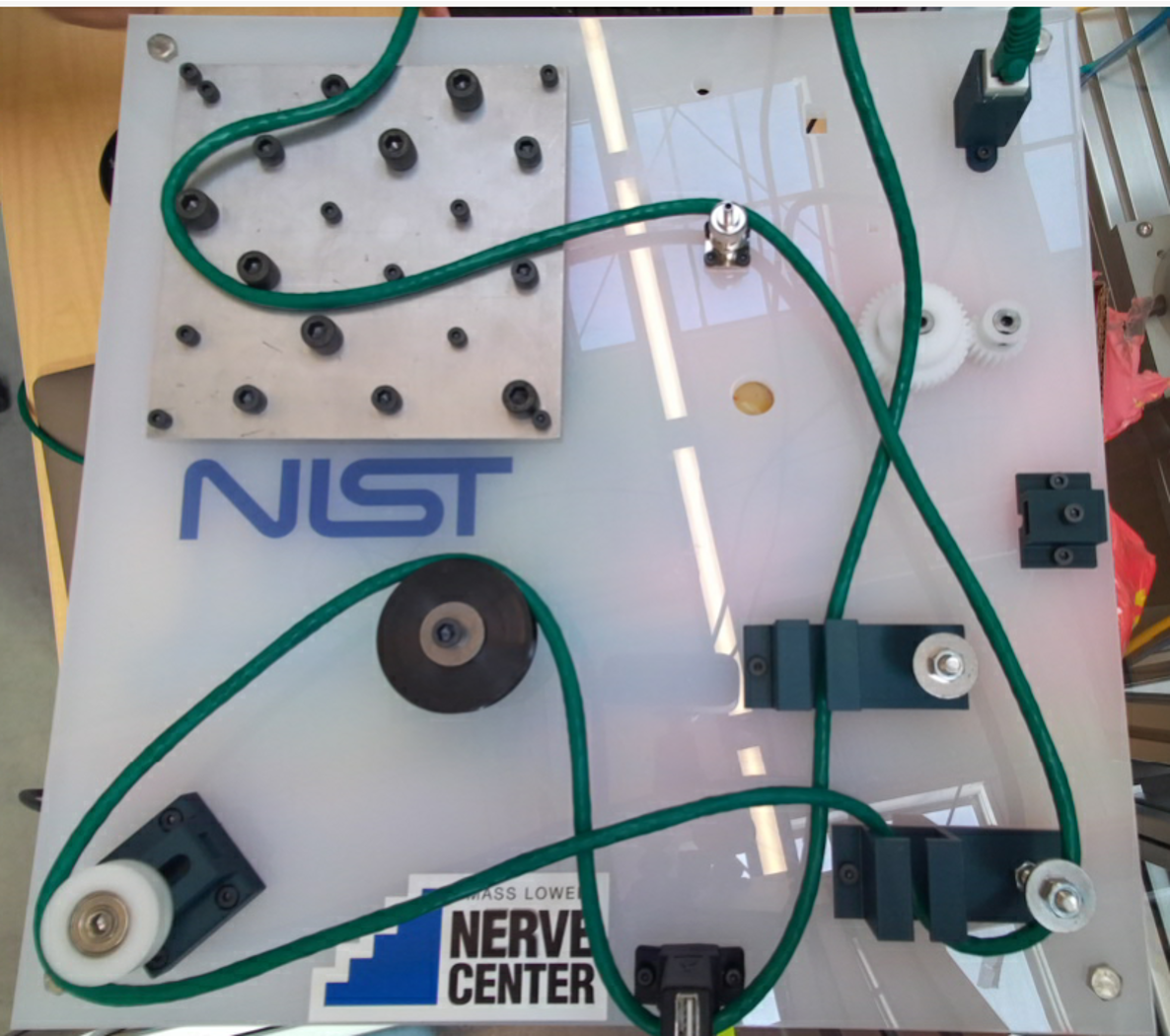}
        \caption{~}
        \label{fig:doo-detection:original-board}
    \end{subfigure}
    \hfill
    \begin{subfigure}[b]{0.48\linewidth}
        \includegraphics[width=\textwidth]{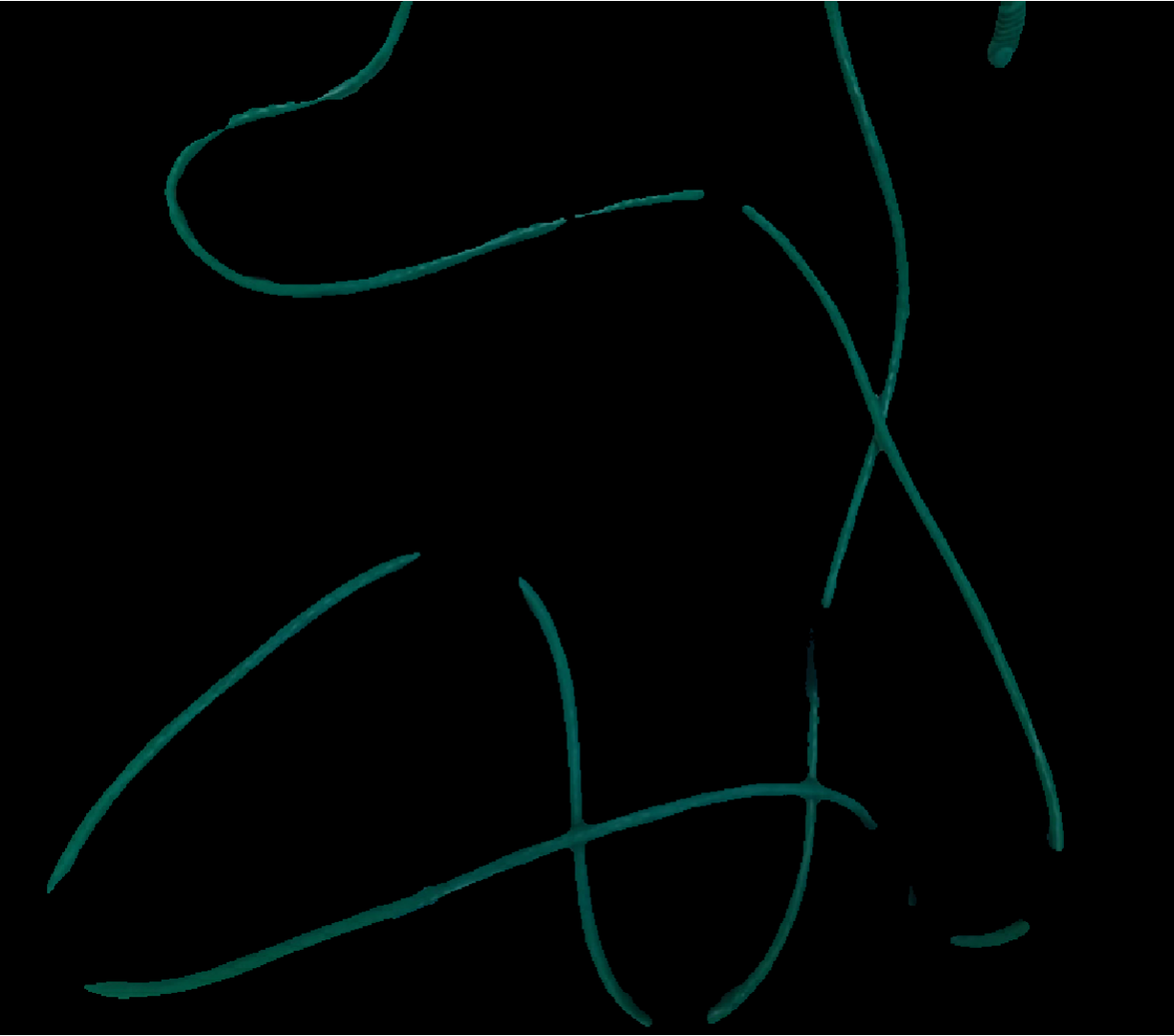}
        \caption{~}
        \label{fig:doo-detection:segmented-board}
    \end{subfigure}
    \caption[DOO segmentation result]{The segmentation of a DOO in a camera frame. (a) The original image. (b) The segmentation result.}
    \label{fig:doo-detection:segmentation}
\end{figure}

The requirement for the segmentation method for our work is to filter the DOO data conservatively, i.e., ideally, it should eliminate all the unrelated data even if it removes some of the DOO data. Figure~\ref{fig:doo-detection:segmentation} illustrates the segmentation of an example cable in the camera frame.

\subsection{Topological Skeletonization}
    
Skeletonization transforms each segmented connected component into a set of connected pixels with single-pixel width called a \textit{skeleton}. It is commonly used in the pre-processing stage of various applications ranging from Optical Character Recognition (OCR) to human motion tracking, fingerprint analysis, and various medical imaging analysis~\cite{SAHA20173, Keipour:2013:arxiv:ocr, Ensafi:2009:ieec:ocr}.

\begin{figure}[!htb]
    \centering
    \includegraphics[width=0.46\textwidth, height=5cm]{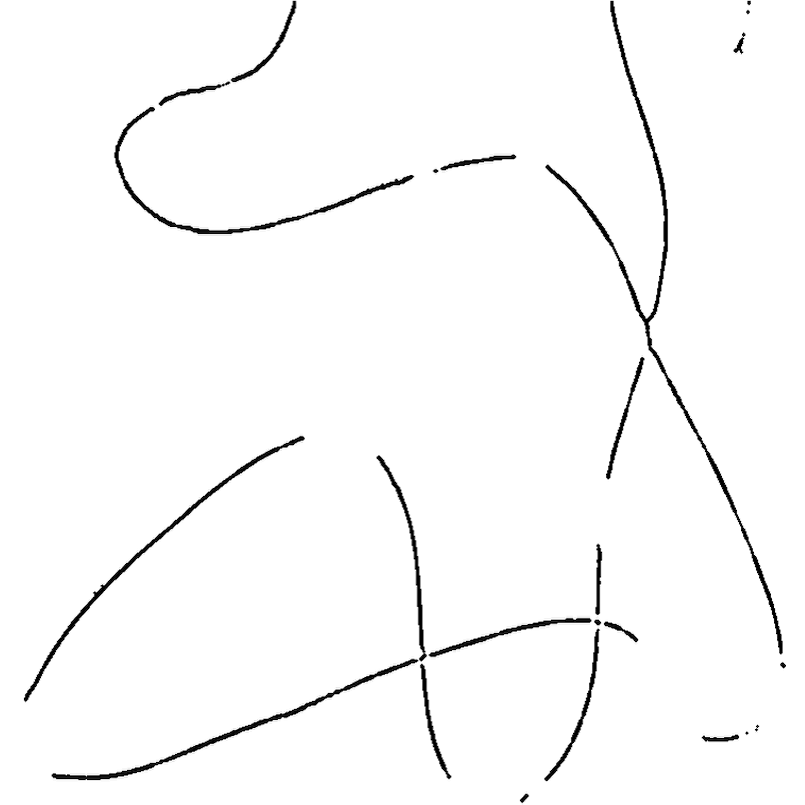}
    \caption[DOO skeletonization result]{The skeletonization of a segmented deformable one-dimensional object.}
    \label{fig:doo-detection:skeletonization}
\end{figure}

Our algorithm has two requirements for choosing the skeletonization method: a) the skeleton of a connected component should remain a connected component; b) only one branch should be returned per actual branch (i.e., multi-branching of a single skeleton branch should be avoided). The skeletonization algorithm chosen for this step should inherently respect the two constraints. Figure~\ref{fig:doo-detection:skeletonization} shows the skeletonization of the segmented example of Figure~\ref{fig:doo-detection:segmentation}.
    
\subsection{Contour Extraction}

A contour (a.k.a., boundary) is an ordered sequence of the pixels around a shape. Extracting contours from an image is utilized in many applications ranging from shape analysis to semantic segmentation and image classification~\cite{Gong2018, Keipour:2021:ral:ellipse}. 

\begin{figure}[!htb]
    \centering
    \includegraphics[width=0.35\textwidth]{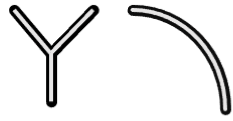}
    \caption[Extracted contour types]{Contour types extracted from a skeletonized image. The black area around the gray skeleton is the contour.}
    \label{fig:doo-detection:contour-types}
\end{figure}

Ideally, the contour extraction method applied to the skeletons should result in one contour per skeleton branch. However, the contour extraction methods can result in several contours per branch and some contours containing multiple branches in practice. Moreover, a contour contains the closed boundary \textit{around} the skeleton and not the actual skeleton pixels. Figure~\ref{fig:doo-detection:contour-types} shows different types of contours that can be extracted from a skeleton piece.

\begin{figure}[!htb]
    \centering
    \includegraphics[width=0.46\textwidth, height=5cm]{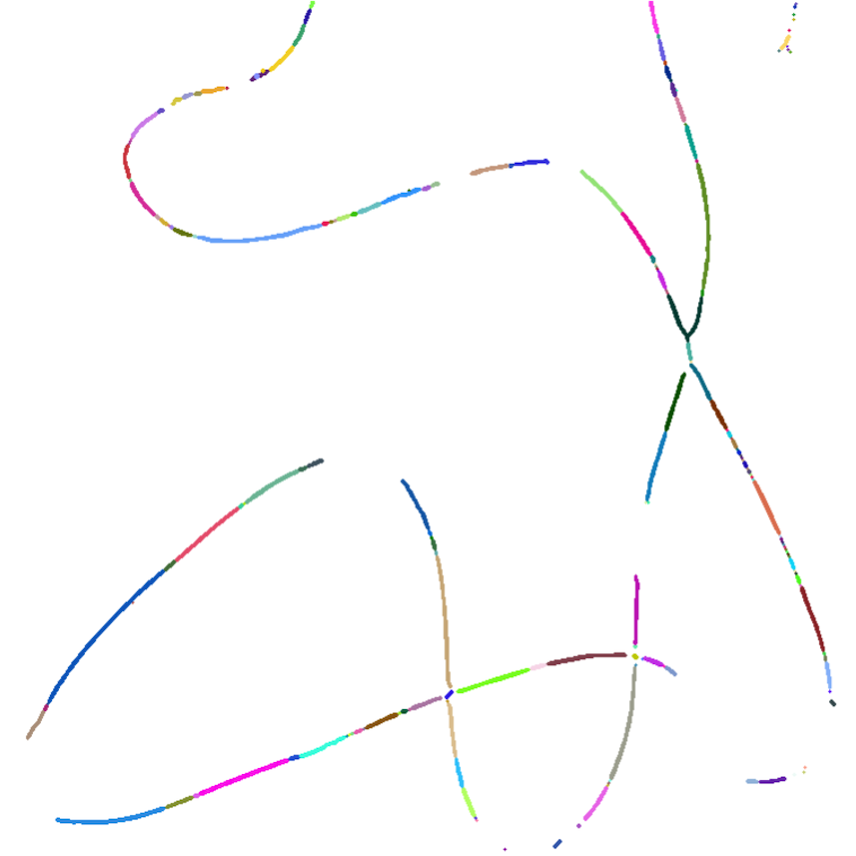}
    \caption[DOO contour extraction result]{The contours extracted from the skeleton. Each contour is drawn with a different color.}
    \label{fig:doo-detection:contour-extraction}
\end{figure}

Our DOO detection method can handle the above-mentioned common issues raised by the contour extraction methods. Therefore, many of the existing contour extraction methods can be used with our algorithm regardless of their output limitations. Figure~\ref{fig:doo-detection:contour-extraction} presents the result of contour extraction.

Contours facilitate traversing points along the skeleton and simplify determining the connections in the branches. The contour extraction step can be skipped if an ordered set of pixels for each skeleton branch is obtained from the skeletonization method or other means.

\subsection{Fitting DOO Segments} \label{sec:doo-detection:segment-fitting}

The next step is to fit a chain of fixed-length segments to each contour. A contour can be a single branch, or it may contain multiple branches (see Figure~\ref{fig:doo-detection:contour-extraction}). The pixel sequence for a contour returned by a typical contour extraction method starts from one of the tips and ends with a sharp turn back at the start point.

Let us call the latest added segment as $s$, the current segment as $s'$, the first point in the contour sequence as the starting point $p_s$ of $s'$, and the point currently being traversed as $p_c$. Let us also define $\Vec{s}$ as the vector in the direction of segment $s$, starting at its start point and pointing towards its end point. Therefore, starting with an empty DOO chain, the points are traversed while the distance $\|p_c - p_s\|_2$ is less than $l_s$. Then a new segment of length $l_s$ is added to the DOO chain from point $p_s$ in the direction of $p_c$, the new segment's end-point $p_e$ is saved as the next segment's start point $p_s$, and the traversal continues.

Three conditions may happen during the traversal:
\begin{enumerate}[leftmargin=*]

    \item Vector $\Vec{s'}$ is close to vector $\Vec{s}$: The new segment is added to the current DOO chain in this case.
    
    \item Vector $\Vec{s}'$ is close to vector $-\Vec{s}$: It means that the traversal has gone over a branch end. In this case, the current DOO is recorded without the new segment, and the new segment is discarded. The traversal continues from the branch tip with a new empty DOO chain.

    \item The last point in the contour sequence is reached: it means that the traverse has returned to the start point, and the traverse can be terminated. There may be cases where the whole contour length is less than the segment size. In these cases, no new DOO chains will be generated.
\end{enumerate}

Algorithm~\ref{alg:doo-detection:contour-traversal} shows the pseudo-code for the described steps.

\begin{algorithm}[!htb]
\caption{Traversing contours for DOO chain creation.}
\label{alg:doo-detection:contour-traversal}
\begin{algorithmic}[1]
\LineComment {Traverses a contour and returns all created DOO chains.}
\Function{TraverseContour}{\textit{contour}}
    \LineComment {Initialize the collection of DOO chains}
    \State $Collection \gets \emptyset$
    \LineComment {Initialize the start point and the next DOO chain}
    \State $p_s \gets contours[0]$\quad,\quad$chain \gets \emptyset$
    \LineComment {Initialize the tip point}
    \State $p_t \gets p_s$\quad,\quad$update\_tip \gets True$
    \LineComment {Traverse over all the points in the contour}
    \ForAll {$p_c \in contour$}
        \LineComment {Create a new segment if $p_c$ is far enough}
        \If {$\textsc{Dist}(p_c, p_s) \geq l_s$}
            \State $p_e \gets p_s + l_s \times (p_c - p_s) / \|p_c - p_s\|$
            \State $s' \gets \textsc{CreateSegment}(p_s, p_e)$
            \If {$chain = \emptyset$ or $\textsc{Angle}(s, s')$ is small}
                \LineComment{Add the segment if it is the first in chain}
                \LineComment {or if the direction has not changed much}
                \State $chain.$Insert$(s')$
                \State $s \gets s'$\quad,\quad$p_s \gets p_e$\quad,\quad$p_t \gets p_e$
            \Else
                \LineComment{Start a new chain if direction has changed}
                \State $Collection.$Insert$(chain)$
                \State $chain \gets \emptyset$\quad,\quad$p_s \gets p_t$
            \EndIf
            \State $update\_tip \gets True$ \quad \NewComment{Start updating tip point}
        \ElsIf {$\textsc{Dist}(p_c, p_s) \geq \textsc{Dist}(p_t, p_s)$}
            \IIf {$update\_tip = True$} {$p_t \gets p_c$} \EndIIf \quad \NewComment{Update the tip point}
        \Else
            \State $update\_tip \gets False$ \quad \NewComment{Stop updating tip point}
        \EndIf
    \EndFor
    \State $Collection.$Insert$(chain)$ \quad \NewComment {Add the last generated chain}
    \State \Return $Collection$ \quad \NewComment {Return all the chains at the end}
\EndFunction
\end{algorithmic}
\end{algorithm}

\subsection{Pruning}

The segment fitting algorithm of Section~\ref{sec:doo-detection:segment-fitting} returns multiple overlapping DOO chains for each part of the object. It is desired to prune the overlapping segments to reduce the total number of segments and simplify the further steps by assuming that no two segments overlap.

\begin{figure}[!htb]
    \centering
    \includegraphics[width=0.3\textwidth]{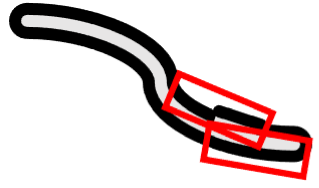}
    \caption[Illustration of overlapping DOO segments]{An illustration of overlapping segments around a skeletonized deformable one-dimensional object.}
    \label{fig:doo-detection:segment-overlap}
\end{figure}

To define the overlap of two segments, each segment can be assumed as a rotated rectangle in the 2-D case and a square cuboid (a cuboid with two square faces) for the 3-D case. The length of the rectangle and cuboid is the segment length $l_s$, and the rectangle's width is 3 pixels or higher. The reasoning for the choice of the width is that the width of the skeleton is generally 1 pixel with occasional width of 2 pixels (depends on the choice of the thinning method). The contour is the boundary around the skeleton, and for the two rectangles on the two sides of the skeleton to overlap, they need to be at least 3 pixels wide. In practice, any small number greater than 3 pixels should work well for pruning the overlapping segments. For the 3-D case, the width of the cuboid can be chosen similarly, i.e., it should be at least the total of the width of the contour layer and the maximum width of the skeleton. Once the segments are defined, the geometric intersection of two rotated rectangles or cuboids is used to find the overlap. Figure~\ref{fig:doo-detection:segment-overlap} shows how two segments can overlap for the same skeleton.

A heuristic that has shown performance improvements in the subsequent stages removes the segment from the shorter chain when two segments overlap. This heuristic will result in many chains being quickly emptied, which reduces the overall number of DOO chains in the collection.

\begin{figure}[!htb]
    \centering
    \includegraphics[width=0.46\textwidth, height=5cm]{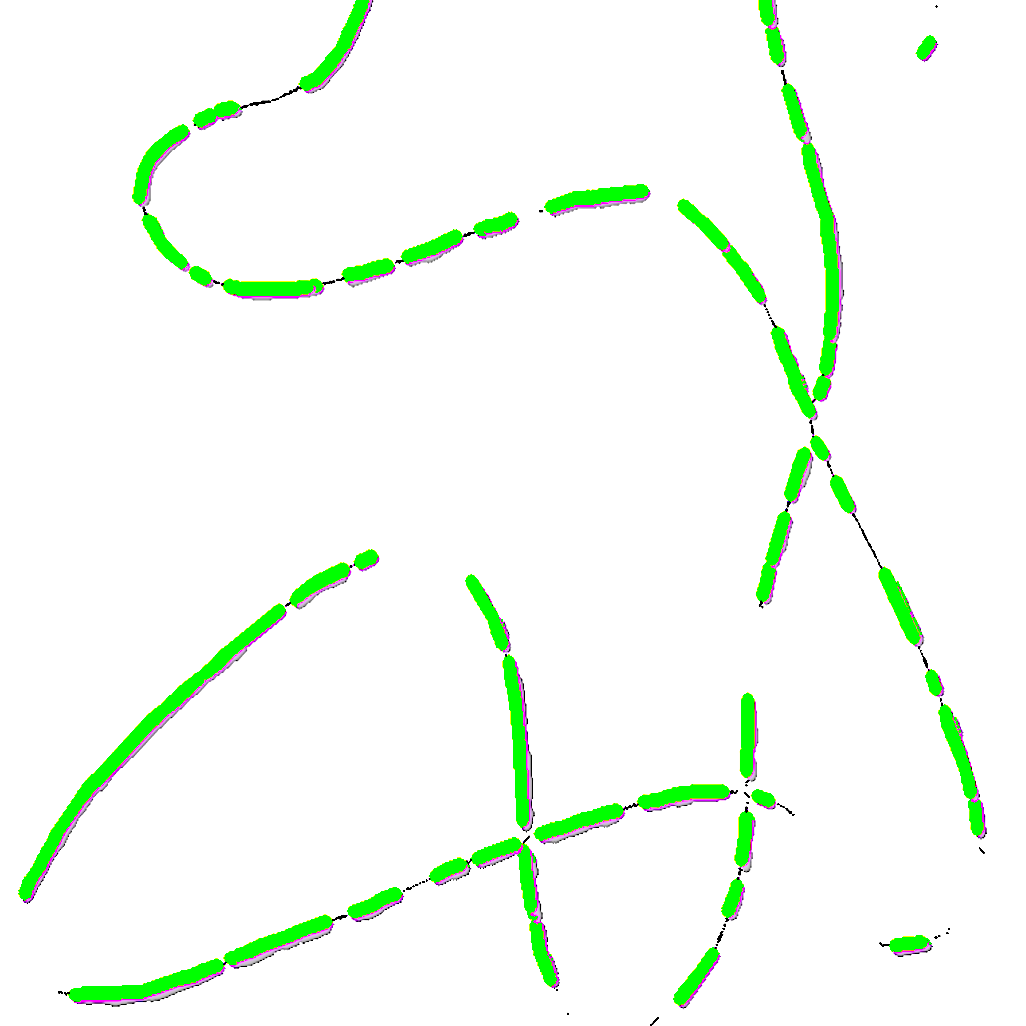}
    \caption[DOO segment and pruning result]{The result of segment fitting from contours and pruning for a deformable one-dimensional object of Figure~\ref{fig:doo-detection:skeletonization}.}
    \label{fig:doo-detection:pruning}
\end{figure}

Figure~\ref{fig:doo-detection:pruning} shows the chains resulting from segment fitting and then pruning of the contours in Figure~\ref{fig:doo-detection:contour-extraction}.

\subsection{Merging}

Once we have a collection of DOO chains, they should be merged to fill the gaps and form a single object. A gap can result from an occlusion or an imperfect segmentation and should be filled in a way that follows the natural curve of the deformable object. 

The merging process can be performed iteratively, connecting two chains at a time until all the chains are merged into a single deformable one-dimensional object. Each iteration can be broken down into two steps:

\begin{enumerate}
    \item Choose the best two chains for merging
    \item Connect the selected chains
\end{enumerate}

The following subsections describe choosing the best chains and properly connecting them with their natural bend.

\subsection{Merging: Choosing the Best Matches} \label{sec:doo-detection:method:merging:choosing}

To choose the best chains to connect, we define a new cost function $C_M(\cdot)$ that calculates the cost of connecting any two chain ends. Considering that there are two ends for each chain, there will be four cost values for connecting the two ends for any two chains. The lowest among the four values is the cost of connecting the two chains. 

Given an end segment $s_1$ of the first chain and an end segment $s_2$ of the second chain, three separate partial costs are defined and then combined to create the total cost function $C_M(\cdot)$:

\begin{itemize}[leftmargin=*]
    \item Euclidean Cost $C_E$: This measure incurs costs to two chain ends based on their Euclidean distance to deter the early connection of far away ends (Figure~\ref{fig:doo-detection:merge-costs-e}). Having the end segments $s_1$ and $s_2$, this cost can be calculated as:
    \begin{equation}
        C_E\left(s_1, s_2\right) = \|s_1.end - s_2.end\|_2,
    \label{eq:doo-detection:cost-euclidean}
    \end{equation}
    \noindent where $\|\cdot\|_2$ is the norm of the resulting vector and $s.end$ is the end point of the segment (end point of the chain).
    
    \begin{figure}[!htb]
    \centering
        \begin{subfigure}[b]{0.23\linewidth}
            \includegraphics[width=\textwidth]{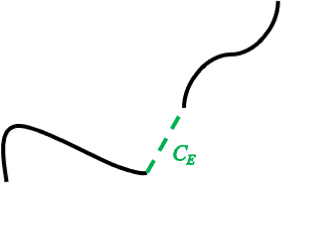}
            \caption{~}
            \label{fig:doo-detection:merge-costs-e}
        \end{subfigure}
        \hfill
        \begin{subfigure}[b]{0.23\linewidth}
            \includegraphics[width=\textwidth]{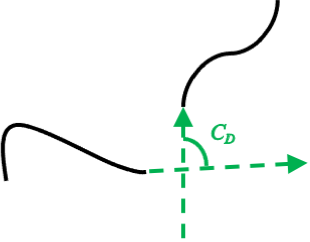}
            \caption{~}
            \label{fig:doo-detection:merge-costs-d}
        \end{subfigure}
        \hfill
        \begin{subfigure}[b]{0.46\linewidth}
            \includegraphics[width=\textwidth]{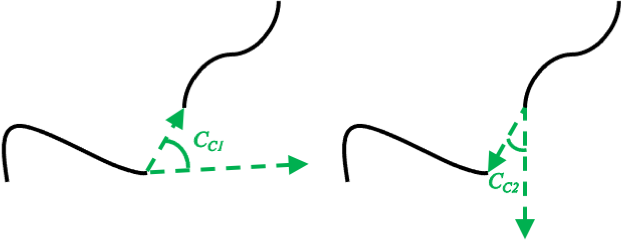}
            \caption{~}
            \label{fig:doo-detection:merge-costs-c}
        \end{subfigure}
    \caption[Illustration of DOO segment merging costs]{Illustration of different partial costs of merging two chain ends. (a) Euclidean cost. (b) Direction cost. (c) Curvature costs from first chain to the second and from the second chain to the first.}
    \label{fig:doo-detection:merge-costs}
    \end{figure}

    \item Direction Cost $C_D$: This measure incurs costs to two chain ends based on the difference between the direction of the first with the opposite direction of the second (Figure~\ref{fig:doo-detection:merge-costs-d}). This cost discourages the connection of chains that are not facing each other. Having the end segments $s_1$ and $s_2$, this cost can be calculated as:
    \begin{equation}
        C_D\left(s_1, s_2\right) = \left|\arccos{\left(\frac{-\Vec{s_1}\cdot \Vec{s_2}}{\|s_1\|\|s_2\|}\right)}\right|,
    \label{eq:doo-detection:cost-direction}
    \end{equation}
    
    \noindent where $|\cdot|$ is the absolute value, $\|\cdot\|$ is the norm of the vector (size of the segment), $\Vec{s}$ is the vector along the segment $s$ starting from the start of the segment and ending at the end of the segment (i.e., the end of the chain), and $\cdot$ is the inner product operator.

    \item Curvature Cost $C_C$: This measure incurs costs to two chain ends based on how much curvature is needed to connect them. It is calculated as the higher cost of bending the first chain's end segment towards the second chain's end segment and vice versa (Figure~\ref{fig:doo-detection:merge-costs-c}). The measure is defined to discourage connections requiring excessive bending and to encourage smooth connections. Having the end segments $s_1$ and $s_2$, this cost can be calculated as:
    \begin{equation}
        \begin{split}
            &C_{C1}\left(s_1, s_2\right) = \left|\arccos{\left(\frac{\Vec{s_1}\cdot \Vec{s_{21}}}{\|s_1\|\|s_{21}\|}\right)}\right| \\
            &C_{C2}\left(s_1, s_2\right) = \left|\arccos{\left(\frac{\Vec{s_2}\cdot \Vec{s_{12}}}{\|s_2\|\|s_{12}\|}\right)}\right| \\
            &C_C\left(s_1, s_2\right) = \max{(C_{C1}, C_{C2})},
        \end{split}
        \label{eq:doo-detection:cost-curvature}
    \end{equation}
    
    \noindent where $s_{nm}$ is a shorthand for $s_n.end - s_m.end$.
\end{itemize}

Having the three cost values for the ends of two chains, the total cost of these ends is computed as:
\begin{equation}
    C_M(s_1, s_2) = \mathscr{F}\Big(C_E\left(s_1, s_2\right), C_D\left(s_1, s_2\right), C_C\left(s_1, s_2\right)\Big),
    \label{eq:doo-detection:cost-merge}
\end{equation}

\noindent where $\mathscr{F}$ is the function combining the three values. In practice, we learned that the weighted sum of the values works well, and even after manually choosing a simple weight set, the algorithm works for almost all kinds of situations (see Section~\ref{sec:doo-detection:tests} for our test values). With the weighted sum, the Equation~\ref{eq:doo-detection:cost-merge} reduces to:
\begin{equation}
    C_M(s_1, s_2) = w_e \cdot C_E\left(s_1, s_2\right) + w_d \cdot C_D\left(s_1, s_2\right) + w_c \cdot C_C\left(s_1, s_2\right)
    \label{eq:doo-detection:cost-merge-weighted-sum}
\end{equation}

After calculating the four costs of all end combinations of the two chains, the minimum of those costs is the cost of merging the two DOO chains. Once the costs for all pairs of chains are calculated, the two chains with the lowest total merging cost are chosen for merging. 

Note that the choice for the cost function of Equation~\ref{eq:doo-detection:cost-merge} is to encourage the connection of closer chains that align well and can connect smoothly. Choosing a single measure such as minimum curvature would result in unwanted connections of farther chains that align perfectly over closer chains that are slightly misaligned.

\subsection{Merging: Connecting Two Chains}

Once two chains $C_1$ and $C_2$ are selected for connection (see Section~\ref{sec:doo-detection:method:merging:choosing}), the gap between the two chains should be filled with a new chain $C_{new}$ in a way that it follows the expected curve of the deformable object. Our experiments show that any deformable object can take almost any curve given different pressure points, forces, tensions, and the object's condition. However, it is possible to have an educated \textit{guess} on how the object behaves. For this purpose, we calculate the "natural" curvature required for the new chain $C_{new}$, which connects the desired end of $C_1$ to the desired end of $C_2$.

\begin{figure}[!htb]
\centering
    \begin{subfigure}[b]{0.3\linewidth}
        \includegraphics[width=\textwidth]{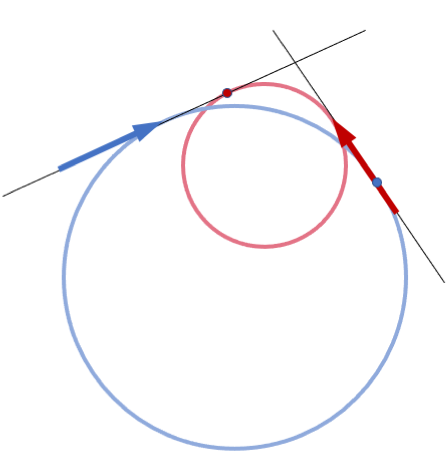}
        \caption{~}
        \label{fig:doo-detection:merge-cases-1}
    \end{subfigure}
    \hfill
    \begin{subfigure}[b]{0.3\linewidth}
        \includegraphics[width=\textwidth]{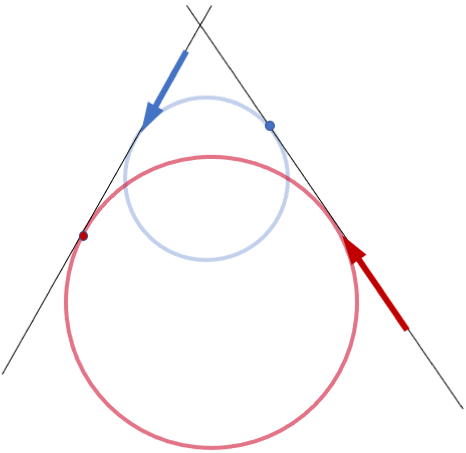}
        \caption{~}
        \label{fig:doo-detection:merge-cases-2}
    \end{subfigure}
    \hfill
    \begin{subfigure}[b]{0.3\linewidth}
        \includegraphics[width=\textwidth]{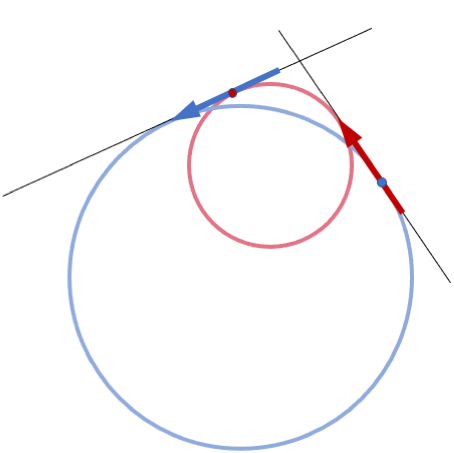}
        \caption{~}
        \label{fig:doo-detection:merge-cases-0}
    \end{subfigure}
\caption[Illustration of DOO segment merging scenarios]{Illustration of different merging scenarios with the two circles tangent to the line passing the end points of the two chains, each circle passing through one of the end points. Arrow ends and directions represent the end points and end directions of the chains. (a) Exactly one of the circles passing through the end point of a chain is touching the other line ahead of the other chain. (b) Both the circles passing through the end points of the chains are touching the other line ahead of the other chain. (c) None of the circles passing through the end point of a chain are touching the other line ahead of the other chain.}
\label{fig:doo-detection:merge-cases}
\end{figure}

To compute the "natural" curvature, we assume that the new chain $C_{new}$ starts in the same direction as the two desired chain ends. In other words, at each end, $C_{new}$ initially follows the direction of the last segment of the chain to which it is connected. On the other hand, we assume that when it is possible, the deformable one-dimensional object will follow a curve with a constant turn rate (i.e., constant radius). With these assumptions, we can find two circles tangent to the lines passing through the two chain ends, each passing through one of the chain end-points. Based on triangle similarity theorems, the radii of the two circles are proportional to the distances of the chain ends from the intersection point. Figure~\ref{fig:doo-detection:merge-cases} illustrates this idea.

Let us define the end-points we desire to connect on chains $C_1$ and $C_2$ as $e_1$ and $e_2$, respectively. We call the lines passing through $e_1$ and $e_2$ in the direction of $C_1$ and $C_2$ ends as $l_1$ and $l_2$, respectively. Finally, we define the circles passing through $e_1$ and $e_2$ as $c_1$ and $c_2$, and the points they touch on the other line as $t_1$ and $t_2$, respectively. Note that points $e_1$ and $t_2$ will be lying on line $l_1$, while points $e_2$ and $t_1$ are on line $l_2$.

Without loss of generality, let us assume that in Figure~\ref{fig:doo-detection:merge-cases}, circle $c_1$ is the red circle, the blue circle is $c_2$, the red dot is $t_1$, the blue dot is $t_2$, the red arrow's end point (arrow side) is $e_1$, the blue arrow's end point is $e_2$, the line passing $e_1$ is $l_1$ and the line passing $e_2$ is $l_2$.

It can be proven that the distance between $e_2$ and $t_1$ is equal to the distance between $e_1$ and $t_2$. However, each $t_1$ and $t_2$ can be lying on lines $l_2$ and $l_1$ ahead or behind $e_2$ and $e_1$, creating three different situations:

\begin{itemize}[leftmargin=*]
    \item Either $t_1$ is ahead of $e_2$ or $t_2$ is ahead of $e_1$, but not both (Figure~\ref{fig:doo-detection:merge-cases-1}). Not surprisingly, a majority of connections in a typical application would be of this type.
    In this case the blue circle $c_2$ that touches $l_1$ at point $t_2$ behind $e_1$ is discarded and we use the radius of the red circle $c_1$ for the turn radius of the new chain $C_{new}$. This chain will be composed of the arc of $c_1$ from $e_1$ to $t_1$ and the line from $t_1$ to $e_2$. Figure~\ref{fig:doo-detection:merge-fill-1} shows this scenario's solution.

    \begin{figure}[!htb]
    \centering
        \begin{subfigure}[b]{0.23\linewidth}
            \includegraphics[width=\textwidth]{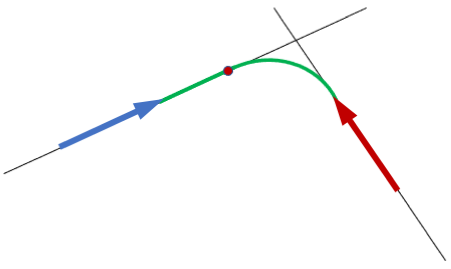}
            \caption{~}
            \label{fig:doo-detection:merge-fill-1}
        \end{subfigure}
        \hfill
        \begin{subfigure}[b]{0.23\linewidth}
            \includegraphics[width=\textwidth]{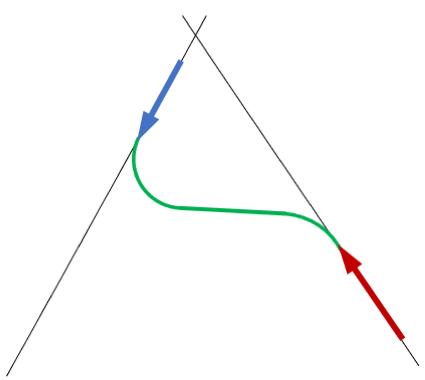}
            \caption{~}
            \label{fig:doo-detection:merge-fill-2}
        \end{subfigure}
        \hfill
        \begin{subfigure}[b]{0.46\linewidth}
            \includegraphics[width=\textwidth]{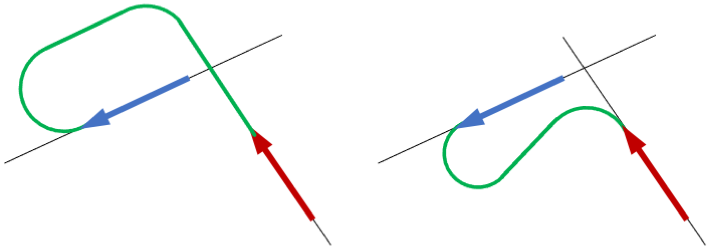}
            \caption{~}
            \label{fig:doo-detection:merge-fill-0}
        \end{subfigure}
    \caption[Illustration of DOO merging suggested solutions]{Suggested solutions for different merging cases illustrated in Figure~\ref{fig:doo-detection:merge-cases}.}
    \label{fig:doo-detection:merge-fill}
    \end{figure}
    
    \item Both $t_1$ and $t_2$ are ahead of $e_2$ and $e_1$ (Figure~\ref{fig:doo-detection:merge-cases-1}). 
    In this case, the new chain $C_{new}$ is composed of an arc on each end ($e_1$ and $e_2$) and a line tangent to the arcs. The turn radius is as desired or can be experimentally determined for the DOO, and it should be large enough to allow the "natural-looking turn." However, the radius should be small enough so that the direction of the line tangent to the two arcs is close to the direction of the line connecting $e_1$ to $e_2$. Finally, we suggest the same turn radius for both ends. Figure~\ref{fig:doo-detection:merge-fill-2} shows this scenario's solution.

    \item Both $t_1$ and $t_2$ are behind of $e_2$ and $e_1$ (Figure~\ref{fig:doo-detection:merge-cases-0}). This case has two suggested solutions that depend on the conditions.
    In both solutions, similar to the previous case, the new chain $C_{new}$ is composed of an arc on each end ($e_1$ and $e_2$) and a line tangent to the arcs. The turn radius is as desired or can be experimentally determined for the DOO. However, depending on external conditions, $C_{new}$ can fill the gap from outside or inside the region between the two chains. Figure~\ref{fig:doo-detection:merge-fill-0} shows this scenario's solutions.

\end{itemize}

Note that, in all scenarios, there can be infinite correct solutions, which depend on the conditions. In practice, the suggested solutions result in good fits with the ground truth and can be used in most conditions without any modification.

\begin{figure}[!htb]
    \centering
    \includegraphics[width=0.6\linewidth]{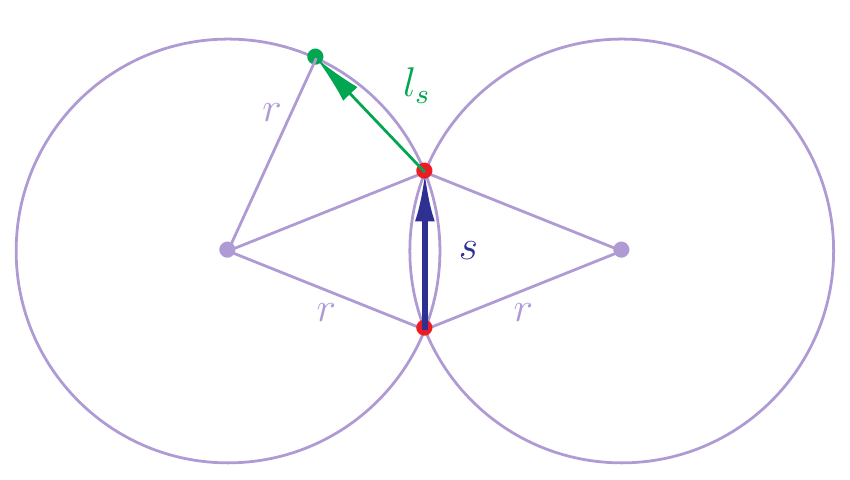}
    \caption[Adding a new DOO segment with a desired turn radius]{Adding a new DOO segment with length $l_s$ with a desired constant turn radius of $r$ to the end of segment $s$.}
    \label{fig:doo-detection:enforce-turn}
\end{figure}

Once the new chain $C_{new}$ is obtained to fill the gap between the two chain ends $C_1$ and $C_2$, it can be added to the ends of the chains to connect them. $C_{new}$ is a combination of constant-radius arcs and lines. Adding a DOO segment for the line sections is trivial and will not be explained. To add a segment that follows the desired turn, we use the last segment $s$ on the chain. Knowing the start and end-points of this segment $s$, we can calculate two circles with the desired radius that pass through these points. Knowing the direction of the turn, one circle is eliminated, and the new point on the remaining circle at the segment distance $l_s$ of the segment's end-point is used to create the new segment $s'$ that is added to the end of the chain. Figure~\ref{fig:doo-detection:enforce-turn} shows how the new segment can be added with the desired turn radius.

\subsection{Notes on the Proposed Method}

The merging process continues connecting the chains, two at a time until all the chains are merged into one single chain, which is the detected DOO represented by the chain of fixed-length cylindrical segments connected by passive spherical joints. This representation can be used as the input to routing and manipulation systems for the desired application. 

Each segmented image will be processed into a single DOO. When there are multiple DOOs present in the camera frame, the easiest way to detect them as separate DOOs is to have separate segmentations for them. For example, if there are multiple cables with different colors in the image, a color-based segmentation can have two segmented images. In case there are multiple DOOs that cannot be segmented separately, the proposed algorithm can still help process them into separate DOO outputs. Note that the algorithm is greedy in the sense that it first goes for the best-matched chains. With multiple DOOs, there is a high chance that the chains related to separate DOOs do not give a good fit. As a result, for example, when there are only two chains left, there is a high chance that the two chains are the two separate DOOs. Therefore, it is enough to stop the merging process when the desired number of chains is left.

The proposed method is general and can be used with both 2-D and 3-D image data to provide the deformable object's representation in 2-D or 3-D.

Finally, the final output of the proposed method is a single chain of segments that does not keep track of the parts seen in the frame vs. the occluded parts. There are two ways to mark the parts of the chain that are related to the occlusions: 

\begin{enumerate}[leftmargin=*]
    \item Map the segmented parts on the final detection to determine the occluded parts of the DOO chain.
    
    \item During the process, when merging two chains, if the gap is equal to or longer than the segment length $l_s$, the newly added chain $C_{new}$ is marked as occluded. The reason for skipping smaller gaps is that many gaps shorter than $l_s$ are created during the pruning process. 
\end{enumerate}

Both approaches ultimately depend on the accuracy of the segmentation. The first approach is simple but may mislabel an occluded part as visible in a multilayer setup. On the other hand, the second approach tends to be more accurate in multilayer settings but may skip more minor occlusions.

\section{Experiments and Results} \label{sec:doo-detection:tests}

The proposed was implemented for 2-D images in Python 3. We used the color-based segmentation of the DOO region. This approach generally tends to include extra areas around the DOO and other regions with similar color hues to the DOO. We chose conservative thresholds to exclude any non-DOO regions. This results in some DOO data being excluded; However, our experiments have shown the DOO detection to have challenges when extra regions are included but to work when some data is lost in segmentation. The same principle is advised for other segmentation methods choices, and those methods' parameters should be chosen conservatively to remove the irrelevant regions.

We used a well-known morphological thinning method for skeletonization~\cite{1164959}. The algorithm proposed by Suzuki and Abe~\cite{suzuki1985topological} and provided in the OpenCV library is used to extract contours. All our tests use $w_e = 1$, $w_d = 100$ and $w_c = 100$ values for the cost function of Equation~\ref{eq:doo-detection:cost-merge-weighted-sum} and the segment length $l_s$ is chosen as 10 pixels. The weights are chosen manually to focus on shorter distances while heavily discouraging non-matching segment directions and excessive bending. Different weights result in some types of incorrect connections increasing while the number of other types decreases. A better set can be found using a more methodical approach and optimization for the desired applications.

Similarly, the segment length was not chosen optimally. In general, a shorter segment length can capture the contour ends better and create segments from smaller contours, leading to better curves in filling gaps; however, it increases the number of segments and the average number of incorrect connections. On the other hand, a longer segment length has the potential of not following the curves well and ignoring small contours. However, it tends to reduce the number of incorrect connections and improve the detection speed by reducing the number of segments.

\begin{figure}[!htb]
\centering
    \begin{subfigure}[b]{\linewidth}
        \includegraphics[width=0.48\textwidth]{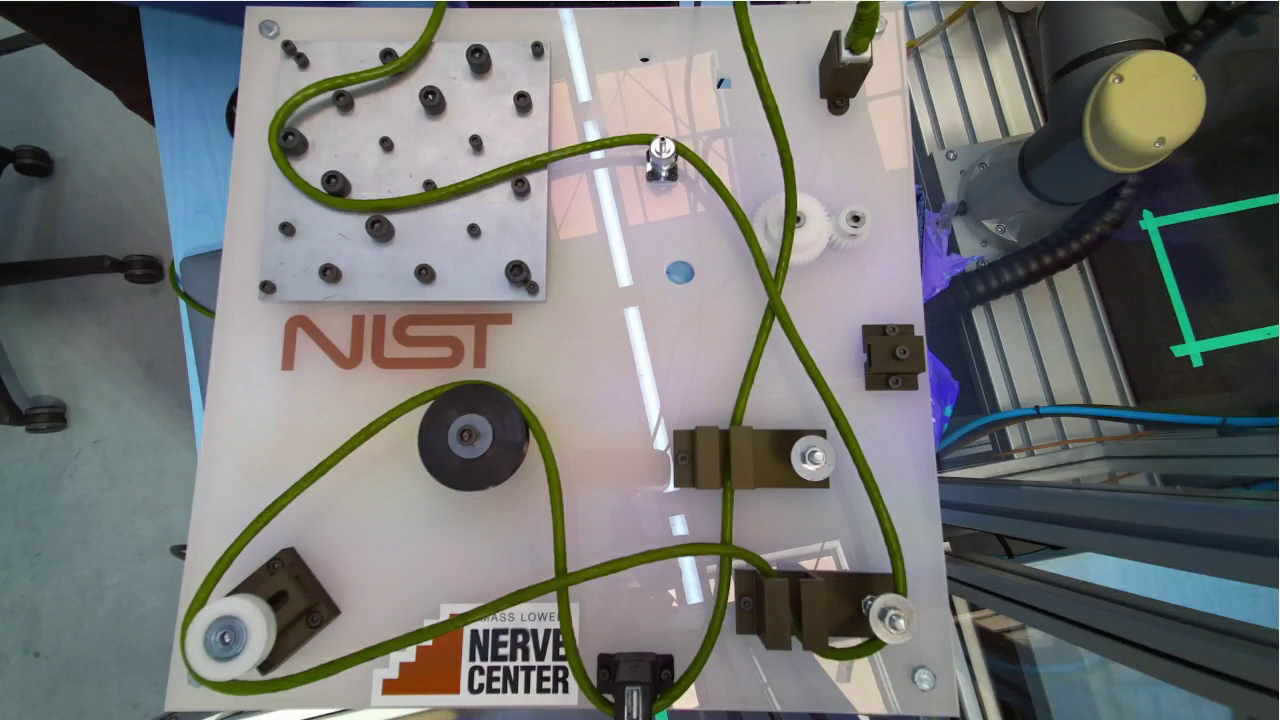}
        \hfill
        \includegraphics[width=0.48\textwidth]{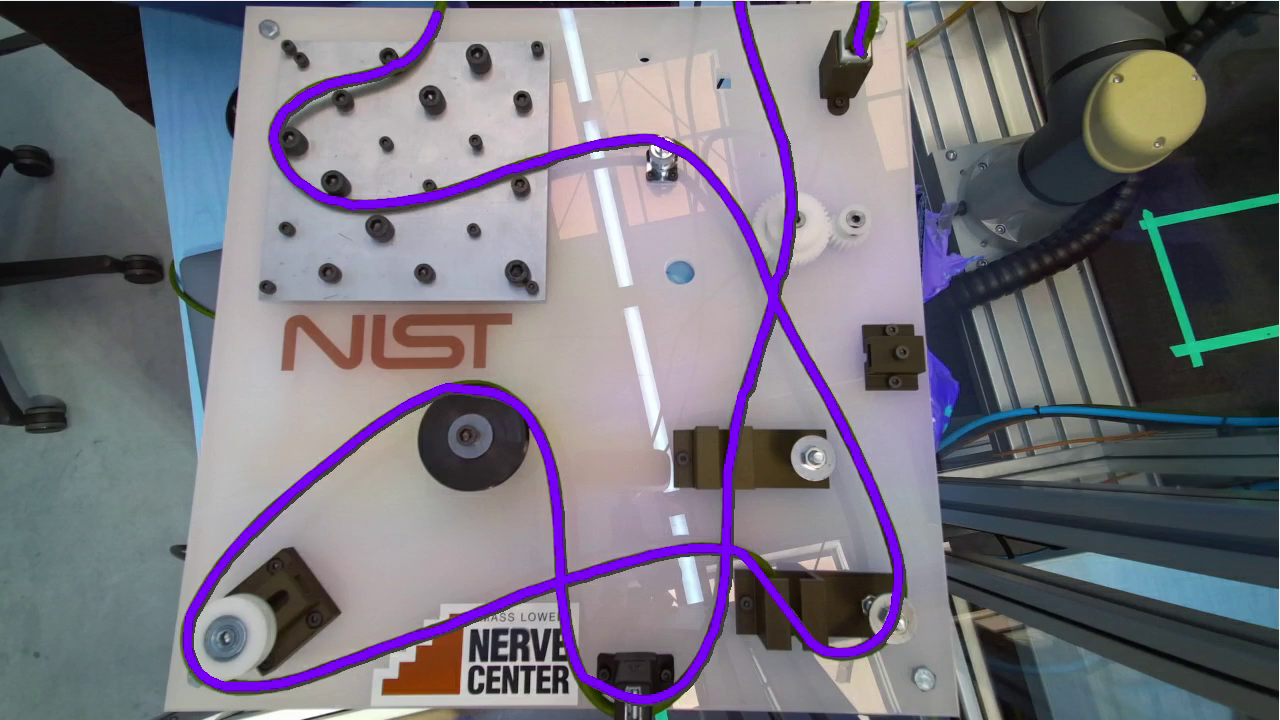}
        \caption{~}
    \end{subfigure}

    \medskip
    \begin{subfigure}[b]{\linewidth}
        \includegraphics[width=0.48\textwidth]{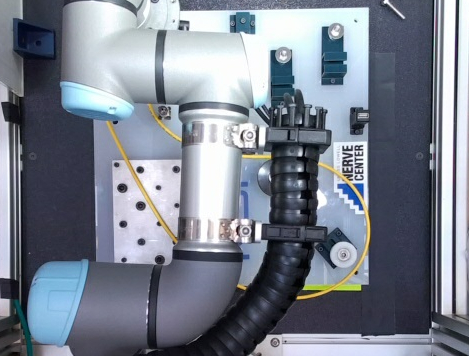}
        \hfill
        \includegraphics[width=0.48\textwidth]{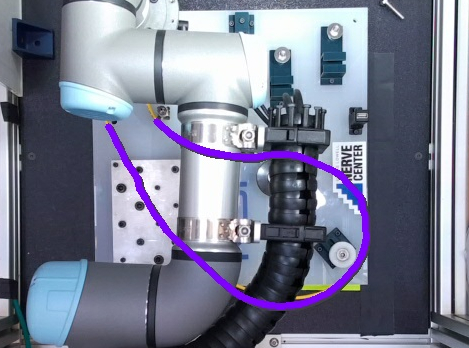}
        \caption{~}
    \end{subfigure}

    \medskip
    \begin{subfigure}[b]{0.48\linewidth}
        \includegraphics[width=0.48\textwidth]{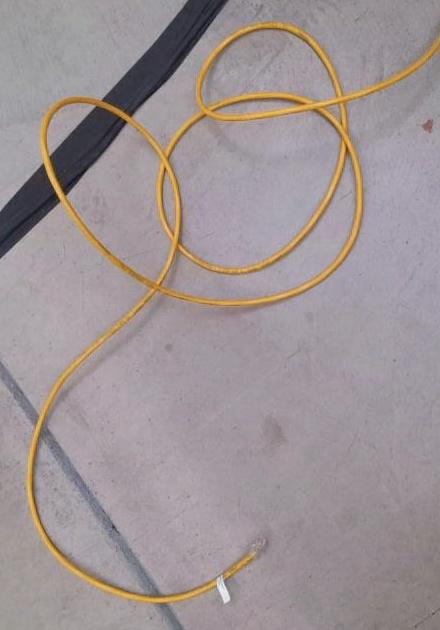}
        \hfill
        \includegraphics[width=0.48\textwidth]{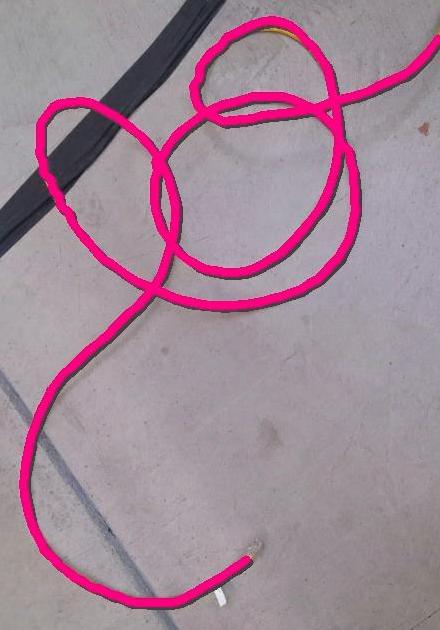}
        \caption{~}
    \end{subfigure}
    \hfill
    \begin{subfigure}[b]{0.48\linewidth}
        \includegraphics[width=0.48\textwidth]{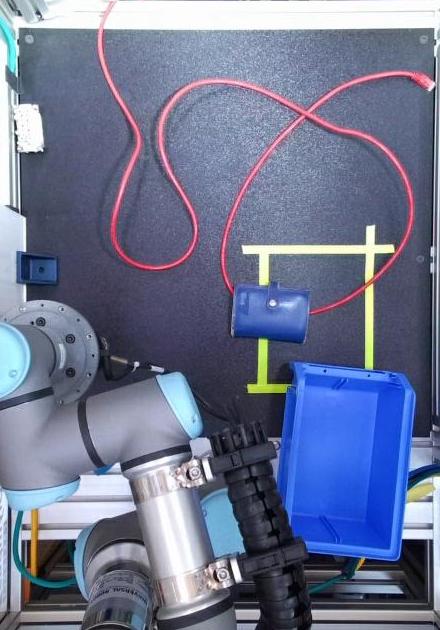}
        \hfill
        \includegraphics[width=0.48\textwidth]{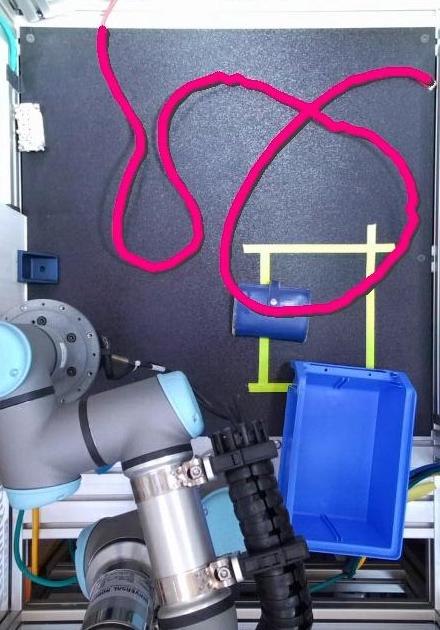}
        \caption{~}
    \end{subfigure}
    \caption[DOO detection results with crossings and occlusions]{The result of the proposed detection method on example inputs with crossings and occlusions. The detected cable (purple and magenta) overlaid on the frames on the right.}
\label{fig:doo-detection:results-1}
\end{figure}

Figure~\ref{fig:doo-detection:results-1} shows results of the detection on inputs with heavy occlusion and several crossings.

We have used the method for cable routing and manipulation tasks~\cite{Keipour:2022:iros:routing}. The viability is tested on 7~video sequences with a total of 4,230~frames of size 1280$\times$720. Table~\ref{tbl:doo-detection:results} shows the quantitative results for the algorithm's accuracy on the whole cable in an image, for the occlusions filled, and for the merges performed. Mengyuan et al.~\cite{8972568} have used the root mean square of the Euclidean distance between their estimated and the ground-truth point positions on the DOO, which they reported as around 23~mm. Note that due to the lack of ground truth for the occluded areas and to focus on testing the key contributions of our proposed approach, we used a stricter measure that even when a single connection is incorrect, we counted the frame as incorrect detection. The occlusions are counted as incorrect when either a wrong connection is made or when the filled connection does not follow the actual cable's path. Finally, we noticed that the incorrect merges only rarely happen in places other than at occlusions, with only 148 cases, almost all of which happened at self-crossings. 

\begin{table}[!htb]
\centering
\caption[DOO detection results]{Detection results on 7 video sequences.}
\label{tbl:doo-detection:results}
\begin{tabular}{|c|c|c|c|c|c|c|c|c|c|}
\hline
\rowcolor[HTML]{EFEFEF} 
~& Total & Correct & Incorrect & Accuracy \\ \hline
Frames         &  4,230 & 3,542 & 688 & 83.7\% \\ \hline
Occlusions & 26,456 & 23,991 & 2,465 & 90.7\% \\ \hline
Merges & 583,743 & 581,130 & 2,613 & 99.6\%\\ \hline
\end{tabular}
\end{table}

\begin{figure}[!htb]
\centering
    \begin{subfigure}[b]{\linewidth}
        \includegraphics[width=0.32\textwidth]{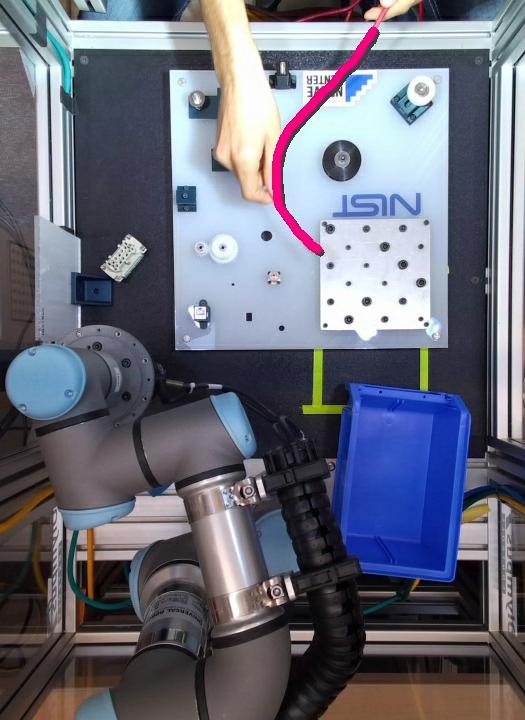}
        \hfill
        \includegraphics[width=0.32\textwidth]{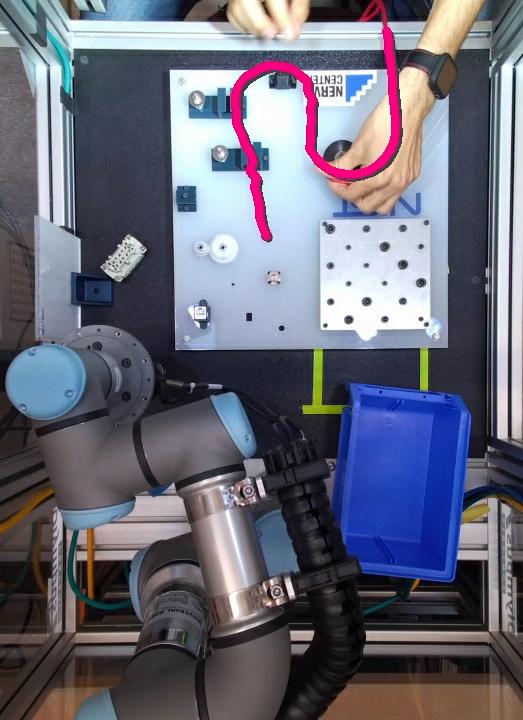}
        \hfill
        \includegraphics[width=0.32\textwidth]{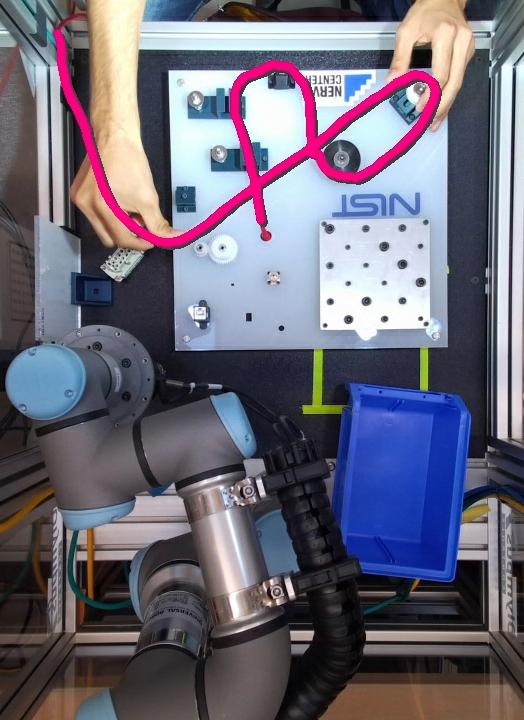}
        \caption{~}
    \end{subfigure}

    \medskip
    \begin{subfigure}[b]{\linewidth}
        \includegraphics[width=0.32\textwidth]{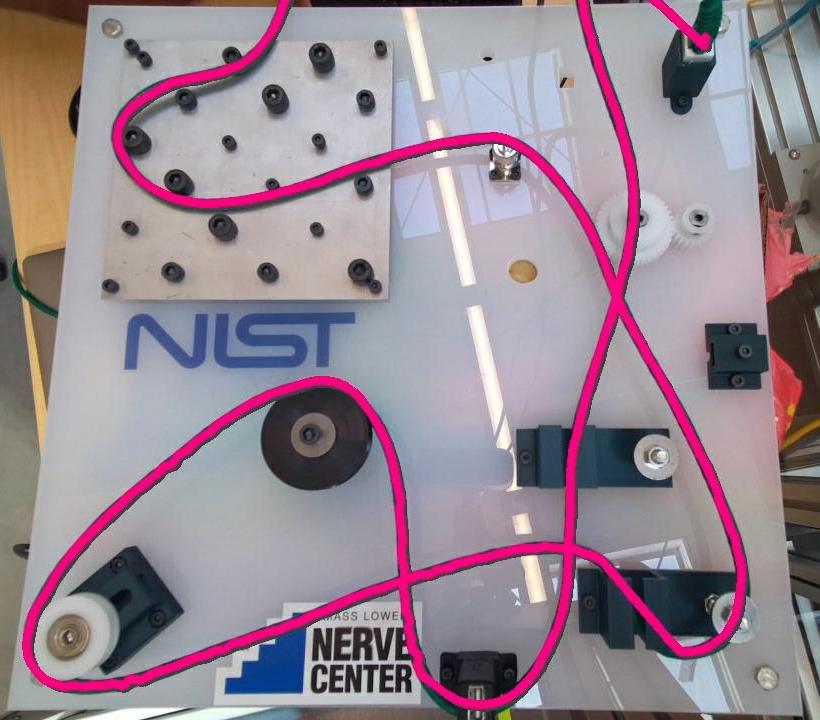}
        \hfill
        \includegraphics[width=0.32\textwidth]{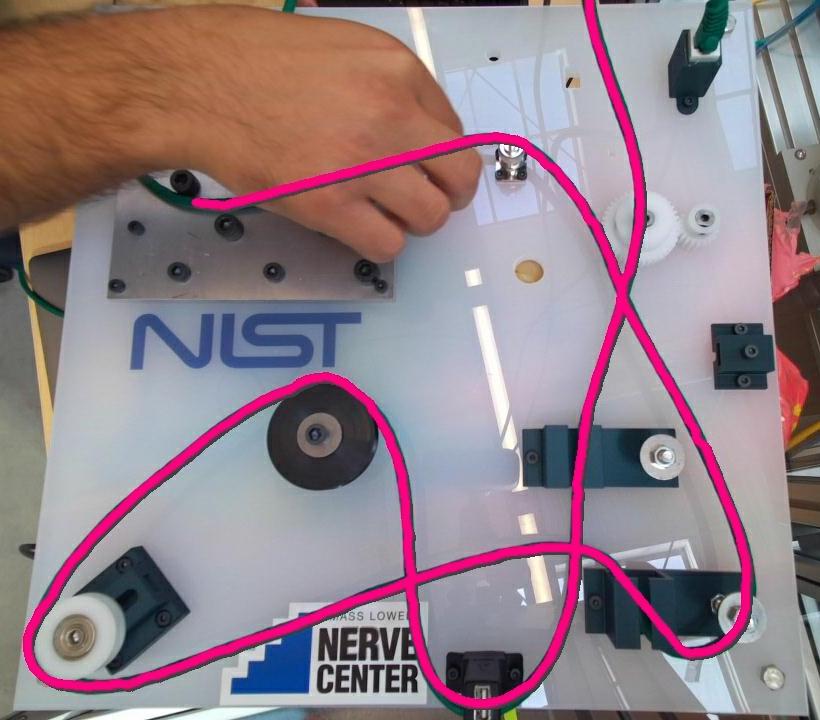}
        \hfill
        \includegraphics[width=0.32\textwidth]{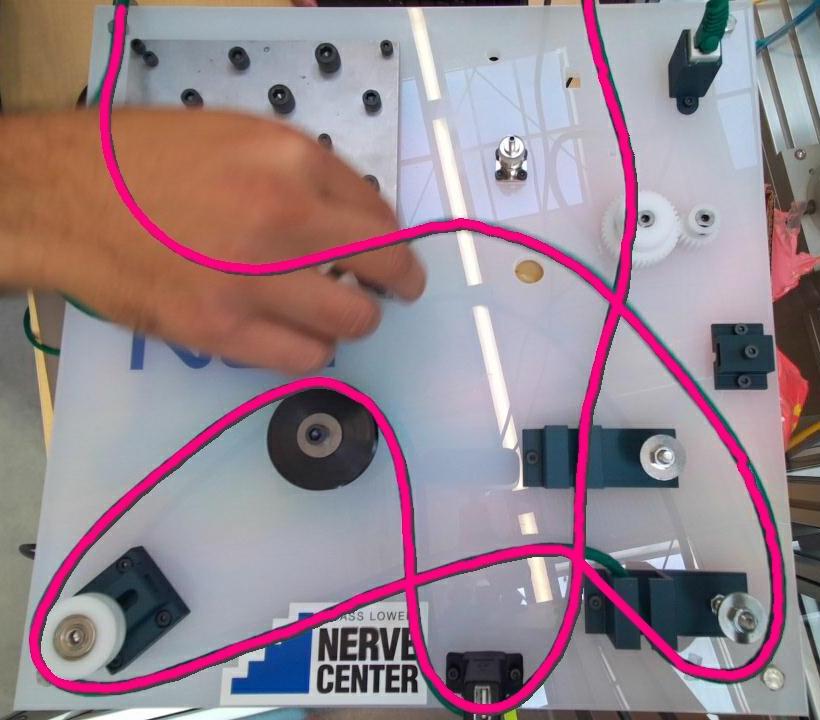}
        \caption{~}
    \end{subfigure}
\caption[DOO detection results on sample video sequences]{Screenshots of example sequences with the overlaid detected cable (in magenta). The third row includes the original frame for comparison.}
\label{fig:doo-detection:results-3}
\end{figure}

Our method's average detection time per frame across all the sequences is 0.537 seconds on a system with Intel® Core™ i9-10885H CPU and 64 GB DDR4 RAM. Figure~\ref{fig:doo-detection:results-3} shows snapshots of some video sequences and the detection results.

\section{Conclusion and Discussion} \label{sec:doo-detection:conclusion}

We presented a novel method for detecting deformable one-dimensional objects (e.g., wires and cables) for robotics applications and showed the results. Our implementation is only 2-D and not tuned towards a specific condition. Choices other than the weighted sum for the total cost function were not researched, and our selection of weights was not made optimally. Nevertheless, the results show promise with an almost 2~Hz detection rate on an HD image input, which is enough to initialize the DOO trackers for inspection and maintenance with UAVs. A more optimized implementation can take advantage of special data structures and parallelization to increase the method's speed by several orders of magnitude.

Furthermore, the proposed method is very general and flexible and can be tuned for specific 2-D and 3-D applications to provide near-perfect results in other settings as well, including surgical and industrial robots. 

The cable segmentation method proposed by Li~\cite{Li1415992} provides a cost function for choosing the best two chains for merging. While we developed our cost function elements independently, the overall cost functions between our work and Li have similar structures, only differing in details. However, the Li method only uses merging for neighbor image patches, and the intention is not to fill the occlusions and gaps but is pure segmentation. We go further by providing different merging solutions for different conditions to deal with occlusions and imperfect segmentation. Our final output is a DOO representation suitable for manipulation instead of the segmentation mask provided by the Li method. 

Note that it is not hard to find unstructured or adversarial situations with entanglements, occlusions, multiple close and parallel DOOs, and other complex scenarios that can easily confuse the proposed algorithm. This work is the first effort to solve the DOO detection problem and was aimed to provide a method that can assist in semi-structured situations rather than addressing those "crazy" scenarios.

The considerations for the 3-D case are provided for each step. However, we did not implement the 3-D case, and there may be unpredicted implementation challenges. In the future, the ideas of the method can be integrated with tracking methods to improve tracking accuracy. Its integration in a robotics pipeline can eventually enable full autonomy in real-world robotics applications working with DOOs such as cables, surgical sutures, and ropes and is a step toward realizing aerial manipulation of these objects.

The following chapter introduces our routing solution for the manipulation of DOOs and explores how aerial manipulators can have physical interaction with the detected DOOs.

\chapter{Deformable One-Dimensional Object Routing and Manipulation} \label{ch:doo-routing}

With the field of rigid-body robotics having matured in the last fifty years, routing, planning, and manipulation of deformable objects have recently emerged as a more untouched research area in aerial robotics and many other fields ranging from surgical robotics to industrial assembly and construction. 

Routing approaches for deformable objects which rely on learned implicit spatial representations (e.g., Learning-from-Demonstration methods) make them vulnerable to changes in the environment and the specific setup. On the other hand, algorithms that entirely separate the spatial representation of the deformable object from the routing and manipulation (often by using a general representation approach independent of planning) result in slow planning in high-dimensional spaces.

This chapter proposes a novel approach to routing deformable one-dimensional objects (e.g., wires, cables, ropes, sutures, threads)~\cite{Keipour:2022:iros:routing}. This approach utilizes a compact representation for the object, allowing efficient and fast online routing. The spatial representation is based on the geometrical decomposition of the space into convex subspaces, resulting in a discrete coding of the deformable object configuration as a sequence. With such a configuration, the routing problem can be solved using a fast dynamic programming sequence matching method that calculates the next routing move. The proposed method couples the routing and efficient configuration for improved planning time. Our simulation and real experiments show the method correctly computing the next manipulation action in sub-millisecond time and accomplishing various routing and manipulation tasks.

Then, analyze the requirements of aerial manipulation of deformable one-dimensional objects~\cite{Keipour:2022:icra-workshop:doo}. We study the feasibility of the physical interaction of UAVs with these objects from the perspective of their end-effector precision and their available wrenches for an example wire manipulation task.

\section{Introduction and Related Work} \label{sec:doo-routing:intro}

Objects such as wires, cables, ropes, threads, and surgical sutures can be found in many industrial, surgical, construction, and everyday settings. In the literature, they are commonly called Deformable One-dimensional Objects (DOOs) or Deformable Linear Objects (DLOs). Automation of tasks involving rigid bodies had been extensively studied in robotics; however, the need for further automation of manual tasks is forcing robotics applications to move towards working with DOOs, raising research interest in aerial as well as other robotics areas~\cite{frobt.2020.00082, 0278364918779698}. 

A crucial part of many robotics applications involving DOOs is routing~\cite{0278364918779698, GUO2020158, 1236303}. \textit{Routing}, also known as \textit{route planning}, finds a viable \textit{path} to change the initial state of a DOO to a goal state. Path in this context depends on the application and may mean a path in Euclidean space, a trajectory in the configuration space of the DOO, or a series of actions performed on the DOO.

Several representation methods have been used to capture the state of a DOO in route planning and manipulation problems. These representations range from spring-mass models~\cite{LV2017385} to linear combination of curves~\cite{1668249, Hirai2000}, and fixed-length segments~\cite{Keipour:2022:ral:doo}.

Planning methods for routing DOOs for manipulation can be categorized into computational algorithms and approaches based on learning. 

Computational routing methods are generally sampling-based approaches. These methods sample the space and use search methods such as Probabilistic Roadmaps (PRM) and Rapidly-exploring Random Trees (RRTs) to find the route from the initial DOO state to the final state. Guo et~al.~\cite{GUO2020158} propose the RRT-BwC (Bi-direction with Constrain) planning algorithm to plan for aircraft cable assembly in narrow cabins with obstacles. Their method is based on the geometric formulation of the objects and the bi-directional RRT search to route in the high dimensional planning space. Amato et~al.~\cite{1236303} find an approximate route by pre-computing a global roadmap using a variant of PRM, then refine the route by constrained sampling and applying adaptive forward dynamics. Moll and Kavraki~\cite{1668249} propose DOO planning using minimal-energy curves and a sampling-based planning method such as PRM. Roussel et~al.~\cite{7139627} use quasi-static and dynamic models coupled with sampling-based methods to plan for an elastic rod manipulation. Koo et~al.~\cite{2011jrm} and Ma et~al.~\cite{ma_liu_ding_lv_2020} apply RRT search-based approaches for routing and manipulation of DOOs. For all these methods, the solution can be statistically guaranteed, but they suffer from the curse of dimensionality, and in complex routing cases, the time and space complexity of the algorithms may make the algorithms slow and infeasible for many practical scenarios.

On the other hand, the learning-based methods primarily consist of learning-from-demonstration approaches, potentially working with any DOO type for any task. However, they are not generalizable and quickly fail when the experiment setup conditions even slightly deviate from the learning data~\cite{6386002, 8256194, wang2019learning}.

We propose a novel approach to solving the DOO routing problem that is suitable for both offline and online routing due to its efficiency and speed~\cite{Keipour:2022:iros:routing}. This approach relies on the geometrical decomposition of the task space into convex regions. It uses this discretized space to describe the DOO's configuration using a compact sequence, which is simplified from the original 3-D continuous space description. Unlike the existing routing methods that have to find a solution by exploring a high-dimensional space, our new spatial representation allows utilizing a high-speed dynamic programming sequence matching method that reduces the planning delay to near zero, making it suitable for online planning for routing and manipulation tasks.

As far as we know, there have not been any studies on the requirements for physical interaction and manipulation of deformable objects using aerial robots. In this chapter, we perform a basic analysis of the requirements of aerial manipulation of deformable one-dimensional objects~\cite{Keipour:2022:icra-workshop:doo}. We study the feasibility of the physical interaction of UAVs with these objects from the perspective of precision of their end-effectors and their available wrenches for a wire manipulation task. 

\section{Problem Definition} \label{sec:doo-routing:problem}

\begin{figure}[!htb]
    \centering
    \includegraphics[width=0.4\linewidth]{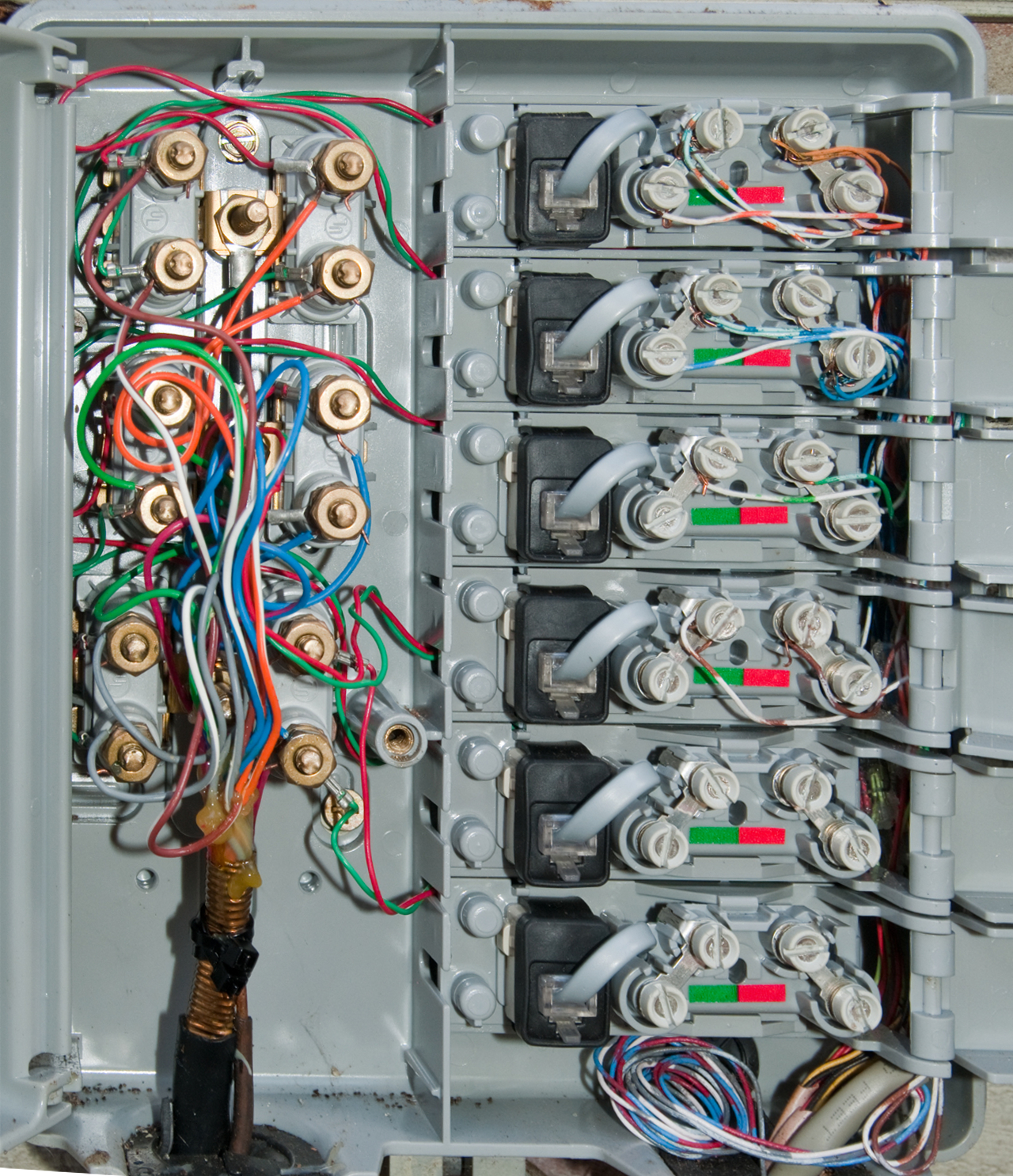}
    \caption[An example of a cable box for wire routing]{An example demarcation box or Network Interface Device (NID)~\cite{cableboxphoto}.}
    \label{fig:doo-routing:cable-box-example}
\end{figure}

In the rest of this chapter, we refer to the area where the task is taking place as \textit{work region}. We assume that the exact positioning of a deformable one-dimensional object is only important inside the work region, and the details of its positioning outside this region are ignored. Additionally, we presume that the work region falls within the workspace of the aerial manipulator, and the robot can access all of the work regions. Moreover, we assume that the work region is a "free space" with different \textit{components} occupying some of its space. All the components' positions, shapes, and dimensions are presumed to be known either beforehand or through processing the perception input. Figure~\ref{fig:doo-routing:cable-box-example} illustrates an example utility box that satisfies all the mentioned assumptions.

The exact positioning of a DOO in the work region is called its \textit{state}, which contains the exact places in space occupied by a DOO. 

Given the initial and desired states of a DOO in a work region, \textit{routing} is finding the sequence of actions required to perform on the DOO to change its state from the initial state to the final state.

The actual DOO state is continuous and needs to be converted to a discrete spatial representation for robotics and computer applications. Common different ways of such representation include sampling equidistant points along the DOO's medial axis, fixed-length B-splines, and fixed-length cylinders connected by spherical joints~\cite{GUO2020158, Keipour:2022:ral:doo}. 

We propose to use a more efficient spatial representation based on the convex decomposition of the work region, combined with a fast sequence matching algorithm to solve the routing problem. Our proposed method is completely independent of the DOO dynamics and tries to embed the dynamics effects in the state representation. We relax the problem assuming that the \textit{slack} of the DOO is not important. In other words, we assume that the application working with the DOO is not affected if the DOO has some extra slack in any area of the work region. For example, suppose a cable is not laying straight in a region and has a rather significant bending in a region. In that case, the extra bending is considered slack, and our algorithm does not consider it. We define the slack as extra curves in DOO that do not pass around any components or anchor points; the slack can virtually be eliminated by creating tension in the DOO. 

The following sections describe our proposed spatial representation and the routing method for DOOs.

\section{Spatial Representation} \label{sec:doo-routing:representation}

Let us introduce a graph $G_s$ (called a \textit{spatial representation graph}) to model the spatial representation of a deformable one-dimensional object passing through a work region. 

\begin{figure}[!htb]
\centering
    \begin{subfigure}[b]{0.48\textwidth}
        \includegraphics[width=\textwidth]{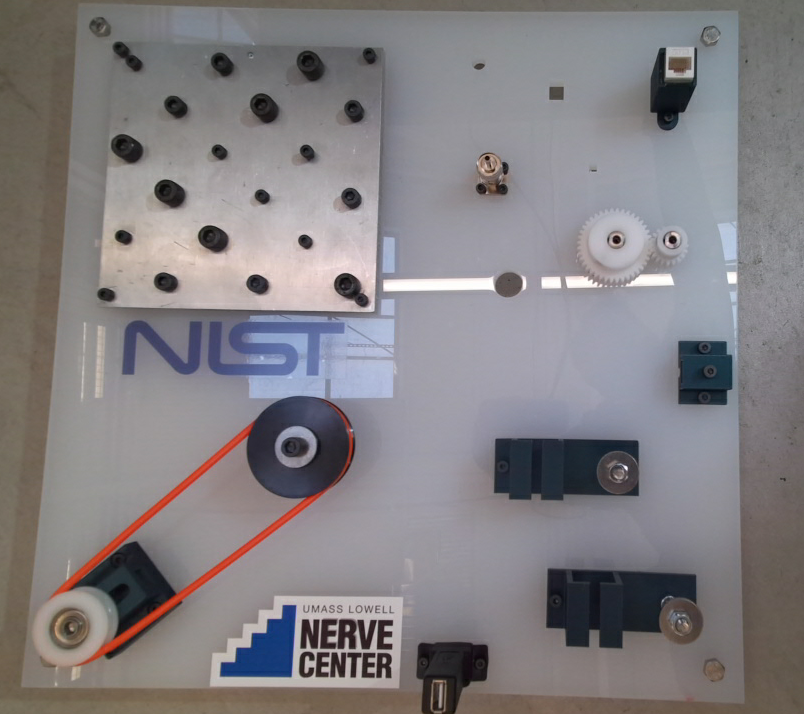}
        \caption{~}
        \label{fig:doo-routing:original-board}
    \end{subfigure}
    ~    
    \begin{subfigure}[b]{0.48\textwidth}
        \includegraphics[width=\textwidth]{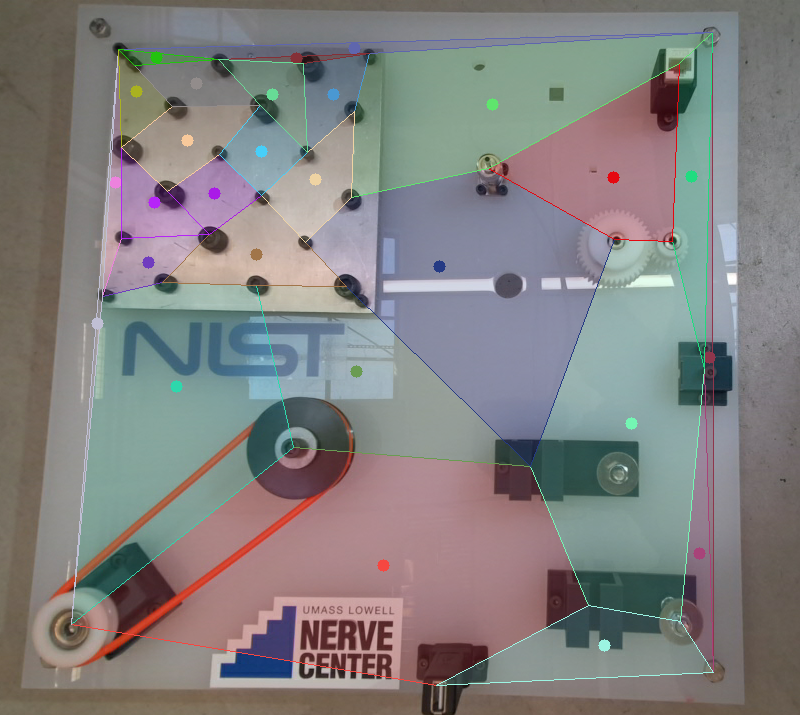}
        \caption{~}
        \label{fig:doo-routing:convex-decomposition}
    \end{subfigure}
    \caption[Circuit board convex decomposition result]{Convex decomposition of an example circuit board. (a) The original board. (b) The result of convex decomposition. Each component on the board is used for defining the convex region vertices. The centroid of each convex region is marked with a dot.}
    \label{fig:doo-routing:board-regions}
\end{figure}

The work region should be decomposed into convex subspaces to generate the vertices $V_s$ of the spatial representation graph $G_s$. These subspaces are called \textit{convex polygons} in 2-D and \textit{convex polytopes} in 3-D spaces. Each of these subspaces is a vertex in the graph $G_s$. If the work region is not enclosed (i.e., if a portion of DOO can lie outside the work region), a new vertex is added to $V_s$ to represent the "outside" region. There are many exact and approximate approaches for convex decomposition, and each can be used for this work~\cite{Deng_2020_CVPR, LIEN2006100, 1236265, 0221025}. Figure~\ref{fig:doo-routing:board-regions} illustrates the convex decomposition on an example circuit board. Each of the components on the board is a node of the convex polygons generated from the board's layout.

\begin{figure}[!htb]
    \centering
    \includegraphics[width=0.5\linewidth]{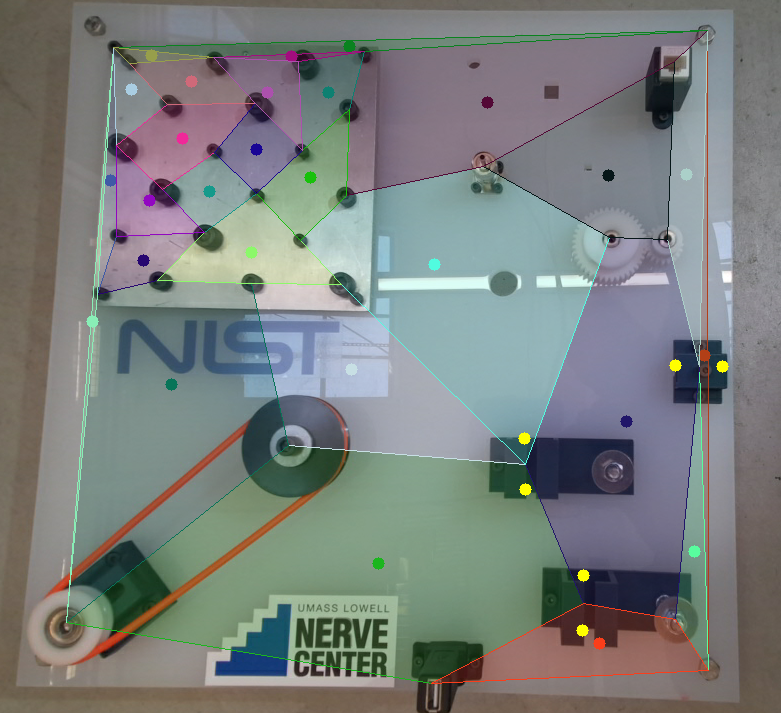}
    \caption[Spatial representation graph vertices of the circuit board]{The vertices $V_s$ of the spatial representation graph $G_s$ constructed from the example board in Figure~\ref{fig:doo-routing:original-board}. The tunnel entrance vertices are depicted by yellow dots. Note that the outside region vertex is omitted in the illustration.}
    \label{fig:doo-routing:spatial-graph-vertices}
\end{figure}

The generated convex regions allow efficiently defining subspaces in both 2-D and 3-D. It is desirable to represent the subspace only in 2-D when possible for simplicity. Much of the workspace in the finish line of industrial robotics is on a tabletop which can be approximated as a 2.5-D space. Meaning two dimensions are far more significant than the third dimension. There are specific scenarios where a 3-D work region can be simplified as a 2-D region with additional 3-D "tunnel"-like components such as bridges, passes, and tunnels. To allow the 2-D representation for these work regions, we can add a vertex to $V_s$ for each of the entrances of these components. Figure~\ref{fig:doo-routing:spatial-graph-vertices} shows all the vertices constructed from the example board of Figure~\ref{fig:doo-routing:original-board} with the yellow dots representing the vertices of the entrances of the tunnel components.  

Once all the vertices $V_s$ are defined, the edges $E_s$ for the spatial representation graph $G_s$ can be computed using the following rules:

\begin{itemize}[leftmargin=*]
    \item Vertices from the neighbor convex regions (convex regions sharing a side) are connected with an edge.
    
    \item Convex regions with a side not shared with any other convex region (i.e., convex regions surrounding the work region) are connected to the outside vertex.
    
    \item Vertices for entrances of a tunnel component are connected to each other.
    
    \item Each vertex for the tunnel entrances is connected to the vertex of the convex (or outside) region that it is lying on.
\end{itemize}

\begin{figure}[!htb]
    \centering
    \includegraphics[width=0.5\linewidth]{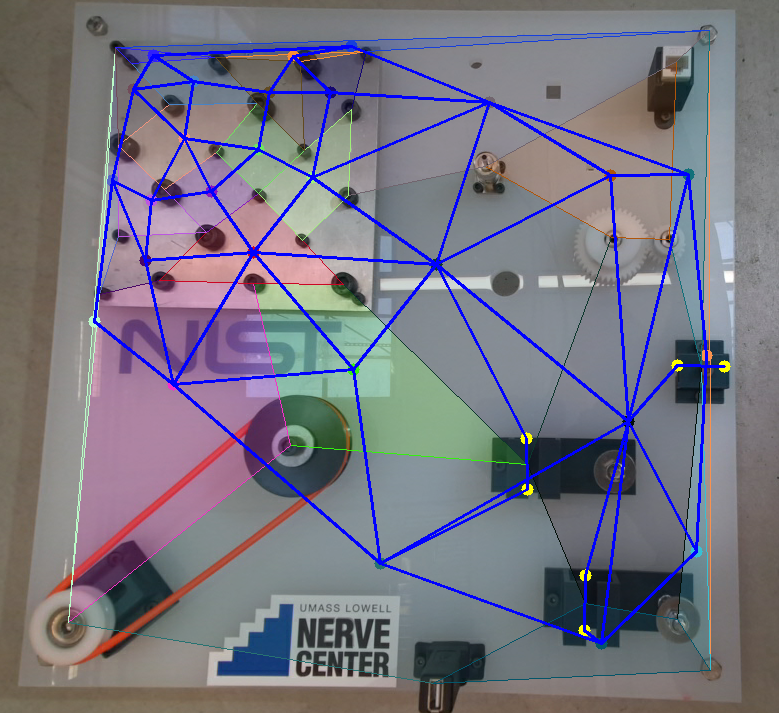}
    \caption[Spatial representation graph for the circuit board]{The spatial representation graph $G_s$ computed from the example board of Figure~\ref{fig:doo-routing:original-board}. For simplicity, the edges connected to the outside vertex are not depicted here.}
    \label{fig:doo-routing:spatial-graph}
\end{figure}

Figure~\ref{fig:doo-routing:spatial-graph} shows the spatial representation graph $G_s$ constructed for the circuit board of Figure~\ref{fig:doo-routing:original-board}.

Without the loss of generality, we assume that the size of $V_s$ is $n + 1$, with the vertices numbered from $-1$ to $n - 1$, and $-1$ reserved for the outside vertex. 

\begin{figure}[!htb]
    \centering
    \includegraphics[width=0.5\linewidth]{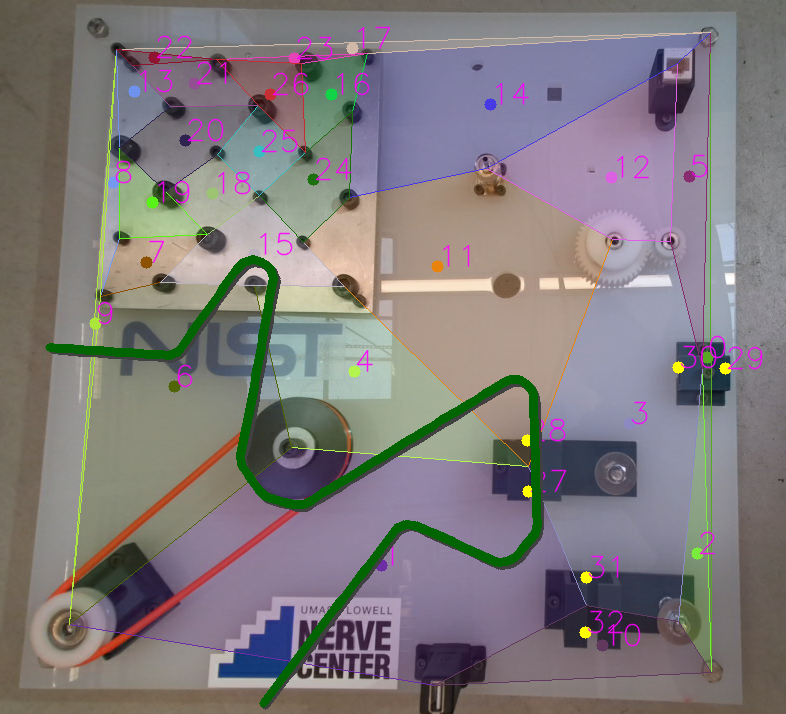}
    \caption[DOO passing through spatial representation graph vertices]{An example DOO passing through the spatial representation graph vertices of Figure~\ref{fig:doo-routing:spatial-graph}.}
    \label{fig:doo-routing:cable-spatial-representation}
\end{figure}

Having computed the graph $G_s$, a DOO lying in the work region or passing through it can be represented by an ordered sequence $C$ of the vertex numbers it is passing through. We call this sequence representing the DOO as \textit{configuration} of DOO. Note that if the DOO is bidirectional (does not have a pre-assigned head and tail), it can have two sequences for the same configuration that are reverse of each other. For example, the configuration of the DOO drawn on the circuit board in Figure~\ref{fig:doo-routing:cable-spatial-representation} is $C = (-1, 1, 27, 28, 11, 4, 1, 6, 4, 15, 6, 9, -1)$ or its reverse. Note that if the graph $G_s$ is computed correctly, every two consecutive vertices in $C$ should have an edge in $E_s$. 

Based on the assumptions of the problem (see Section~\ref{sec:doo-routing:problem}), the extra slack of a DOO in each region is not encoded into its configuration. However, the extra slack is encoded if the DOO goes through some neighboring regions and comes back (e.g., if the DOO "touches" the neighbor convex region while passing through a convex region). Such instances are encoded as palindrome subsequences (i.e., subsequences that are the same if read backward or forward). Removing such subsequences may be desirable depending on the application and simplifies the configuration $C$ of a DOO in the work region.

We should note that the idea of convex decomposition in planning has been explored in other contexts before~\cite{1729881419894787, s19194165, 3022571}. However, it is used differently here to define the DOO configuration rather than planning itself.

\section{Routing Approach} \label{sec:doo-routing:routing}

Assume that the current configuration $C_0$ of a deformable one-dimensional object in a work region is provided along with the desired goal configuration $C_g$ of the DOO. 

The problem is to route the DOO in the work region from the current configuration $C_0$ to the goal configuration $C_g$. A naive solution to the routing problem is to completely undo $C_0$ into a "free" DOO, then apply $C_g$ configuration by passing through all the vertices in $C_g$. However, this solution is inefficient and requires the maximum number of manipulative actions. A more efficient approach is to keep the matching areas between the current and goal configurations and only manipulate what is necessary to reduce the number of manipulative actions.

We propose utilizing the sequence matching algorithms to minimize the number of actions required to change from $C_0$ into $C_g$. Let us assume that the manipulator supports two motion primitives: 1) pick a DOO at a specific point, and 2) place the picked DOO at a specific point in the work region. Then the following actions on the configuration sequences can be applied:

\begin{itemize}
    \item Replacing the \nth{i} element $s_{i}$ in $C_0$ with the \nth{j} element $g_{j}$ in $C_g$: Pick the DOO where it is passing through vertex $s_{i}$ and place it at vertex $g_{j}$.
    
    \item Removing the \nth{i} element $s_{i}$ in $C_0$ that does not correspond to an element in $C_g$: Remove the DOO from region $s_i$.
    
    \item Inserting the \nth{j} element $g_j$ in $C_g$ that does not correspond to an element in $C_0$: Adding (i.e., stretching) the DOO to region $g_j$.
\end{itemize}

With these three actions, we propose modifying the well-known Levenshtein sequence distance algorithm~\cite{Lev65} to obtain the manipulation actions required for routing. 

The original Levenshtein algorithm computes the minimum required edits (i.e., replacement, deletion, and insertion) to convert the initial sequence to the final sequence. To return this minimum distance, Levenshtein's dynamic programming method computes a matrix that retains the minimum number of edits required for converting the first $i$ elements of the initial sequence to the first $j$ elements of the final sequence. While the algorithm itself only computes the minimum number of edits, the types of edits can be extracted by backtracing this matrix once the algorithm is finished. Note that these edits are not unique, and backtracing will only output only one of the feasible solutions with the minimum number of actions.

To use the Levenshtein algorithm for the routing problem, the following modifications are required:

\begin{enumerate}[leftmargin=*]
    \item When comparing two elements $s_{i}$ and $g_{i}$, they match if they are the same vertex number (i.e., $s_{i}=g_{i}$). However, if they are both $-1$ (the "outside" vertex), they only match if there is a common neighbor for them in the sequence (e.g., if $s_{i-1}=g_{i+1}$). In other words, the two outside regions are considered the same only if they are next to the same vertices. That same vertex may occur before or after $-1$.
    
    \item The cost for each action is set to $1$. However, for a tunnel-like component, the action cost of either of the operations depends on how many vertices come before and after it. In other words, for the \nth{i} element in the sequence of size $n$, the cost will be $2\times\min(i-1, n-i) + 1$. For example, to remove the DOO from a tunnel-like region, it must free either the start of the DOO or the end of the DOO and put everything back again, bypassing the tunnel.
\end{enumerate}



\begin{figure}[!htb]
\centering
    \begin{subfigure}[b]{0.48\textwidth}
        \includegraphics[width=\textwidth]{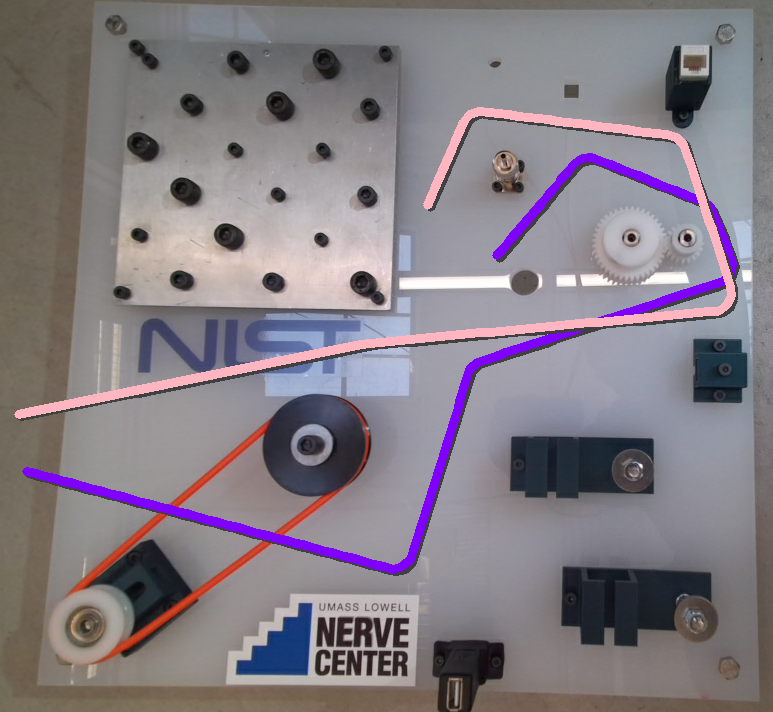}
        \caption{~}
        \label{fig:doo-routing:routing-original}
    \end{subfigure}
    ~    
    \begin{subfigure}[b]{0.48\textwidth}
        \includegraphics[width=\textwidth]{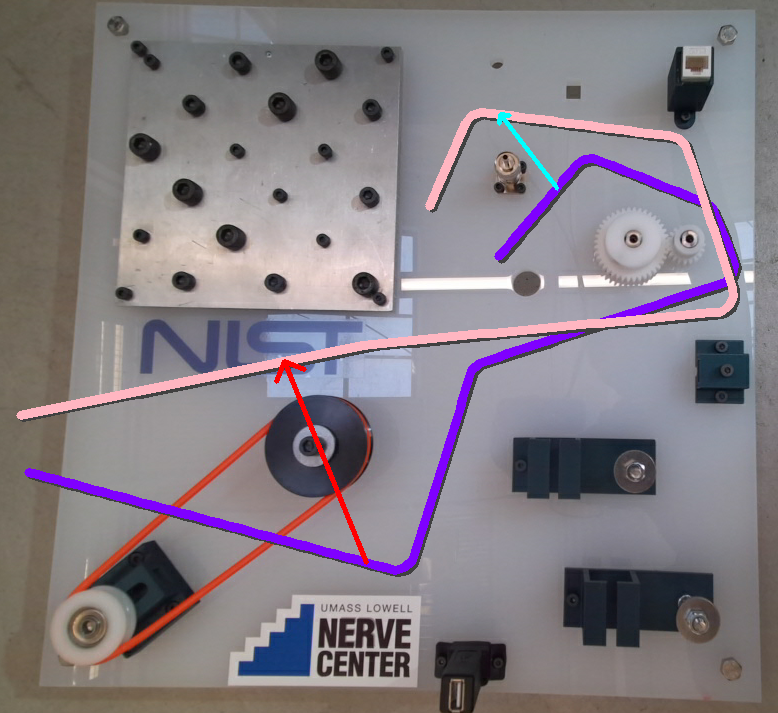}
        \caption{~}
        \label{fig:doo-routing:routing-result}
    \end{subfigure}
    \caption[DOO routing from initial to goal configuration]{Routing of a deformable one-dimensional object from the current configuration to a goal configuration. (a) The current (purple) and goal  (pink) deformable objects drawn on the example board of Figure~\ref{fig:doo-routing:original-board}. (b) The result of the routing from current to final configuration. Red arrow shows a removal action and cyan arrow shows a move (replacement) action.}
    \label{fig:doo-routing:routing}
\end{figure}

If the DOO is bi-directional, the algorithm should be repeated with one of the sequences reversed to get the least number of actions. Then, backtracing can give the actions needed to perform on the DOO to change its configuration from $C_0$ to $C_g$. The time and space complexities of the algorithm are $\mathcal{O}(nm)$, where $n$ and $m$ are the lengths of the current configuration ($|C_0|$) and goal ($|C_g|$) configuration sequences. Figure~\ref{fig:doo-routing:routing} shows the routing actions for a DOO to get from its current configuration to the goal configuration.

To realize the computed actions, a single manipulator can act as below:

\begin{itemize}
    \item For replacement of element $s_{i}$ in $C_0$ with element $g_{j}$ in $C_g$: Pick the DOO where it is passing through vertex $s_{i}$ and place it at vertex $g_{j}$.
    
    \item Removing element $s_{i}$ in $C_0$: Pick the DOO where it is passing through vertex $s_{i}$, and place it at the point where the goal DOO crosses from $g_{j-1}$ to $g_{j}$.
    
    \item Inserting element $g_j$ in $C_g$: Pick the DOO where it is crossing $s_{i-1}$ into $s_i$, and place it at (i.e., stretch it to) vertex $g_{j}$.
\end{itemize}

The details of how these actions are implemented depend on the motion primitives of the robot and the environment. 

We call the point on the DOO where the picking happens as \textit{picking point} and the points where the two sequences match (i.e., no action is required) as \textit{fixed points}. If more than one manipulator is available, the second manipulator can grab the closest fixed point before the picking point, and the third manipulator can grab the closest fixed point after the picking point to prevent these points from moving. 

The proposed routing algorithm does not incorporate the DOO dynamics and, therefore, cannot understand the result of the actions taken by the manipulator on the whole DOO configuration. To mitigate the lack of dynamics knowledge, routing and manipulation action can be performed iteratively until the current configuration matches the goal configuration. A single routing and then a manipulative action is performed at each iteration to get the DOO closer to the goal configuration. 

\section{Experiments and Results} \label{sec:doo-routing:tests}

We implemented it for a routing and manipulation task to test the proposed method. The task includes a single-arm manipulating a cable on the circuit board of Figure~\ref{fig:doo-routing:original-board} to change its current configuration to a goal configuration. This board was originally designed for task~\#3 of the Assembly Performance Metrics and Test Methods by the National Institute of Standards and Technology (NIST) to measure the capability of robotics systems for performing advanced manipulation on cables~\cite{nist}. The board was also adopted for the Robotic Grasping and Manipulation Competition in IROS 2020.

We performed many simulation experiments and several experiments with different settings on our robot. Each experiment included several iterations of routing and a manipulative action (i.e., pick and place actions) until the cable configuration had matched the given goal configuration. 

We performed 200 simulation experiments, where we randomly placed a 0.3-0.5~$\unit{m}$ cable on the 0.38~$\unit{m}$ NIST board and randomly (in 170 tests) or manually (in 30 tests) placed a cable of the same length on the board as the goal configuration. The average number of actions over all the experiments was 4.34, and the maximum number of actions was 9. The processing time for each routing step was less than 1~$\unit{ms}$ for all the experiments. Figure~\ref{fig:doo-routing:routing-result} illustrates an iteration of our simulated routing experiments.

One of the fastest DOO routing methods applicable in our scenario is provided by Guo et al.~\cite{GUO2020158}. They propose a bi-directional RRT-based method called RRT-BwC (Bi-direction with Constrain) to route DOOs. They report the execution time of 14.7~$\unit{s}$ for a DOO routing task, which is a significant improvement over the other existing routing methods but several orders of magnitude slower than our method. Compared to our solution, their method uses a higher number of anchor (sampled) points on the cable and a higher dimension for the space, leading to a much higher dimensionality of the configuration space and a much slower planning problem. However, note that most routing solutions are designed for more general scenarios than ours and can work for scenarios where our solution would either fail or needs to be extended.

\begin{figure}[!htb]
\centering
    \begin{subfigure}[b]{0.32\textwidth}
        \includegraphics[width=\textwidth]{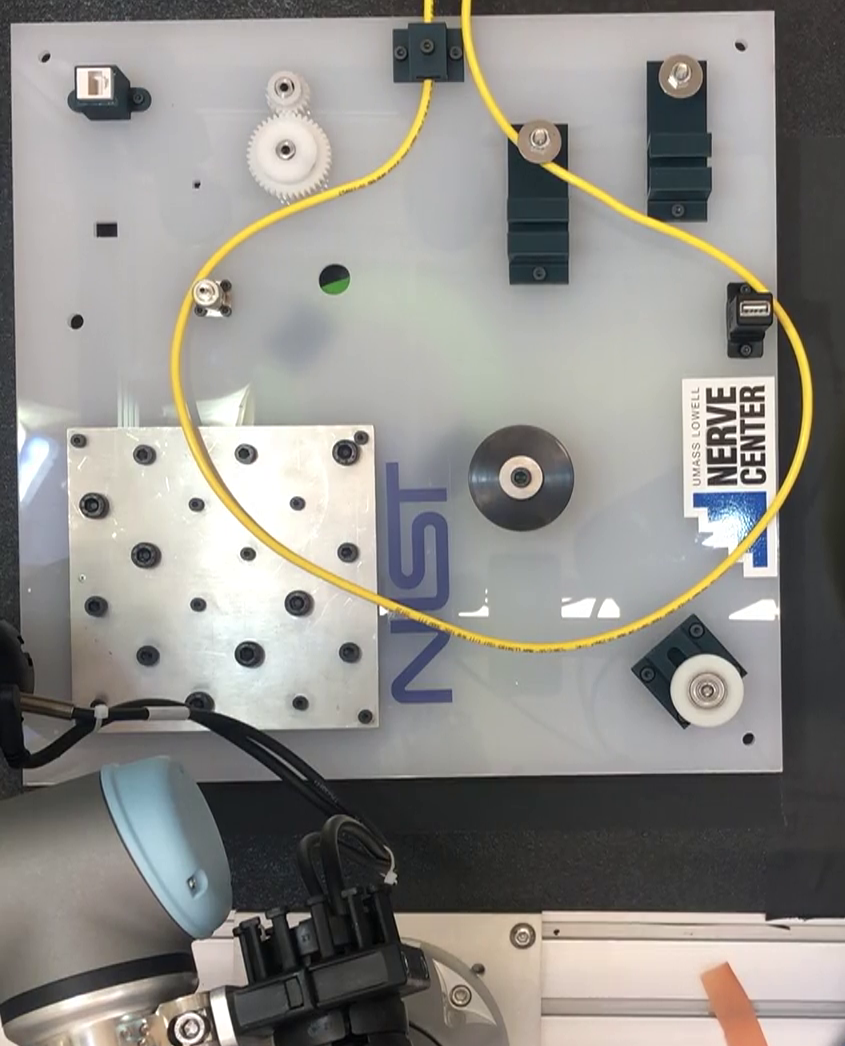}
        \caption{~}
    \end{subfigure}
    \hfill    
    \begin{subfigure}[b]{0.32\textwidth}
        \includegraphics[width=\textwidth]{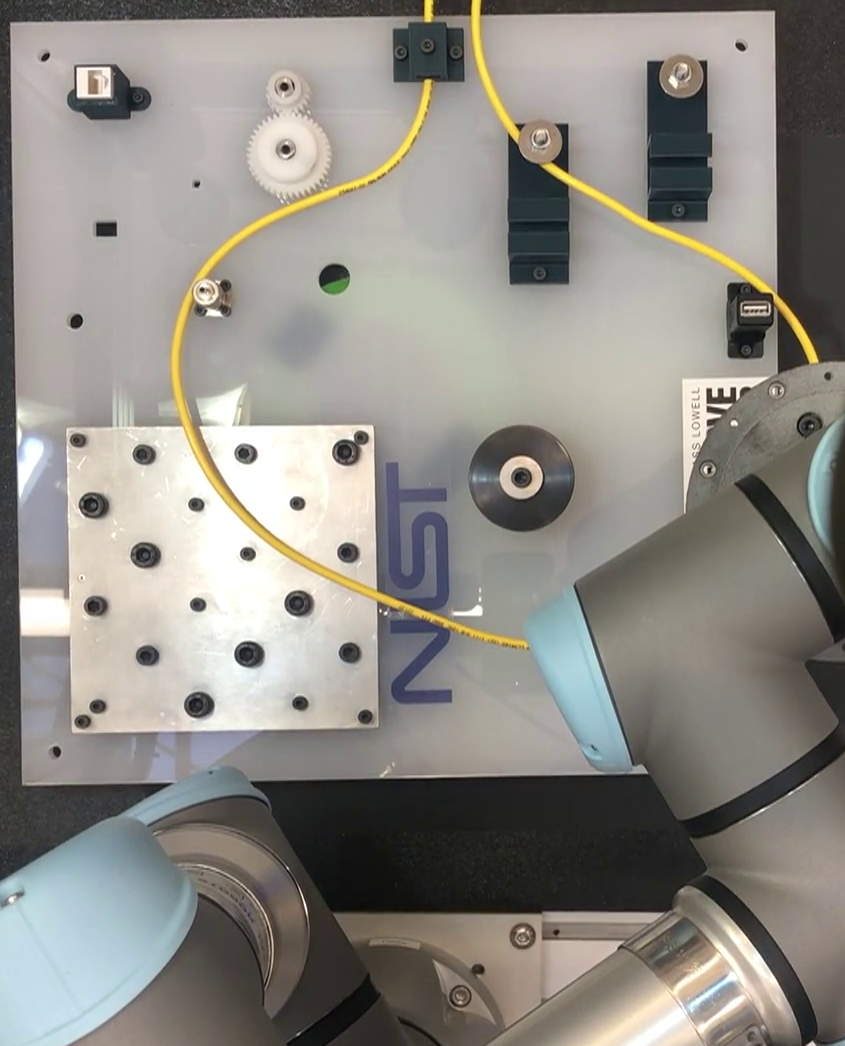}
        \caption{~}
    \end{subfigure}
    \hfill    
    \begin{subfigure}[b]{0.32\textwidth}
        \includegraphics[width=\textwidth]{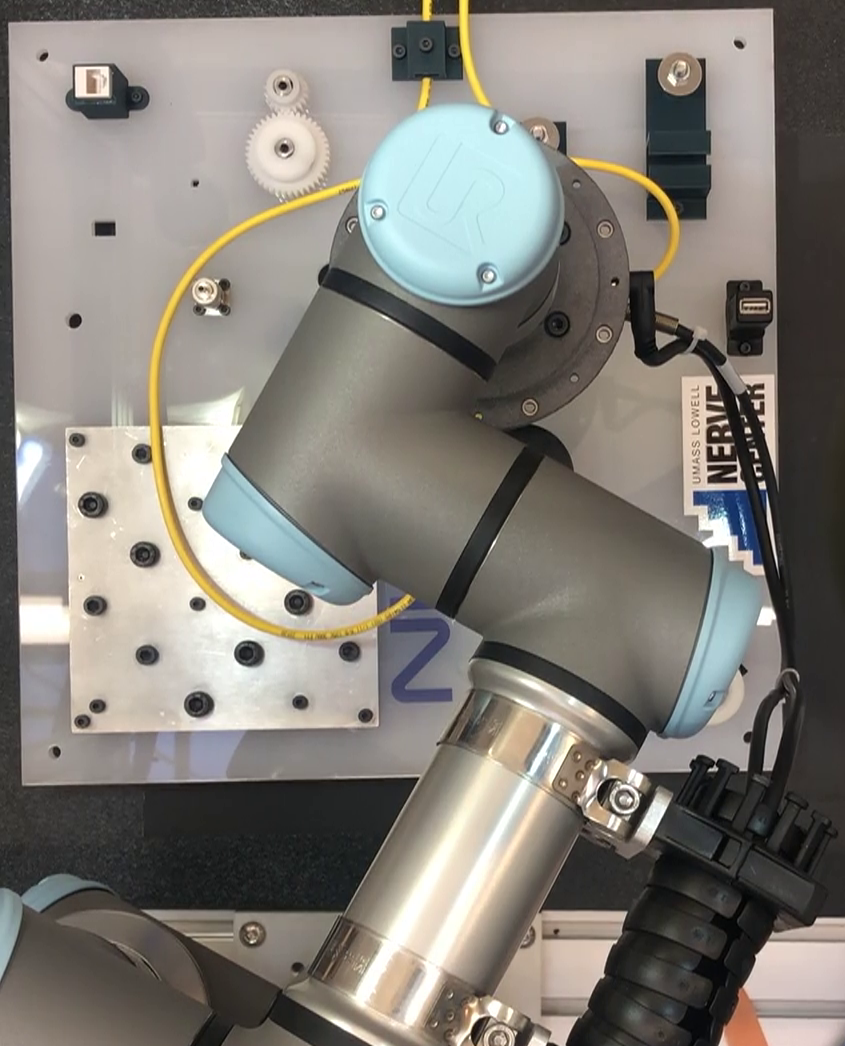}
        \caption{~}
    \end{subfigure}
    
    \medskip
    
    \begin{subfigure}[b]{0.32\textwidth}
        \includegraphics[width=\textwidth]{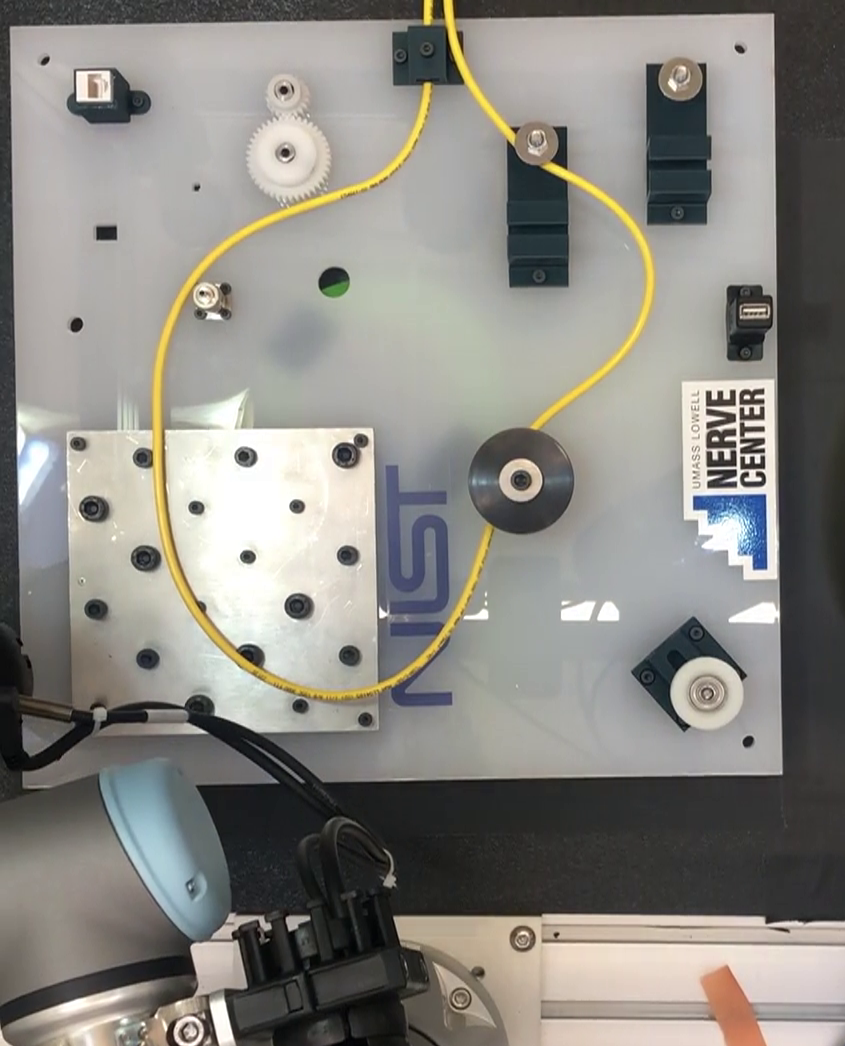}
        \caption{~}
    \end{subfigure}
    \hfill
    \begin{subfigure}[b]{0.32\textwidth}
        \includegraphics[width=\textwidth]{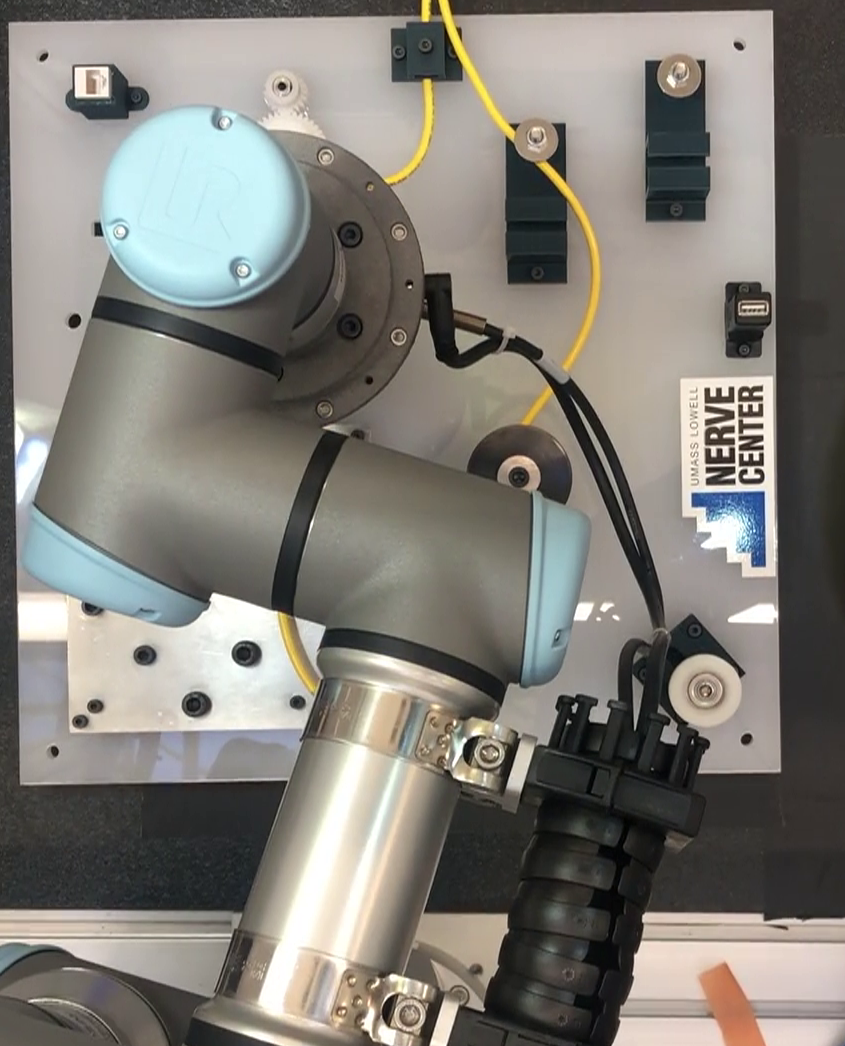}
        \caption{~}
    \end{subfigure}
    \hfill    
    \begin{subfigure}[b]{0.32\textwidth}
        \includegraphics[width=\textwidth]{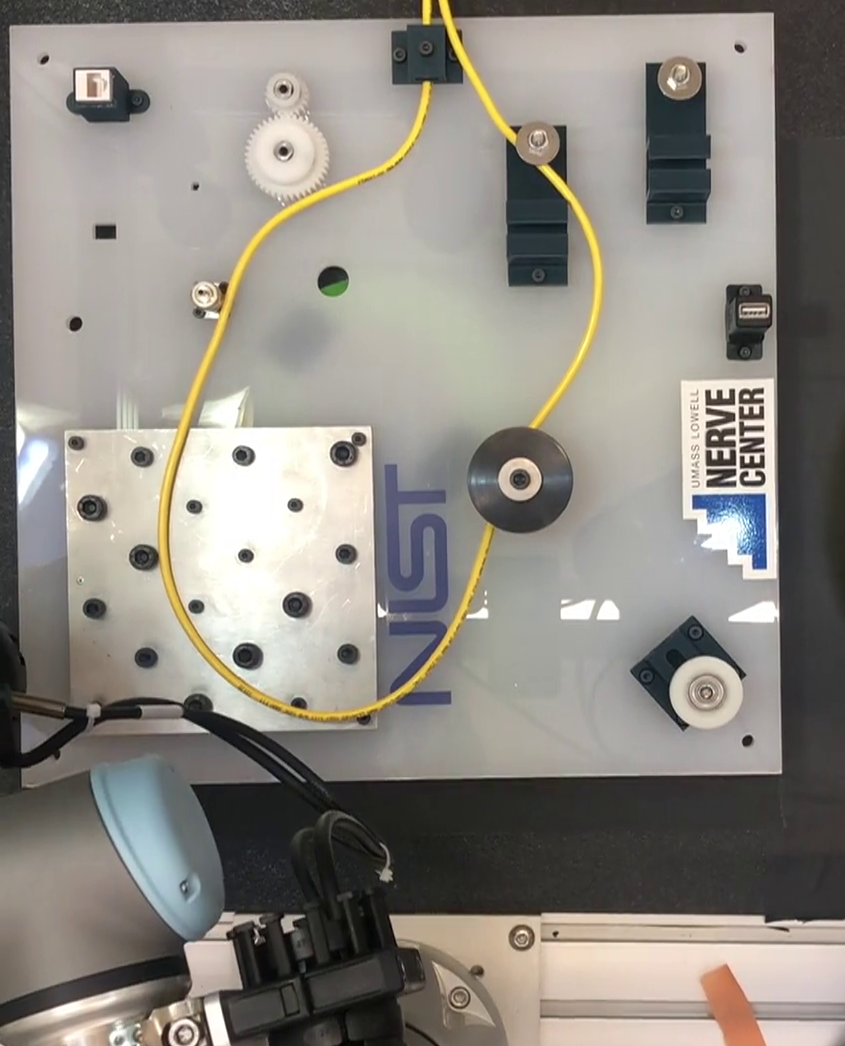}
        \caption{~}
    \end{subfigure}
    \caption[Video sequence of a two-step DOO routing and manipulation]{Routing and manipulation of a deformable one-dimensional object from the initial configuration shown in figure~(a) to the goal configuration shown in figure~(f). (a) Initial configuration. (b-c) Manipulating the cable based on the first iteration of routing. (d) The result configuration after the first manipulative task. (e) Manipulating the cable based on the second iteration of routing.}
    \label{fig:doo-routing:manipulation-res01}
\end{figure}

\begin{figure}[!htb]
\centering
    \begin{subfigure}[b]{0.40\textwidth}
        \includegraphics[width=\textwidth]{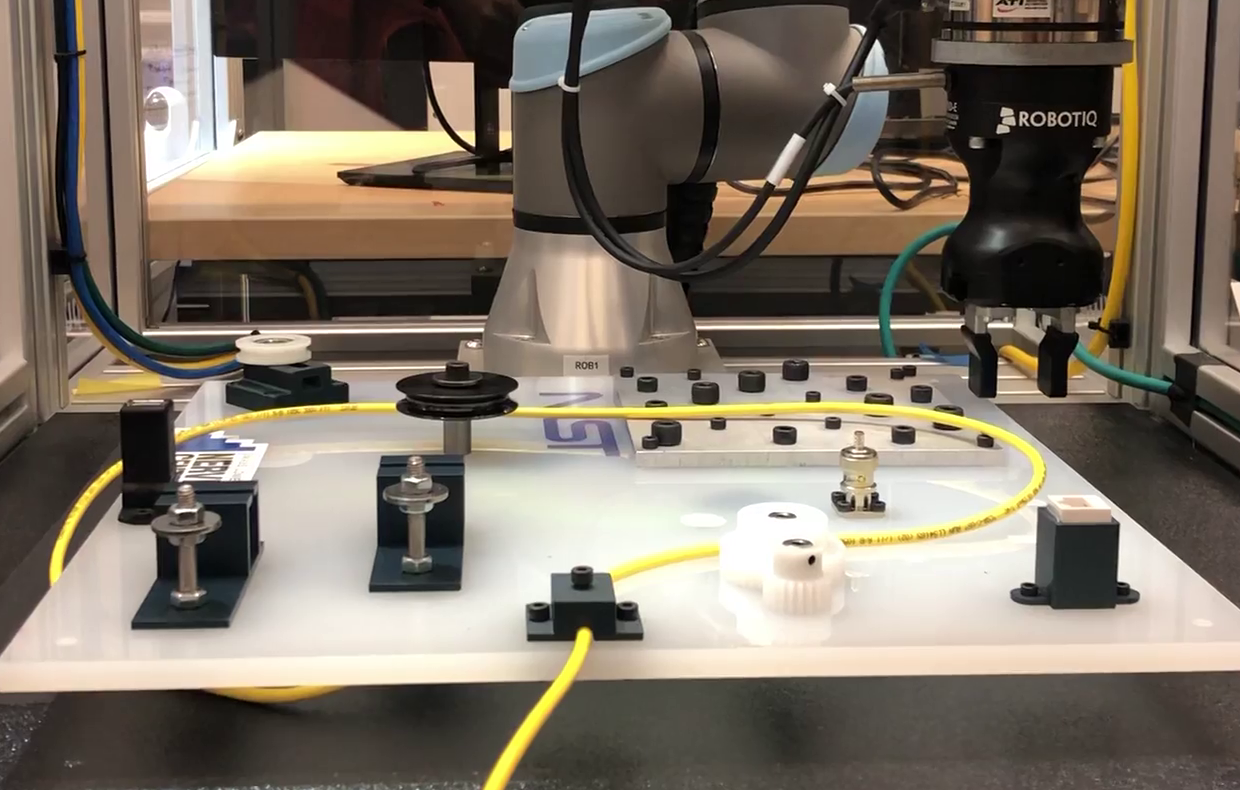}
        \caption{~}
    \end{subfigure}
    ~
    \begin{subfigure}[b]{0.40\textwidth}
        \includegraphics[width=\textwidth]{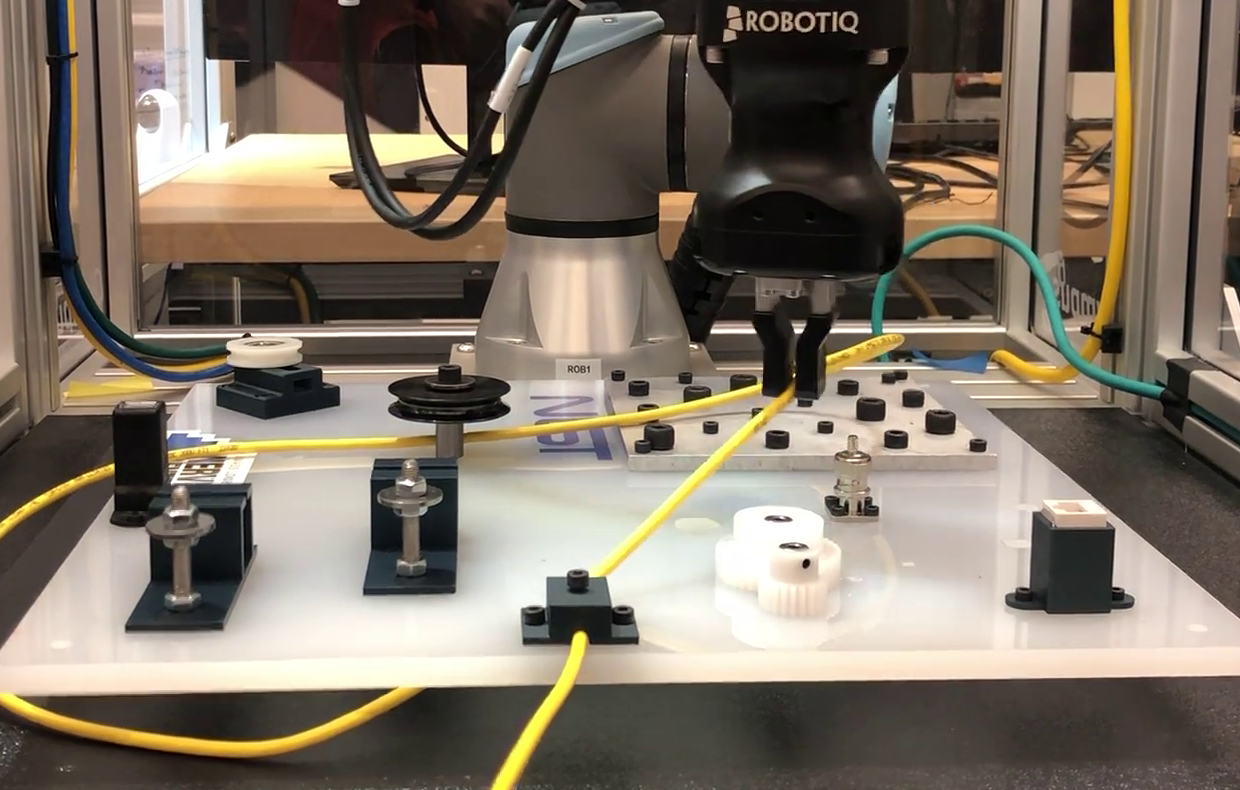}
        \caption{~}
    \end{subfigure}
    
    \medskip
    
    \begin{subfigure}[b]{0.40\textwidth}
        \includegraphics[width=\textwidth]{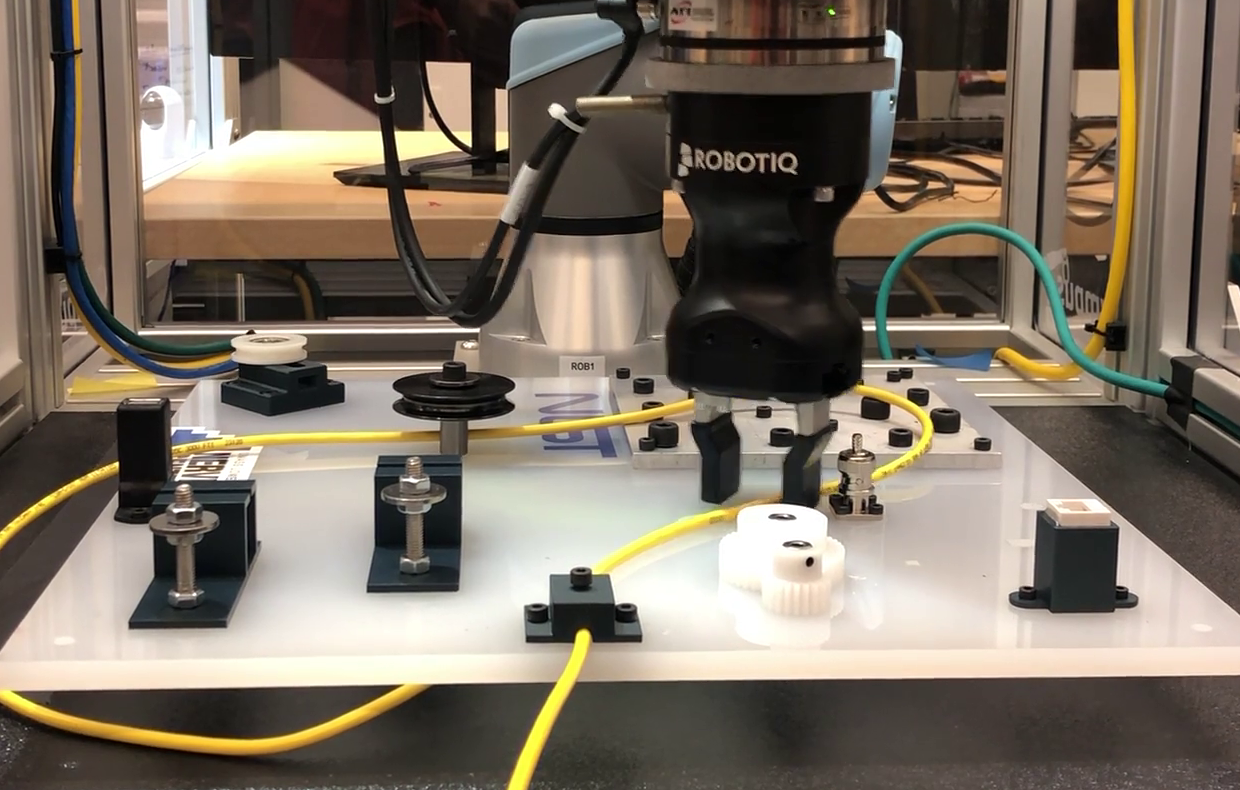}
        \caption{~}
    \end{subfigure}
    ~
    \begin{subfigure}[b]{0.40\textwidth}
        \includegraphics[width=\textwidth]{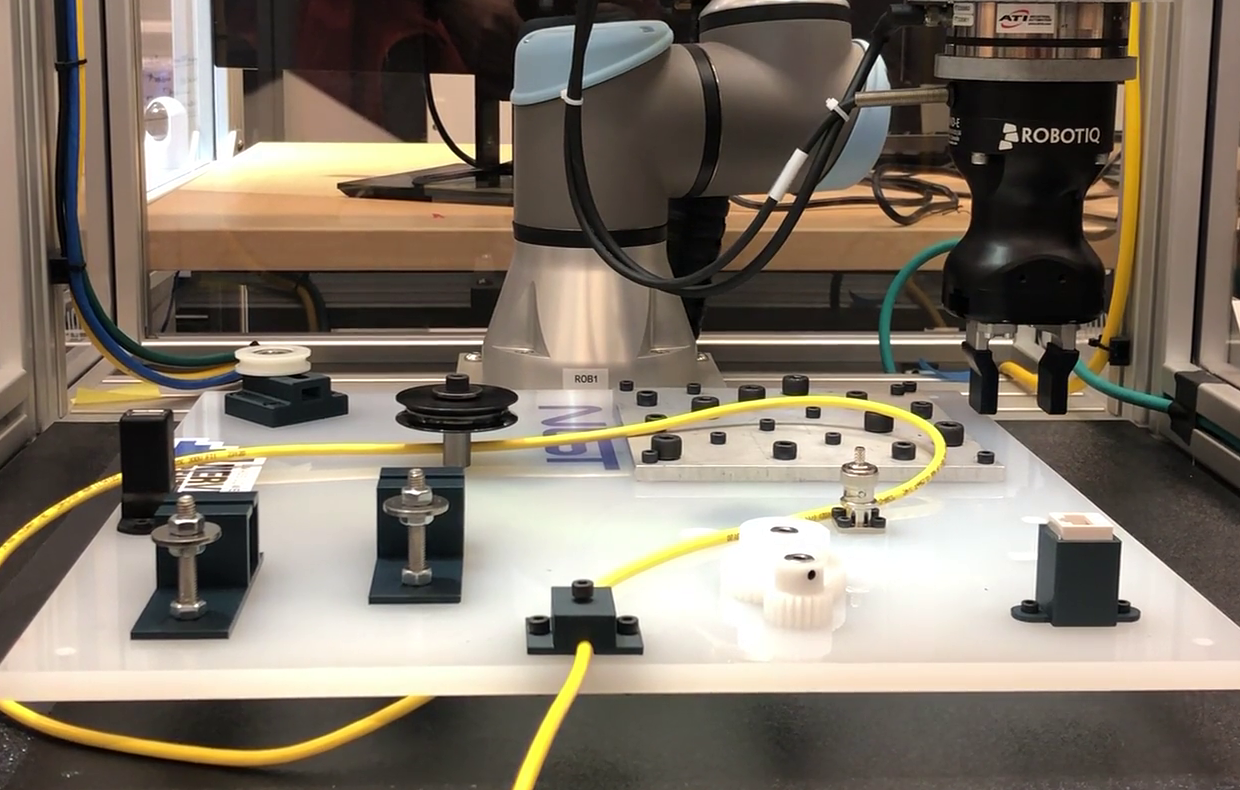}
        \caption{~}
    \end{subfigure}
    \caption[Video sequence of a one-step DOO routing and manipulation]{Routing and manipulation of a deformable one-dimensional object. (a - d) Screenshots from a single iteration of routing and manipulation sequence for a cable on the example circuit board of Figure~\ref{fig:doo-routing:original-board}.}
    \label{fig:doo-routing:manipulation-res02}
\end{figure}

We manually placed the cables on the NIST board for the robot experiments. At each step of the planning, we utilized the DOO detection algorithm proposed in Chapter~\ref{ch:doo-detection} to automatically detect the cable and extract its configuration on the circuit board. Then the goal configuration was manually given to the system. Our experiments showed that the method could also extend to real systems. Figures~\ref{fig:doo-routing:manipulation-res01} and~\ref{fig:doo-routing:manipulation-res02} show the routing and manipulation experiments using Universal Robots UR3 arm robot.

\section{Analysis for Aerial Robot Manipulation} \label{sec:doo-routing:uav}

In general, ground manipulators have higher position precision compared to aerial robots. During a free flight or even physical interaction tasks such as painting a wall or pushing a box, the lower precision of the aerial robots may not affect the performance or the task's feasibility. However, many physical interaction tasks require more position precision, such as maintenance and cable manipulation at the top of utility poles with many wires and components around. For such tasks, the lower precision of the aerial manipulator end-effector can become a significant issue.

\begin{figure}[!htb]
\centering
    \begin{subfigure}[b]{0.37\textwidth}
        \includegraphics[width=\textwidth]{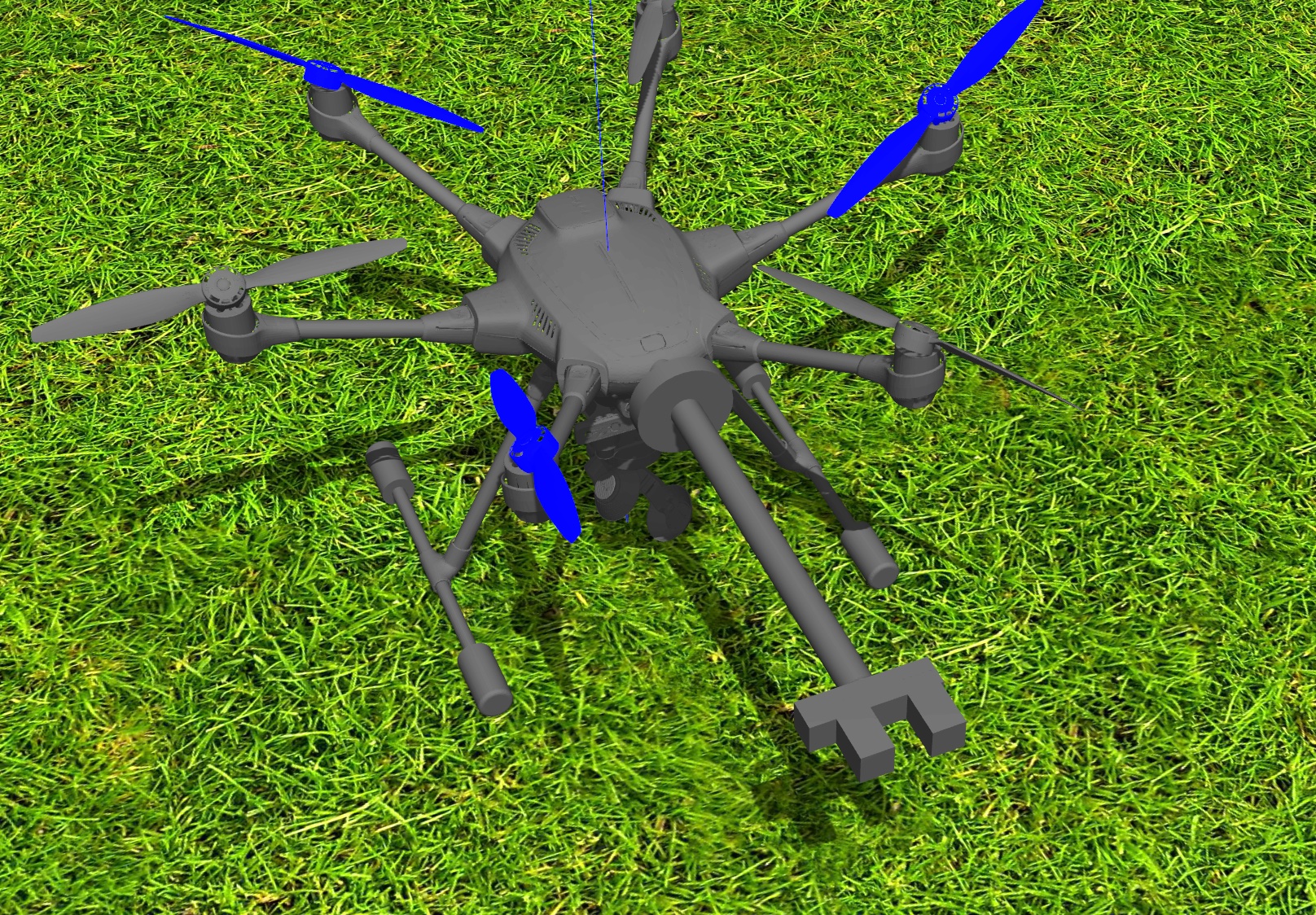}
        \caption{~}
    \end{subfigure}
    \hfill    
    \begin{subfigure}[b]{0.60\textwidth}
        \includegraphics[width=0.59\textwidth]{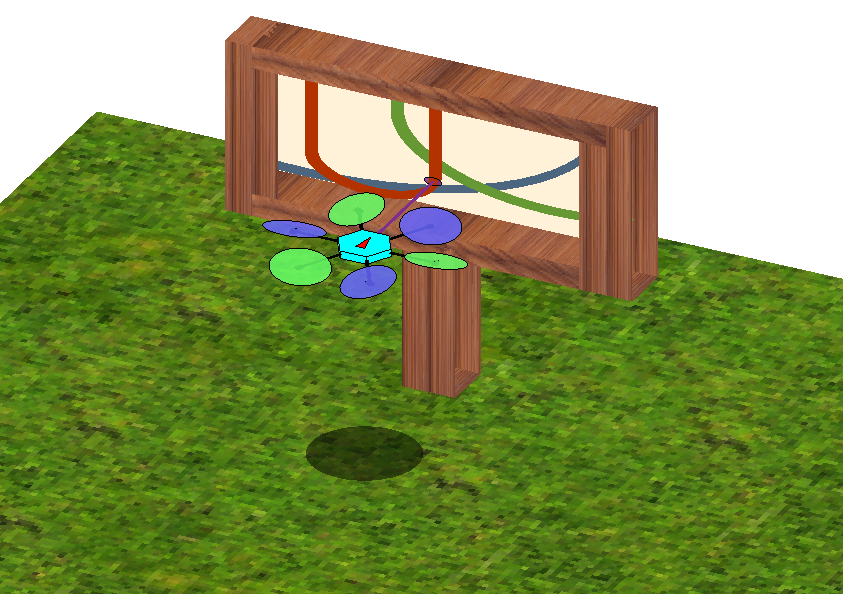}
        \includegraphics[width=0.38\textwidth]{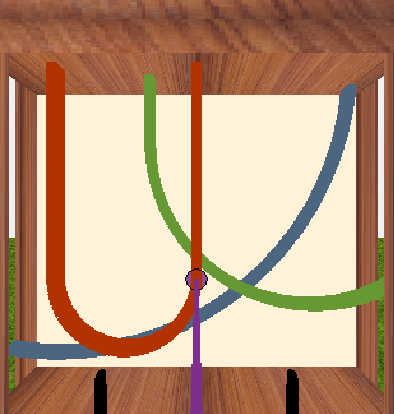}
        \caption{~}
    \end{subfigure}
    \caption[Setup for feasibility analysis of the aerial DOO manipulation]{The setup used for feasibility tests of DOO manipulation using aerial robots in simulation. (a) Gazebo model of our robot with a gripper for cable manipulation. (b) A screenshot of the MATLAB simulator used for analyzing the feasibility of aerial DOO manipulation and the accuracy of the end-effector for grasping and manipulating the cables. (Left) An external view of grasping the red wire using our hexarotor with tilted arms. (Right) First-person view of the moment of grasping the red wire.}
    \label{fig:doo-routing:uav:aerial-manipulation}
\end{figure}

We analyzed the feasibility of manipulating the detected cable in both Gazebo and MATLAB simulation environments for our fully-actuated hexarotor with tilted arms (see Section~\ref{sec:control:tests:hardware-software}) controlled with the controller system developed in Chapter~\ref{ch:control}. We measured the position error for grasping a specific point on the cable. Figure~\ref{fig:doo-routing:uav:aerial-manipulation} shows our setup for testing the feasibility of the task. 

For each experiment, the robot first flies to around 0.5~$\unit{m}$ distance from the cable, then moves forward to grasp the cable segment. We measure the distance of the end-effector from the desired point on the cable at the moment when it is the closest to it. 

The results from the MATLAB simulation were unrealistically perfect, and since the Gazebo simulator tends to give more realistic results, we only report the Gazebo experiments. 

\begin{figure}[!htb]
    \centering
    \includegraphics[width=0.65\textwidth]{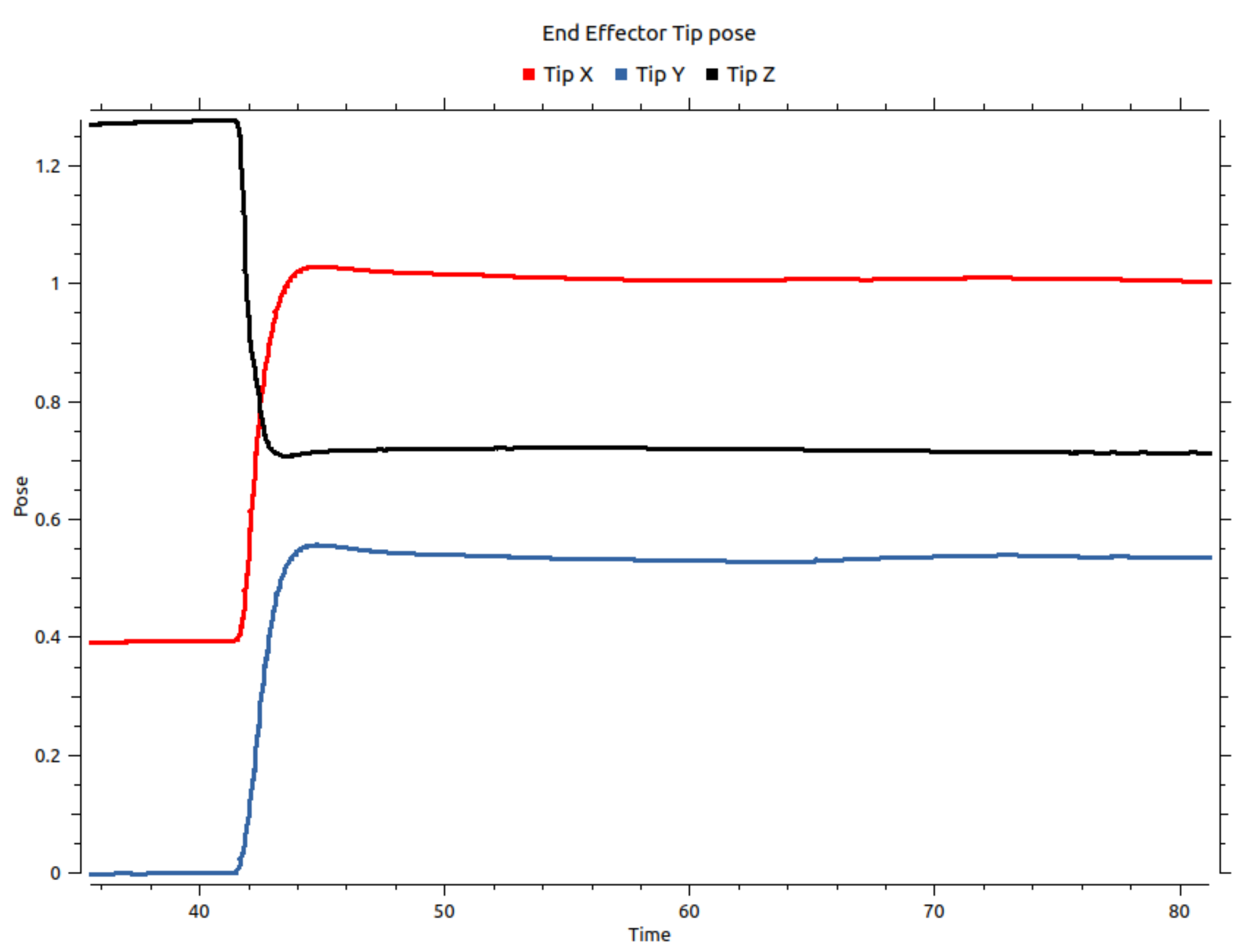}
    \caption[End-effector's position to grasp the desired cable segment]{Feasibility tests of aerial DOO manipulation in Gazebo: End-effector's position to grasp the desired cable segment at $\matrice{1.0 & 0.5 & 0.7}\T$.}
    \label{fig:doo-routing:uav:end-effector-plot}
\end{figure}

Figure~\ref{fig:doo-routing:uav:end-effector-plot} illustrates how our end-effector's position can reach the target cable point. Table~\ref{tbl:doo-routing:uav:end-effector-error} shows the viability of the physical interaction with the perceived DOOs in the simulation if at least 13~$\unit{mm}$ position error in grasping can be tolerated in the application. The next future step would be to perform the analysis on the real robot.

\begin{table}[!htb]
\centering
\caption[End-effector position error test results]{The end-effector's position error (in~$\unit{mm}$) for grasping a cable segment. Trials done in the Gazebo simulator.}
\label{tbl:doo-routing:uav:end-effector-error}
\begin{tabular}{|c|c|c|c|c|c|c|c|c|c|}
\hline
\rowcolor[HTML]{EFEFEF} 
\# of Tests & Max. Error & Mean Error & Std. Dev. \\ \hline
20 & 12.92 & 7.84 & 2.91 \\ \hline
\end{tabular}
\end{table}

At the same time, aerial robots have limited wrenches compared to ground robots. It is imperative to analyze the feasibility of the physical interaction tasks from the manipulability perspective as well. 

\begin{figure}[!htb]
        \centering
        \includegraphics[width=0.65\textwidth]{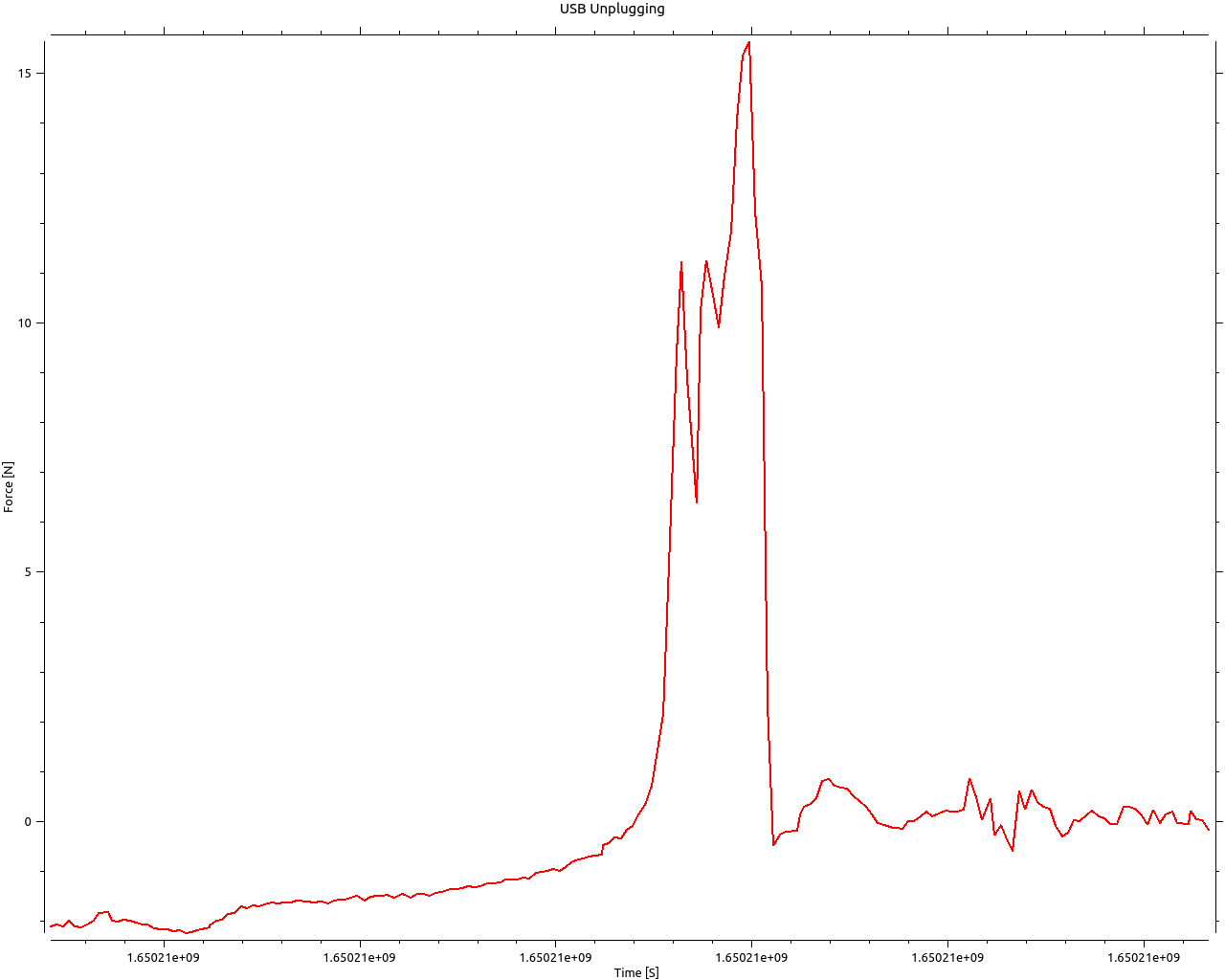}
        \caption[The required forces to unplug a USB cable]{Feasibility tests of aerial DOO manipulation in Gazebo: Forces measured for the task of unplugging a USB Type-A cable.}
        \label{fig:doo-routing:uav:usb-unplug-plot}
    \end{figure}

We measured the forces required for simple cable-related tasks, such as plugging and unplugging cables in the slots on a board. Figure~\ref{fig:doo-routing:uav:usb-unplug-plot} shows the example forces measured for unplugging a USB Type-A cable. In this specific experiment, the maximum measured required force is 15.84~$\unit{N}$ at the peak. 

\begin{figure}[!htb]
    \centering
        \begin{subfigure}[b]{0.30\linewidth}
            \includegraphics[width=\textwidth]{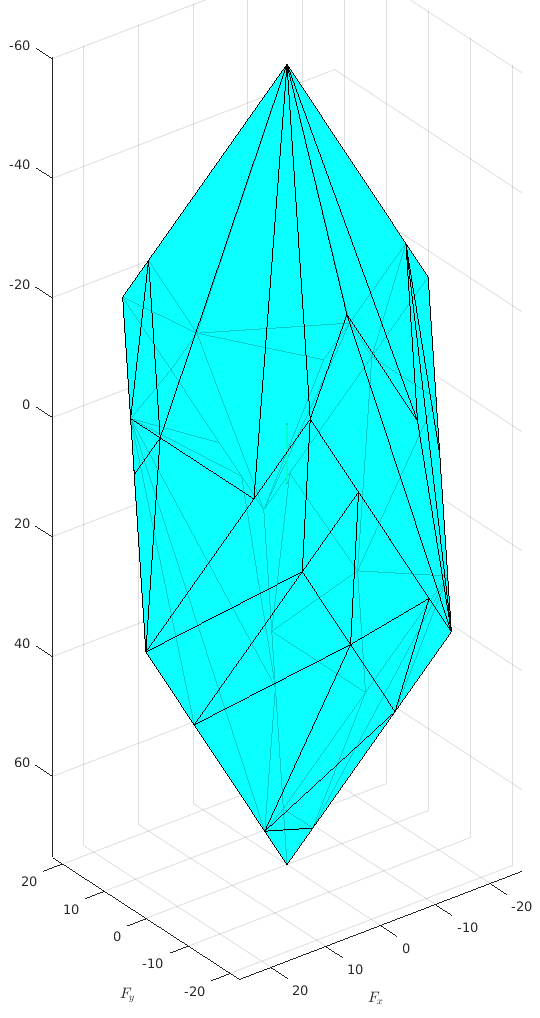}
            \caption{~}
            \label{fig:doo-routing:uav:thrust-set}
        \end{subfigure}
        ~~
        \begin{subfigure}[b]{0.41\linewidth}
            \includegraphics[width=\textwidth]{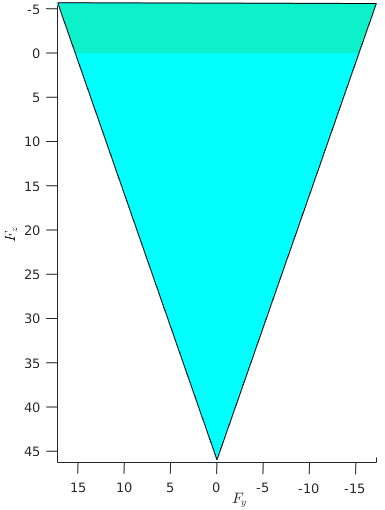}
            \caption{~}
            \label{fig:doo-routing:uav:thrust-set-usb-unplug}
        \end{subfigure}
    \caption[Feasibility tests of DOO manipulation forces]{Feasibility tests of DOO manipulation forces. (a) Our aerial platform. (b) Force polytopes for our platform. (c) Remaining $y$ and $z$ forces when unplugging a USB cable.}
    \label{fig:doo-routing:uav:usb-analysis}
    \end{figure}

Figure~\ref{fig:doo-routing:uav:usb-analysis} compares the available thrust set for our UAV during hovering (Figure~\ref{fig:doo-routing:uav:thrust-set}) vs. when it is pulling the aforementioned USB Type-A cable directly in its backward direction (i.e., $-\XB$ direction) at the peak moment when the required force to unplug the cable is 15.84~$\unit{N}$ (Figure~\ref{fig:doo-routing:uav:thrust-set-usb-unplug}). The sets are computed using Algorithms~\ref{alg:wrench:decoupled:thrust-set-estimation} and~\ref{alg:wrench:coupled:wrench-set-estimation-with-desired-components} proposed in Chapter~\ref{ch:wrench}.

The green region in Figure~\ref{fig:doo-routing:uav:thrust-set-usb-unplug} shows the remaining forces that still allow the robot to keep its altitude during the maximum force required for the unplugging task. This analysis shows that our aerial robot would be able to unplug the cable in this case, but it is very close to its limits. For example, if the goal is to perform the task while keeping the altitude, it may not be able to achieve a more demanding task.

\section{Conclusion and Discussion} \label{sec:doo-routing:conclusion}

We presented a novel method for the spatial representation and routing of a deformable one-dimensional object that is efficient and fast. The proposed method decomposes the work region into convex polygons and polytopes. Then it uses the decomposition to encode the configuration of a DOO in this work region. The resulting configuration is a sequence of the regions the DOO passes from, effectively simplifying the routing algorithm to a modified sequence matching method with a quadratic time and space complexity. The iteration of our routing algorithm with manipulative actions can accomplish a desired routing and manipulation task. The low planning time and overhead make it ideal for offline and online planning problems for routing and manipulation. Our experiments showed that the proposed method could correctly plan the manipulation actions and achieve goal configurations of the DOO from various initial configurations. 

The proposed approach is still in its infancy and can be extended further to cover many real-world tasks that are currently being addressed using slower and less efficient methods such as sampling-based planners. In its current iteration, this method can be used in tasks where the environment can be divided to separate convex regions and for a manipulator with two primitive actions: pick a point on the DOO, and place it in a specific point. The algorithm can be easily extended to include more manipulator motion primitives, such as wrapping the DOO around a component or passing through loops to create knots.

The proposed routing and manipulation algorithm ignores the dynamics of a cable. Although the routing algorithm itself can work well independent from the dynamics, in our real-world tests, we realized that considering the dynamics during manipulative tasks can help the system with performing the pick/place tasks. Additionally, further incorporating simple dynamics into the routing method's cost calculation in the future can reduce the number of actions (i.e., iterations) required for performing a routing/manipulation task by more accurately predicting the result of the manipulation task at each iteration.

We further analyzed the feasibility of DOO manipulation tasks using aerial robots. The analysis demonstrated our achieved precision and the required wrenches for a sample wire manipulation task. It provides a good stepping stone to having a more comprehensive and powerful study to identify the challenges and achieve helpful insights on deformable object manipulation using aerial robots.

\chapter{Conclusion and Future Work} \label{ch:conclusion}

This work aimed to improve the state-of-the-art in physical interaction and manipulation of the environment using aerial robots and further extend such interactions to deformable objects. 

We introduced a novel controller design that can extend most existing designs to provide faster integration of the new fully-actuated multirotors into existing flight stacks and allow them to work with commonly-available software and hardware tools without any modification. 

We further extended the controller design to control both the positions and forces applied to the contact point during physical interactions. In addition to extensive simulation tests in different simulators, we showed the viability of our design for real-world free-flight and physical interaction tasks, such as contact inspection and drawing on the whiteboard, using our fully-actuated multirotor. 

\begin{figure}[t]
    \centering
    \includegraphics[width=\linewidth]{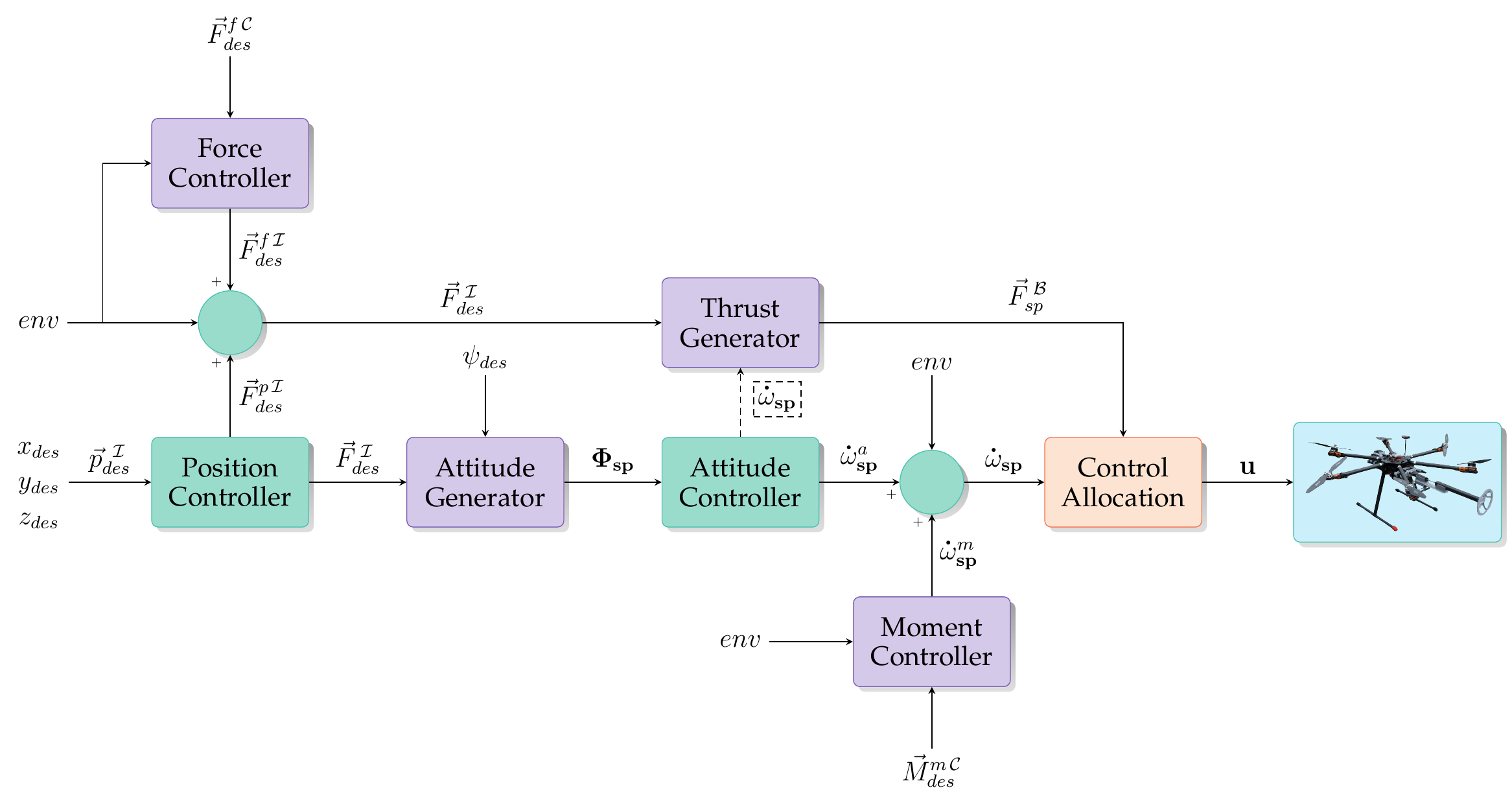}
    
    \caption[Hybrid Motion-Wrench Controller architecture]{A possible extension of our proposed controller architecture of Chapter~\ref{ch:control} into a Hybrid Motion-Wrench controller. The position, attitude, force, and moment control modules independently calculate the necessary linear and angular accelerations to achieve the desired inputs, then are combined based on their respective subspaces. All the modules also receive the state feedback $\mat{x}$ from the multirotor, which is omitted here for better illustration.}
    
    \label{fig:conclusion:hmw-controller}
\end{figure}

A possible next step for taking this Hybrid Position-Force controller further would be to extend it into a full Hybrid Motion-Wrench controller (HMWC), allowing full control of both translational and rotational motions and wrenches at the point of contact. A new Moment Controller module can be added to control the moment during the contact, similar to the Force Controller module. Figure~\ref{fig:conclusion:hmw-controller} illustrates a possible architecture for the hybrid motion-wrench controller architecture.

This thesis provided a real-time method for estimating the dynamic manipulability polytopes (i.e., complete wrench set) for multirotors. It further extended the estimation to controlled physical interaction scenarios, allowing accurate estimation of the wrenches when some of the desired wrenches are already known. Extensive analysis and experiments illustrated how the methods work and can be used in different tasks. 

We also showed how a rough estimate of the wrench set could be computed for the variable-pitch multirotors by sampling different pitches. The next step to improve the wrench-set estimation methods for aerial robots would be to solve the problem for variable-pitch rotors analytically instead of sampling. This can result in a real-time solution that can expand the benefits of the wrench estimation methods to a new group of aerial manipulators. Additionally, the proposed methods can be extended to arm manipulators to provide the real-time estimation of the wrench polytopes for an even more extensive set of robots.

We illustrated the benefits of the real-time wrench set estimation methods in planning the physical interaction tasks, enhancing the control allocation module making it more flexible and more accurate, and computing the optimal tilt and thrust setpoint in the presence of external forces. We also enumerated other applications such as failure recovery and optimization of multirotor designs.

We mostly only explored the mentioned applications, and the future possible research directions would be developing and illustrating these applications on real UAVs in real-world tasks. 

We presented a novel method for detecting deformable one-dimensional objects (e.g., wires and cables) for robotics applications and illustrated its effectiveness using real experiments.

A possible future enhancement of the method would be exploring choices other than the weighted sum for the total cost function. A more optimized implementation would take advantage of special data structures and parallelization to increase the method's speed by several orders of magnitude. Furthermore, while the considerations for the 3-D case are provided for each step, we did not implement the 3-D case, and there may be unpredicted implementation challenges. In the future, the ideas of the method can be integrated with tracking methods to improve tracking accuracy. Finally, the method's handling of the cable crossings does not detect which cable was on top. Another future step would be detecting the order of the cables in the crossings.

We presented a novel method for the spatial representation and routing of a deformable one-dimensional object that is efficient and fast. The low planning time and overhead make it ideal for offline and online planning problems for routing and manipulation. Our experiments showed its effectiveness for environments such as a wire board.

The proposed routing and manipulation algorithm ignores the dynamics of a cable. Although the routing algorithm itself can work well independent from the dynamics, in our real-world tests, we realized that considering the dynamics during manipulative tasks can help the system with performing the pick/place tasks. Additionally, further incorporating simple dynamics into the routing method's cost calculation in the future can reduce the number of actions (i.e., iterations) required for performing a routing/manipulation task by more accurately predicting the result of the manipulation task at each iteration.

Finally, we presented the initial feasibility analysis for manipulating the deformable one-dimensional objects using aerial robots. We measured the accuracy of the end-effector in simulation and measured the real forces required for some everyday wire manipulation tasks. 

\begin{figure}[!htb]
    \centering
    \includegraphics[width=\linewidth]{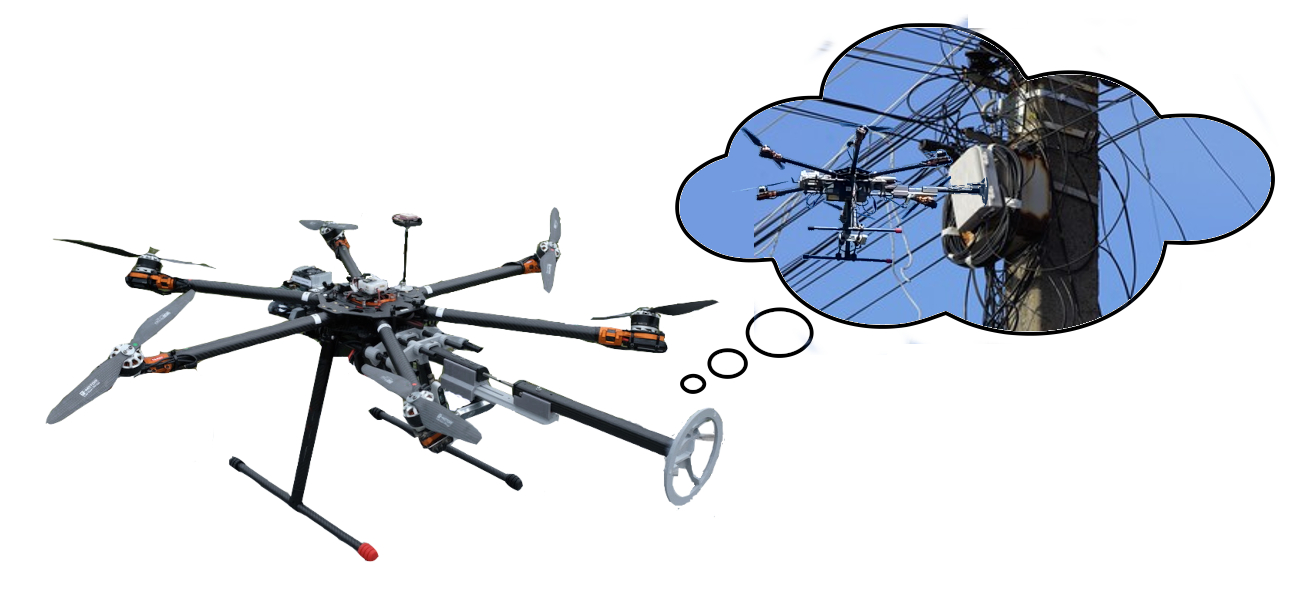}
    \caption[Future of aerial DOO manipulation]{Illustration of a possible future aerial manipulation task performed on deformable objects.}
    \label{fig:conclusion:future-of-aerial-doo-manipulation}
\end{figure}

With the steps we took in this work and the future advances, we believe one day it will be possible to have aerial manipulation applications involving deformable objects (such as Figure~\ref{fig:conclusion:future-of-aerial-doo-manipulation}) that can help bring operational and maintenance costs while reducing the risks associated with these tasks.

The next step would be integrating the different parts of this thesis into a real aerial manipulator to perform fully-automated cable manipulation in realistic scenarios. However, to achieve this goal, there is a need to study the aerial manipulation of deformable objects in more realistic scenarios. Some challenges include understanding the effects of gravity and the applied forces on the manipulation task and analyzing the optimal grasping strategy of these objects in 3-D space. The planner and controller for the tasks involving deformable objects should consider the dynamics of the objects and account for the precision of the end-effector and the required wrenches. While grasping any part of an object may work for a rigid object, the physical interaction with a deformable object should be carefully planned to contact the right spot on the object with correct poses and wrenches.

\appendix
\chapter{Symbols and Notation} \label{appendix:notation}


The symbols in the document that have special meanings are listed in Table~\ref{tbl:notation:symbols}.

\begin{longtable}{ll}\toprule 
Symbol(s) & Meaning\\ \midrule
\arrayrulecolor{lightgray}

$\FI, \FB, \FR{i}$              & Inertial, body-fixed and $\nth{i}$ rotor coordinate frames \\ \hline
$\OI, \OB, \OR{i}$              & Inertial, body-fixed and $\nth{i}$ rotor frame origins \\ \hline
$\ORB{i}$                       & Origin of $\nth{i}$ rotor described in body-fixed frame \\ \hline
$\XI, \YI, \ZI$                 & Axes of the inertial frame \\ \hline
$\XB, \YB, \ZB$                 & Axes of the body-fixed frame \\ \hline
$\XR{i}, \YR{i}, \ZR{i}$        & Axes of the $\nth{i}$ rotor frame \\ \hline
$\RBI, \RIB$                    & Rotation from inertial frame to body-fixed frame and vice-versa \\ \hline
$\RBR{i}, \RRB{i}$              & Rotation from body-fixed frame to $\nth{i}$ rotor frame and vice-versa \\ \hline

$\vec{p}^\frm{I}, \vec{q}$   & The position and attitude of the robot in the inertial frame \\ \hline
$x, y, z$                       & The robot position elements in the inertial frame \\ \hline
$\mat{\Phi}$                    & The set of Euler angles of the robot \\ \hline
$\phi, \theta, \psi$            & Roll, pitch and yaw of the robot \\ \hline
$\omega$                        & Body angular velocity \\ \hline
$p, q, r$                       & Elements of the angular velocity \\ \hline
$\vec{r}_i$                     & The vector from the robot's origin to the origin of the $\nth{i}$ rotor \\ \hline
$r_{i_x}, r_{i_y}, r_{i_z}$     & The $\nth{i}$ rotor position elements in the body-fixed frame \\ \hline
$\alpha_i$                      & The angle between $\vec{r}_i$ and $\vec{r}_{i+1}$ projections on $\XB\YB$ plane \\ \hline
$\phi_{dih_i}$                  & The angle between $\vec{r}_i$ and its projection on $\XB\YB$ plane \\ \hline
$\mu_i$                         & The rotation angle of $\vec{r}_i$ projection on $\XB\YB$ plane \\ \hline
$\phix{i}, \phiy{i}$            & Inward and sideward angles of $\nth{i}$ rotor frame \\ \hline
$\ell_i$                        & The length of the $\nth{i}$ rotor arm \\ \hline
$n_r$                           & The number of rotors in the multirotor \\ \hline

$m, m_{rotor_i}, m_{leg_i}$     & The total mass of the multirotor, $\nth{i}$ rotor and $\nth{i}$ rotor arm \\ \hline
$\InertB$                       & The body-frame inertia tensor of the multirotor \\ \hline
$g$                             & Gravitational acceleration \\ \hline
$c_{Fi}, c_{\tau i}$            & Thrust and moment constants of the $\nth{i}$ rotor \\ \hline
$\Omega_i$                      & Rotational (angular) velocity of the $\nth{i}$ rotor \\ \hline
$d_i$                           & Spinning direction of the $\nth{i}$ rotor \\ \hline

$\vec{F}, \vec{M}$              & Total force and moment applied to robot \\ \hline
$\vec{F}_{grav}, \vec{F}_{thr}$ & Gravitational and the total rotor thrust forces applied to robot \\ \hline
$\vec{M}_{grav}, \vec{M}_{thr}$ & Robot moments resulted from weight distribution and rotor thrusts \\ \hline
$\vec{M}_{reac}$                & Reaction moment as a result of the rotors spinning \\ \hline
$\vec{F}_{app}, \vec{M}_{app}$  & External force and moment applied by the end-effector \\ \hline
$\vec{F}_{des}, \vec{M}_{des}$  & Desired force and moment applied by the end-effector \\ \hline

$\mat{x}$                       & The state of the system \\ \hline
$\mat{u}$                       & Vector of squared rotor angular velocities (robot system input) \\ \hline
$\mat{u'}$                      & The inputs to the extended robot/end-effector system \\ \hline
$\mat{y}$                       & The system output vector \\ \hline
$\mat{y_d}, \mat{y_m}$          & The set of the desired and the measured variables of the system \\ \hline
$\vec{p_d}, \vec{q_d}$          & The desired end-effector position and attitude \\ \hline
$\vec{p_e}, \vec{q_e}$          & The position and attitude of the end-effector \\ \hline
$\mat{f}(\mat{x})$              & System state drift due to gravity and rotational inertia\\ \hline
$\mat{J}(\mat{x})$              & Decoupling matrix mapping the input $\mat{u}$ to the state space\\ \hline

$\mat{\eta}(\mat{\Phi})$        & Matrix mapping angular velocity to Euler angular rates \\ \hline
$\mat{L}$                       & Matrix mapping input $\mat{u}$ to total thrust force $\vec{F}_{{thr}}^\frm{B}$ \\ \hline
$\mat{G}$                       & Matrix mapping input $\mat{u}$ to the reaction moment $\vec{M}_{reac}^\frm{B}$ \\ \hline
$\mat{F}$                       & Matrix mapping input $\mat{u}$ to the thrust moment $\vec{M}_{thr}^\frm{B}$ \\ \hline
$\mat{M}$                       & The sum of $\mat{F}$ and $\mat{G}$ matrices \\
\arrayrulecolor{black} \bottomrule
\caption{List of the symbols}
\label{tbl:notation:symbols}
\end{longtable}

\clearpage
A summary of notation used in this document is as shown in Table~\ref{tbl:notation:notations}.

\begin{longtable}{l L{3.2in}  L{1.5in}} \toprule
  Symbol   & Quantity &  Comments\\ \midrule \arrayrulecolor{lightgray}


  $\numset{R}{n}{m}$ & Real space of dimension $n \times m$ \\ \hline

  $SE(3)$ & Special Euclidean group & \\ \hline
  $SO(3)$ & Special orthogonal group & \\ \hline

  $s, S$                    & scalar & \\ \hline
  $\vec{v}, \vec{V}$        & Geometric vector & can also be shown as matrix $\mat{v}$ and $\mat{V}$\\ \hline
  $\dvec{v}, \ddvec{v}$     & First and second time derivatives of vector $\vec{v}$\\ \hline
  $\vecelem{V}{x}$, $\vecelem{V}{y}$, $\vecelem{V}{z}$    & The $x$, $y$ and $z$ components of vector $\vec{V}$ \\ \hline
  $\mat{m}, \mat{M}$        & Matrix & \\ \hline
  $\mat{m}\T, \mat{M}\T$    & Transpose of matrices $\mat{m}$ and $\mat{M}$ & \\ \hline
  $\matcomp{\mat{M}}{ij}$   & Matrix element at row $i$ and column $j$ & \\ \hline
  $\frm{F}$                 & Reference frame & \\ \hline
  $\axis{X}$                & Axis or unit vector & \\ \hline
  $\matcomp{\axis{X}}{i}$   & \nth{i} element of the unit vector & \\ \hline
  $\vec{v}^\frm{F}$         & Vector $\vec{v}$ described in frame $\frm{F}$ \\ \hline
  $\unit{U}$                & Unit of measurement & \\ \hline

  $\sine{(\cdot)}, \cosine{(\cdot)}$  & Abbreviations for $\sin{(\cdot)}$ and $\cos{(\cdot)}$ & \\ \hline
  
  \arrayrulecolor{black}\bottomrule
\caption{List of the notations}
\label{tbl:notation:notations}
\end{longtable}


\backmatter

\singlespace


\bibliographystyle{plainnat}

\bibliography{bibliography/references}
\end{document}